\documentclass[mnsc,nonblindrev]{informs4}

\OneAndAHalfSpacedXII 
\usepackage{amsmath,amssymb,amsfonts}

\usepackage{natbib}
 \bibpunct[, ]{(}{)}{,}{a}{}{,} \def\bibfont{\small}    
\usepackage{microtype}

\usepackage{rotating}
\usepackage{fancyvrb}
\usepackage{subfig}

\usepackage{enumerate}
\usepackage{multirow}
\usepackage[ruled,linesnumbered]{algorithm2e}
\usepackage{bm}
\usepackage{tabularx}
\usepackage{makecell}
\usepackage{bbm}
\usepackage[hidelinks]{hyperref}
\usepackage[normalem]{ulem}
\usepackage{bibunits}

\usepackage{endnotes}
\let\footnote=\endnote

\usepackage[dvipsnames]{xcolor}
\SetKwInput{KwInit}{Initialization}
\newcommand{\phat}{\hat{\mathbb{P}}}
\newcommand{\supp}[1]{\mathrm{supp}(#1)}
\newcommand{\dom}[1]{\mathrm{dom}(#1)}
\newcommand{\ptrue}{\mathbb{P}^{*}}

\DeclareMathOperator*{\esssup}{ess\,sup}

\newtheorem{lemmaAp}{Lemma}[section]
\newtheorem{theoremAp}{Theorem}[section]
\newtheorem{propositionAp}{Proposition}[section]
\newtheorem{corollaryAp}{Corollary}[section]
\newtheorem{assumptionAp}{Assumption}[section]
\newtheorem{definitionAp}{Definition}[section]

\defaultbibliographystyle{informs2014} 
\defaultbibliography{reference} 
\TheoremsNumberedThrough     \JOURNAL{}

\EquationsNumberedThrough    
\MANUSCRIPTNO{MS-DSC-2025-00661.R2}

\makeatletter
\renewcommand{\theARTICLEABSTRACT}{  \HOOKb
  \vspace*{18pt}
  \noindent
  \begin{minipage}[t]{\textwidth}
    \parindent1em
    \ABSfont
    \noindent\theABSTRACT\endgraf
    \vskip5pt
    \theFUNDING
    \theKEYWORDS
    \theSUBJECTCLASS
    \theAREAOFREVIEW
    \theMSCCLASS
    \theORMSCLASS
    \if@BLINDREV\else\theHISTORY\fi
    \noindent\hrulefill
  \end{minipage}
  \vspace*{0pt}
}
\makeatother

\renewcommand{\theARTICLETOP}{}

\begin{document}

\RUNAUTHOR{Cui et al.} 
\RUNTITLE{Fragility-aware Classification for Risk Characterization and Robust Learning} 

\TITLE{Fragility-aware Classification for Risk Characterization and Robust Learning}

\ARTICLEAUTHORS{\AUTHOR{Zheng Cui\textsuperscript{1}, Daniel Zhuoyu Long\textsuperscript{2}, Jin Qi\textsuperscript{3}, Chen Yang\textsuperscript{1*}, Ruohan Zhan\textsuperscript{4}}
\AFF{\textsuperscript{1} School of Management, Zhejiang University, \EMAIL{chenyang12@zju.edu.cn, zhengcui@zju.edu.cn}}
\AFF{\textsuperscript{2} Department of Systems Engineering and Engineering Management, The Chinese University of Hong Kong, \EMAIL{zylong@se.cuhk.edu.hk}}
\AFF{\textsuperscript{3} Department of Industrial Engineering and Decision Analytics, Hong Kong University of Science and Technology, \EMAIL{jinqi@ust.hk}}
\AFF{\textsuperscript{4} London Business School, \EMAIL{rzhan@london.edu}}
}

\ABSTRACT{{Classification models play a central role in data-driven decision-making applications such as medical diagnosis, recommendation systems, and risk assessment. Traditional performance metrics, such as accuracy and AUC, quantify classification errors and pairwise ranking performance but do not account for the magnitude of errors, i.e., the risk of confident misjudgments. This limitation is particularly consequential in safety-critical and cost-sensitive settings, where overconfident errors can lead to severe outcomes. To address this issue, we propose the Fragility Index (FI), a novel performance metric that evaluates classifiers from a risk-averse perspective by capturing the tail risk of confident misjudgments. We formulate FI within a robust satisficing (RS) framework to quantify classifier sensitivity under distributional uncertainty. Building on this, we develop a tractable surrogate-based training framework and establish provable bounds connecting training fragility to ranking-error FI. We further derive tractable reformulations for a broad class of loss functions, including cross-entropy, hinge-type, and Lipschitz losses, and identify specific conditions under which these reformulations are exact. We also introduce an FI-induced regularizer for deep neural networks. Empirical results on real-world medical diagnosis tasks demonstrate that FI complements existing metrics by revealing error tail risk not reflected in accuracy or AUC. FI-based models achieve competitive predictive performance while generally attaining lower FI. In the simulated human--AI diagnostic workflow, they also achieve lower average operational costs than empirical risk minimization.}}


\maketitle

\begin{bibunit}

\section{Introduction}\label{sec:introduction}

Classification is a fundamental machine learning task with broad applications in operations management, including supply chain management \citep{kumar2022role}, healthcare operations \citep{pianykh2020improving}, and empirical theory building \citep{chou2023supervised}. The selection of appropriate performance metrics to evaluate classification models (classifiers) is crucial and remains a key focus in both academia and industry \citep{powers2020evaluation}.

Many popularly used performance metrics focus on the rate of \emph{misjudgment}, namely, the frequency of false predictions \citep{grandini2020metrics}. Accuracy aggregates pointwise errors across samples, while AUC, the area under the Receiver Operating Characteristic (ROC) curve, captures pairwise ranking errors between classes. Metrics derived from the confusion matrix, such as precision, recall, and F$_\beta$ score, similarly quantify discrete misclassification counts. These measures are effective at evaluating how often a classifier is wrong, but largely ignore the \emph{risk of misjudgment}, that is, the severity of errors when they occur~\citep{norton2019maximization}.

In many operational contexts, however, the magnitude of confidence behind a wrong prediction critically determines its consequence \citep{vaicenavicius2019evaluating}. In disease detection and medical diagnosis, the confidence of a classifier influences treatment decisions and allocation of medical resources \citep{huang2020tutorial}. In financial risk management, confidence levels directly affect credit approval and fraud investigation processes \citep{moscatelli2020corporate}. In autonomous driving, overconfident object misclassification can lead to catastrophic system failures \citep{bojarski2016end}. In these applications, a hesitant mistake and a highly confident mistake carry fundamentally different risks, even if both are counted equally under standard metrics.

Traditional metrics fail to capture this distinction. Consider AUC in binary classification. AUC measures the probability that a randomly chosen positive sample receives a higher score than a randomly chosen negative sample \citep{hanley1982meaning}. It therefore captures the frequency of ranking errors but remains indifferent to the \emph{magnitude} of those ranking violations \citep{norton2019maximization}\endnote{This magnitude is also related to the classification margin and generalization performance \citep{vapnik2013nature}.}. To illustrate, consider the following toy example.

\begin{example}
\label{exp:1}
Consider a training dataset with two positive samples and two negative samples. A classifier assigns a probability of being positive to each sample. Compare two classifiers A and B in Table \ref{tab:example1}: Classifier A assigns probabilities $\{0.6, 0.9\}$ to positive samples and $\{0.2, 0.7\}$ to negative samples; Classifier B assigns the same probabilities to positive samples, but $\{0.1, 0.8\}$ to negative samples. Both classifiers incorrectly rank one negative sample above one positive sample and yield identical Accuracy (0.75) and identical AUC (0.75). However, classifier B assigns a higher probability (0.8 versus 0.7) to the misclassified negative sample, representing a more severe and confident error. From a risk-management perspective, classifier B is strictly more dangerous, while conventional metrics cannot differentiate between them.
\begin{table}[htbp]
    \centering
    \begin{tabular}{ccccc}
        \hline
         Classifier & Prob. on positive samples & Prob. on negative samples & Accuracy & AUC  \\
         \hline
         A & $\{0.6, 0.9\}$ & $\{0.2, 0.7\}$ & $0.75$ & $0.75$  \\
         B & $\{0.6, 0.9\}$ & $\{0.1, 0.8\}$ & $0.75$ & $0.75$  \\
         \hline
    \end{tabular}
    \caption{The predicted probability and performance metrics of the two classifiers in Example \ref{exp:1}.}
    \label{tab:example1}
\end{table}
\end{example}

This limitation reflects a broader gap. Existing performance metrics emphasize error frequency but overlook error severity and tail risk. Simply replacing error counts with the expected magnitude of error is insufficient, because expectation remains risk-neutral and does not penalize tail risk that is also associated with rare but catastrophic failures. This calls for a principled way to quantify and control the \emph{risk of confident misjudgment}.

Parallel to this concern is the growing interest in the \emph{generalizability} of classifiers. In practice, models are deployed in environments that differ from the training distribution due to covariate shifts, label noise, or adversarial perturbations. Improving generalization has thus become a central theme in machine learning. The conventional empirical risk minimization (ERM) framework optimizes performance on the training data but provides limited guarantees under distributional shifts \citep{vapnik2013nature}. Various approaches, such as regularization, data augmentation, and distributionally robust optimization (DRO) \citep{kuhn2025distributionally}, have been proposed to mitigate this issue.

Among these, robust satisficing (RS) has emerged as a promising target-oriented framework for handling distributional ambiguity \citep{long2023robust}. RS specifies a reference performance level and minimizes the model \emph{fragility}, defined as the sensitivity of the model to distributional shifts. Compared with DRO, which optimizes worst-case performance, RS is typically less conservative and more interpretable in terms of acceptable risk levels \citep{long2023robust}. However, existing RS-based supervised learning frameworks, such as \cite{sim2021new}, primarily focus on out-of-sample performance under generic loss functions, without explicitly capturing the risk of confident misjudgment and its consequences. They also commonly assume perfectly reliable labels, thereby overlooking the joint presence of covariate and label shifts that is prevalent in practice.

These observations suggest that the risk of confident misjudgment and generalization are intrinsically linked. Samples on which a model is overconfident yet fragile are precisely those that are most vulnerable under distributional shifts. Hence, controlling such errors requires a risk-aware performance metric and a robustness-oriented learning framework.

To this end, we propose the Fragility Index (FI), a new performance metric that captures the tail risk of confident misjudgments while accounting for joint distributional ambiguity. FI explicitly penalizes large error magnitudes and incorporates a robust satisficing perspective, providing a risk-aware complement to existing metrics. {Building on this metric, we develop a computationally tractable training framework {based on a surrogate objective and establish provable FI bounds of the resulting models}.} We further derive closed-form reformulations for a broad class of loss functions and extend the approach to deep neural networks through a lightweight regularization scheme. Our results show that FI-based evaluation reveals important differences between models that are indistinguishable under standard metrics, and that FI-based training improves robustness and generalization while reducing the risk of confident misjudgments. These properties make FI particularly valuable for high-stakes applications, where reliable and risk-aware decision-making is essential.

\subsection{Contributions}

In this work, we develop a novel classification framework that addresses two critical challenges: (i) controlling the risk of misjudgment, which conventional performance metrics such as accuracy and AUC fail to capture, and (ii) improving the generalizability of classifiers to unseen or adversarial samples. We address these challenges through the following key contributions.

\begin{itemize}
    \item \emph{A new performance metric---Fragility Index (FI)}: We introduce the Fragility Index, which quantifies the risk and generalizability of a classifier under a robust satisficing framework. By explicitly penalizing large error magnitudes, FI provides a tractable and risk-averse measure of confident misjudgments and reflects the tail risk of the error distribution. {In particular, we {define the }multi-class FI {as the equal-weight }macro-average {of the FIs over all ordered one-versus-one class pairs, thereby reducing the masking of minority-class risks }by majority classes.}

    \item {\emph{A classifier training framework controlling FI}: We develop a computationally efficient training framework that minimizes a loss-based surrogate fragility parameter.     We prove that the resulting model's FI is effectively bounded by this surrogate, 
    connecting the training objective and risk control. The framework accommodates widely used loss functions, including hinge-type and cross-entropy losses, and admits exact reformulations under commonly used ambiguity sets. We further examine its behavior under distributional shifts such as label noise and establish corresponding statistical learning guarantees.}

    \item {\emph{Extensive empirical evaluations and managerial insights}: {We validate FI and FI-based training through real-world medical-diagnosis and complementary synthetic experiments, showing improved robustness while retaining competitive predictive performance. 
In a simulated human--AI diagnostic workflow, we establish sufficient conditions for alignment between FI and expected operational cost. Our experiments in this workflow also show that FI-based models achieve lower average costs than ERM and other considered models.}}

    \item \emph{Extension to deep neural networks}: We extend our framework to deep neural networks by introducing a lightweight FI-based regularizer derived from an augmented Lagrangian formulation. The resulting objective is smooth and computationally efficient, making it scalable to complex architectures. Empirical results on image-based medical diagnosis tasks show that FI-regularized models outperform ERM baselines using the same weight decay.
\end{itemize}

\subsection{Notations}
Vectors are denoted by boldface lowercase letters (e.g., $\BFbeta$), while matrices are represented by boldface uppercase letters (e.g., $\BFB$). Sets are denoted by calligraphic letters (e.g., $\mathcal{X}$). We use $\mathcal{P}(\mathcal{X})$ to denote the set of all distributions of an $n$-dimensional random vector with support $\mathcal{X} \subseteq \mathbb{R}^n$. For a distribution $\mathbb{P}$, $\supp{\mathbb{P}}$ denotes the support of $\mathbb{P}$. For a positive integer $m$, let $[m] = \{1, \dots, m\}$. We follow the convention that $\inf \emptyset = + \infty$. 

The norm of the vector $\BFx$ is denoted by $\|\BFx\|$, and its dual norm is defined by $\|\BFx\|_* = \sup_{\|\BFz\| \leq 1} \BFz^T\BFx$. For a function $f:\mathcal{X}\to \mathbb{R}$, the effective domain of $f$ is denoted as $\dom{f}:=\{\BFx\in\mathcal{X}\mid f(\BFx)<\infty\}$. The conjugate of a convex function $\rho(\BFx)$ is defined as $\rho^*(\BFz) = \sup_{\BFx \in \dom{\rho}}\{ \BFz^T \BFx - \rho(\BFx)\}$. For a two-variable function $\ell(\BFx, \BFy)$, $\ell^{1*}(\BFz, \BFy) = \sup_{\BFx \in \dom{\ell(\cdot, \BFy)}} \{\BFz^T\BFx - \ell(\BFx, \BFy)\}$ denotes the convex conjugate to the first variable $\BFx$. For convenience, we use $\dom{\ell}$ to denote $\dom{\ell(\cdot, \BFy)}$ and $\dom{\ell^{1*}}$ to denote $\dom{\ell^{1*}(\cdot, \BFy)}$.
Finally, following the convention in convex analysis, the
extended-valued indicator function of a set $\mathcal{X}$ is
defined as $\delta_{\mathcal{X}}(\BFx)=0$ if
$\BFx\in\mathcal{X}$ and $+\infty$ otherwise.
Its convex conjugate is the support function of $\mathcal{X}$,
given by
$\delta_{\mathcal{X}}^*(\BFz)
=\sup_{\BFx\in\mathcal{X}}\BFz^T\BFx$. The positive part is denoted by $(\cdot)_+ = \max\{0, \cdot\}$.

\section{Related Literature}\label{sec:literature}

\subsection{Risk-Aware Performance Metrics in Classification}
Our work relates to the design of performance metrics that capture uncertainty in classification. Conventional metrics such as accuracy and AUC are based on error rates, measuring the probability of misclassification \citep{powers2020evaluation}. Building on these, refined metrics such as the $F_\beta$-score and the Matthews correlation coefficient have been proposed \citep{grandini2020metrics}. However, these metrics rely on discrete error counts at fixed decision thresholds, treating borderline errors and confident misjudgments identically. As a result, they ignore error magnitude and are limited in assessing risk severity in safety-critical settings.

To address this limitation, \cite{norton2019maximization} extended AUC to buffered AUC using the buffered probability of exceedance, and \cite{chaudhuri2022certifiable} developed risk measures based on superquantiles and buffered probabilities. These approaches incorporate information about large errors and better capture tail risk. More recently, \cite{yang2023fragility} introduced the notion of fragility in binary classification and studied its role in model evaluation.

Our work builds on these foundations in three key ways. First, we propose a new, fragility-based performance metric to capture the risk of confident misjudgments in multi-class settings. Second, we provide a unified theoretical framework that connects fragility to distributional robustness. {Third, we develop a tractable {robust-satisficing training framework based on a loss-based surrogate and establish conditions under which its training fragility controls ranking-error FI}.}

\subsection{Uncertainty Quantification and Model Calibration}
The magnitude of prediction errors is closely tied to estimated probabilities and the risk of misjudgment, linking our work to uncertainty quantification and model calibration in machine learning \citep{ghanem2017handbook}. This issue has received increasing attention, particularly as deep neural networks are known to exhibit overconfidence \citep{guo2017calibration}. Traditional approaches focus on post-hoc calibration, evaluating the alignment between predicted probabilities and empirical accuracy \citep{abdar2021review}. More recent work incorporates uncertainty directly into training, including Bayesian methods, ensemble learning, and data augmentation \citep{vaicenavicius2019evaluating, nemani2023uncertainty}. 

In contrast, our approach takes a different perspective. Rather than focusing solely on calibration, we propose a metric that explicitly penalizes large, confident errors and accounts for distributional ambiguity through the robust satisficing (RS) framework. Unlike risk measures such as buffered probability or superquantile, which rely on the empirical data distribution and accurate estimation of tail expectations, our metric is designed to be robust to distributional shifts. {As a result, {FI }captures error severity {under distributional ambiguity, while the associated robust-satisficing training framework supports generalization under distributional shifts}.}

\subsection{Distributionally Robust Optimization and Robust Satisficing}
Our work is closely related to robust optimization, which integrates distributional ambiguity into optimization and machine learning models. A widely used approach in this area is distributionally robust optimization (DRO) \citep{kuhn2025distributionally}. A key insight in this literature is the theoretical equivalence between distributional robustness and regularization. Various studies have established this connection, demonstrating that data-driven robust learning inherently induces a regularization effect under specific conditions 
\citep{shafieezadeh2019regularization, gao2024wasserstein}. DRO is also computationally tractable in many settings \citep{kuhn2019wasserstein}, which has led to broad applications in classification \citep{wang2024wasserstein}, reinforcement learning \citep{liu2022distributionally}, and deep learning \citep{bui2022unified}.

A closely related and increasingly popular framework is robust satisficing (RS), introduced by \cite{long2023robust}, which is a target-oriented framework that minimizes the target violations under distributional shift. While RS and DRO differ in their formulations, \cite{wang2025equivalence} established their mathematical equivalence under certain conditions. Recent work further extends RS into an end-to-end data-driven pipeline that bridges prediction and decision-making \citep{sim2025analytics}. Driven by the interpretability of the target parameter and its theoretical alignment with DRO, RS has been widely applied across various fields, such as supervised learning \citep{sim2021new}, reinforcement learning \citep{ruan2023robust}, and online learning \citep{saday2025robust}.

Building upon this, our work is most closely related to \citet{sim2021new}, which introduces the RS framework to supervised learning to provide a perspective on standard regularization. However, our approach differs from theirs in three main aspects. First, our primary objective is not to bound a generic empirical loss, but to control the risk of confident misjudgments via the ranking-error FI. Consequently, we utilize the loss-based RS formulation strictly as a computational surrogate, establishing a theoretical bridge to bound the intractable ranking error. Second, while their explicit reformulations primarily focus on covariate shifts with deterministic labels, we model joint distributional ambiguity to accommodate multi-class label noise. Third, we provide exact, finite convex reformulations for key settings, most notably multi-class cross-entropy, which remained structurally uncharacterized in their work, and extend the RS optimization to deep neural networks. Collectively, these distinctions tailor the RS framework specifically to the risk-aware demands of high-stakes classification.

\subsection{Domain Adaptation and Generalization}
Finally, our work is related to domain adaptation literature, which aims to mitigate the performance degradation caused by distributional shifts between training (source) and testing (target) data. Foundational approaches in this field focus on minimizing statistical discrepancies across domains, such as \emph{Maximum Mean Discrepancy} \citep{long2015learning}, or employing adversarial learning to align feature distributions \citep{ganin2016domain}. More recent advancements have moved towards source-free domain adaptation, which adapts pre-trained models to target domains without accessing source data \citep{fang2024source}.

A common limitation of these approaches is the reliance on target-domain data (unsupervised or semi-supervised) to guide adaptation. While domain generalization \citep{zhou2022domain} alleviates this requirement by learning invariant representations, our approach takes a complementary perspective. Rather than adapting to a specific target distribution, we control the model's fragility over a worst-case ambiguity set, providing robustness guarantees against unforeseen shifts without requiring target data or multiple source domains.

\section{Fragility Index: A New Performance Metric}
\label{sec:FI}

We consider stochastic classifiers that generate a probability vector for each input.{\endnote{For deterministic classifiers with scores, the probability vector can be generated through the softmax function or proper calibration methods such as Platt scaling \citep{platt1999probabilistic}.}} { Although our primary focus is multi-class classification, we begin with  the binary FI and its computation to fix notation, and then construct the multi-class FI by aggregating pairwise binary FIs and establish its properties.} 
\subsection{{FI in Binary Classification}}\label{sec:defining_FI}

Consider binary classification with input $\BFx \in \mathcal{X}$ and label $y \in \mathcal{Y}$. Let $p(\BFx)$ denote the estimated probability that the sample with feature $\BFx$ belongs to the positive class. Define the ranking error as
$$
    \varepsilon(p) = p(\BFx^-) - p(\BFx^+),
$$
where $\BFx^+$ and $\BFx^-$ are positive and negative samples, respectively.

A common performance metric is AUC, which measures the probability of correct ranking for a random positive--negative pair \citep{hanley1982meaning}: $    AUC(p) = \mathbb{P} (p(\BFx^-) \leq p(\BFx^+)) = \mathbb{P} (\varepsilon(p) \leq 0) $. However, AUC only captures the frequency of correct rankings and ignores the magnitude of errors. Large positive values of $\varepsilon(p)$ correspond to confident misjudgments and should be penalized more heavily. A natural alternative is to consider the expectation $\mathbb{E}[\varepsilon(p)]$, which accounts for magnitude but is risk-neutral, treating correct and incorrect rankings symmetrically.

These limitations motivate a risk-averse metric, the Fragility Index (FI), which captures both confident misjudgment and robustness to distributional shifts. {Let $\tau$ be a target level for the expected ranking error. Assume that the empirical sample contains observations from both classes. Let $\widehat{\mathbb{P}}_+$ and $\widehat{\mathbb{P}}_-$ denote the empirical class-conditional feature distributions for the positive and negative classes, respectively, and define the nominal ordered-pair distribution $ \widehat{\mathbb Q}:=\widehat{\mathbb{P}}_+\otimes\widehat{\mathbb{P}}_- \in\mathcal P(\mathcal X^2). $
}\begin{definition}[Fragility Index]
   {{Given a target value $\tau$ for the expected ranking error, }the Fragility Index is defined as
    \begin{equation} \label{eq:fi_def}
        \mathrm{FI}(p; \tau)=\inf\left\{ k>0 \mid  \mathbb{E}_{\mathbb{Q}}[\varepsilon(p)]\leq \tau + k D (\mathbb{Q},\widehat{\mathbb{Q}}),\ \forall \mathbb{Q}\in \mathcal{P}(\mathcal{X}^2)\right\},
\end{equation}
where $D(\mathbb Q,\widehat{\mathbb Q})$ is a divergence between distributions on the sample-pair space $\mathcal X^2$.} 
\end{definition}

{
The pairwise ranking performance depends only on their class-conditional feature distributions. Hence, we formulate distributional uncertainty directly on the ordered-pair space, rather than on the joint distribution of features and labels.\footnote{{Allowing $\mathbb Q$ to range over all of $\mathcal P(\mathcal X^2)$ provides a conservative relaxation of product-form shifts induced by independently drawn class-conditional populations. Accordingly, the FI in Equation \eqref{eq:fi_def} should be interpreted as a measure of robustness of pairwise ranking comparisons, rather than as an exact characterization of every shifted classification population.}}}

\phantomsection\label{loc:tau_zero_interpretation}
{FI quantifies the smallest sensitivity coefficient needed to meet the target $\tau$ under distributional shifts; smaller values indicate greater robustness. Since a ranking error of zero marks the boundary between correctly and incorrectly ordered pairs, a natural choice is $\tau=0$. Alternatively, the target can be tied to empirical pairwise performance by setting $\tau=\lambda\mathbb{E}_{\widehat{\mathbb Q}}[\varepsilon(p)]$, with $\lambda\geq 0$ serving as a scaling parameter relative to this nominal baseline.}

{
\paragraph{Calculating Fragility Index}

For the ranking error $\varepsilon(p)$, define the function $G$ by
\begin{equation}
    \label{eq:def_gk}
    G(k) = \sup_{\mathbb{Q} \in \mathcal{P}(\mathcal{X}^2)} \left\{
        \mathbb{E}_{\mathbb{Q}}[\varepsilon(p)] - k D (\mathbb{Q},\widehat{\mathbb{Q}})
    \right\} - \tau, \qquad k>0,
\end{equation}
where we {suppress the dependence of $G(k)$ on $p$ and $\tau$ for notational simplicity}. The following result {characterizes FI through a monotone feasibility condition.
}}

\begin{lemma}
   \label{lemma:root_finding}
    {The function $G(k)$ in Equation \eqref{eq:def_gk} is {non-increasing in }$k$, and the FI defined in Equation \eqref{eq:fi_def} satisfies
    {\[
        \mathrm{FI}(p;\tau)
        =
        \inf\{k>0:G(k)\leq0\}.
\]}}
\end{lemma}

{Since $G(k)$ is monotone, {the smallest $k$ satisfying $G(k)\leq0$ }can be efficiently computed via bisection. When $G$ is continuous and strictly decreasing, this reduces to finding its unique root.}

{ We next derive an explicit expression for $G(k)$ under KL divergence. Let $\mathrm{FI}_{\mathrm{KL}}(p; \tau)$ denote FI computed with $D=D_{\mathrm{KL}}$, where}

{\begin{equation}
    \label{eq:kl_def}
    D_{\mathrm{KL}}(\mathbb{Q} \Vert {\widehat{\mathbb{Q}}}) = \left\{ \begin{array}{ll}
        \mathbb{E}_{\mathbb{Q}}\Big[\ln\left( \frac{\mathrm{d}\mathbb{Q}}{\mathrm{d}{\widehat{\mathbb{Q}}}}\right)\Big] & \text{if $\mathbb{Q} \ll \widehat{\mathbb{Q}}$},\\
        \infty & \text{otherwise},
    \end{array}\right.
\end{equation}
and {$\mathbb{Q} \ll \widehat{\mathbb{Q}}$ denotes that $\mathbb{Q}$ }is absolutely continuous with respect to $\widehat{\mathbb{Q}}$. We obtain the following reformulation.}
\begin{lemma}\label{lemma:fi_kl}
    {Suppose $\mathbb{E}_{\widehat{\mathbb Q}}[\varepsilon(p)] < \tau < \esssup_{\widehat{\mathbb Q}} \varepsilon(p)$. Then, $\mathrm{FI}_{\mathrm{KL}}(p; \tau)$ is determined by the unique root of $G_{\mathrm{KL}}(k)$ on $k>0$, where
    \begin{equation*}
        G_{\mathrm{KL}}(k) = k \ln \left(\mathbb{E}_{\widehat{\mathbb Q}}[\exp(\varepsilon(p)/k)]\right) - \tau.
\end{equation*}}
\end{lemma} 

{This closed-form expression for $G_{\mathrm{KL}}(k)$ reduces the computation of $\mathrm{FI}_{\mathrm{KL}}(p;\tau)$ from an infinite-dimensional optimization over $\mathbb Q$ to a one-dimensional root-finding problem. It also highlights the role of tail risk. A heavier right tail in $\varepsilon(p)$ leads to a larger $\mathrm{FI}_{\mathrm{KL}}(p;\tau)$, reflecting stronger penalties for confident misjudgment. The next result shows that FI controls large errors, a property not captured by AUC.}

\begin{proposition}\label{prop:fi_kl_ub}
    {Under the conditions of Lemma \ref{lemma:fi_kl}, for any $\theta\in\mathbb R$, we have
    }{\[
    \widehat{\mathbb Q}(\varepsilon(p)\geq \theta)\leq \exp\left(- \frac{\theta - \tau}{\mathrm{FI}_{\mathrm{KL}}(p;\tau)}\right).
\]}
\end{proposition}
Since positive $\varepsilon(p)$ corresponds to misjudgment, Proposition \ref{prop:fi_kl_ub} bounds the right tail of the error distribution. {{For thresholds $\theta>\tau$, }a smaller FI {yields a tighter exponential upper bound, implying a faster  tail-decay rate}.} 
{We conclude with examples illustrating how $\mathrm{FI}_{\mathrm{KL}}$ captures risk magnitude not reflected in AUC.}

\noindent \paragraph{Revisiting Example \ref{exp:1}} While Accuracy and AUC cannot distinguish classifiers A and B, FI does: $\mathrm{FI}_{\mathrm{KL}}(p_A; 0) = 0.073$ and $\mathrm{FI}_{\mathrm{KL}}(p_B; 0) = 0.163$. This correctly identifies classifier B as more prone to confident misjudgment.

\begin{example}
\label{exp:2}
We next consider a more realistic setting. We generate a binary classification dataset with informative Gaussian features and noise, and train logistic regression and random forest models on 100 balanced samples. Figure \ref{fig:fi_example} reports the ranking error and confidence of incorrect predictions. Despite having similar AUC ($0.904$), the two models differ substantially in tail behavior. Logistic regression exhibits a heavier right tail in $\varepsilon(p)$, indicating more severe misjudgments, which is also reflected in higher confidence on incorrect predictions. Accordingly, FI distinguishes the models: $\mathrm{FI}_{\mathrm{KL}}=0.211$ for logistic regression versus $0.125$ for random forest. In contrast, AUC fails to capture this difference.

\begin{figure}[!htbp]
    
    \centering
    \includegraphics[width=0.4\linewidth]{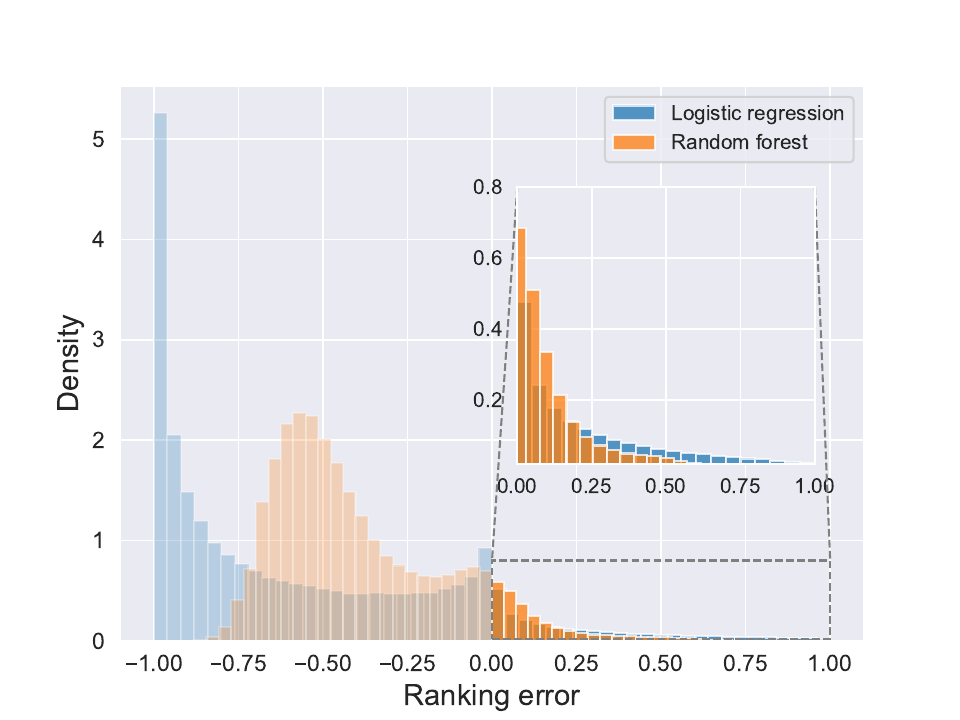}
    \includegraphics[width=0.4\linewidth]{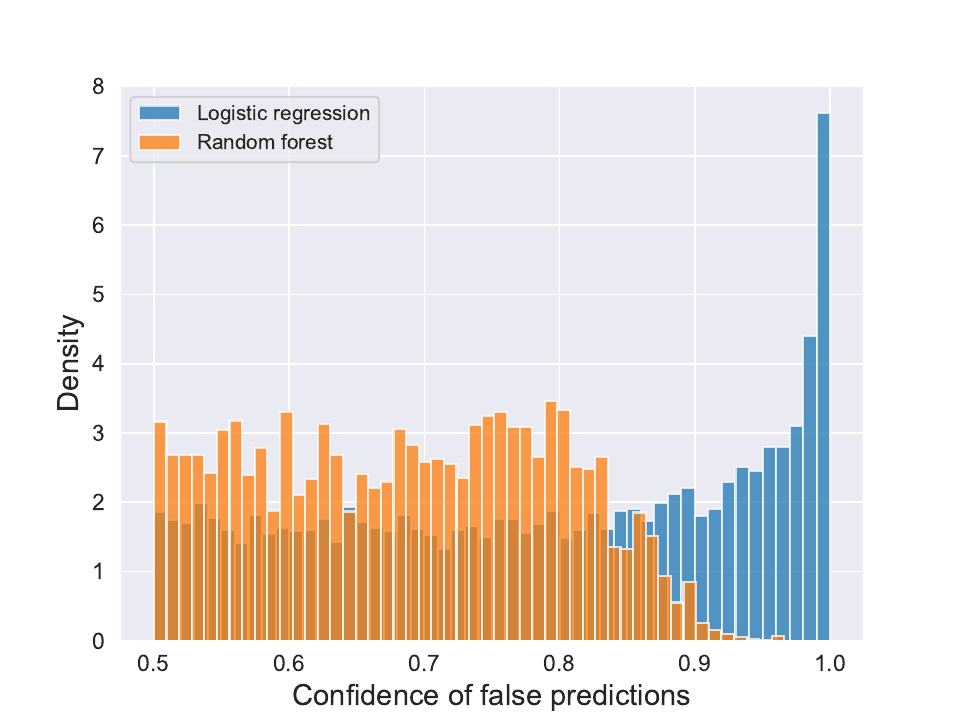}
    \caption{{Distributions of pairwise ranking errors and confidence on incorrect predictions for the two classifiers in Example \ref{exp:2}.}}
    \label{fig:fi_example}
    
\end{figure}
\end{example}

\subsection{{FI in Multi-Class Classification}}\label{subsec:multi_class_FI}
{Consider $C$ classes with input $\BFx \in \mathcal{X}$ and label $y \in \mathcal{Y} = [C]$. 
Let $\boldsymbol p=(p_1,\ldots,p_C)$ denote a classifier,
where $p_i(\BFx)$ is the predicted probability of class $i$.
We adopt a one-versus-one scheme to decompose the task into binary problems, which mitigates dominance by majority classes }\citep{galar2011overview}{. Specifically, for an ordered class pair $(i,j)$, let $\BFx^i$ and $\BFx^j$ denote feature vectors drawn from classes $i$ and $j$, respectively, and define the associated ranking error as $\varepsilon_{i|j}(p_i)=p_i(\BFx^j) - p_i(\BFx^i)$.}

{{Assume that every class is represented in the empirical sample. For each $c\in[C]$, let $\widehat{\mathbb{P}}_c$ denote the empirical distribution of $\BFx$ conditional on $Y=c$. For every ordered pair $(i,j)$ with $i\neq j$, define the nominal distribution $ \widehat{\mathbb Q}_{ij}=\widehat{\mathbb{P}}_i\otimes\widehat{\mathbb{P}}_j \in\mathcal P(\mathcal X^2). $ For each ordered pair, we instantiate the binary FI in Equation \eqref{eq:fi_def} under $\widehat{\mathbb Q}_{ij}$ as
}\begin{equation*}
    \mathrm{FI}_{i|j}(p_i; \tau) = \inf\left\{ k>0 \middle| \mathbb{E}_{\mathbb{Q}}[\varepsilon_{i|j}(p_i)] \leq \tau + k D(\mathbb{Q}, \widehat{\mathbb{Q}}_{ij}),\ \forall \mathbb{Q} \in \mathcal{P}(\mathcal{X}^2) \right\}.
\end{equation*}}

\begin{definition}[Multi-class Fragility Index]
    {Given a target value $\tau$, the multi-class Fragility Index
of $\boldsymbol p$ is defined as
\begin{equation}
    \label{eq:def_multiclass_fi}
    \mathrm{FI}(\boldsymbol p;\tau)
    =
    \frac{1}{C(C-1)}
    \sum_{i\in[C]}
    \sum_{\substack{j\in[C]\\j\neq i}}
    \mathrm{FI}_{i|j}(p_i;\tau).
\end{equation}}
\end{definition}

{
For each pair of classes, the definition includes both ordered comparisons, with $(i,j)$ evaluated using $p_i$ and $(j,i)$ using $p_j$. Their FIs can differ because these probabilities need not be complementary when $C>2$.
We aggregate the resulting $C(C-1)$ pairwise FIs through an equal-weight macro-average.
This summarizes overall ranking fragility while giving comparisons involving minority classes the same aggregation weight as other class pairs, regardless of their sample sizes.
The multi-class FI can be computed by applying the binary characterization in Lemma \ref{lemma:root_finding} separately under each nominal distribution $\widehat{\mathbb Q}_{ij}$ and averaging the resulting $C(C-1)$ pairwise FI values.

Building on the robust satisficing framework
\citep{long2023robust}, we next establish the properties
of the resulting multi-class FI.

\begin{theorem}[Properties of the multi-class FI]
    \label{theorem:multi_fi_properties}
    Let $\boldsymbol p$ and $\widetilde{\boldsymbol p}$ be classifiers with the same $C$ classes.
    The multi-class FI has the following properties.
    \begin{enumerate}[(a)]

        \item (Convexity under probability mixtures)
        For any $\lambda\in(0,1)$,
        $
            \mathrm{FI}\bigl(
                \lambda\boldsymbol p
                +(1-\lambda)\widetilde{\boldsymbol p};
                \tau
            \bigr)
            \leq
            \lambda\,\mathrm{FI}(\boldsymbol p;\tau)
            +(1-\lambda)\,
            \mathrm{FI}(\widetilde{\boldsymbol p};\tau).
        $
        Here, the mixture is taken componentwise
in the predicted probability vectors.

        \item (Prorobustness)
        If
        $
            \tau
            \geq
            \max_{i\neq j}
            \sup_{\mathbb Q\in\mathcal P(\mathcal X^2)}
            \mathbb E_{\mathbb Q}
            [\varepsilon_{i|j}(p_i)],
        $
        then $\mathrm{FI}(\boldsymbol p;\tau)=0$.

        \item (Antifragility)
        If there exists an ordered class pair $(i,j)$
        such that
        $
            \mathbb E_{\widehat{\mathbb Q}_{ij}}
            [\varepsilon_{i|j}(p_i)]>\tau,
        $
        then $\mathrm{FI}(\boldsymbol p;\tau)=+\infty$.

        \item ($\tau$-FI tradeoff)
        For any $\tau_1\geq\tau_2$,
        $
            \mathrm{FI}(\boldsymbol p;\tau_1)
            \leq
            \mathrm{FI}(\boldsymbol p;\tau_2).
        $

        \item (Monotonicity)
        Suppose $D$ is the KL divergence.
        For every ordered class pair $(i,j)$, if $\varepsilon_{i|j}(p_i)$ first-order stochastically dominates $\varepsilon_{i|j}(\widetilde p_i)$ under $\widehat{\mathbb Q}_{ij}$, then
        $
            \mathrm{FI}_{\mathrm{KL}}(\boldsymbol p;\tau)
            \geq
            \mathrm{FI}_{\mathrm{KL}}
            (\widetilde{\boldsymbol p};\tau).
        $
    \end{enumerate}
\end{theorem}

Convexity shows that averaging predicted probability vectors does not increase FI beyond the corresponding average of the constituent classifiers' FIs.
Prorobustness and antifragility identify sufficient conditions for the two extreme values: meeting the worst-case target for every class pair yields zero FI, whereas violating the nominal target for even one pair makes the aggregate FI infinite.
The $\tau$-FI tradeoff clarifies the effect of relaxing the target.
Finally, monotonicity ensures that a classifier with stochastically larger ranking errors for every class pair cannot have a smaller KL-based multi-class FI.

We leave extensions and additional discussions in Appendix \ref{appe:fi}: (i) while we focus on ranking error to align with AUC, the framework applies to other metrics such as 0--1 loss;  (ii) while we adopt a one-versus-one formulation, a one-versus-all formulation is also possible but may introduce class imbalance and obscure risks in minority classes; and (iii) we examine the relationship between confident-misjudgment mitigation and calibration.
}

\section{FI-Based Model Training}

\label{sec:fi_training}

{Having introduced FI as a performance metric, we now turn to model training. We refer to models learned through the proposed framework as \emph{FI-based models}. We establish a formal connection between the optimal training fragility and the ranking-error FI and derive tractable reformulations under common divergences and loss functions (see Table \ref{tab:summary_reformulation}). }

\begin{table}[htbp]
    \centering
    \begin{tabular}{c|c|c}
        \hline
        \textbf{Distance metric }                                              & {\textbf{Distributional support}}                        & \textbf{Convex reformulation results}         \\
        \hline
        KL-divergence                                                 & $\supp{\hat{\mathbb{P}}}$                               & Convex loss (Theorem \ref{theorem:kl_reformulation})                  \\
        \hline
        1-Wasserstein distance                    & {$\mathbb{R}^M\times[C]$ }& \makecell{Cross-entropy loss (Theorem \ref{theorem:cross_entropy_reformulation}) \\
        Hinge-type loss (Theorem \ref{theorem:hinge_type_reformulation}) \\
        Lipschitz loss (Theorem \ref{theorem:lipschitz_approx}) 
        } \\ 
        \hline
    \end{tabular}
    \caption{Summary of the reformulation results for different distance metrics.}
    \label{tab:summary_reformulation}
\end{table}

\subsection{Training via Surrogate Objective}\label{sec:model_learning_framework}
Directly optimizing the ranking-error FI is challenging for two reasons. First, the ranking error $\varepsilon(p)$ is \emph{non-convex} in the model parameters, complicating conventional min-max reformulations. Second, $\varepsilon(p)$ is defined via pairwise comparisons, inducing $\mathcal{O}(N^2)$ complexity for sample size $N$, which limits scalability for large datasets. Similar issues arise in AUC-based and contrastive learning, where surrogate losses are commonly used \citep{yang2021learning, chen2020simple}.

These challenges motivate a tractable loss-based surrogate. Let $\BFx \in \mathcal{X} \subseteq \mathbb{R}^M$ and $y \in \mathcal{Y} = [C]$. {Given training data $\{(\BFx_n,y_n)\}_{n=1}^N$, let $\hat{\mathbb P}$ denote their empirical distribution. The ERM objective is
\begin{equation}
    \label{eq:loss_erm}
    L_{ERM}(\BFB) = \frac{1}{N} \sum_{n \in [N]} \ell(\BFB^T\BFx_n, y_n) + R(\BFB) = \mathbb{E}_{\phat}[\ell(\BFB^T\BFx, y)] + R(\BFB),
\end{equation}
where $\BFB\in \mathbb{R}^{M \times C}$ is the parameter matrix to be learned with each column $\BFbeta_i$ determining the score for class $i$, and $R(\BFB)$ is a pre-specified convex nonnegative regularization term.\endnote{The framework accommodates linear models, sieve approximations, and last-layer neural network fine-tuning.} The model induces probabilities via softmax as\phantomsection\label{loc:softmax_probability_mapping}
{\[
   p_i(\BFx;\BFB) = \frac{\exp(\BFbeta_i^T \BFx)}{\sum_{c\in[C]} \exp(\BFbeta_c^T \BFx)}.
\]}
Instead of minimizing \eqref{eq:loss_erm}, we optimize a robust satisficing surrogate:
\begin{equation}\label{eq:prob_general}
    \min_{k,\BFB}\left\{k\geq 0 \middle| \mathbb{E}_{\mathbb{P}}\!\left[\ell(\BFB^T\BFx,y)\right]+R(\BFB)\leq\tau+kD(\mathbb{P},\hat{\mathbb{P}}),\ \forall\mathbb{P}\in\mathcal{P}(\mathcal X\times\mathcal Y),\ \BFB\in\mathcal B\right\}.
\end{equation}
{Here, $\mathcal{P}(\mathcal{X}\times\mathcal{Y})$ }denotes the set of joint distributions {of }the feature $\BFx \in \mathcal{X}$ and the label $y \in \mathcal{Y}$, {and $D$ is a divergence between joint distributions on $\mathcal{X}\times\mathcal{Y}$.} The target $\tau$ is a constant reference value that regulates the expected {surrogate }loss, and the {convex }feasible region of the weight matrix $\BFB$ is {denoted by $\mathcal B$}.
\phantomsection\label{loc:loss_target_distinction}
{Throughout this section, $\tau$ denotes the target level for the expected surrogate loss in the training framework, which is distinct from the target level for the expected ranking error in Section \ref{sec:FI}.} For notational simplicity, we retain the same symbol $\tau$ for these two distinct targets.} This formulation minimizes fragility with respect to a loss-based surrogate.

\begin{remark}[Regularization] The regularization term $R(\BFB)$ is included to maintain modeling flexibility. While \cite{shafieezadeh2019regularization} connect regularization with distributional robustness, the regularization induced by our formulation (shown later) specifically depends on differences between columns of $\BFB$. Retaining $R(\BFB)$ therefore allows us to decouple structural priors from distributional uncertainty. For example, sparsity can be enforced via an $\ell_1$ penalty while fragility is controlled independently through the robust satisficing constraint.
\end{remark}

\begin{remark}[Target level $\tau$] The choice of $\tau$ is critical. If $\tau$ is too large, the solution becomes trivial ($k^*=0$ with a naive classifier $\BFB = \BFzero$). To avoid this, we impose:
\begin{assumption}[Nontriviality of $\tau$]
    \label{asmp:tau}
    {There exist $\delta_1 > 0, \delta_2 > 0$ such that the target $\tau$ satisfies 
    $$
        \tau \leq \inf_{\BFB\in \mathcal{B}} \sup_{\mathbb{P} \in \mathcal{P}({\mathcal{X}\times\mathcal{Y}}), \ D (\mathbb{P},\hat{\mathbb{P}}) \leq \delta_2} \mathbb{E}_{\mathbb{P}} \left[\ell(\BFB^T\BFx, y) \right] + R(\BFB) - \delta_1.
    $$}
\end{assumption}

The right-hand side is a DRO problem. This ensures nontrivial solutions that are instrumental in the subsequent reformulation and analysis:
\begin{lemma}
    \label{lemma:lower_bound_k}
    Under Assumption \ref{asmp:tau}, the optimal $k^*$ of Problem \eqref{eq:prob_general} satisfies
    $ k^* \geq \frac{\delta_1}{\delta_2} > 0. $
\end{lemma}

\end{remark}

\subsection{{Connection between the Surrogate Objective and Ranking-Error FI}}
\label{sec:training_fragility_fi_connection}

{To relate the surrogate objective in Problem \eqref{eq:prob_general}
to the ranking-error FIs defined in Equations \eqref{eq:fi_def}
and \eqref{eq:def_multiclass_fi}, we impose a condition that relates the predicted true-class probability to the training loss.
}\begin{assumption}[Probability--loss dominance]
    \label{asmp:probability_loss_dominance}
    {{There exist constants $a>0$ and $b\geq0$, which may depend on }the loss function {and the number of classes but not on $\BFB$, $\BFx$, or $y$, such that
    }\begin{equation*}
        1-p_y(\BFx;\BFB)
        \leq
        a\ell(\BFB^T\BFx,y)+b,
        \qquad
        \forall \BFB\in\mathcal B,\ \BFx\in\mathcal X,\ y\in[C].
\end{equation*}}
\end{assumption}

{Assumption \ref{asmp:probability_loss_dominance} holds for commonly used losses, including cross-entropy loss and the multi-class hinge loss (Appendix \ref{appe:fi_training}); in particular, cross-entropy loss satisfies it with $a=1$ and $b=0$.}

\begin{theorem}    \label{theorem:kl_wasserstein_ovo_bound}
    {{For each $c\in[C]$, let $q_c=\hat{\mathbb P}(Y=c)>0$. Under Assumptions \ref{asmp:tau} and \ref{asmp:probability_loss_dominance}, let $(k,\BFB)$ be any feasible pair for Problem \eqref{eq:prob_general}, with $D$ given by either the KL divergence or the Wasserstein distance. Define
    }{\[
        t_{ij}(\BFB)
        =
        b
        +\frac{a}{q_i+q_j}
        \left[
        \tau-R(\BFB)
        -\sum_{c\in[C]\setminus\{i,j\}}q_c
        \mathbb E_{\widehat{\mathbb{P}}_c}[\ell(\BFB^T\BFx,c)]
        \right],
        \qquad
        t_{\max}(\BFB)
        =
        \max_{\substack{i,j\in[C]\\i\neq j}}t_{ij}(\BFB).
\]}
    {Then, for every $i\neq j$,
    }\begin{equation*}
        \mathrm{FI}_{i|j}
        \left(p_i(\cdot;\BFB);t_{ij}(\BFB)\right)
        \leq
        \frac{a\max\{q_i,q_j\}}{q_i+q_j}k.
\end{equation*}
    {Consequently,
    }\begin{equation}
        \label{eq:kl_wasserstein_ovo_bound}
        \mathrm{FI}
        \left(
        p_1(\cdot;\BFB),\ldots,p_C(\cdot;\BFB);
        t_{\max}(\BFB)
        \right)
        \leq
        \frac{ak}{C(C-1)}
        \sum_{\substack{i,j\in[C]\\i\neq j}}
        \frac{\max\{q_i,q_j\}}{q_i+q_j}
        \leq ak.
\end{equation}}
\end{theorem}

{
Theorem \ref{theorem:kl_wasserstein_ovo_bound} formally establishes the connection between the training fragility $k$ and the ranking-error FI. Specifically, for any feasible pair $(k,\BFB)$ of Problem \eqref{eq:prob_general}, the resulting classifier $\BFB$ has a bounded ranking-error FI under the target $t_{ij}(\BFB)$. The bound is proportional to $k$, the objective minimized in the training framework. Therefore,  solving the surrogate objective \eqref{eq:prob_general} yields an  optimal training fragility $k^*$ that, up to this proportionality constant, upper bounds  the ranking-error FI of the learned classifier with parameters $\BFB^*$.

We now discuss the bound coefficients. In binary classification, $q_1+q_2=1$ and there are no classes outside the selected pair. Hence, we have $\mathrm{FI}( p(\cdot;\BFB);b+a[\tau-R(\BFB)] )\leq a\max\{q_1,q_2\}k$.
In particular, for balanced classes and cross-entropy loss, we have $\mathrm{FI}(p(\cdot;\BFB);\tau-R(\BFB)) \leq k/2$.
In multi-class classification, the specific target $t_{ij}(\BFB)$ depends on the nominal losses of all other classes. The common target $t_{\max}(\BFB)$ further aggregates these pairwise guarantees into the macro-average bound in \eqref{eq:kl_wasserstein_ovo_bound}. Moreover, the coefficient $\max\{q_i,q_j\}/(q_i+q_j)$ always lies in $[1/2,1)$, so remains bounded under class imbalance. }

\subsection{Training under Kullback-Leibler Divergence}\label{sec:training_kl}
We instantiate the framework under the KL-divergence in \eqref{eq:kl_def}, setting $D(\mathbb{P}, \hat{\mathbb{P}}) = D_{\mathrm{KL}} (\mathbb{P} || \phat)$. Since the empirical distribution $\hat{\mathbb{P}}$ is discrete with finite support, the absolute continuity condition $\mathbb{P} \ll \hat{\mathbb{P}}$ is equivalent to $\mathrm{supp} (\mathbb{P}) \subseteq \mathrm{supp} (\hat{\mathbb{P}})$. Accordingly, we restrict $\mathbb{P}$ to $\mathcal{P} (\mathrm{supp} (\hat{\mathbb{P}}))$. Problem \eqref{eq:prob_general} then becomes
\begin{equation}\label{eq:prob_general_kl}
    \min_{k,\BFB}\left\{k\geq 0 \middle| \mathbb{E}_{\mathbb{P}}\!\left[\ell(\BFB^T\BFx,y)\right]+R(\BFB)\leq\tau+kD_{\mathrm{KL}}(\mathbb{P}\,\|\,\phat),\ \forall\mathbb{P}\in\mathcal{P}(\mathrm{supp}(\hat{\mathbb{P}})),\ \BFB\in\mathcal B\right\}.
\end{equation}

We provide a closed-form reformulation below.
\begin{theorem}
    \label{theorem:kl_reformulation}
    {Under Assumption \ref{asmp:tau}, Problem \eqref{eq:prob_general_kl} is equivalent to
    \begin{equation*}
        \min_{k,\BFB}\left\{k>0 \middle| k\ln\!\left(\mathbb{E}_{\phat}\!\left[\exp\!\left(\frac{\ell(\BFB^T\BFx,y)}{k}\right)\right]\right)+R(\BFB)-\tau\leq 0,\ \BFB\in\mathcal B\right\}.
    \end{equation*}}
    The problem is convex if the loss function $\ell(\BFB^T\BFx, y)$ is convex in $\BFB$ for any $y\in\mathcal{Y}$. 
\end{theorem}
This reformulation highlights a loss reweighting mechanism. Samples with a larger loss receive exponentially higher weight in the worst-case distribution \citep{donsker1975asymptotic}. As a result, the model emphasizes high-loss samples, improving robustness to misjudgment. This behavior is reminiscent of boosting, which prioritizes hard examples \citep{vapnik2013nature}. From a computational perspective, when $\ell$ is convex, the problem is convex and can be solved efficiently.

\subsection{Training under Wasserstein Distance}
\label{sec:training_wass}

{We now instantiate the framework under the Wasserstein distance, defined via the optimal transport~(OT) discrepancy between two distributions $\mathbb{P}$ and $\hat{\mathbb{P}}$:
\begin{equation}
    \label{eq:def_ot}
    W(\mathbb{P}, \hat{\mathbb{P}}) = \inf_{\pi \in \Pi(\mathbb{P}, \hat{\mathbb{P}})} \mathbb{E}_{\pi} [c(\BFx, y, \hat{\BFx}, \hat{y})],
\end{equation}
where $c(\cdot)$ is the transport cost, and $\Pi(\mathbb{P}, \hat{\mathbb{P}})$ denotes the set of couplings with marginals $\mathbb{P}$ and $\hat{\mathbb{P}}$. The corresponding optimization problem is
\begin{equation*}
    \min_{k,\BFB}\left\{k\geq 0 \middle| \mathbb{E}_{\mathbb{P}}\!\left[\ell(\BFB^T\BFx,y)\right]+R(\BFB)\leq\tau+kW(\mathbb{P},\hat{\mathbb{P}}),\ \forall\mathbb{P}\in\mathcal{P}(\mathcal X\times\mathcal Y),\ \BFB\in\mathcal B\right\}.
\end{equation*}}

Compared to KL divergence, Wasserstein ambiguity allows shifts to unseen samples but introduces additional computational challenges. In particular, convex reformulation is nontrivial due to the maximization of the convex conjugate of the cost function, which can induce nonconvexity \citep{shafieezadeh2023nash}.\endnote{Convex reformulation is possible when $\ell(\BFB^T\BFx, y)$ is convex in $\BFB$ and concave in $\BFx$; otherwise, a nonconvex counterpart may arise \citep{shafieezadeh2023nash}.}

{We focus on the commonly used 1-Wasserstein distance with cost
\begin{equation}
    \label{eq:cost_1wass}
    c(\BFx, y, \hat{\BFx}, \hat{y}) = \|\BFx - \hat{\BFx}\| + \gamma \mathbb{I}(y \neq \hat{y}),
\end{equation}
where {$\gamma\geq 0$ }controls the relative cost of label versus feature perturbations and {may be selected using domain knowledge or }cross-validation \citep{shafieezadeh2015distributionally}.}

\begin{lemma}
    \label{lemma:1wass_reformulation}
    Suppose the loss function $\ell(\BFu, y)$ is convex in $\BFu$ for every $y \in \mathcal{Y}$ and Assumption \ref{asmp:tau} holds. {Consider the Wasserstein distance in {Equation }\eqref{eq:def_ot} with cost $c(\BFx, y, \hat{\BFx}, \hat{y})$ specified in {Equation \eqref{eq:cost_1wass} and }support $\mathcal{X} = \mathbb{R}^M$. Problem \eqref{eq:prob_general} is equivalent to
    \begin{equation}
        \label{eq:1wass_reformulation}
        \begin{aligned}
            \min_{k\geq0, \BFB \in \mathcal{B}} & \ k \\
            \text{s.t.} \hspace*{10pt} & \frac{1}{N} \sum_{n \in [N]} \max_{y_n \in \mathcal{Y}} \{ \ell(\BFB^T\hat{\BFx}_n, y_n) - k \gamma \mathbb{I}(y_n \neq \hat{y}_n)\} + R(\BFB) - \tau \leq 0,\\
            & \sup_{\substack{y\in\mathcal Y\\
            \BFzeta \in \dom{\ell^{1*}(\cdot,y)}}}
            \|\BFB\BFzeta\|_* \leq k.
        \end{aligned}
\end{equation}}
\end{lemma}

{The constraint {$\sup_{\substack{y\in\mathcal Y,\ \BFzeta\in\dom{\ell^{1*}(\cdot,y)}}} \|\BFB \BFzeta\|_* \leq k$ }connects our framework to sharpness-aware minimization \citep{foret2021sharpness}, as minimizing $k$ controls the worst-case gradient norm and limits how rapidly the loss can increase under
feature perturbations, which is known to improve generalization \citep{andriushchenko2022towards}.}

However, although this constraint is convex in \((\BFB,k)\), evaluating its supremum generally involves a nonconvex inner maximization problem.
{A tractable reformulation arises when the label-specific domains $\mathrm{dom}(\ell^{1*}(\cdot,y))$, $y\in\mathcal Y$, have finitely many extreme points, allowing reduction to finitely many constraints.
The following results exploit the structure of cross-entropy, hinge-type, and Lipschitz-continuous losses to obtain tractable formulations.}

\subsubsection{Cross-Entropy Loss}\label{subsubsec:cross_entropy_loss}
The cross-entropy loss for the $n$-th sample is defined as
\begin{equation}
    \label{eq:cross_entropy_loss}
    \ell_{CE}(\BFB^T\hat{\BFx}_n, \hat{y}_n) 
    = \ln \left(\sum_{i \in [C]} \exp(\BFbeta_i^T \hat{\BFx}_n)\right) - \BFe_{\hat{y}_n}^T \BFB^T \hat{\BFx}_n,
\end{equation}
where $\BFe_{\hat{y}_n}$ is the one-hot vector of the label $\hat{y}_n$. 
\begin{theorem}
    \label{theorem:cross_entropy_reformulation}
    {Supposing the loss function $\ell$ is the cross-entropy loss in Equation \eqref{eq:cross_entropy_loss}, the problem \eqref{eq:1wass_reformulation} {admits the following conservative approximation:
    }\begin{equation}
        \label{eq:cross_entropy_reformulation}
        \begin{aligned}
            \min_{k\geq0, \BFB \in \mathcal{B}} & \ k \\
            \text{s.t.} \hspace*{10pt} & \frac{1}{N} \sum_{n \in [N]} \left[\ell(\BFB^T\hat{\BFx}_n, \hat{y}_n) + k (\|\hat{\BFx}_n\| - \gamma)_+\right] + R(\BFB) - \tau \leq 0,\\
            & \|\BFbeta_i - \BFbeta_j\|_* \leq k, \ \forall i, j \in [C] \ \text{and}\ i < j.
        \end{aligned}
\end{equation}}
    Any solution of the problem \eqref{eq:cross_entropy_reformulation} is feasible for the problem \eqref{eq:1wass_reformulation}. When $\gamma \geq \max_{n\in[N]} \|\hat{\BFx}_n\|$, the problem \eqref{eq:cross_entropy_reformulation} is exactly equivalent to the problem \eqref{eq:1wass_reformulation}.

\end{theorem}

{
Theorem \ref{theorem:cross_entropy_reformulation}
yields a compact exact reformulation when
$\gamma\geq\max_{n\in[N]}\|\hat{\BFx}_n\|$.
In this regime, label perturbations are relatively more
costly than feature perturbations, and the term
$k(\|\hat{\BFx}_n\|-\gamma)_+$ vanishes,
further simplifying the formulation.
For arbitrary $\gamma\geq0$,
Proposition \ref{prop:cross_entropy_exact_all_gamma}
provides an exact finite convex formulation with
$NC$ additional label-wise inequalities and the objective gaps and computational costs are compared. At or above the sufficient exactness threshold, the compact formulation \eqref{eq:cross_entropy_reformulation} preserves exactness while avoiding the label-wise constraints.
To the best of our knowledge, this general formulation
is the first exact reformulation for multiclass
cross-entropy under the Wasserstein ambiguity considered
here, in contrast to existing relaxed formulations
\citep{chen2023distributionally}.
}

The formulation in Theorem \ref{theorem:cross_entropy_reformulation} also has several useful implications. First, cross-entropy loss is invariant to a common shift $\BFdelta$ applied to all weight vectors $\BFbeta_i$, so the absolute norm of $\BFB$ is not meaningful. Accordingly, the constraints in \eqref{eq:cross_entropy_reformulation} regulate only the pairwise differences $\|\BFbeta_i-\BFbeta_j\|_*$. Second, these pairwise constraints effectively bound the loss induced by misclassification between any two classes. {This is particularly relevant {to }label-flipping attacks \citep{cina2023wild}, where an adversary injects label noise {into the training data. Let $\BFy^{\mathrm{true}}\in\mathcal{Y}^N$ }denote the true {underlying labels and let $\hat{\BFy}\in\mathcal{Y}^N$ denote the corrupted empirical labels used for training}. The resulting {clean-to-corrupted empirical loss gap of the fitted model is }controlled by the optimal fragility $k^*$, as shown below.}

\begin{corollary}
    \label{coro:flipping_label_attack}
    {{
    Let $(k^*,\BFB^*)$ be an }optimal solution of {Problem \eqref{eq:cross_entropy_reformulation} trained on the corrupted empirical data $(\hat{\BFx}_n,\hat{y}_n)_{n\in[N]}$, and let $\phi = \max_{n\in[N]}\|\hat{\BFx}_n\|$. Define $N_{\mathrm{flip}} = \sum_{n\in[N]} \mathbb{I}(y_n^{\mathrm{true}} \neq \hat{y}_n)$. We use $L_{CE}(\BFB;\BFz)$ to denote the empirical cross-entropy loss of $\BFB$ evaluated on the fixed feature sample with label vector $\BFz$. }Then, the loss deviation {between the true and corrupted labels satisfies
    }{$$
        \left|L_{CE}(\BFB^*; \BFy^{\mathrm{true}}) - L_{CE}(\BFB^*; \hat{\BFy})\right| \leq \phi k^* \frac{N_{\mathrm{flip}}}{N}.
$$}}
\end{corollary} 

{Corollary \ref{coro:flipping_label_attack} {bounds the absolute empirical loss deviation between the true and corrupted label evaluations by the product of the optimal pairwise sensitivity }$k^*${, the maximum feature norm, and the fraction of flipped labels}.}

\subsubsection{Hinge-Type Loss}
We follow \cite{glasmachers2016unified} and define the hinge-type losses for multi-class classification as follows:
\begin{equation}
    \label{eq:hinge_type_loss}
    \ell_{hinge}(\BFB^T\BFx, y) = \max_{y' \neq y} \rho((\BFbeta_{y} - \BFbeta_{y'})^T \BFx),
\end{equation}
where $(\BFbeta_{y} - \BFbeta_{y'})^T \BFx$ refers to the relative margin between classes; $\rho$ is a binary hinge-type loss, such as the hinge loss $\rho(u) = \max\{0, 1 - u\}$, the logistic loss $\rho(u) = \log(1 + e^{-u})$, and the smoothed hinge loss \citep{luo2021learning}. 

\begin{theorem}
    \label{theorem:hinge_type_reformulation}
    Consider the hinge-type loss function in the equation \eqref{eq:hinge_type_loss}. {If function $\rho$ is convex and subdifferentiable, and {for some $\theta>0$, }$\sup_{u\in\mathbb{R}} \partial \rho(u) = 0$ and $\inf_{u\in\mathbb{R}} \partial \rho(u) = - \theta$, the problem \eqref{eq:1wass_reformulation} {admits the following conservative approximation:
    }\begin{equation}
        \label{eq:hinge_type_reformulation}
        \begin{aligned}
            \min_{k\geq0, \BFB \in \mathcal{B}} & \ k \\
            \text{s.t.} \hspace*{10pt} & \frac{1}{N} \sum_{n \in [N]} \left[\ell(\BFB^T\hat{\BFx}_n, \hat{y}_n) + k (2 \|\hat{\BFx}_n\| - \gamma)_+\right] + R(\BFB) - \tau \leq 0,\\
            & \|\BFbeta_i - \BFbeta_j\|_* \leq \frac{k}{\theta}, \ \forall i, j \in [C] \ \text{and}\ i < j.
        \end{aligned}
\end{equation}}
    Any solution to the problem \eqref{eq:hinge_type_reformulation} is feasible for the problem \eqref{eq:1wass_reformulation}. When $\gamma \geq 2\max_{n\in[N]} \|\hat{\BFx}_n\|$, the problem \eqref{eq:hinge_type_reformulation} is exactly equivalent to the problem \eqref{eq:1wass_reformulation}.
\end{theorem}

{ The formulation depends only on the closed subgradient range of $\rho$ and therefore remains unchanged under smoothing or piecewise linear approximations that preserve the range $[-\theta,0]$. Similar to the cross-entropy case, the pairwise constraints $\|\BFbeta_i - \BFbeta_j\|_* \leq k/\theta$ ensure that the model is robust to label noise. Regarding the exact reformulation, Proposition \ref{prop:hinge_exact_all_gamma} provides a formulation with $NC(C-1)$ convex inequalities. Therefore, similar to the cross-entropy case, it is also preferable to solve the conservative approximation \eqref{eq:hinge_type_reformulation} when $\gamma$ is comparable to the feature norms $2\|\hat{\BFx}_n\|$.}

\subsubsection{{Lipschitz-Continuous Loss}}
\label{paragraph:Lipschitz_continuous_loss}
We move on to Lipschitz-continuous loss functions. Formally, we consider functions satisfying the following assumption.
\begin{assumption}[Lipschitz-continuous Loss Functions]
    \label{asmp:lipschitz}
    There {exist $\omega_1>0$ and $\omega_2 \geq 0$ }such that the loss function $\ell(\BFu, y)$ satisfies\endnote{Notice that the vector norm $\|\cdot\|$ in Assumption \ref{asmp:lipschitz} is in the same order as the norm in $c(\BFx, y, \hat{\BFx}, \hat{y})$. }:
    \begin{align*}
        & |\ell(\BFu_1, y) - \ell(\BFu_2, y)| \leq \omega_1 \|\BFu_1 - \BFu_2\|, &\forall \BFu_1, \BFu_2 \in {\mathbb{R}^C}, y \in \mathcal{Y}, \\
        & |\ell(\BFu, y_1) - \ell(\BFu, y_2)| \leq \omega_2 \|\BFu\|, &\forall \BFu \in {\mathbb{R}^C}, y_1, y_2 \in \mathcal{Y}.
\end{align*}
\end{assumption}
{We consider the matrix norm induced by this vector norm as $ \|\BFB\|_* :=\sup_{\|\BFzeta\|_*\leq1}\|\BFB\BFzeta\|_* $ }. The following states the reformulation result for Lipschitz-continuous loss functions.
\begin{theorem}
    Suppose\label{theorem:lipschitz_approx} Assumption \ref{asmp:lipschitz} holds, and $\ell(\BFu, y)$ is convex and subdifferentiable in $\BFu$ for any $y\in\mathcal{Y}$. Then, the problem \eqref{eq:1wass_reformulation} can be approximated by the following problem
    \begin{equation}
        \label{eq:lipschitz_reformulation}
        \begin{aligned}
            \min_{k\geq0, \BFB \in \mathcal{B}} & \ k \\
            \text{s.t.} \hspace*{10pt}
            & \frac{1}{N}\sum_{n\in[N]}
            \left[
                \ell(\BFB^T\hat{\BFx}_n,\hat y_n)
                +k\left(
                    \frac{\omega_2}{\omega_1}\|\hat{\BFx}_n\|-\gamma
                \right)_+
            \right]
            +R(\BFB)-\tau\leq0,\\
            &  \|\BFB\|_* \leq \frac{k}{\omega_1}.
        \end{aligned}
\end{equation}
    {Every feasible pair for Problem }\eqref{eq:lipschitz_reformulation} is feasible for {Problem }\eqref{eq:1wass_reformulation}. Moreover, the {two problems are exactly equivalent }when the following {sufficient }conditions hold:
    \begin{enumerate}[(a)]
        \item $\gamma \geq \max_{n\in[N]} \frac{\omega_2}{\omega_1}\|\hat{\BFx}_n\|$,
        \item {For any $\BFv \in \mathcal{V} := \{\BFv \in \mathbb{R}^C \colon \|\BFv\|_* = 1\}$}, there exist {$y \in \mathcal{Y}$ and $\BFu \in \dom{\ell(\cdot,y)}$} such that {$\BFv \in \frac{1}{\omega_1} \partial_{\BFu} \ell(\BFu, y)$}.
    \end{enumerate}
\end{theorem}

{The sufficient equivalence conditions, particularly condition (b), may be difficult to verify. For example, the 2-norm regression loss function $\ell(\BFu, \BFy) = \|\BFu - \BFy\|_2$ is symmetric with respect to the rotation around $\BFy$, which ensures condition (b) can hold. Without these conditions, Problem \eqref{eq:lipschitz_reformulation} remains a tractable conservative reformulation of Problem \eqref{eq:1wass_reformulation}. }

For the Euclidean norm, the formulation in Theorem \ref{theorem:lipschitz_approx} can be conservatively approximated using the Frobenius norm. The matrix 2-norm satisfies the inequality of $\|\BFB\|_2 \leq \|\BFB\|_F$, where $\|\BFB\|_F = \sqrt{\sum_{i \in [M]} \sum_{j \in [C]} B_{ij}^2}$ is the Frobenius norm. We therefore can derive a convex approximation of the problem \eqref{eq:lipschitz_reformulation} by replacing the matrix 2-norm with the Frobenius norm in Theorem \ref{theorem:lipschitz_approx}. Since the Frobenius norm is widely used as a regularization term in neural network training \citep{tian2022comprehensive}, this result provides a theoretical foundation for its application in modern neural network optimization.

{
To complete the discussion of FI-based training, we provide several additional analyses and extensions in Appendix \ref{appe:fi_training}. In particular, 
for cross-entropy loss, we refine the bound linking training fragility to ranking-error FI and numerically assess its tightness; we also establish finite-sample generalization bounds and exponential convergence guarantees for the learned parameters; we also discuss extensions to more general transport costs and connections to DRO.
}

\section{Case Study: Heart Failure Prediction}
\label{sec:numerical_experiment}

In this section, we empirically evaluate the proposed FI metric and FI-based learning framework, focusing on real-world data and operational implications. {We consider a heart failure prediction task {using the Heart Failure Prediction dataset }\citep{fedesoriano2021heart}, along with a simulated human--AI diagnostic workflow with manual review to illustrate downstream decision impact.} 
Complementary results on synthetic data and additional real-world datasets, together with implementation details for the baselines, are provided in Appendix \ref{appe:experiment}.

Our results highlight the diagnostic value of FI in distinguishing models by their risk of confident misjudgments. {{The }FI-based models {remain competitive in }accuracy and AUC {while generally }attaining lower FI {under label noise. In the operational-cost experiment, both FI-based models also yield lower average operational cost than ERM throughout the considered cost configurations}.}

\subsection{Setup}

\subsubsection{Data}
We use the Heart Failure Prediction dataset \citep{fedesoriano2021heart}, a safety-critical setting where risk control is essential. The task is to predict the presence of heart disease from patient attributes (e.g., age, sex, blood pressure). The dataset contains 918 samples with a positive-class ratio of 0.55. After one-hot encoding categorical variables, the feature dimension is 20. 

We formulate the task as binary classification and randomly split the data into equal-sized training and testing sets. To simulate distributional shift, we introduce label noise by flipping training labels with probability $p_{flip}$. The testing set remains clean, so $p_{flip}$ controls the degree of mismatch between training and deployment environments.

\subsubsection{Compared Methods}
We use ERM with hinge loss \eqref{eq:hinge_type_loss} as the baseline. The regularization $R(\BFB)$ is selected via 5-fold cross-validation, considering both $\ell_1$ ($R_1(\BFB)=\alpha\sum_{ij}|B_{ij}|$) and $\ell_2$ ($R_2(\BFB)=\alpha\sum_{ij}B_{ij}^2$) penalties with $\alpha \in \{10^{-4}, 10^{-3}, 10^{-2}, 10^{-1}\}$. {Let $\widehat{\BFB}_{\mathrm{ERM}}$ denote the fitted ERM solution and define $\hat L_{\mathrm{ERM}}:=L_{\mathrm{ERM}}(\widehat{\BFB}_{\mathrm{ERM}})$, where $L_{\mathrm{ERM}}$ is defined in Equation \eqref{eq:loss_erm}.
}

For comparability, FI-based models adopt the same loss and regularization settings. We consider two variants: one under KL divergence (Theorem \ref{theorem:kl_reformulation}) and one under Wasserstein distance (Theorem \ref{theorem:hinge_type_reformulation}). The {FI-based models are trained using the loss target }$\tau = \lambda \hat{L}_{\mathrm{ERM}}$, where $\lambda \in \{1.05, 1.1, 1.15\}$ is chosen via cross-validation. {A larger }$\lambda$ {relaxes this training-loss target and thereby allows a smaller training fragility}.

We also include XGBoost \citep{chen2016xgboost} as a nonlinear baseline. In addition, we consider {robust training baselines, including} trimmed loss \citep{shen2019learning}, generalized cross-entropy (GCE) \citep{zhang2018generalized}{, }symmetric cross-entropy (SCE) \citep{wang2019symmetric}{, }ERM with data purification \citep{huang2019o2u}{, mixup }\citep{zhang2017mixup}, and label smoothing \citep{szegedy2016rethinking}.

\subsubsection{Evaluation Metrics}
\phantomsection\label{loc:heart_fi_evaluation_target}
We report averages over 100 independent trials with random data splits. Performance is evaluated on the testing set using accuracy, AUC, and FI. {All reported FI values are computed post-training using the KL-based ranking-error FI with target $0$.
}

\subsection{Model Training Results}\label{num:experiment_real_result}
\begin{figure}[htbp]
    \centering
    \subfloat[Accuracy]{\includegraphics[width=0.34\textwidth]{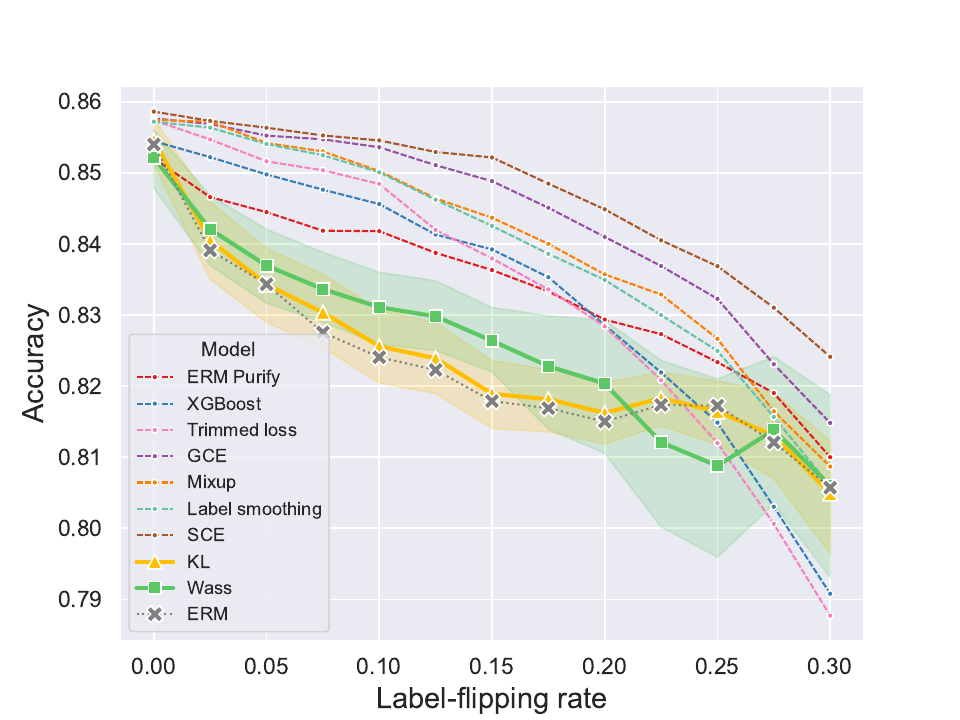}}
    \subfloat[AUC]{\includegraphics[width=0.34\textwidth]{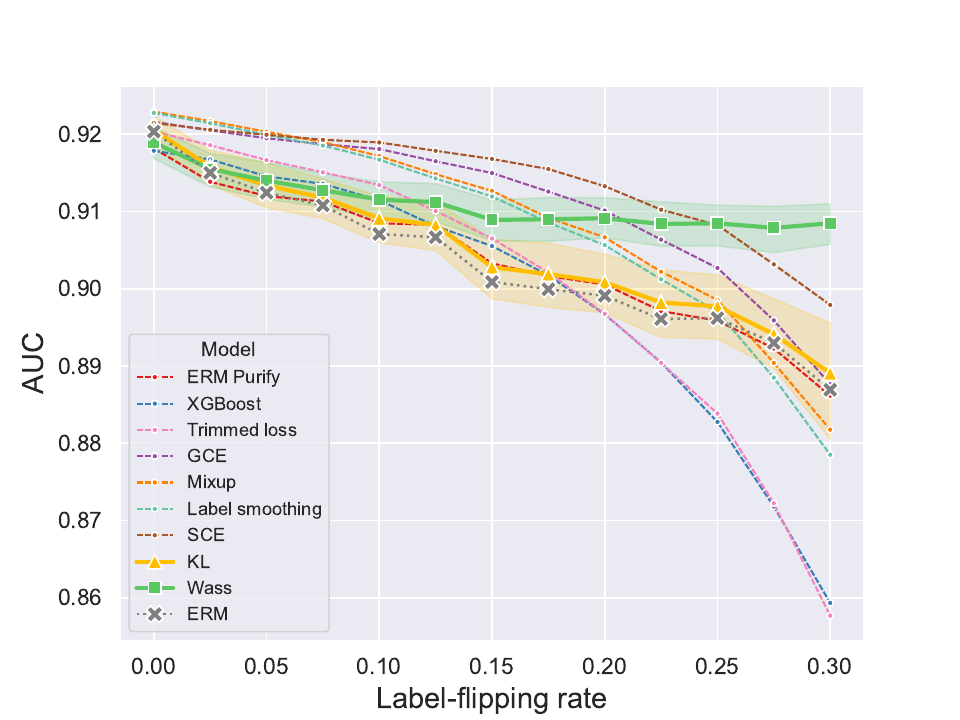}}
    \subfloat[FI]{\includegraphics[width=0.34\textwidth]{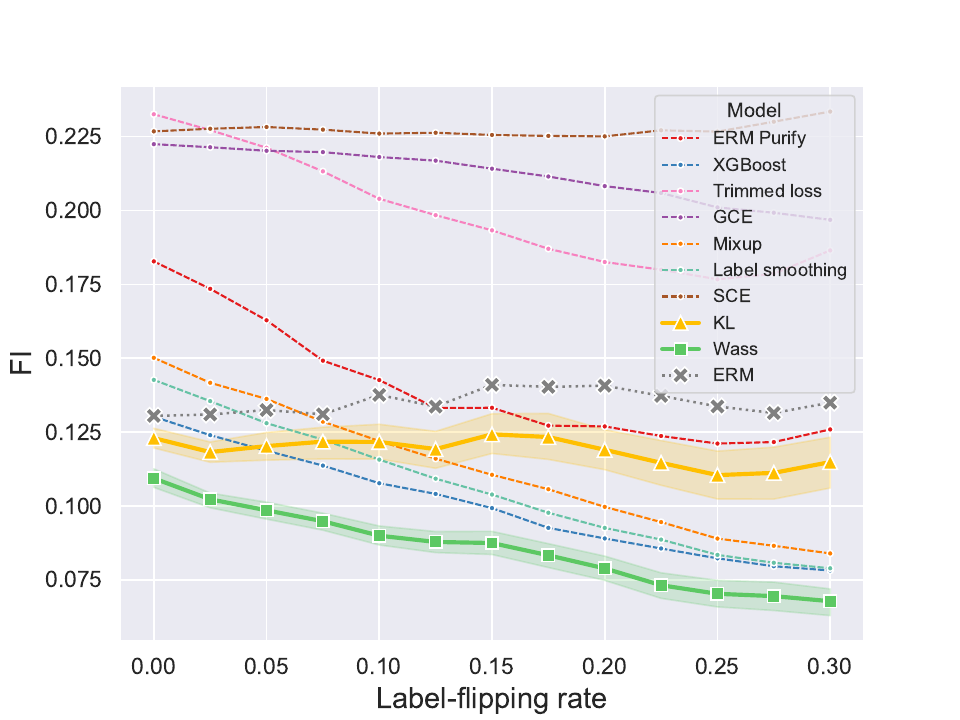}}
    \caption{The average accuracy, AUC, and FI on the heart failure prediction dataset. The error bands are calculated by $95\%$ confidence intervals. }
    \label{fig:heartattack}
\end{figure}

Figure \ref{fig:heartattack} reports model performance in terms of accuracy, AUC, and FI. Higher values indicate better performance for accuracy and AUC, while lower values are preferred for FI.
We highlight three main findings. First, FI-based models {achieve accuracy and AUC comparable to ERM while generally attaining lower FI}. Second, the Wasserstein FI-based model consistently achieves the lowest FI, indicating the strongest control of fragility. Third, the Wasserstein FI-based model maintains relatively stable AUC as the noise level $p_{flip}$ increases and achieves the highest AUC at higher noise levels. {In our multi-class experiments, however, the Wasserstein FI-based model can become overly conservative and underfit the data at high label-noise levels (see Appendix~\ref{appe:experiment_synthetic_multi_class}).}

\begin{figure}[htbp]
    \centering
    \subfloat[Ranking error]{\includegraphics[width=0.34\textwidth]{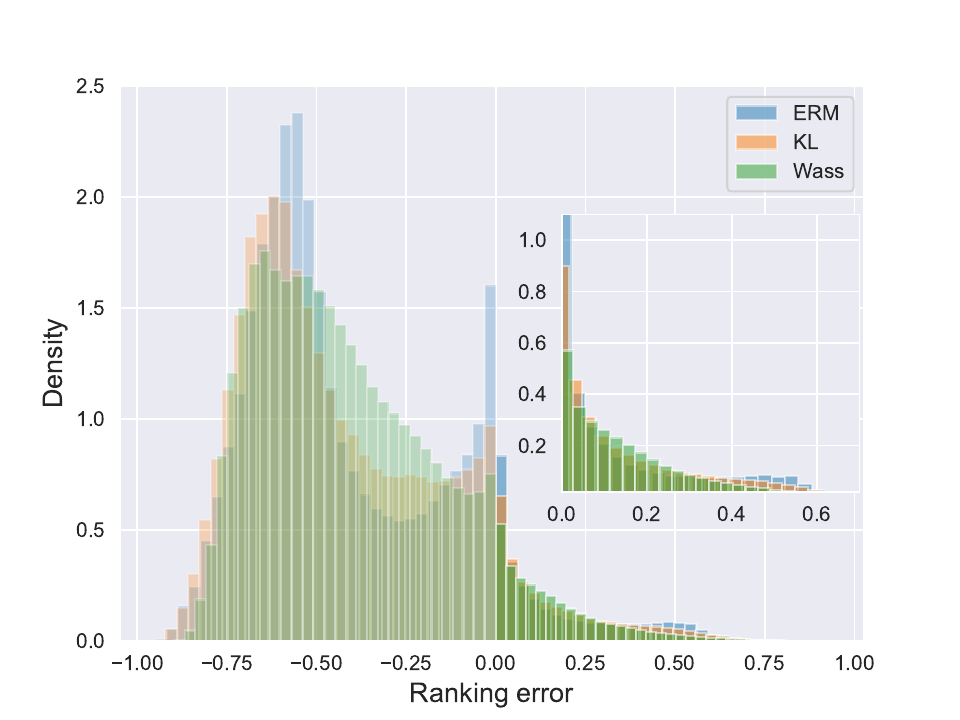}}
    \subfloat[Confidence of false predictions]{\includegraphics[width=0.34\textwidth]{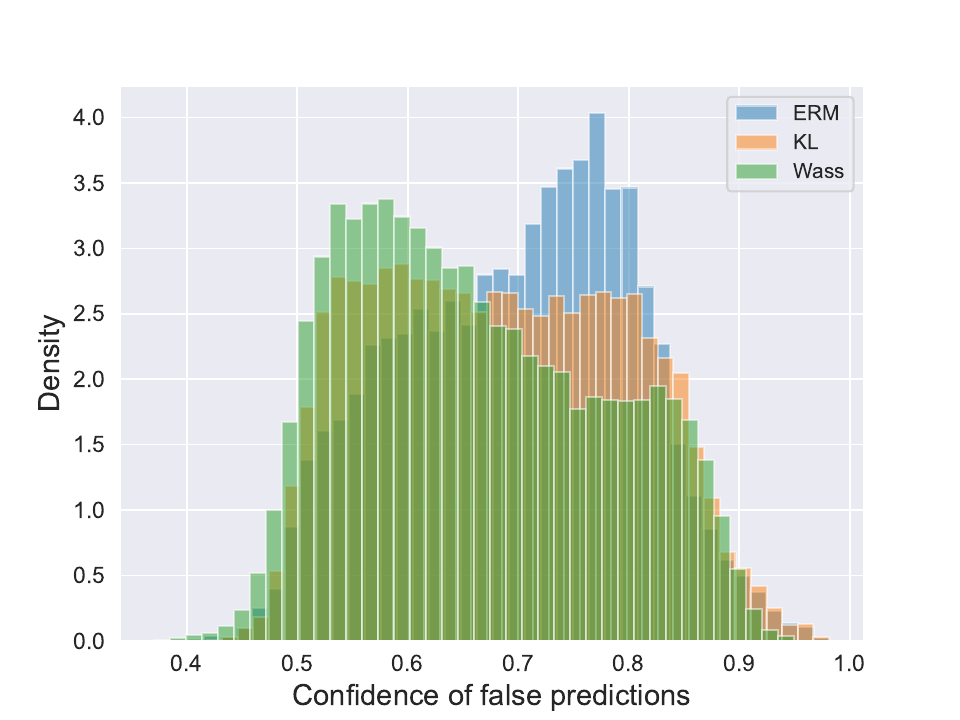}}
    \subfloat[Confidence of true predictions]{\includegraphics[width=0.34\textwidth]{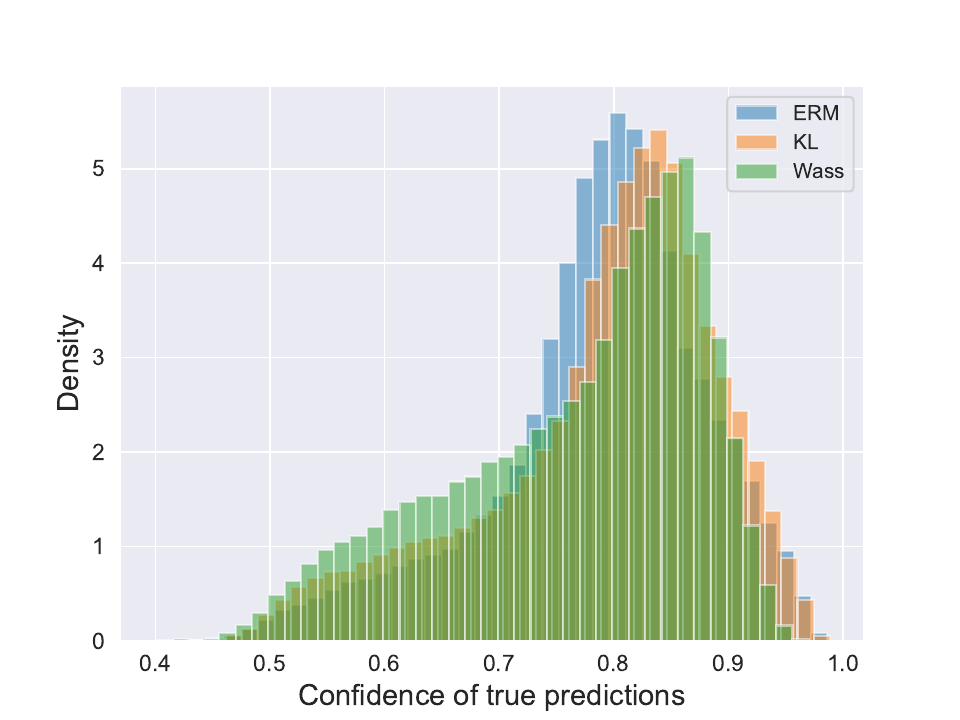}}
    \caption{The ranking error and classifiers' estimated probability of the three models on the heart failure prediction dataset when $p_{flip} = 0.1$. (a) Ranking errors: focus on the positive region (error $> 0$) and a lighter tail indicates fewer confident misjudgments. (b) Confidence of false predictions: smaller is better. (c) Confidence of true predictions: larger is better.}
    \label{fig:heartattack_error_confidence}
\end{figure}

To better understand how FI reflects the risk of misjudgment, we examine the distributions of ranking errors and confidences on the predictions, using Platt scaling for calibration \citep{platt1999probabilistic}. {Here, confidence refers to the predicted probability assigned to the predicted class. At $p_{flip}=0.1$, Figure \ref{fig:heartattack_error_confidence} compares ERM, the KL-based model, and the Wasserstein-based model, which have similar accuracy and AUC but distinct FI values and are ordered by decreasing FI.}

Figure \ref{fig:heartattack_error_confidence}(a) shows that {ERM places more mass in the large positive tail, whereas the two FI-based models reduce the occurrence of large positive ranking errors}. Figures \ref{fig:heartattack_error_confidence}(b){--}(c) show a consistent pattern: the Wasserstein model {tends to assign }lower confidence to incorrect predictions and higher confidence to correct ones, indicating better separation. ERM exhibits the opposite pattern, with the KL model in between. {The ordering of these distributions is consistent with the FI ordering in Figure \ref{fig:heartattack}(c), confirming that lower FI corresponds to lower risk of confident misjudgment.}

From a managerial perspective, these results suggest that FI provides a practical criterion for model selection in safety-critical settings. Even when models achieve similar accuracy or AUC, {a lower FI identifies models that }are less prone to highly confident mistakes{. The associated operational implications are examined in Section \ref{sec:cost_analysis}. }

\subsection{{Operational Cost Analysis in a {Human--AI Diagnostic Workflow}}}
\label{sec:cost_analysis}

{We consider a human–AI diagnostic workflow, where a classifier estimates the probability of heart disease, and its predictions guide the choice between automated decisions and manual review. Cases with sufficiently low or high predicted probabilities receive automated negative or positive decisions, respectively, while intermediate cases are referred for manual review. This workflow balances the cost of manual review against the potentially asymmetric costs of false negatives and false positives, providing an operational setting for examining the relationship between FI and expected cost.
}

\subsubsection{{Threshold-Based Referral Policy}}
{{
Let $p(\BFx)\in[0,1]$ denote the predicted probability of the positive class. 
Given lower and upper decision thresholds $\theta_{\mathrm{neg}}$ and $\theta_{\mathrm{pos}}$, where $0\leq\theta_{\mathrm{neg}}\leq\theta_{\mathrm{pos}}\leq1$, the system automatically classifies a case as negative when $p(\BFx)<\theta_{\mathrm{neg}}$, refers it for manual review when $\theta_{\mathrm{neg}}\leq p(\BFx)<\theta_{\mathrm{pos}}$, and automatically classifies it as positive when $p(\BFx)\geq\theta_{\mathrm{pos}}$.
Each referred case incurs a fixed total cost $C_{\mathrm{man}}>0$, and the cost of a correct automated decision is normalized to zero. Let $C_{\mathrm{FP}}\geq C_{\mathrm{man}}$ and $C_{\mathrm{FN}}\geq C_{\mathrm{man}}$ denote the costs of an automated false positive and false negative, respectively. In many diagnostic settings, $C_{\mathrm{FN}}>C_{\mathrm{FP}}$ reflects the greater harm of failing to detect disease}.}

{Let $\mathbb{P}_+$ and $\mathbb{P}_-$ denote the positive- and negative-class feature distributions in the operational population, and let $q_+$ and $q_-$ be the corresponding class proportions, where $q_++q_-=1$. The expected operational cost per patient is
}{$$
\begin{aligned}
    C(p; \theta_{\mathrm{neg}},\theta_{\mathrm{pos}})
    =\;&q_+\mathbb E_{\mathbb{P}_+}\Big[C_{\mathrm{man}}\mathbb{I}\{\theta_{\mathrm{neg}}\leq p(\BFx)<\theta_{\mathrm{pos}}\}
    +C_{\mathrm{FN}}\mathbb{I}\{p(\BFx)<\theta_{\mathrm{neg}}\}\Big]\\
    &+q_-\mathbb E_{\mathbb{P}_-}\Big[C_{\mathrm{man}}\mathbb{I}\{\theta_{\mathrm{neg}}\leq p(\BFx)<\theta_{\mathrm{pos}}\}
    +C_{\mathrm{FP}}\mathbb{I}\{p(\BFx)\geq\theta_{\mathrm{pos}}\}\Big].
\end{aligned}
$$}
{In practice, the thresholds $(\theta_{\mathrm{neg}},\theta_{\mathrm{pos}})$ can be selected on validation data by minimizing empirical operational cost over an admissible set of threshold pairs.  
Alternatively, when the predicted probabilities are calibrated and manual review is error-free, cost-optimal thresholds admit closed-form expressions (see Appendix \ref{appe:experiment}).
}

\begin{proposition}
    \label{prop:fi_operational_cost_certificate}
    {Let $p_1$ and $p_2$ be two classifiers. The FIs are defined by Equation \eqref{eq:fi_def} with $\mathbb P_+\otimes\mathbb P_-$ as the nominal distribution. Suppose that, for some $0\leq\ell<u\leq1$,
    \[
    \begin{aligned}
        \mathbb{P}_+\bigl(p_2(\BFx)<t\bigr)
        \leq \mathbb{P}_+\bigl(p_1(\BFx)<t\bigr),
        \forall t\in[0,u], \quad
        \mathbb{P}_-\bigl(p_2(\BFx)\geq t\bigr)
        \leq \mathbb{P}_-\bigl(p_1(\BFx)\geq t\bigr),
        \forall t\in[\ell,1],
    \end{aligned}
    \]
    with each inequality holding strictly at some point of its respective interval. Then, for every common threshold pair satisfying $\ell\leq\theta_{\mathrm{neg}}\leq\theta_{\mathrm{pos}}\leq u$,
    \(
        C(p_2;\theta_{\mathrm{neg}},\theta_{\mathrm{pos}})
        \leq
        C(p_1;\theta_{\mathrm{neg}},\theta_{\mathrm{pos}}).
    \)
    Moreover, suppose that $\varepsilon(p_1)$ is nonconstant
under $\mathbb{P}_+\otimes\mathbb{P}_-$, and let
$M_1=\esssup_{\mathbb{P}_+\otimes\mathbb{P}_-}\varepsilon(p_1)$.
Then there exists $\tau_0<M_1$ such that, for every
$\tau\in[\tau_0,M_1)$,
\[
    0\leq \mathrm{FI}_{\mathrm{KL}}(p_2;\tau)
    <
    \mathrm{FI}_{\mathrm{KL}}(p_1;\tau)
    <\infty.
\]}
\end{proposition}

{ Proposition \ref{prop:fi_operational_cost_certificate} provides conditions for the alignment between operational cost and FI. It establishes cost ordering for common thresholds in the specified interval and FI ordering at sufficiently high targets. In this task, confident mistakes are particularly consequential because they can bypass human review and trigger costly automated decisions. This supports considering FI alongside accuracy and AUC when selecting models for automation. 

}

\subsubsection{{Empirical Operational-Cost Comparison}}
{
We set the label-flipping probability to $p_{flip}=0.2$ to induce a moderate distributional shift and the manual-review cost to $C_{\mathrm{man}}=1$. For each $C_{\mathrm{FP}}\in\{4,8,12\}$, we vary $C_{\mathrm{FN}}/C_{\mathrm{FP}}$ from 2 to 10. In each of the 100 trials, we split the testing sample into disjoint validation and evaluation subsets that are equal in size and stratified by class. For every model and cost configuration, the thresholds $(\theta_{\mathrm{neg}},\theta_{\mathrm{pos}})$ are selected to minimize empirical operational cost on the validation subset and then applied unchanged to the evaluation subset. We compare ERM, the KL- and Wasserstein-based models, XGBoost, and SCE.
}

\begin{figure}[htbp]
    \centering
    \includegraphics[width=\textwidth]{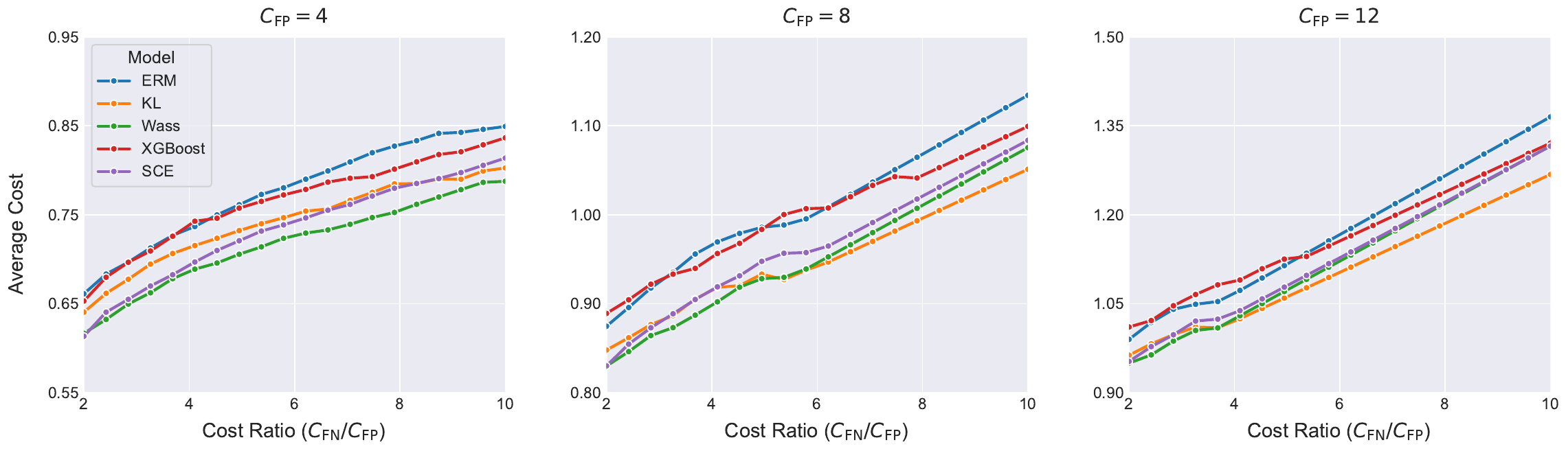}
    \caption{{{Average operational }cost per patient on the heart failure prediction dataset when {$p_{flip}=0.2$ and $C_{\mathrm{man}}=1$}.}}
    \label{fig:heartattack_cost_analysis}
\end{figure}

{Figure \ref{fig:heartattack_cost_analysis} reports the {resulting average operational }cost per patient. {Across all five methods, an }FI-based {model achieves the lowest average cost at almost every evaluated cost ratio. Within the FI-based pair, the relative advantage varies with cost asymmetry. The Wasserstein }model performs better {for nearly all cost ratios when $C_{\mathrm{FP}}=4$ and at lower cost ratios in the other two panels, whereas the KL model performs better as $C_{\mathrm{FN}}/C_{\mathrm{FP}}$ increases.
}}

{{The selected policies increasingly rely on human review as automated errors become more costly relative to the fixed review cost. Across models and cost ratios, the average manual-review rates are approximately $28\%$--$43\%$, $48\%$--$60\%$, and $60\%$--$71\%$ for $C_{\mathrm{FP}}=4$, $8$, and $12$, respectively. 
These results show that the FI-based models also achieve lower operational costs in the evaluated threshold-based referral workflow}.}

\section{FI-Based Deep Neural Network Training}
\label{sec:fi_deep_learning}
We extend our FI-based model to neural networks, demonstrating the potential of our framework for deep learning. Unlike the linear model considered in the problem \eqref{eq:prob_general}, incorporating FI into a network structure is not straightforward. The primary challenge lies in the backpropagation and optimization process. Most neural network training processes rely on stochastic gradient descent (SGD) and its extensions, which are designed primarily for optimizing constraint-free loss functions. In contrast, our FI formulations are constrained optimization problems. While constrained SGD techniques can be employed, their practical convergence and stability are often questionable due to the inherent complexities and non-convexity of neural networks.

To address this, we propose an approximate regularization scheme based on the Lagrangian multiplier method. Consider the multi-class setting with $y \in [C]$ and a neural network whose last layer is fully connected. The output of the penultimate layer is treated as the feature $\BFx$, and the final-layer weights $\BFB$ correspond to the model parameters. In this view, the network before the last layer serves as a feature map.

Building on the cross-entropy reformulation in Theorem \ref{theorem:cross_entropy_reformulation} (under the equivalence condition), the Lagrangian can be written as
{
$$
    \mathcal{L}(\BFB, k) = k + \lambda_0 (L_{ERM}(\BFB) - \tau) + \sum_{i,j\in[C], i>j} \lambda_{ij} \left(\|\BFbeta_{i} - \BFbeta_{j}\|_* - k\right),
$$}
where $L_{ERM}(\BFB)$ is the empirical loss defined in the equation \eqref{eq:loss_erm}, $\lambda_0$ and $\lambda_{ij}$ are the Lagrangian multipliers, and recall that $\BFbeta_i$ denotes the weights to obtain the score for the $i$-th class. 

To improve optimization, we adopt an augmented Lagrangian approach, replacing function $\|\BFbeta_{i} - \BFbeta_{j}\|_* - k$ with $(\|\BFbeta_{i} - \BFbeta_{j}\|_* - k)_+^2$ to enhance its convexity. We further replace the original multipliers $\lambda_{ij}$ by another predetermined large hyperparameter $\alpha$ such that $(\|\BFbeta_{i} - \BFbeta_{j}\|_* - k)_+^2$ is always close to 0. Dividing the whole formula by $\lambda_0$
and omitting $\tau$, as it does not affect optimization, we obtain the final training objective:
\begin{equation}
    \label{eq:fi_def_loss_nn}
    L_{ERM}(\BFB) + \underbrace{\frac{1}{\lambda_0} \left(k + \frac{\alpha}{2} \sum_{i,j\in[C], i>j}(\|\BFbeta_{i} - \BFbeta_{j}\|_* - k)_+^2 \right)}_{\text{FI-inducing regularizer}}.
\end{equation}
This reformulation converts the constrained problem into a regularized objective that is compatible with SGD-based optimization. It also remains compatible with any additional regularization $R(\BFB)$ already included in the loss $L_{ERM}(\BFB)$. The following result characterizes how hyperparameters in the new objective \eqref{eq:fi_def_loss_nn} control robustness.

\begin{proposition}
    \label{prop:nn_fi_tradeoff}
    {For a fixed $\alpha>0$, let $(B^*(\lambda_0),k^*(\lambda_0))$ denote a global minimizer of objective \eqref{eq:fi_def_loss_nn} over $\BFB$ and $k\geq0$ for each $\lambda_0>0$, and let $L^*(\lambda_0)=L_{ERM}(B^*(\lambda_0))$. Then $L^*(\lambda_0)$ is non-increasing in $\lambda_0$, and $k^*(\lambda_0)\geq k^*(\lambda_0')-1/(2\alpha)$ for any $\lambda_0>\lambda_0'>0$.}
\end{proposition}

This result provides guidance for hyperparameter selection. The parameter $\lambda_0$ {controls the trade-off between empirical fit and the FI-inducing regularizer. }A smaller $\lambda_0$ places greater weight on the {regularizer, whereas a larger }$\lambda_0$ {places greater weight on $L_{ERM}$. For a fixed $\alpha$, Proposition \ref{prop:nn_fi_tradeoff} also shows that optimal fragility is non-decreasing in $\lambda_0$ up to an error of $1/(2\alpha)$, which vanishes as $\alpha$ is large. The target $\tau$ can be omitted, as its role is effectively absorbed by $\lambda_0$}. In contrast, $\alpha$ should not be tuned but set sufficiently large to {reduce violations of the constraints $\|\BFbeta_i - \BFbeta_j\|_* \leq k$}. In practice, values of $\alpha$ above 50 are sufficient to keep violations negligible, as observed in our empirical results.

\subsection{Image Diagnosis with FI-Based ResNet}

We apply the FI-induced neural network training objective \eqref{eq:fi_def_loss_nn} to {nine selected single-label datasets from MedMNIST, }an MNIST-like collection of standardized {biomedical images }\citep{medmnistv2}{. These datasets cover diverse tasks}, such as {classifying colorectal tissue types }and diagnosing pneumonia. Each sample is a $28 \times 28$ biomedical image, presented in either grayscale or RGB. Given the high cost of misclassification in medical diagnosis, robustness and sensitivity to risks and outliers are critical. The FI-based model, with its emphasis on risk control and robustness, is particularly well-suited for these challenges in medical image diagnosis.

We adopt the ResNet architecture as the backbone for neural network training \citep{he2016deep}, specifically using the ResNet-18 model for the 2D-image classification task. Following \cite{medmnistv2}, we employ the cross-entropy loss and the Adam optimizer, with a batch size of 128. We evaluate both the ERM model and the Wasserstein FI-based model. For regularization, we implement conventional weight decay for all parameters in both models. The Wasserstein FI-based model further incorporates an FI-induced regularizer from objective function \eqref{eq:fi_def_loss_nn}, which applies only to the weight matrix of the last fully connected layer. {The regularization coefficient is determined based on Bayesian optimization.} We also apply a conventional early stopping strategy and an adaptive learning rate scheduler.
\phantomsection\label{loc:medmnist_fi_evaluation_target}
Further training details are provided in Appendix \ref{appe:experiment}.

Since the training, validation, and testing sets are predefined in the MedMNIST dataset, we report the average accuracy, AUC, cross-entropy, and FI on the testing set over 10 repetitions. The results are summarized in Table \ref{tab:medmnist}\endnote{As a sanity check, the AUC and accuracy values in Table \ref{tab:medmnist} are comparable to the benchmarks reported in \cite{medmnistv2}.}.

\begin{table}[htbp]
    \small
    \setlength{\tabcolsep}{2pt}
    \begin{tabular}{cccccccccc}
    \hline
    \multicolumn{1}{c}{\multirow{2}{*}{\textbf{Dataset(\# classes)}}}  & \multicolumn{1}{c}{\multirow{2}{*}{\textbf{\#Train/\#Val/\#Test}}} & \multicolumn{4}{c}{\textbf{ERM}}                                                                      & \multicolumn{4}{c}{\textbf{Wasserstein FI}}                                                              \\
    \multicolumn{1}{c}{}                                  & \multicolumn{1}{c}{}                                               & \multicolumn{1}{c}{\textbf{ACC}} & \multicolumn{1}{c}{\textbf{AUC}} & \multicolumn{1}{c}{\textbf{CE}} & \multicolumn{1}{c}{\textbf{FI}} & \multicolumn{1}{c}{\textbf{ACC}} & \multicolumn{1}{c}{\textbf{AUC}} & \multicolumn{1}{c}{\textbf{CE}} & \multicolumn{1}{c}{\textbf{FI}} \\
    \hline

    BloodMNIST(8)     & 11,959/1,712/3,421    & 0.9464                           & 0.9958                           & 0.1836                           & 0.0585 & 0.9501                           & 0.9947                           & \textbf{0.1770} & \textbf{0.0552} \\
    BreastMNIST(2)    & 546/78/156            & 0.8391                           & 0.8638                           & 0.5832                           & 0.2525 & \textbf{0.8635} & \textbf{0.8778} & \textbf{0.3661} & \textbf{0.1844} \\
    OrganAMNIST(11)   & 34,561/6,491/17,778   & 0.9098                           & 0.9922                           & \textbf{0.3839} & 0.0731 & 0.9107                           & 0.9888                           & 0.3996                           & 0.0729                           \\
    OrganCMNIST(11)   & 12,975/2,392/8,216    & 0.8941                           & 0.9902                           & 0.4566                           & 0.0827 & 0.9028                           & 0.9831                           & \textbf{0.4291} & \textbf{0.0769} \\
    OrganSMNIST(11)   & 13,932/2,452/8,827    & 0.7474                           & \textbf{0.9615} & 1.0165                           & 0.2244 & \textbf{0.7612} & 0.9505                           & \textbf{0.8325} & \textbf{0.1951} \\
    PathMNIST(9)      & 89,996/10,004/7,180   & 0.8322                           & \textbf{0.9684} & 0.7686                           & 0.0855 & 0.8403                           & 0.9568                           & \textbf{0.6425} & 0.0856                           \\
    PneumoniaMNIST(2) & 4,708/524/624         & 0.8588                           & \textbf{0.9556} & 0.6262                           & 0.1341 & 0.8572                           & 0.9343                           & \textbf{0.4238} & \textbf{0.1281} \\
    RetinaMNIST(5)    & 1,080/120/400         & 0.4890                           & \textbf{0.7158} & 1.2542                           & 2.4161 & \textbf{0.5052} & 0.7031                           & 1.2643                           & \textbf{1.6505} \\
    TissueMNIST(8)    & 165,466/23,640/47,280 & \textbf{0.6507} & 0.9023                           & 0.9841                           & 0.1534 & 0.6442                           & 0.9013                           & 0.9872                           & \textbf{0.1511} \\ 
    \hline
    \end{tabular}
    \caption{The results of the ResNet-18 model on the MedMNIST dataset. For the performance metrics, ACC means accuracy, and CE means cross-entropy. The values are the average of 10 repetitions. {Bold indicates }the {better value when }the {relative difference between the two models is }at least $1\%$.}
    \label{tab:medmnist}
\end{table}

Table \ref{tab:medmnist} shows that the ERM model and the Wasserstein FI-based model perform comparably in terms of accuracy and AUC. The FI-based model achieves lower cross-entropy on six of the nine datasets and lower FI on eight datasets. On the downside, the Wasserstein model requires more training epochs, making it more computationally expensive. On average, the Wasserstein {FI-based }model requires 43\% more epochs {before early stopping }than the ERM model.

\section{Conclusion}

We study classification with a focus on the risk of misjudgment and the generalization of classifiers under distributional shifts. To bridge the gap between conventional metrics and misjudgment risk, we propose the Fragility Index (FI), a risk-averse performance metric that captures large pairwise ranking errors. Unlike metrics such as AUC, which emphasize error rates, FI quantifies the fragility of errors and their sensitivity to distributional shifts. 
FI complements existing metrics by providing insight into ranking-error tail risk. We also identify conditions under which FI and expected operational cost improve together, and observe lower operational costs for FI-based models in a simulated diagnostic workflow.

{We further develop a model training framework {based on }a tractable surrogate loss, leveraging robust satisficing to control fragility. We show that models trained under this framework admit effective bounds on FI. {In addition, we derive tractable convex formulations for cross-entropy, hinge-type, and Lipschitz losses, characterize when their compact forms are exact, and establish generalization and finite-sample guarantees.} We also extend FI as a regularization principle for deep neural networks, providing a scalable approach compatible with standard training pipelines.}

More broadly, classification models are widely used as decision-support tools, where understanding prediction risk is as important as predictive accuracy. Our results suggest that incorporating FI into model evaluation and training can lead to more reliable and cost-efficient decisions, particularly in safety-critical and cost-sensitive settings such as healthcare and finance. An important direction for future work is to extend this risk-aware framework to more complex systems, including large language models, where overconfidence remains a central challenge.

\theendnotes

\putbib

\end{bibunit}

\newpage
\begin{bibunit}

\begin{APPENDIX}{}

\renewcommand*{\theHsection}{appendix.\Alph{section}}
\renewcommand*{\theHsubsection}{\theHsection.\arabic{subsection}}
\renewcommand*{\theHsubsubsection}{\theHsubsection.\arabic{subsubsection}}

\makeatletter
\renewcommand{\paragraph}{\@startsection{paragraph}{4}{10pt}  {0.1pt plus 0.1pt minus 0.1pt}{-0.5em}  {\AppendixFontSize\itshape\sffamily}}
\makeatother

\numberwithin{equation}{section}

{The supplementary material is organized into four sections. Section \ref{appe:fi} provides supplementary results for FI; Section \ref{appe:fi_training} presents additional results for FI-based training, including reformulation details, finite-sample guarantees, and connections to DRO; Section \ref{appe:experiment} reports supplementary experiments; and Section \ref{appe:proof} contains all proofs.}

\section{Supplementary to FI (Section \ref{sec:FI})}
\label{appe:fi}
{This section provides supplementary details on the Fragility Index (FI) introduced in Section \ref{sec:FI}. Section \ref{appe:fi_upon_loss} extends FI to loss functions and discusses its relationship with accuracy and classification quality. Section \ref{appe:fi_ovo_ova} compares the OVA and OVO formulations of multi-class FI and analyzes their behavior under class imbalance. Section \ref{appe:calibration_vs_confident_misjudgment} examines the relationship between calibration and confident-misjudgment mitigation across several illustrative settings. Section \ref{appe:fi_calculation_algorithm} provides a bisection algorithm for computing FI. Finally, Section \ref{appe:fi_calculation_wasserstein} derives an exact computation of Wasserstein FI for linear logistic classifiers under norm transport costs.}

\subsection{FI upon the Loss Function}
\label{appe:fi_upon_loss}
{In classification, the loss function is the objective we minimize during the training process. Therefore, its magnitude also reveals the classification quality. {The empirical expectation of the $0$--$1$ loss equals the misclassification rate, that is, one minus accuracy}. However, due to tractability issues, we usually adopt surrogate loss functions, such as hinge loss. For example, consider the linear classifier $\BFw^T \BFx$ with the hinge loss 
$$
    \ell(\BFx, y) = \max\{0, 1-y \BFw^T \BFx\},
$$ 
where $y \in \{-1, 1\}$ is the label. For observation $\BFx$, if $\ell(\BFx, y) > 1$, the classifier makes false predictions for this sample. Moreover, the hinge loss is also interpreted as the margin-maximization loss notably for SVM. Therefore, the value of the hinge loss also reveals the distance between the sample and the decision boundary and reflects the confidence level of the false prediction. Therefore, we can also define the FI based on the hinge loss as 
\begin{equation}
    \label{eq:fi_hinge}
    \mathrm{FI}(\ell; \tau) = \inf\left\{ k>0 \middle| \mathbb{E}_{\mathbb{P}}[\ell] \leq \tau + k D_{\mathrm{KL}}(\mathbb{P}, \hat{\mathbb{P}}), \ \forall \mathbb{P}\in \mathcal{P}(\mathcal{X}\times\mathcal{Y})\right\}.
\end{equation}
The distribution $\phat$ is the empirical distribution of data. The FI based on the hinge loss inherits all the properties of risk aversion and tail risk from the original FI. It indicates the risk of large hinge loss values, which means the risk of large margin violations. Regarding the target $\tau$, it can be set based on $\mathbb{E}_{\phat}[\ell]$ for sure. Since the threshold of correct and incorrect prediction is 1, $\tau = 1$ is also a natural choice.} 

Unlike the FI based on the ranking error, the FI based on the hinge loss is sample-wise instead of sample-pair-wise. Since the basic criterion changes, it is hard to compare the FI in \eqref{eq:fi_hinge} with AUC or ranking-error-based FI. {However, recall that the hinge loss is a surrogate {for the $0$--$1$ loss, whose empirical expectation equals one minus }accuracy.} Therefore, we can make a connection between the hinge loss and accuracy. As mentioned,
$$
    \mathrm{Accuracy} = \mathbb{P}(y \BFw^T \BFx > 0) = \mathbb{P} (\ell(\BFx, y) < 1).
$$
Therefore, the hinge loss plays the same role as the ranking error in AUC. This also implies that the relationship between hinge-loss-based FI and accuracy is similar to the relationship between ranking-error-based FI and AUC. 

Even though we mainly focus on hinge loss as an example, for many general loss functions like cross-entropy loss or square loss, their values also convey information about the quality of prediction. Therefore, a similar argument can be applied to the FI based on these loss functions.

\subsection{FI for Different Averaging Schemes in Multi-Class Classification}
\label{appe:fi_ovo_ova}

In the main text, we focus on the one-vs-one (OVO) scheme for defining FI in multi-class classification because it is robust to rare classes. However, another widely used scheme is the one-vs-all (OVA) scheme. In this section, we provide a detailed discussion of OVO and OVA schemes. We formally define OVA FI, analyze its theoretical relationship with OVO FI, in the spirit of \citep{yang2022auc}, and show that, similar to AUC, OVO FI exhibits greater robustness to rare classes than OVA FI.

{The OVA and OVO schemes are two standard approaches for reducing a multi-class classification problem to a series of binary classification tasks. In the OVA scheme, each class $i$ is treated as the positive class, while all other classes are grouped as negative. In contrast, the OVO scheme considers every pair of classes separately. Formally, let $\BFx^i$ denote {a sample from class $i$ and let }$p_i$ be the score function {for }class $i$. The ranking error for class $i$ versus class $j$ is defined as
\begin{equation*}
    \varepsilon_{i|j}(p_i) = p_i(\BFx^j) - p_i(\BFx^i).
\end{equation*}
Similarly, let $\BFx^{\lnot i}$ denote {a sample from all classes other than }class $i$. The OVA ranking error for class $i$ is
\begin{equation*}
    \varepsilon_i(p_i) = p_i(\BFx^{\lnot i}) - p_i(\BFx^i).
\end{equation*}
{For each $c\in[C]$, let $\widehat{\mathbb{P}}_c$ denote the empirical class-conditional feature distribution and let $N_c$ denote the number of observations in class $c$. Define $N=\sum_{c\in[C]}N_c$ and $N_{\lnot i}=N-N_i$. The nominal ordered-pair distributions }for OVO and OVA {are}, respectively,
\begin{align*}
    \widehat{\mathbb Q}_{ij}
    &:=\widehat{\mathbb{P}}_i\otimes\widehat{\mathbb{P}}_j, \qquad i\neq j,\\
    \widehat{\mathbb Q}_i
    &:=\widehat{\mathbb{P}}_i\otimes\widehat{\mathbb{P}}_{\lnot i},
    \qquad
    \widehat{\mathbb{P}}_{\lnot i}
    :=\sum_{j\in[C],\,j\neq i}\frac{N_j}{N_{\lnot i}}\widehat{\mathbb{P}}_j.
\end{align*}
{The pairwise and one-versus-all FIs are then defined as
}\begin{align*}
    \mathrm{FI}_{i|j}(p_i;\tau)
    &:=\inf\left\{k>0\ \middle|\ 
    \mathbb E_{\mathbb Q}[\varepsilon_{i|j}(p_i)]
    \leq\tau+kD(\mathbb Q,\widehat{\mathbb Q}_{ij}),\ 
    \forall\mathbb Q\in\mathcal P(\mathcal X^2)\right\},\\
    \mathrm{FI}_i(p_i;\tau)
    &:=\inf\left\{k>0\ \middle|\ 
    \mathbb E_{\mathbb Q}[\varepsilon_i(p_i)]
    \leq\tau+kD(\mathbb Q,\widehat{\mathbb Q}_i),\ 
    \forall\mathbb Q\in\mathcal P(\mathcal X^2)\right\}.
\end{align*}
The overall FI for OVO and OVA are then defined as
\begin{align*}
    \mathrm{FI}_{\text{OVO}}(p_1,\ldots,p_C;\tau)
    &:= \frac{1}{C(C-1)} \sum_{i \in [C]} \sum_{j \in [C], j \neq i} \mathrm{FI}_{i|j}(p_i;\tau), \\
    \mathrm{FI}_{\text{OVA}}(p_1,\ldots,p_C;\tau)
    &:= \frac{1}{C} \sum_{i \in [C]} \mathrm{FI}_i(p_i;\tau).
\end{align*}
{Since
}\begin{equation*}
    \widehat{\mathbb Q}_i
    =\sum_{j\in[C],\,j\neq i}\frac{N_j}{N_{\lnot i}}\widehat{\mathbb Q}_{ij},
\end{equation*}
{the OVA ranking-error distribution for }class $i$ {is the corresponding }mixture of the OVO {ranking-error distributions},
weighted by the class sizes. This mixture structure implies that OVA FI is sensitive to the relative sizes of the negative classes, potentially causing minority class risks to be underrepresented.}

\subsubsection{Theoretical Relationship between OVO FI and OVA FI}
Unlike AUC, for which \citep{yang2022auc} provides an equation relationship between OVO and OVA, the relationship between OVO and OVA FIs is more complex due to the non-linear risk measure. Nevertheless, we can establish tight bounds between them. For computational clarity, we focus on the FI with KL-divergence as the distance metric. 
Then, the connection of OVO and OVA FI is described as follows:
\begin{propositionAp}
    \label{prop:fi_sum_bound}

    {For each $i\in[C]$, the OVA FI satisfies
    }\begin{enumerate}[(a)]
        \item {{$\mathrm{FI}_i(p_i;\tau)$ }is bounded below by
        {$$
            \mathrm{FI}_i(p_i;\tau) \geq \min_{j \in [C], j \neq i} \mathrm{FI}_{i|j}(p_i;\tau).
$$}
        This bound is tight: there exist {nominal ordered-pair distributions $\widehat{\mathbb Q}_{ij}$ }and targets $\tau$ such that {$\mathrm{FI}_i(p_i;\tau) = \min_{j \in [C], j \neq i} \mathrm{FI}_{i|j}(p_i;\tau)$}.}

        \item {{$\mathrm{FI}_i(p_i;\tau)$ }is bounded above by
        {$$
            \mathrm{FI}_i(p_i;\tau) \leq \max_{j \in [C], j \neq i} \mathrm{FI}_{i|j}(p_i;\tau).
$$}
        This bound is also tight: there exist {nominal ordered-pair distributions $\widehat{\mathbb Q}_{ij}$ }and targets $\tau$ such that {$\mathrm{FI}_i(p_i;\tau) = \max_{j \in [C], j \neq i} \mathrm{FI}_{i|j}(p_i;\tau)$}.}
    \end{enumerate}

\end{propositionAp}
The proof is provided in Appendix \ref{appendix:proof_fi_sum_bound}. {Proposition \ref{prop:fi_sum_bound} shows that {$\mathrm{FI}_i(p_i;\tau)$ }is bounded by the minimum and maximum of the {pairwise FIs $\mathrm{FI}_{i|j}(p_i;\tau)$}, and both bounds are tight. Unlike the AUC case, {OVA }FI does not {arise as }a simple average {of the pairwise FIs}, but rather {lies }between the best and worst pairwise cases depending on the class weights.}

\subsubsection{Quantitative Analysis of the Effect of Class Imbalance on OVA FI}
{{We next }provide quantitative insight into their relationship. {Under KL divergence, the OVA FI feasibility condition for class $i$ }is}

{\[
k \ln \left( \mathbb{E}_{\widehat{\mathbb{Q}}_i} \left[ \exp \left(\frac{\varepsilon_i(p_i)}{k}\right)\right] \right) = k \ln \left( \sum_{j \in [C], j \neq i} \frac{N_j}{N_{\lnot i}} \mathbb{E}_{\widehat{\mathbb{Q}}_{ij}} \left[ \exp \left(\frac{\varepsilon_{i|j}(p_i)}{k}\right)\right] \right) \leq \tau.
\]}

{{Thus, each pairwise exponential moment enters the OVA feasibility condition with the class-size weight $N_j/N_{\lnot i}$. A }high-risk {pair $(i,j)$ can therefore have limited influence on OVA FI when }the opposing class $j$ is a minority. Conversely, the OVO scheme treats all {ordered }class pairs equally {through }macro-averaging, preventing majority classes from obscuring risks in minority {subpopulations}.}

{To illustrate this effect, {fix class $i$ and consider two opposing classes $j_1$ and $j_2$. Suppose the nominal ordered-pair distributions $\widehat{\mathbb Q}_{ij_1}$ and $\widehat{\mathbb Q}_{ij_2}$ induce ranking-error samples with $\mathrm{FI}_{i|j_1}(p_i;0)=0.077$ and $\mathrm{FI}_{i|j_2}(p_i;0)=0.055$, respectively. Their average pairwise OVO FI is }$0.066$. {For
}{\[
    \widehat{\mathbb Q}_i(w)
    =w\widehat{\mathbb Q}_{ij_1}+(1-w)\widehat{\mathbb Q}_{ij_2},
    \qquad w\in[0,1],
\]}
the resulting OVA FI values are shown in Table \ref{tab:fi_weight}.}
\begin{table}[htbp]
    \small
    \centering
    \caption{{OVA FI {$\mathrm{FI}_i(p_i;0)$ under }different {mixture }weights $w$.}}
    \label{tab:fi_weight}
    \begin{tabular}{cccccccccc}
        \hline
        $w$  & 0.1   & 0.2   & 0.3   & 0.4   & 0.5   & 0.6   & 0.7   & 0.8   & 0.9   \\
        {$\mathrm{FI}_i(p_i;0)$ }& 0.061 & 0.064 & 0.066 & 0.068 & 0.070 & 0.072 & 0.074 & 0.075 & 0.076 \\
        \hline
    \end{tabular}
\end{table}

{{In this example, the OVA FI increases as more weight is assigned to $\widehat{\mathbb Q}_{ij_1}$, which has the larger pairwise FI. This illustrates }that OVA FI {depends on }class sample sizes {and can diminish }the influence of minority classes.} Hence, OVA and OVO FI may yield different results when class imbalance is severe.

\paragraph{Example: OVA vs OVO FI under Class Imbalance} We show an example where OVO and OVA FI provide conflicting assessments. {Suppose we evaluate class $1$ with respect to classes $2$ and $3$, {whose sample-size ratio }is $49:1${. Consider two classifiers $p^a$ and $p^b$. Their pairwise FIs are
}\begin{align*}
    \mathrm{FI}_{1|2}(p_1^a;0)&=0,
    &\mathrm{FI}_{1|3}(p_1^a;0)&=0.0824,\\
    \mathrm{FI}_{1|2}(p_1^b;0)&=0.0554,
    &\mathrm{FI}_{1|3}(p_1^b;0)&=0.
\end{align*}}
{For $x\in\{a,b\}$, define the class-$1$ macro-averaged OVO FI as
}{\[
    \overline{\mathrm{FI}}^{\mathrm{OVO}}_1(p_1^x;0)
    :=\frac{1}{2}\left(
        \mathrm{FI}_{1|2}(p_1^x;0)
        +\mathrm{FI}_{1|3}(p_1^x;0)
    \right).
\]}
{{This gives $\overline{\mathrm{FI}}^{\mathrm{OVO}}_1(p_1^a;0)=0.0412$ and $\overline{\mathrm{FI}}^{\mathrm{OVO}}_1(p_1^b;0)=0.0277$, so OVO prefers classifier $p^b$}. However, {aggregating the nominal pairwise ranking-error distributions with weights $49/50$ and $1/50$ gives $\mathrm{FI}_1(p_1^a;0)=0.0531$ and $\mathrm{FI}_1(p_1^b;0)=0.0552$. Thus, OVA gives the opposite assessment and prefers classifier $p^a$}.}

{{Following }the reasoning in \citep{yang2022auc}, classifier {$p^b$ is preferable because both classifiers have zero FI for one pair, while $p^b$ has the smaller FI }for the other {pair ($0.0554<0.0824$)}. The OVO FI {reflects this comparison, whereas the class-size-weighted OVA FI favors $p^a$}. This example {illustrates }that OVA FI {can overlook }risk associated with {a minority class}.}

\subsubsection{{{Incorporating Heterogeneous }Misclassification {Costs into }FI and FI-Based {Training}}}

In multi-class settings, misclassification costs are often heterogeneous across class pairs. A standard way to account for this is cost-sensitive learning~\citep{elkan2001foundations}, which uses weighted losses based on domain costs.

{Let {$\mathbf C\in\mathbb R^{C\times C}$ }be a cost matrix with {$\mathbf C_{ii}=0$, where $\mathbf C_{ij}$ is }the cost of predicting class $j$ when the true class is $i$. {For a classifier $\hat y$, its expected misclassification cost is
}{\[
    \operatorname{Cost}(\hat y)
    :=\mathbb E_{(\BFx,y)}
    \left[\mathbf C_{y,\hat y(\BFx)}\right].
\]}}

{{To prioritize costly error directions in FI evaluation, let $w_{i|j}\geq0$ denote the weight assigned to the ordered }class pair $(i,j)$ {and assume that $\sum_{i\neq j}w_{i|j}>0$. A normalized cost-weighted FI is
}\begin{equation*}
    \mathrm{FI}_{\mathrm{weighted}}(p_1,\ldots,p_C;\tau)
    :=
    \frac{
        \displaystyle\sum_{i\in[C]}\sum_{j\in[C],\,j\neq i}
        w_{i|j}\,\mathrm{FI}_{i|j}(p_i;\tau)
    }{
        \displaystyle\sum_{i\in[C]}\sum_{j\in[C],\,j\neq i}w_{i|j}
    }.
\end{equation*}
{When all weights equal one, this definition reduces to the macro-averaged OVO FI in \eqref{eq:def_multiclass_fi}.
}}

{{Because $\varepsilon_{i|j}(p_i)$ measures the tendency of a sample from true }class $j$ to {receive a high class-}$i$ {score, $w_{i|j}$ should reflect the cost $\mathbf C_{ji}$. Thus, larger cost for misclassifying class $j$ as class $i$ }corresponds to {a }larger $w_{i|j}${. For example}, if misclassifying {``no creature'' as ``person'' }is costlier than {misclassifying it as ``animal,'' one should set
}{\[
    w_{\mathrm{person}\mid\mathrm{no\ creature}}
    >
    w_{\mathrm{animal}\mid\mathrm{no\ creature}}.
\]}
{The weights }can encode economic {or }safety priorities and {may also incorporate }class-imbalance {considerations}~\citep{zhang2023deep,wang2023unified}. {The FI-based training framework can likewise incorporate these costs by using a convex cost-sensitive surrogate loss in Problem \eqref{eq:prob_general}.
}}

\subsection{{Calibration and Confident Misjudgment{: Alignment and Conflict Regimes}}}
\label{appe:calibration_vs_confident_misjudgment}

{We next {clarify }when mitigating confident misjudgment {is aligned with calibration and when the two objectives may conflict. They measure different properties: calibration concerns whether predicted probabilities agree with conditional outcome frequencies, whereas FI concerns the magnitude and distributional fragility of pairwise ranking errors. When a }model is overconfident relative to the true distribution{, mitigating confident misjudgment can improve probability fit and calibration. If a model is already calibrated or underconfident, however, further confidence reduction can induce or exacerbate underconfidence and worsen calibration, even though it may lower FI}.}

{Overfitting {can make }a model match training data too closely {and thereby hurt }generalization, calibration, and confidence reliability. To {illustrate how empirical fitting can induce excessive confidence}, consider cross-entropy (negative log-likelihood) as an example. Let the true distribution be $\pi(\BFx, y)$ and the model be $\hat{\pi}(y|\BFx)$. The population objective is
$$
	\min_{\hat{\pi}} \mathbb{E}_{(\BFx, y) \sim \pi}[-\log \hat{\pi}(y|\BFx)] = \min_{\hat{\pi}} \mathbb{E}_{\BFx} \left[ -\sum_{y \in \mathcal{Y}} \pi(y|\BFx) \log \hat{\pi}(y|\BFx) \right],
$$
which is minimized {by the true conditional probabilities, }$\hat{\pi}(y|\BFx) = \pi(y|\BFx)${, and hence yields perfect calibration}. For a finite training set $\mathcal{D} = \{(\BFx_i, y_i)\}_{i=1}^n$, however, the empirical objective is
$$
	\min_{\hat{\pi}} \mathbb{E}_{(\BFx, y) \sim \mathcal{D}}[-\log \hat{\pi}(y|\BFx)] = \min_{\hat{\pi}} -\frac{1}{n} \sum_{i=1}^n \log \hat{\pi}(y_i|\BFx_i).
$$
{When the training inputs are distinct and the predictor class is unrestricted, this empirical objective is }minimized by $\hat{\pi}(y|\BFx_i) = \mathbb{I}\{y = y_i\}$ for all $i \in [n]$, i.e., confidence 1 on {the }observed label $y_i$ for sample {$\BFx_i$}.}

{When overfitting {manifests as excessive confidence from }matching the empirical distribution too closely, mitigating confident misjudgment acts like regularization against overfitting and {can be }aligned with better calibration and generalization.}

{This alignment is regime-dependent rather than universal. In binary classification, even the perfectly calibrated oracle predictor $p^*(\BFx)=\eta(\BFx):=\mathbb P(Y=1\mid\BFx)$ can make errors because labels are stochastic. For example, observations with $\eta(\BFx)=0.99$ and $0.01$ may, each with conditional probability $0.01$, receive labels $0$ and $1$, respectively, producing a ranking error of $0.99-0.01=0.98$. Thus, irreducible label randomness can generate a nonzero FI even under perfect calibration.
}

\subsubsection{{Numerical Illustrations of Alignment and Conflict}}

{{We conduct three experiments on synthetic }data. First, we compare models {ranging }from logistic regression to neural networks and gradient-boosted trees and show that {more severe overfitting worsens probability estimation and increases }FI. Second, we fix the model class and vary {the degree of }regularization, showing that reducing overfitting improves {probability estimation and decreases FI. Third, we start from a perfectly calibrated oracle classifier and contract its probabilities toward $0.5$, thereby illustrating a regime in which FI decreases while calibration worsens.
}}

{The data are generated from a {binary }logistic model. For a {feature vector }$\BFx$, {let
}{$$
	\eta(\BFx):=\pi(Y=1\mid\BFx)
	= \frac{1}{1 + \exp\!\left(-\theta \BFw^T(\BFx-\mathbb E[\BFx])\right)}.
$$}
{The coefficient vector }$\BFw$ is randomly {drawn, the feature vector }$\BFx$ is sampled from a {uniform distribution, and the label $Y$ is sampled from $\operatorname{Bernoulli}(\eta(\BFx))$}. The parameter $\theta$ controls class separability: {a }larger $\theta$ yields a steeper {decision }boundary and, under perfect calibration, higher average prediction confidence.}

{For {Experiments 1 and 2}, we generate training, validation, and test sets. We {fit each classifier }on the training set, tune {the temperature }on the validation set, and evaluate {the resulting classifier }on the test set. Given {the logit }$h_j(\BFx)$ for class $j$ and {a temperature $T>0$, temperature scaling gives
}{$$
	\hat{\pi}_T(Y=k\mid\BFx)
	=\frac{\exp(h_k(\BFx)/T)}{\sum_{j\in\mathcal Y}\exp(h_j(\BFx)/T)},
	\qquad k\in\mathcal Y.
$$}
We choose $T$ by minimizing {the }negative log-likelihood {on the validation set and write $\hat p_T(\BFx):=\hat\pi_T(Y=1\mid\BFx)$ for the resulting positive-class probability}.}

{We report {accuracy}, AUC, and {$\mathrm{FI}_{\mathrm{KL}}(\hat p_T;0)$. Since the true conditional probability $\eta(\BFx)$ }is known, we {use Probability RMSE to measure the pointwise deviation of the predicted probabilities from the perfectly calibrated oracle probabilities:
}{$$
	\text{Probability RMSE}
	=\sqrt{\frac{1}{N}\sum_{i=1}^{N}
	\left(\hat p_T(\BFx_i)-\eta(\BFx_i)\right)^2}.
$$}
{A smaller Probability RMSE indicates closer agreement with the oracle probabilities}.}

\paragraph{Experiment 1: Comparison across Classification Models} We evaluate five models: logistic regression (LR), support vector machine (SVM), multilayer perceptron (MLP), XGBoost, and LightGBM. {All {reported }metrics are computed on the test set, {except for training accuracy, which is included as an indicator of }overfitting. Table \ref{tab:calibration_performance_models} reports averages over 50 runs {with $\theta=5.0$}.}
\begin{table}[htbp]
	\centering
	\small
	\caption{Performance comparison across different classification models.}
	\label{tab:calibration_performance_models}
	\begin{tabular}{ccccccc}
		\hline
		\textbf{Classifiers} & \textbf{Accuracy (Train)} & \textbf{Accuracy} & \textbf{AUC}  & \makecell{$\mathbf{\mathrm{FI}_{\mathrm{KL}}}$\\$\mathbf{(\hat p_T;0)}$}   & \makecell{\textbf{Probability}\\\textbf{RMSE}}  \\
		\hline
		LR                   & 0.867 $\pm$ 0.016             & 0.734 $\pm$ 0.032     & 0.814 $\pm$ 0.038 & 0.194 $\pm$ 0.049 & 0.279 $\pm$ 0.029 \\
		SVM                  & 0.942 $\pm$ 0.010             & 0.727 $\pm$ 0.030     & 0.806 $\pm$ 0.037 & 0.195 $\pm$ 0.047 & 0.286 $\pm$ 0.026 \\
		MLP                  & 0.984 $\pm$ 0.012             & 0.712 $\pm$ 0.029     & 0.788 $\pm$ 0.036 & 0.203 $\pm$ 0.047 & 0.299 $\pm$ 0.024 \\
		XGBoost              & 1.000 $\pm$ 0.000             & 0.703 $\pm$ 0.029     & 0.778 $\pm$ 0.036 & 0.208 $\pm$ 0.050 & 0.306 $\pm$ 0.023 \\
		LightGBM             & 1.000 $\pm$ 0.000             & 0.701 $\pm$ 0.029     & 0.775 $\pm$ 0.037 & 0.210 $\pm$ 0.052 & 0.309 $\pm$ 0.023\\
		\hline
	\end{tabular}
\end{table}
{Training accuracy rises from LR to LightGBM, while test accuracy falls, indicating {increasingly severe }overfitting. At the same time, AUC {decreases, probability RMSE increases, and $\mathrm{FI}_{\mathrm{KL}}(\hat p_T;0)$ increases. These results show }that overfitting degrades ranking quality {and probability estimation and increases fragility }to confident misjudgment.}

{{Temperature }scaling can improve {probability }calibration but cannot undo ranking damage {caused by overfitting. Because $T>0$, }temperature scaling is {monotone in the logits and therefore preserves the ordering of the samples and the AUC. Although it rescales the predicted probabilities and can consequently change probability RMSE and FI, it cannot repair ranking errors or recover ranking information lost during model fitting}. Hence, overfitting can harm {probability estimation in ways that }post-hoc {temperature scaling cannot fully correct}.}

\paragraph{{Experiment 2: {Effect of }Regularization}} {We {fix the model class to be an MLP and vary its }$\ell_2$ regularization coefficient $\alpha$ from $0.001$ to $1.0${. Table \ref{tab:calibration_performance_regularization} reports the results under the same 50-run setting with $\theta=5.0$}. As $\alpha$ increases, training accuracy {decreases }while test accuracy {increases}, indicating reduced overfitting. {At the same time}, test AUC {increases, probability RMSE decreases, and $\mathrm{FI}_{\mathrm{KL}}(\hat p_T;0)$ }decreases. These {results show }that reducing overfitting {moves the predictions closer to the calibrated oracle probabilities while reducing fragility to confident misjudgment in this regime.
}}

\begin{table}[htbp]
	\centering
	\small
	\caption{Performance comparison of MLP across different regularization coefficients.}
	\label{tab:calibration_performance_regularization}
	\begin{tabular}{ccccccc}
		\hline
		\makecell{\textbf{MLP}\\\textbf{regularization}} & \textbf{Accuracy (Train)} & \textbf{Accuracy} & \textbf{AUC}  & \makecell{$\mathbf{\mathrm{FI}_{\mathrm{KL}}}$\\$\mathbf{(\hat p_T;0)}$}   & \makecell{\textbf{Probability}\\\textbf{RMSE}}  \\
		\hline
		MLP $\alpha=0.001$      & 0.985 $\pm$ 0.012             & 0.713 $\pm$ 0.029     & 0.788 $\pm$ 0.036 & 0.203 $\pm$ 0.047 & 0.300 $\pm$ 0.023 \\
		MLP $\alpha=0.01$       & 0.984 $\pm$ 0.012             & 0.713 $\pm$ 0.029     & 0.789 $\pm$ 0.036 & 0.202 $\pm$ 0.047 & 0.299 $\pm$ 0.023 \\
		MLP $\alpha=0.1$        & 0.946 $\pm$ 0.015             & 0.722 $\pm$ 0.030     & 0.800 $\pm$ 0.037 & 0.199 $\pm$ 0.047 & 0.291 $\pm$ 0.026 \\
		MLP $\alpha=1.0$        & 0.872 $\pm$ 0.015             & 0.728 $\pm$ 0.032     & 0.809 $\pm$ 0.039 & 0.197 $\pm$ 0.050 & 0.284 $\pm$ 0.029\\
		\hline
	\end{tabular}
\end{table}

\paragraph{{Experiment 3: Probability Shrinkage from a Perfectly Calibrated Oracle Classifier}} {{We use the same logistic data-generating process as in Experiments }1 {and }2{, set $\theta=5.0$, and repeat the experiment over 50 independent runs. Unlike the previous two experiments, this experiment involves neither model fitting nor temperature scaling. We begin with the perfectly calibrated oracle probability $p^*(\BFx)=\eta(\BFx)=\pi(Y=1\mid\BFx)$, sample binary labels according to this probability, and define the contracted predictions
}{\[
    p_r(\BFx)=\frac{1}{2}+r\left(p^*(\BFx)-\frac{1}{2}\right),
    \qquad r\in\{1.00,0.90,0.75,0.50\}.
\]}
{Here, $r$ is the contraction factor. $r=1$ recovers the calibrated oracle, whereas $r<1$ moves every probability toward $0.5$ and thereby creates underconfidence.
}}

{This construction gives an analytical benchmark. First, $p_r(\BFx)\geq0.5$ if and only if $p^*(\BFx)\geq0.5$, and $p_r$ is strictly increasing in $p^*$; hence, accuracy and AUC are unchanged. Second, for every positive--negative pair, $ \varepsilon(p_r) =p_r(\BFx^-)-p_r(\BFx^+) =r\varepsilon(p^*). $ With $\tau=0$, the change of variable $k=r\widetilde k$ in the definition of FI in Equation \eqref{eq:fi_def} yields $\mathrm{FI}_{\mathrm{KL}}(p_r;0)=r\,\mathrm{FI}_{\mathrm{KL}}(p^*;0)$. 
In contrast, pointwise probability estimation worsens. Specifically,
}{\[
    \text{Probability RMSE}(p_r)
    =(1-r)\sqrt{\frac{1}{N}\sum_{i=1}^{N}
    \left(\eta(\BFx_i)-\frac{1}{2}\right)^2}.
\]}
{Because $p_r$ is a one-to-one transformation of $\eta$ for $r>0$,
}{\[
    \mathbb E[Y\mid p_r(\BFx)]
    =\eta(\BFx)
    =\frac{1}{2}+\frac{p_r(\BFx)-\frac{1}{2}}{r}.
\]}
{Hence, if $r<1$ and $\mathbb P(\eta(\BFx)\neq 0.5)>0$, then $p_r$ is miscalibrated and systematically underconfident. Thus, the contraction worsens calibration while preserving both the classification decisions and the ranking.
}

\begin{table}[htbp]
    \centering
    \small
    \caption{{Probability shrinkage from a perfectly calibrated oracle classifier.}}
    \label{tab:calibration_oracle_shrinkage}

    \begin{tabular}{ccccc}
        \hline
        {\textbf{{Shrinkage $r$}}} & {\textbf{{Accuracy}}} & {\textbf{{AUC}}} & \makecell{$\mathbf{\mathrm{FI}_{\mathrm{KL}}}$\\$\mathbf{(p_r;0)}$}
        & \makecell{\textbf{Probability}\\\textbf{RMSE}}\\
        \hline
        {1.00 }& {0.860 $\pm$ 0.014 }& {0.939 $\pm$ 0.011 }& {0.146 $\pm$ 0.006 }& {0.000 $\pm$ 0.000}\\
        {0.90 }& {0.860 $\pm$ 0.014 }& {0.939 $\pm$ 0.011 }& {0.131 $\pm$ 0.006 }& {0.039 $\pm$ 0.001}\\
        {0.75 }& {0.860 $\pm$ 0.014 }& {0.939 $\pm$ 0.011 }& {0.109 $\pm$ 0.005 }& {0.097 $\pm$ 0.003}\\
        {0.50 }& {0.860 $\pm$ 0.014 }& {0.939 $\pm$ 0.011 }& {0.073 $\pm$ 0.003 }& {0.195 $\pm$ 0.006}\\
        \hline
    \end{tabular}

\end{table}

{Table \ref{tab:calibration_oracle_shrinkage} shows that decreasing $r$ from $1$ to $0.5$ leaves accuracy and AUC unchanged and halves $\mathrm{FI}_{\mathrm{KL}}(p_r;0)$, whereas probability RMSE increases from $0$ to $0.195$. The nonzero $\mathrm{FI}_{\mathrm{KL}}(p^*;0)$ at $r=1$ is not evidence of miscalibration. It arises because labels are stochastic. Even when the oracle assigns a probability close to $1$ or $0$, the opposite label can still be realized with small probability, and such rare realizations can generate large ranking errors.
}

{At the metric level, because the oracle FI is positive in this experiment, using zero-target ranking-error FI as the sole model-selection criterion over this family would prefer every listed $r<1$ to the calibrated oracle $r=1$, despite the resulting underconfidence. 
Taken together, these experiments show that FI provides risk information beyond calibration. Experiments 1 and 2 show that mitigating overfitting can improve probability estimation and reduce FI simultaneously. Experiment 3 shows that a perfectly calibrated predictor can still make high-confidence errors because labels are random.
Although these errors are rare, they can produce large positive ranking errors. FI measures the associated risk and its sensitivity to distributional shifts, even when the predicted probabilities are perfectly calibrated.
Its preference for more conservative predictions in this example reflects its objective of reducing the magnitude of confident misjudgments.
}

\subsection{Algorithm and Reformulation for Calculating FI}
\label{appe:fi_calculation_algorithm}

{Whenever $G(k)$ can be evaluated and has a finite positive feasibility threshold, the monotonicity in Lemma \ref{lemma:root_finding} yields the bisection procedure in Algorithm \ref{alg:solve_k}. In particular, the conditions of Lemma \ref{lemma:fi_kl} guarantee these requirements for KL divergence.
}

\begin{algorithm}[H]
    \caption{{Bisection search for FI.}}
    \label{alg:solve_k}
    {\KwIn{$G(\cdot)$, an initial $k_0>0$, and tolerance $\epsilon>0$}
    \KwOut{A feasible $\epsilon$-approximation of FI}
    \While{$G(k_0)\leq0$}{
        $k_0=k_0/2$\;
    }
    \While{$G(2k_0)>0$}{
        $k_0=2k_0$\;
    }
    {$k_{\min}=k_0$, $k_{\max}=2k_0$\;
    }\While{$k_{\max} - k_{\min} > \epsilon$}{
        $k = \frac{k_{\max} + k_{\min}}{2}$\;
        \If{$G(k) \leq 0$}{
            $k_{\max} = k$\;
        }
        \Else{
            $k_{\min} = k$\;
        }
    }
    \Return{$k_{\max}$}}
\end{algorithm}

\subsection{{Exact Computation of Wasserstein FI for Logistic Classifiers}}
\label{appe:fi_calculation_wasserstein}

{
For binary FI, let $\widehat{\mathbb Q}=\widehat{\mathbb{P}}_+\otimes\widehat{\mathbb{P}}_-$ and $\varepsilon(p)=p(\BFx^-)-p(\BFx^+)$. We take the nominal class distributions to be
\[
    \widehat{\mathbb{P}}_-=\frac{1}{N_-}\sum_{r=1}^{N_-}\delta_{\widehat{\BFx}^{\, -}_r},
    \qquad
    \widehat{\mathbb{P}}_+=\frac{1}{N_+}\sum_{s=1}^{N_+}\delta_{\widehat{\BFx}^{\, +}_s},
    \qquad N_-,N_+\geq1.
\]
Let $y_+$ and $y_-$ denote the two class labels, and write $c_s(\BFx,\widehat{\BFx})=c(\BFx,y_s,\widehat{\BFx},y_s)$ for $s\in\{+,-\}$. The additive pairwise Wasserstein distance in \eqref{eq:pair_wasserstein_distance} is
\[
    W(\mathbb Q,\widehat{\mathbb Q})
    =\inf_{\pi\in\Pi(\mathbb Q,\widehat{\mathbb Q})}
    \mathbb E_\pi\!\left[
        c_+(\BFx^+,\widehat{\BFx}^{\, +})
        +c_-(\BFx^-,\widehat{\BFx}^{\, -})
    \right].
\]
Let $\mathrm{FI}_{\mathrm W}(p;\tau)$ denote \eqref{eq:fi_def} with $D=W$. For every $k>0$, the additive cost separates the function $G(k)$ in \eqref{eq:def_gk} into two classwise terms,
\begin{equation}
    \label{eq:binary_pair_wasserstein_reformulation}
    \begin{aligned}
        G(k)
        ={}&\mathbb E_{\widehat{\mathbb{P}}_-}\!\left[
            \sup_{\BFx^-\in\mathcal X}
            \left\{p(\BFx^-)-kc_-(\BFx^-,\widehat{\BFx}^{\, -})\right\}
        \right]+\mathbb E_{\widehat{\mathbb{P}}_+}\!\left[
            \sup_{\BFx^+\in\mathcal X}
            \left\{-p(\BFx^+)-kc_+(\BFx^+,\widehat{\BFx}^{\, +})\right\}
        \right]-\tau.
    \end{aligned}
\end{equation}
This identity removes the distributional optimization, but evaluating its pointwise suprema still requires the structure of the classifier and transport cost. For a linear logistic classifier on $\mathbb R^M$ with norm transport costs, both suprema can be evaluated explicitly.

Write $\sigma(z)=(1+\exp(-z))^{-1}$. For $\lambda>0$, define
\[
    H_\lambda(z)=
    \begin{cases}
        \max\{\sigma(z),\,\sigma(a_\lambda)-\lambda(a_\lambda-z)\},
            &0<\lambda<\frac14,\ z<a_\lambda,\\[1mm]
        \sigma(z),&\text{otherwise},
    \end{cases}
    \qquad z\in\mathbb R,
\]
where
\[
    a_\lambda=
    \log\frac{1+\sqrt{1-4\lambda}}{1-\sqrt{1-4\lambda}},
    \qquad 0<\lambda<\frac14.
\]
The two candidates in $H_\lambda$ correspond to retaining the nominal logit or moving it to $a_\lambda$ after accounting for the transport penalty.

\begin{propositionAp}
    \label{prop:fi_wass_reformulation_binary_calculation}
    Suppose $\mathcal X=\mathbb R^M$, $c_s(\BFx,\widehat{\BFx})=\|\BFx-\widehat{\BFx}\|$ for $s\in\{+,-\}$, and $p(\BFx)=\sigma(\BFw^\top\BFx+b)$ with $\BFw\neq\BFzero$. Let $L=\|\BFw\|_*$. For every $k>0$,
    \begin{equation}
        \label{eq:logistic_wasserstein_fi_explicit}
        \begin{aligned}
            G(k)
            ={}&\frac{1}{N_-}\sum_{r=1}^{N_-}
                H_{k/L}\!\left(\BFw^\top\widehat{\BFx}^{\, -}_r+b\right)
            +\frac{1}{N_+}\sum_{s=1}^{N_+}
                H_{k/L}\!\left(-\BFw^\top\widehat{\BFx}^{\, +}_s-b\right)
                -1-\tau,
        \end{aligned}
    \end{equation}
    and
    \[
        \mathrm{FI}_{\mathrm W}(p;\tau)=\inf\{k>0:G(k)\leq0\}.
    \]
    If $\tau<\mathbb E_{\widehat{\mathbb Q}}[\varepsilon(p)]$, then $\mathrm{FI}_{\mathrm W}(p;\tau)=+\infty$; if $\tau\geq1$, then $\mathrm{FI}_{\mathrm W}(p;\tau)=0$. For
    $\mathbb E_{\widehat{\mathbb Q}}[\varepsilon(p)]\leq\tau<1$, the index is attained and satisfies
    \[
        0<\mathrm{FI}_{\mathrm W}(p;\tau)
        =\min\{k\in(0,L/4]:G(k)\leq0\}\leq L/4.
    \]
\end{propositionAp}

Equation \eqref{eq:logistic_wasserstein_fi_explicit} evaluates the original ranking-error FI by a one-dimensional search over $k$ and $N_-+N_+$ closed-form evaluations of $H_{k/L}$. After precomputing the sample logits, each evaluation of $G(k)$ requires $O(N_-+N_+)$ scalar operations. In the finite positive regime, extend $G$ continuously to $k=0$ by $G(0)=1-\tau$ and apply bisection on $[0,L/4]$, retaining an infeasible lower endpoint and a feasible upper endpoint. The bracket width bounds the absolute error of the returned feasible upper endpoint. When $\tau=\mathbb E_{\widehat{\mathbb Q}}[\varepsilon(p)]$, $G$ may have a zero plateau, so the search must locate the smallest feasible value rather than assume a unique root.

The full-space assumption makes every logit displacement attainable at its norm-duality cost. If the admissible feature domain is restricted, the full-space expression instead upper-bounds the corresponding $G(k)$ and yields a conservative upper bound on FI. For $\BFw=\BFzero$, the score is constant and the ranking error is identically zero, giving $\mathrm{FI}_{\mathrm W}(p;\tau)=0$ for $\tau\geq0$ and $+\infty$ otherwise.
}

\section{Supplementary to FI-Based Training (Section \ref{sec:fi_training})}
\label{appe:fi_training}
{{This section provides supplementary details for }the FI-based training framework. {Section \ref{appe:common_losses_probability_dominance} verifies the probability--loss dominance condition for common classification losses. Section \ref{appe:ce_entropy_refinement} sharpens the generic OVO FI certificate for cross-entropy and gives a controlled numerical illustration. Section \ref{appe:fi_reformulation} develops }the Wasserstein reformulation{, including exact finite formulations valid for arbitrary $\gamma$ and numerical comparisons with their compact counterparts; Section \ref{appe:finite_sample_guarantee} establishes generalization, convergence, }and finite-sample guarantees{, and Section \ref{appe:connection_dro} connects }our framework with DRO and {discusses their integration to mitigate }potential over-conservativeness.}

\subsection{{Common Losses Satisfying Probability--Loss Dominance}}
\label{appe:common_losses_probability_dominance}

\begin{lemmaAp}[Common Losses Satisfying Probability--Loss Dominance]
    \label{lemma:common_losses_probability_dominance}
    {The following losses satisfy Assumption \ref{asmp:probability_loss_dominance}:
    }\begin{enumerate}[(a)]
        \item {The softmax cross-entropy loss
        }{\[
            \ell_{CE}(\BFB^T\BFx,y)
            :=
            \ln\left(\sum_{i\in[C]}\exp(\BFbeta_i^T\BFx)\right)
            -\BFbeta_y^T\BFx
            =-\ln p_y(\BFx;\BFB)
\]}
        {satisfies the assumption with $a=1$ and $b=0$. }
        \item {For any $\kappa>0$, the margin-$\kappa$ multi-class hinge loss
        }{\[
            \ell_{hinge,\kappa}(\BFB^T\BFx,y)
            :=
            \max_{y'\neq y}
            \left(\kappa-(\BFbeta_y-\BFbeta_{y'})^T\BFx\right)_+
\]}
        {satisfies the assumption with $a=1/4$ and $b=(C-1)e^{-\kappa}/\left(1+(C-1)e^{-\kappa}\right)$.
    }\end{enumerate}
\end{lemmaAp}

\subsection{{A Specific Sharpening of the OVO Certificate}}
\label{appe:ce_entropy_refinement}

{The surrogate-to-ranking step in Theorem \ref{theorem:kl_wasserstein_ovo_bound} uses only the generic probability--loss dominance condition. For cross-entropy, its logarithmic structure yields the following sharper induced targets without changing the training problem}.

\begin{theoremAp}[Cross-entropy OVO FI bound]
    \label{theorem:ce_entropy_ovo_bound}
    {{For each $c\in[C]$, let $q_c:=\hat{\mathbb P}(Y=c)>0$. Consider Problem \eqref{eq:prob_general} under either the KL or Wasserstein setting of Theorem \ref{theorem:kl_wasserstein_ovo_bound}. Suppose that the loss is the softmax cross-entropy loss $\ell_{\mathrm{CE}}$, and let $(k,\BFB)$ be any feasible pair with $k>0$. For every $i\neq j$, define
    }{\[
        w_{ij}:=\frac{q_i}{q_i+q_j},\qquad
        H_{\mathrm b}(w):=-w\ln w-(1-w)\ln(1-w),
\]}
    \begin{equation}
        \label{eq:ce_entropy_target}
        t^{\mathrm{CE}}_{ij}(\BFB)
        :=
        \frac{1}{q_i+q_j}
        \left[
        \tau-R(\BFB)
        -\sum_{c\in[C]\setminus\{i,j\}}q_c
        \mathbb E_{\widehat{\mathbb{P}}_c}
        [\ell_{\mathrm{CE}}(\BFB^T\BFx,c)]
        \right]
        -H_{\mathrm b}(w_{ij}),
        \qquad
        t^{\mathrm{CE}}_{\max}(\BFB)
        :=
        \max_{\substack{i,j\in[C]\\i\neq j}}
        t^{\mathrm{CE}}_{ij}(\BFB).
\end{equation}
    {Then, for every $i\neq j$,
    }\begin{equation}
        \label{eq:ce_entropy_pair_bound}
        \mathrm{FI}_{i|j}
        \left(p_i(\cdot;\BFB);t^{\mathrm{CE}}_{ij}(\BFB)\right)
        \leq
        \frac{\max\{q_i,q_j\}}{q_i+q_j}k.
\end{equation}
    {Consequently,
    }\begin{equation}
        \label{eq:ce_entropy_macro_bound}
        \mathrm{FI}
        \left(
        p_1(\cdot;\BFB),\ldots,p_C(\cdot;\BFB);
        t^{\mathrm{CE}}_{\max}(\BFB)
        \right)
        \leq
        \frac{k}{C(C-1)}
        \sum_{\substack{i,j\in[C]\\i\neq j}}
        \frac{\max\{q_i,q_j\}}{q_i+q_j}
        \leq k.
\end{equation}}
\end{theoremAp}

{For cross-entropy, the generic target $t_{ij}(\BFB)$ in Theorem \ref{theorem:kl_wasserstein_ovo_bound} equals the first term on the right-hand side of \eqref{eq:ce_entropy_target}. Thus, $t^{\mathrm{CE}}_{ij}(\BFB)=t_{ij}(\BFB)-H_{\mathrm b}(w_{ij})$. Theorem \ref{theorem:ce_entropy_ovo_bound} therefore retains the same right-hand-side coefficient while certifying FI at a lower target. In the balanced binary case, the target reduction is $H_{\mathrm b}(1/2)=\ln 2$.
}

\paragraph{{Numerical Illustration}} {We examine how closely the entropy-sharpened KL certificate is attained in a fixed balanced binary design. For each run, 100 observations per class are sampled independently from $\mathcal N((-0.75,0)^T,I_2)$ and $\mathcal N((0.75,0)^T,I_2)$, respectively. We fit a logistic classifier with an $\ell_2$ regularization. The training target is $\tau=\lambda L_{\mathrm{ERM}}$ for $\lambda\in\{1.2,1.3,1.4\}$, where $L_{\mathrm{ERM}}$ denotes the minimized regularized empirical cross-entropy. Each setting uses the 50 fixed random seeds. Let $p^*(\cdot):=p(\cdot;\BFB^*)$ denote the probability function induced by the fitted parameter matrix $\BFB^*$. In this case, Theorem \ref{theorem:ce_entropy_ovo_bound} gives
}{\[
    \mathrm{FI}_{\mathrm{KL}}
    \left(p^*;t^{\mathrm{CE}}\right)
    \leq \frac{k^*}{2},
    \qquad
    t^{\mathrm{CE}}=\tau-R(\BFB^*)-\ln 2.
\]}
{We compute the left-hand side exactly from the empirical KL log-moment generating function and report the certificate ratio
}{\[
    \rho_{\mathrm{CE}}
    :=
    \frac{\mathrm{FI}_{\mathrm{KL}}(p^*;t^{\mathrm{CE}})}{k^*/2}.
\]}
{We call a run nontrivial when $k^*>0$ and
$\mathrm{FI}_{\mathrm{KL}}(p^*;t^{\mathrm{CE}})>0$, so that this ratio is
well-defined and informative. A run with $k^*=0$ and
$\mathrm{FI}_{\mathrm{KL}}(p^*;t^{\mathrm{CE}})=0$ is a trivial case. We therefore compute the ratio summaries only over
the nontrivial runs.}

\begin{table}[htbp]
    \small
    \centering
    \caption{{Numerical sharpness of the entropy-sharpened cross-entropy certificate}}
    \label{tab:ce_entropy_bound_sharpness}
    \begin{tabular}{c|c|c|c|c}
        \hline
        {Target ratio $\lambda$ }& {Nontrivial runs }& {Median $\rho_{\mathrm{CE}}$
        }& {5th--95th percentile }& {Maximum $\rho_{\mathrm{CE}}$ }\\
        \hline
        {$1.2$ }& {$50/50$ }& {$0.563$ }& {$0.452$--$0.687$ }& {$0.903$ }\\
        {$1.3$ }& {$48/50$ }& {$0.750$ }& {$0.633$--$0.886$ }& {$0.945$ }\\
        {$1.4$ }& {$38/50$ }& {$0.876$ }& {$0.791$--$0.985$ }& {$0.999$ }\\
        \hline
    \end{tabular}
\end{table}

{Table \ref{tab:ce_entropy_bound_sharpness} shows that the certificate is not uniformly tight, but can be close to attained in nontrivial instances of moderate size.
The maximum ratios are $0.903$, $0.945$, and $0.999$ for $\lambda=1.2$, $1.3$, and $1.4$, respectively. In particular, the last value shows that the bound can be essentially tight in at least one nontrivial run.
}

\subsection{Supplementary to Wasserstein Reformulation Results}\label{appe:fi_reformulation}
We first introduce the $c$-transformation of the loss function as {a }building block for {the }reformulation results in Section \ref{sec:training_wass}. {{We then explain the role of the feature-support constraint in Lemma \ref{lemma:1wass_reformulation}}, {give exact finite formulations for cross-entropy and hinge-type losses for arbitrary $\gamma$, present numerical comparisons with their compact conservative approximations}, {and provide }reformulation results for {piecewise linear loss functions under general convex OT costs}.}

\subsubsection{The $c$-Transformation}
We start from the general case of OT discrepancy. Consider the OT transportation cost function as 
$$
    c(\BFx, y, \hat{\BFx}, \hat{y}) = c(\BFx, \hat{\BFx}) + \gamma \mathbb{I}(y \neq \hat{y}).
$$ 
We next make the following mild assumptions.
\begin{assumptionAp} (Convexity)
    \label{asmp:wass_convexity}
    \begin{enumerate}[(a)]
        \item The uncertainty set {$\mathcal{X} \subseteq \mathbb{R}^M$ }is convex and closed. Moreover, the transportation cost function $c(\BFx, \hat{\BFx})$ is proper, {closed, and }convex in $\BFx$ for every {$\hat{\BFx}\in\mathcal X$, and $\operatorname{ri}(\dom{c(\cdot,\hat{\BFx})})\cap\operatorname{ri}(\mathcal X)\neq\emptyset$}.
        \item The loss function {$\ell(\BFu,y)$ is proper, closed, and convex in $\BFu$ for every $y\in\mathcal Y$, and $\ell(\BFB^T\BFx,y)$ is }integrable with respect to {$\mathbb{P}\in\mathcal{P}(\mathcal X\times\mathcal Y)$ for any $\BFB\in\mathcal B$}.
    \end{enumerate}
\end{assumptionAp}
We {first apply the standard OT transformation used }in DRO and RS to handle the worst-case {distribution.
}

\begin{align*}
    &\phantom{=} \sup_{\mathbb{P}\in \mathcal{P}(\mathcal{X}\times\mathcal{Y})}
    \left\{\mathbb{E}_{\mathbb{P}}[\ell(\BFB^T\BFx,y)]-kW(\mathbb{P},\hat{\mathbb{P}})\right\} \\
    &=\sup_{\mathbb{P}\in \mathcal{P}(\mathcal{X}\times\mathcal{Y})}
    \sup_{\pi\in\Pi(\mathbb{P},\hat{\mathbb{P}})}
    \mathbb{E}_{\pi}\!\left[\ell(\BFB^T\BFx,y)-kc(\BFx,y,\hat{\BFx},\hat y)\right] \\
    &=\mathbb{E}_{(\hat{\BFx},\hat y)\sim\hat{\mathbb{P}}}
    \left[\sup_{(\BFx,y)\in\mathcal X\times\mathcal Y}
    \left\{\ell(\BFB^T\BFx,y)-kc(\BFx,y,\hat{\BFx},\hat y)\right\}\right] \\
    &=\frac{1}{N}\sum_{n\in[N]}\sup_{y\in\mathcal Y}
    \left\{\sup_{\BFx\in\mathcal X}
    \left\{\ell(\BFB^T\BFx,y)-kc(\BFx,\hat{\BFx}_n)\right\}
    -k\gamma\mathbb I(y\neq\hat y_n)\right\}.
\end{align*}
{The second equality follows by disintegrating each coupling with respect to its fixed second marginal $\hat{\mathbb P}$ and optimizing the conditional distribution of $(\BFx,y)$ pointwise for each $(\hat{\BFx},\hat y)$. }The last equality {uses the empirical representation $\hat{\mathbb P}=N^{-1}\sum_{n\in[N]}\delta_{(\hat{\BFx}_n,\hat y_n)}$}.

The {inner maximization $\sup_{\BFx\in\mathcal X}\{\ell(\BFB^T\BFx,y)-kc(\BFx,\hat{\BFx})\}$
}is {the $c$-transformation }with respect to the cost function {$c(\BFx,\hat{\BFx})$ }\citep{tacskesen2023semi}. {We incorporate the feature-support constraint $\BFx\in\mathcal X$ }and call the following expression the {$c$-transformed loss }in our work.

\begin{equation}
    \label{eq:c_transformed_loss}
    \ell_c(\BFB, k, \hat{\BFx}, y) = \sup_{\BFx \in \mathcal{X}} \left\{ \ell(\BFB^T\BFx, y) - k c(\BFx, \hat{\BFx}) \right\}.
\end{equation}

In the context of traditional DRO, \cite{shafieezadeh2023nash} proposed {conditions under which the $c$-transformation admits }a finite convex reformulation. {These conditions require the loss to be }convex in the {decision variable }$\BFB$ but concave in the uncertain {feature }$\BFx$. {For a loss that }is convex in {its first argument, however, the composition $\ell(\BFB^T\BFx,y)$ is convex in }$\BFx$ {for fixed }$\BFB$ {and therefore generally fails the required concavity condition. Consequently, evaluating the inner maximization over }$\BFx$ {can be difficult. Nevertheless, $\ell_c(\BFB,k,\hat{\BFx},y)$ is convex in $\BFB$ because it is the pointwise supremum of functions that are convex in $\BFB$. The following proposition gives a conjugate representation that facilitates }further analysis.

\begin{propositionAp}
    \label{prop:convex_ot_reformulation}
    Under Assumptions \ref{asmp:tau} and \ref{asmp:wass_convexity}, {for every $k>0$, }the c-transformed loss function $\ell_c(\BFB, k, \hat{\BFx}, y)$ defined in equation \eqref{eq:c_transformed_loss} is convex in $\BFB$. Moreover,
    \begin{equation}
        \label{eq:c_transformed_loss_dual}
        \ell_c(\BFB, k, \hat{\BFx}, y) = \sup_{\BFzeta \in \dom{\ell^{1*}}} \inf_{\BFtheta \in \dom{\delta^*_{\mathcal{X}}}} \left\{
            kc^{1*}((\BFB\BFzeta - \BFtheta)/k, \hat{\BFx}) + \delta_{\mathcal{X}}^*(\BFtheta) - \ell^{1*}(\BFzeta, y) 
        \right\}.
    \end{equation}
    In particular, when $\mathcal{X} = \mathbb{R}^M$, meaning that {there is no feature-support constraint on }$\BFx$, we have
    $$
        \ell_c(\BFB, k, \hat{\BFx}, y) = \sup_{\BFzeta \in \dom{\ell^{1*}}} \left\{
            kc^{1*}((\BFB\BFzeta)/k, \hat{\BFx}) - \ell^{1*}(\BFzeta, y)
        \right\}.
    $$
\end{propositionAp}
Recall that the superscript ``1'' in $c^{1*}$ and $\ell^{1*}$ means the convex conjugate to the first variable. Proposition {\ref{prop:convex_ot_reformulation} }is the main building block for Lemma \ref{lemma:1wass_reformulation}, which further helps to derive exact convex reformulations.
The term $\delta_\mathcal{X}(\BFx)$ represents the {indicator }function of $\mathcal{X}$ and $\delta_\mathcal{X}^*(\BFtheta)$ 
is its convex conjugate. For more examples of the {indicator }function and its convex conjugate, we refer to Table B.2 of \cite{kuhn2019wasserstein} for more details.  

\subsubsection{Remark on Lemma \ref{lemma:1wass_reformulation}}
{Proposition \ref{prop:convex_ot_reformulation} allows a general feature-support set $\mathcal X$, whereas Lemma \ref{lemma:1wass_reformulation} takes $\mathcal X=\mathbb R^M$, thereby removing the feature-support constraint to obtain its tractable reformulation. To see why a proper subset $\mathcal X\subsetneq\mathbb R^M$ complicates that reformulation, consider the }1-Wasserstein {cost. Following the proof of Lemma \ref{lemma:1wass_reformulation}, we obtain
}$$
    \ell_c(\BFB, k, \hat{\BFx}, y) =   \sup_{\BFzeta \in \dom{\ell^{1*}}} \inf_{\substack{\BFtheta \in \dom{\delta^*_{\mathcal{X}}} \\ \|\BFB\BFzeta - \BFtheta\|_* \leq k}} \left\{
        (\BFB\BFzeta - \BFtheta)^T\hat{\BFx} + \delta_\mathcal{X}^*(\BFtheta) - \ell^{1*}(\BFzeta, y)
    \right\} 
$$
{Here, the constraint $\|\BFB\BFzeta-\BFtheta\|_*\leq k$ jointly depends on the outer variable $\BFzeta$ and the inner variable $\BFtheta$. Such coupled minimax problems are generally difficult }\citep{tsaknakis2023minimax}{, and the standard conditions for interchanging the supremum and infimum do not directly apply. Thus, to obtain the simple reformulation in \eqref{eq:1wass_reformulation}, we remove the feature-support constraint by taking $\mathcal X=\mathbb R^M$}.

\subsubsection{\texorpdfstring{Exact Finite Formulations for Arbitrary $\gamma$}{Exact Finite Formulations for Arbitrary Gamma}}
\label{appe:exact_wass_reformulations}

{The compact formulations in Theorems \ref{theorem:cross_entropy_reformulation} and \ref{theorem:hinge_type_reformulation} replace the finite maximization over candidate labels in Problem \eqref{eq:1wass_reformulation} by norm-based upper bounds. Retaining that finite maximization through label-wise epigraph inequalities gives exact convex formulations for every nonnegative label transportation cost $\gamma$. We state the two loss-specific formulations separately below.
}

\begin{propositionAp}[Exact cross-entropy formulation for arbitrary $\gamma$]
    \label{prop:cross_entropy_exact_all_gamma}
    {{Suppose the loss is the cross-entropy loss in Equation \eqref{eq:cross_entropy_loss}. For every $\gamma\geq0$, Problem \eqref{eq:1wass_reformulation} is exactly equivalent to the finite convex program
    }\begin{equation}
        \label{eq:cross_entropy_exact_all_gamma}
        \begin{aligned}
            \min_{k\geq0,\,\BFB\in\mathcal B,\,\BFs\in\mathbb R^N}
            \quad & k\\
            \mathrm{s.t.}\quad
            & \frac{1}{N}\sum_{n\in[N]}s_n+R(\BFB)-\tau\leq0,\\
            & \ell_{CE}(\BFB^T\hat{\BFx}_n,y)
            -k\gamma\mathbb I(y\neq\hat y_n)\leq s_n,
            && \forall n\in[N],\ y\in[C],\\
            & \|\BFbeta_i-\BFbeta_j\|_*\leq k,
            && \forall i,j\in[C],\ i<j.
        \end{aligned}
\end{equation}
    {In particular, the formulation contains $NC$ epigraph inequalities and $C(C-1)/2$ pairwise norm inequalities.
}}\end{propositionAp}

\begin{propositionAp}[Exact hinge-type formulation for arbitrary $\gamma$]
    \label{prop:hinge_exact_all_gamma}
    {{Suppose the loss is the hinge-type loss in Equation \eqref{eq:hinge_type_loss}, where $\rho$ satisfies the conditions of Theorem \ref{theorem:hinge_type_reformulation}. For every $\gamma\geq0$, Problem \eqref{eq:1wass_reformulation} is exactly equivalent to the finite convex program
    }\begin{equation*}
        \begin{aligned}
            \min_{k\geq0,\,\BFB\in\mathcal B,\,\BFs\in\mathbb R^N}
            \quad & k\\
            \mathrm{s.t.}\quad
            & \frac{1}{N}\sum_{n\in[N]}s_n+R(\BFB)-\tau\leq0,\\
            & \ell_{hinge}(\BFB^T\hat{\BFx}_n,y)
            -k\gamma\mathbb I(y\neq\hat y_n)\leq s_n,
            && \forall n\in[N],\ y\in[C],\\
            & \|\BFbeta_i-\BFbeta_j\|_*\leq\frac{k}{\theta},
            && \forall i,j\in[C],\ i<j.
        \end{aligned}
\end{equation*}
    {The compact representation contains $NC$ epigraph inequalities. If the inner maximum in Equation \eqref{eq:hinge_type_loss} is fully expanded, these become $NC(C-1)$ scalar convex loss inequalities.
}}\end{propositionAp}

\subsubsection{\texorpdfstring{Numerical Comparison of the Exact and Compact Formulations}{Numerical Comparison of the Exact and Compact Formulations}}
\label{appe:exact_wass_numerical}

{{We compare the exact formulations in Propositions \ref{prop:cross_entropy_exact_all_gamma} and \ref{prop:hinge_exact_all_gamma} with the cross-entropy and hinge-type formulations in Theorems \ref{theorem:cross_entropy_reformulation} and \ref{theorem:hinge_type_reformulation}, respectively. We generate balanced four-class datasets with $N=80$ observations. For class $c$, the two-dimensional features are drawn from $\mathcal N(\BFmu_c,0.9^2I_2)$, where the four class means $\BFmu_c$ are equally spaced on a circle of radius $1.5$. For each loss, we conduct 50 independent replications. In each replication, we first solve the $\ell_2$-regularized ERM problem with $R(\BFB)=0.01\|\BFB\|_F^2$ and set $\tau=1.2L_{\rm ERM}$. We then vary the normalized label transportation cost over
}\begin{equation*}
    r_\gamma\in\{0.70,0.75,\ldots,1.10\},\qquad
    \gamma=
    \begin{cases}
        r_\gamma\max_{n\in[N]}\|\hat{\BFx}_n\|_2,
            & \text{cross-entropy loss},\\
        2r_\gamma\max_{n\in[N]}\|\hat{\BFx}_n\|_2,
            & \text{hinge loss}.
    \end{cases}
\end{equation*}
{By construction, $r_\gamma=1$ corresponds to the sufficient exactness condition in Theorems \ref{theorem:cross_entropy_reformulation} and \ref{theorem:hinge_type_reformulation}. We solve the exact and compact models on the same sample and with the same $(\tau,\gamma)$ using MOSEK 11.2.2.
}}

{{Figure \ref{fig:exact_compact_numerical} compares objective values and feasibility. The upper panels show the median relative objective gap
}\begin{equation*}
    100\times\frac{k_{\rm compact}-k_{\rm exact}}{\max\{k_{\rm exact},10^{-8}\}},
\end{equation*}
{among replications for which both formulations return an optimal solution; the shaded regions show the empirical 10th--90th percentile intervals. The lower panels report the fraction of replications in which each formulation is feasible.
}}

\begin{figure}[htbp]
    \centering
    \includegraphics[width=0.8\textwidth]{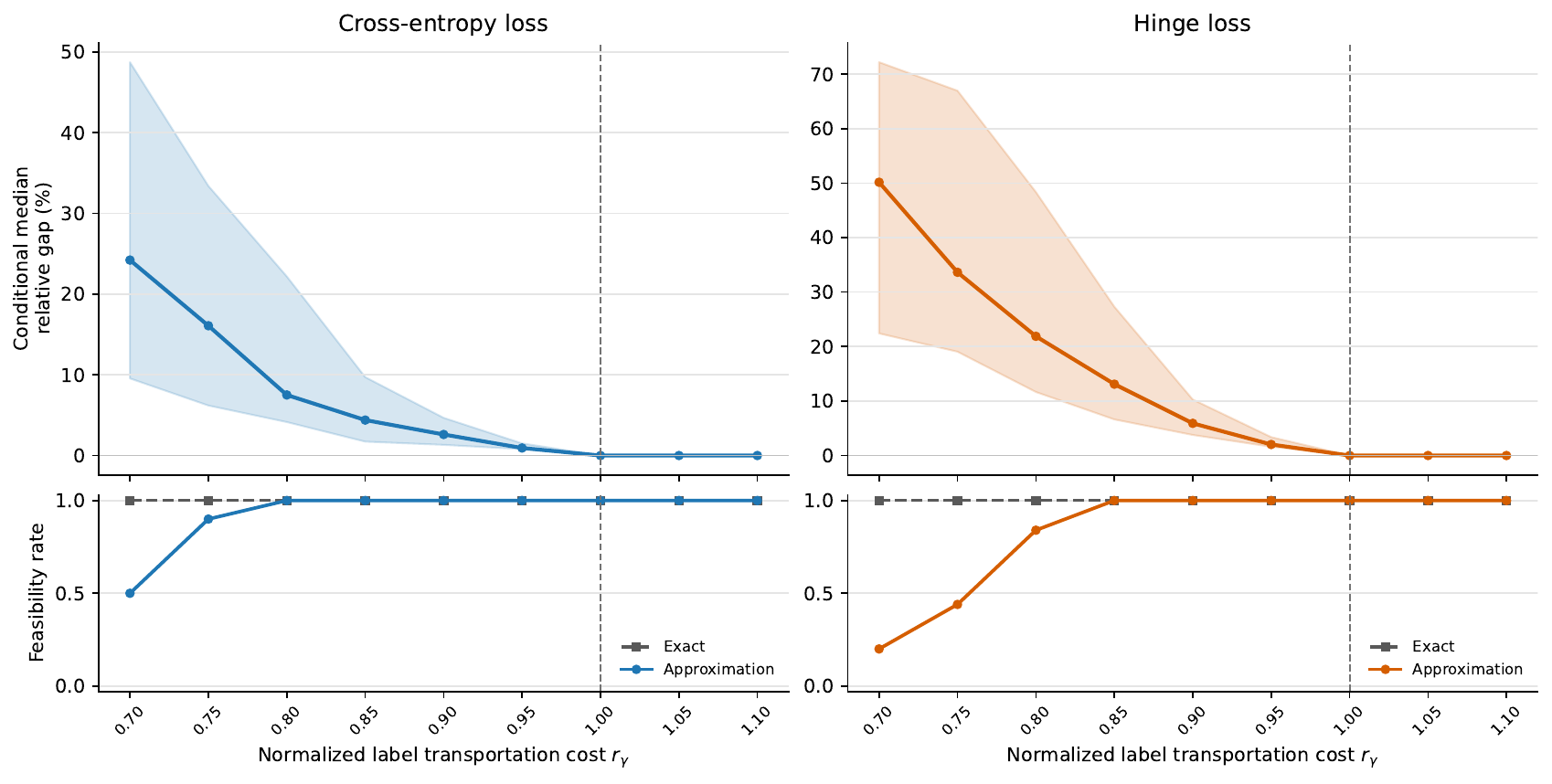}
    \caption{{Comparison of the exact and compact formulations for cross-entropy loss (left) and hinge loss (right). The upper panels report the conditional median relative objective gap and its empirical 10th--90th percentile interval; the lower panels report the exact and compact feasibility rates separately. The vertical dashed line marks the theoretical exactness threshold $r_\gamma=1$.}}
    \label{fig:exact_compact_numerical}
\end{figure}

{The exact formulation is feasible in every replication. The compact formulation is less frequently feasible at smaller values of $r_\gamma$. Among jointly feasible replications, the relative objective gap decreases as $r_\gamma$ approaches $1$. At $r_\gamma=0.80$, the median gaps are $7.50\%$ for cross-entropy loss and $21.88\%$ for hinge loss; at $r_\gamma=0.95$, they fall to $0.93\%$ and $2.03\%$, respectively. At and above $r_\gamma=1$, the two formulations have the same objective value up to solver tolerance.
}

\paragraph{{Computational Scalability}} {We next compare solution times at $r_\gamma=1$, where both formulations are exact. We use the same data-generating process and regularization scheme and consider two scaling paths. In the first, we fix $C=10$ and vary $N\in\{1000,2000,3000,4000,5000,6000\}$; in the second, we fix $N=2000$ and vary $C\in\{5,10,15,20,25,30\}$. For every configuration and loss, we generate 10 independent samples, set $\tau=1.2L_{\rm ERM}$, and solve both formulations using MOSEK 11.2.2. 
}

\begin{figure}[htbp]
    \centering
    \includegraphics[width=0.8\textwidth]{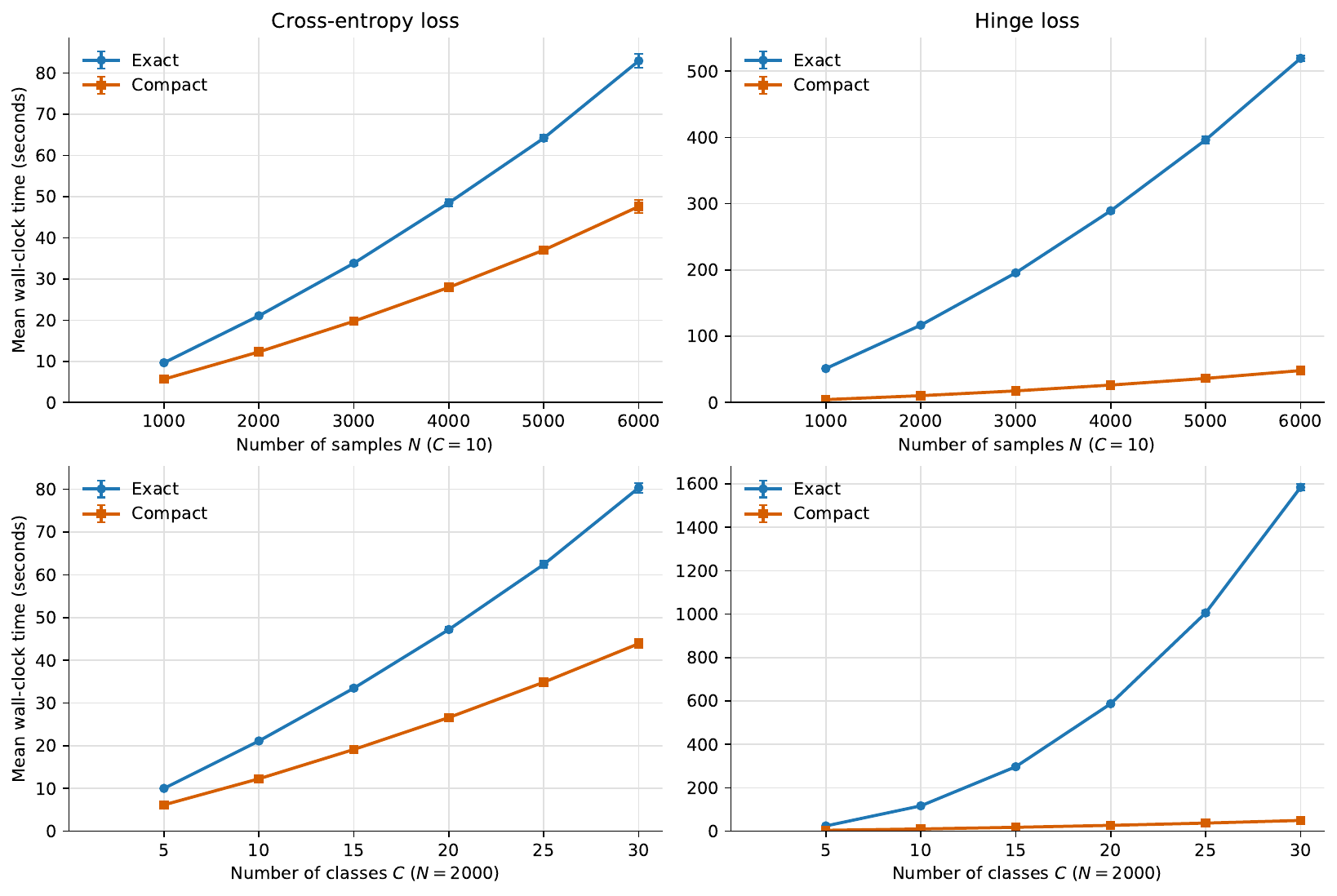}
    \caption{{Solution times of the exact and compact formulations for cross-entropy loss (left) and hinge loss (right). The top panels vary the sample size $N$ with $C=10$, whereas the bottom panels vary the number of classes $C$ with $N=2000$. Markers report the mean over 10 independent replications.}}
    \label{fig:exact_compact_runtime}
\end{figure}

{Figure \ref{fig:exact_compact_runtime} shows that the compact formulation is faster in every setting. For cross-entropy loss, the speedup is approximately $1.7$ along the sample-size path and increases from $1.63$ to $1.83$ as $C$ increases from 5 to 30. The difference is larger for hinge loss: the speedup is $10.76$--$11.46$ along the sample-size path. At $C=30$, the exact hinge formulation contains $1{,}800{,}440$ constraints and requires $1{,}584.53$ seconds on average, whereas the compact formulation contains $58{,}440$ constraints and requires $49.59$ seconds. This difference reflects the $NC(C-1)$ scalar loss inequalities in the fully expanded exact hinge formulation.
}

{Together, the two comparisons show that the exact formulation accommodates arbitrary $\gamma$, whereas the compact formulation offers a substantial computational advantage when its exactness condition holds.
}

\subsubsection{Extending to General Convex OT Costs}\label{appe:general_convex_OT}
We now extend our discussion to general convex OT costs:
$$
    c(\BFx, y, \hat{\BFx}, \hat{y}) = c(\BFx, \hat{\BFx}) + \gamma \mathbb{I}(y \neq \hat{y}).
$$ 
As with previous reformulation results, linearity is key to our analysis. Following the literature, we restrict our discussion to \emph{piecewise linear convex loss functions} \citep{mohajerin2018data,sim2021new}, which include widely used loss functions such as the hinge loss. The following result provides an exact convex reformulation under this setting.

\begin{theoremAp}
    \label{theorem:piecewise_reformulation}
    Suppose {Assumptions \ref{asmp:tau} and \ref{asmp:wass_convexity} hold}, and the loss function $\ell$ is piecewise linear convex as {$\ell(\BFu,y)=\max_{i\in[K_y]}\{\BFa_{yi}^T\BFu+b_{yi}\}$. Then}, the problem \eqref{eq:prob_general} is equivalent to
    \begin{equation*}
        \begin{aligned}
            \min_{k>0, \BFB \in \mathcal{B}, \BFs, \BFv_{iny}} & \ k \\
            \text{s.t.} \hspace*{20pt} & \frac{1}{N} \sum_{n \in [N]} s_n + R(\BFB) - \tau \leq 0, \\
            & b_{yi} + \delta_\mathcal{X}^*(\BFv_{iny}) + k c^{1*}((\BFB \BFa_{yi} - \BFv_{iny})/k, \hat{\BFx}_n ) -  k \gamma \mathbb{I}(y \neq \hat{y}_n) \leq s_n, \\
            &\hspace{200pt} \forall i \in [K_y], n \in [N], y \in \mathcal{Y}.
        \end{aligned}
\end{equation*}
    where the function {$\delta_\mathcal{X}(\BFx)$ represents the indicator }function of $\mathcal{X}$.
\end{theoremAp}
{The following corollary instantiates Theorem \ref{theorem:piecewise_reformulation} under two common forms of the feature-support set.
}

\begin{corollaryAp}
    \label{cor:eg_uncertainty}
    Consider the piecewise linear loss function $\ell(\BFu, y) = \max_{i\in[K_y]} \{\BFa_{yi}^T \BFu + b_{yi}\}$.
    \begin{enumerate}[(a)]
        \item Suppose the uncertainty set is a {polyhedron given by $\mathcal{X} = \{\BFx \in \mathbb{R}^M: \BFA \BFx \leq \BFd \}$}. Then, the problem \eqref{eq:prob_general} is equivalent to 
        \begin{align*}
            \min_{k>0,\BFB \in \mathcal{B}, \BFs, \BFlambda_{iny}} & \ k\\
            \text{s.t.} &\frac{1}{N} \sum_{n \in [N]} s_n + R(\BFB) - \tau \leq 0\\
            & \ b_{yi} + \BFd^T \BFlambda_{iny}  + k c^{1*}((\BFB \BFa_{yi} - \BFA^T \BFlambda_{iny})/k, \hat{\BFx}_n ) -  k \gamma \mathbb{I}(y \neq \hat{y}_n) \leq s_n, & \forall i \in [K_y], n \in [N], y \in \mathcal{Y}, \\
            & \BFlambda_{iny} \geq 0, & \forall i \in [K_y], n \in [N], y \in \mathcal{Y}.
\end{align*}
        \item Suppose the uncertainty set is given by $\mathcal{X} = \{\BFx \in \mathbb{R}^M: f_j(\BFx) \leq 0, j \in [J] \}$, where $f_j(\BFx)$ are convex proper functions{, and there exists $\bar{\BFx}$ such that $f_j(\bar{\BFx})<0$ for every $j\in[J]$}. Then, the problem \eqref{eq:prob_general} is equivalent to
        \begin{align*}
            \min_{k>0,\BFB \in \mathcal{B}, \BFs, \BFlambda_{iny}, \BFz_{inyj}, \BFv_{iny}} & \ k\\
            \text{s.t.} & \frac{1}{N} \sum_{n \in [N]} s_n + R(\BFB) - \tau \leq 0\\
            & \ b_{yi} + \sum_{j\in[J]} \lambda_{inyj} f_j^*\left(\frac{\BFz_{inyj}}{\lambda_{inyj}}\right) \\
            & \quad + k c^{1*}((\BFB \BFa_{yi} - \BFv_{iny})/k, \hat{\BFx}_n ) -  k \gamma \mathbb{I}(y \neq \hat{y}_n) \leq s_n,
            && \forall i \in [K_y], n \in [N], y \in \mathcal{Y}, \\
            & \sum_{j\in[J]} \BFz_{inyj} = \BFv_{iny},
            && \forall i \in [K_y], n \in [N], y \in \mathcal{Y},\\
            & \BFlambda_{iny} \geq 0,
            && \forall i \in [K_y], n \in [N], y \in \mathcal{Y}.
\end{align*}
    \end{enumerate}
\end{corollaryAp}

In addition, Proposition \ref{prop:convex_ot_reformulation} {generalizes }Theorem 3.8(i) of \cite{shafieezadeh2023nash}. {Although evaluating the conjugate representation may still require solving a nonconvex inner problem, its dimension can be smaller}. For example, {when $\mathcal X=\mathbb R^M$, the optimization is over $\BFzeta$, whose dimension is the output dimension of the loss rather than the feature dimension $M$}. For binary classification, $\BFB$ can be reduced to a single vector $\BFbeta$, and {the resulting inner problem is one-dimensional, which may permit }an analytical solution{. Given }an oracle or algorithm {for this inner problem, the convexity of $\ell_c$ in $\BFB$ can be used to construct }a stochastic subgradient for the minimization {over }$\BFB$ in {the training problem. For further details on handling }\eqref{eq:c_transformed_loss_dual}, we refer to \cite{shafieezadeh2023nash}.

\subsection{Finite-Sample Guarantee}
\label{appe:finite_sample_guarantee}

For {a model trained from finite data, a central question is how well its empirical performance and learned parameters reflect the underlying population}. Let $\ptrue$ denote the {data-generating distribution }and $\phat_N$ the empirical distribution of $N$ samples. {Although }$\phat_N$ converges to $\ptrue$ {under standard conditions, this alone does not guarantee }convergence of the {learned objective or classifier parameters without additional regularity conditions}.

{In }high-dimensional {feature spaces, convergence in Wasserstein distance can be slow, so asymptotic consistency alone may provide limited guidance at practical sample sizes }\citep{gao2023finite}{. This motivates finite-sample guarantees with explicit probability and rate statements}.

{We first establish a }finite-sample {generalization bound for the loss and then study the convergence of the learned classifier parameters}.

\subsubsection{Generalization Guarantee on Loss Function}\label{sec:guarantee_on_loss_function}
{An important feature }of the FI-based training framework \eqref{eq:prob_general} is that {its robust-satisficing constraint directly yields a distribution-shift bound for the learned loss}.
\begin{lemmaAp}
    \label{lemma:generalization}
    {Let $(k^*,\BFB^*)$ be an optimal solution to Problem }\eqref{eq:prob_general}. Then, for any distribution {$\mathbb{P} \in \mathcal{P}(\mathcal{X}\times\mathcal{Y})$, we have
    }{$$
    \mathbb{E}_{\mathbb{P}}[\ell \left(\BFB^{*T}\BFx, y\right)] - \mathbb{E}_{\hat{\mathbb{P}}}[\ell \left(\BFB^{*T}\BFx, y\right)] \leq \tau -R(\BFB^*) - \mathbb{E}_{\hat{\mathbb{P}}}[\ell \left(\BFB^{*T}\BFx, y\right)] + k^* D (\mathbb{P}, \hat{\mathbb{P}}).
$$}
\end{lemmaAp}
This bound {follows directly from feasibility in Problem \eqref{eq:prob_general}. The optimal fragility $k^*$ controls how rapidly the loss bound increases with the discrepancy from the empirical distribution.
}

{For a statistical generalization guarantee, we compare the empirical loss under }$\phat_N$ with the expected loss under $\ptrue$ {and impose the following }light-tail {assumption}.
\begin{assumptionAp}
    \label{asmp:light_tail}
    (Light-tailed distribution) There is an exponent $a > 1$ such that $\mathbb{E}_{\ptrue}[\exp((\|(\BFx, y)\|)^a)] < \infty${, where $\|(\BFx,y)\|:=\|\BFx\|+|y|$}.
\end{assumptionAp}
{Under the boundedness conditions stated below, we }obtain the following {high-probability }generalization guarantee of order $O(N^{-\frac{1}{M + 1}})$.
\begin{propositionAp}
    \label{prop:generalization_error}
    Suppose Assumptions \ref{asmp:lipschitz} and \ref{asmp:light_tail} hold, {$M\geq2$, $\gamma>0$, $\sup_{(\BFx,y)\in\supp{\ptrue}}\|\BFx\|<\infty$, and $\mathcal B$ is compact. Let $(\hat k_N,\hat{\BFB}_N)$ be an optimal solution to Problem \eqref{eq:prob_general} }under the empirical distribution $\phat_N$ and {the }1-Wasserstein distance in {Equation }\eqref{eq:def_ot}. For {any $\epsilon\in(0,1)$ and all sufficiently large $N$}, with probability at least $1 - \epsilon$, we have
    \begin{equation*}
        \begin{aligned}
            & \mathbb{E}_{\ptrue}[\ell (\hat{\BFB}_N^T\BFx, y)] - \mathbb{E}_{\phat_N}[\ell (\hat{\BFB}_N^T\BFx, y)] \leq 
            \min \left\{
                (1+\gamma)\omega \|\hat{\BFB}_N^T\|_* \left(
                \frac{1}{C_2 N} \log \left(\frac{C_1}{\epsilon}\right)
            \right)^{\frac{1}{M+1}}, \right.\\
            &\hspace{150pt} \left.
            \tau -R(\hat{\BFB}_N)- \mathbb{E}_{\phat_N}[\ell (\hat{\BFB}_N^T\BFx, y)] + (1+\gamma) \hat k_N \left(
                \frac{1}{C_2 N} \log \left(\frac{C_1}{\epsilon}\right)
            \right)^{\frac{1}{M+1}}
            \right\},
        \end{aligned}
\end{equation*}
    where $C_1$ and $C_2$ are {positive }constants depending on the feature dimension $M${, the light-tail }exponent $a${, and the exponential moment }in Assumption \ref{asmp:light_tail}, and {$\omega = \max\left\{\omega_1, \omega_2 \frac{\sup_{(\BFx, y) \in \supp{\ptrue}} \|\BFx\|}{\gamma}\right\}$}.
\end{propositionAp}
The bound can be improved in constant factors by appropriately shrinking the ambiguity set $\mathcal{P}(\mathcal{X}, \mathcal{Y})$, and the details are deferred to Appendix \ref{appe:connection_dro}.

\subsubsection{Finite-Sample Guarantee of the Learned Classifier}\label{sec:Guarantee_of_learned_classifier} \hfill

\paragraph{Setup} {We study }the convergence and {finite-sample behavior of classifiers trained under the FI framework}.
{We define the population counterpart of Problem \eqref{eq:prob_general} by replacing its nominal empirical distribution }$\phat_N$ with {the data-generating distribution }$\ptrue${.
}\begin{equation}
    \label{eq:prob_true}
    \begin{aligned}
        \min_{k\geq 0, \BFB \in \mathcal{B}} & \ k\\
        \text{s.t.}\  &\ \mathbb{E}_{\mathbb{P}}[\ell \left(\BFB^T\BFx, y\right)] + R(\BFB)\leq \tau + k D (\mathbb{P}, \ptrue), &\ \forall \mathbb{P}\in \mathcal{P}(\mathcal{X}\times\mathcal{Y}).\\
    \end{aligned}
\end{equation}
Let {$(k^*,\BFB^*)$ be an optimal solution to Problem \eqref{eq:prob_true}, and let $(\hat{k}_N,\hat{\BFB}_N)$ be an optimal solution to its empirical counterpart }under $\phat_N$. {Because }$\ptrue$ is generally {unknown, we use the empirical solution }to estimate the {population solution. This is a }sample average approximation (SAA) {problem }\citep{shapiro2021lectures}.

{We restrict attention to cases in which the FI-based training problem }admits a finite convex reformulation. {Let $\mathbb Q$ denote the nominal distribution, with $\mathbb Q=\ptrue$ for the population problem and $\mathbb Q=\phat_N$ for the empirical problem, and let $\BFbeta:=\operatorname{vec}(\BFB)\in\mathbb R^{MC}$. We consider reformulations of the form
}\begin{equation}
    \label{eq:reformulation_for_convergence}
    \begin{aligned}
        \min_{k,\BFbeta} & \ k\\
        \text{s.t.}\  & \mathbb{E}_{\mathbb Q} [g_1(k,\BFbeta;\BFx,y)] \leq 0, \\
        & g_i(k,\BFbeta) \leq 0, &\forall i = 2, \dots, S,
    \end{aligned}
\end{equation}
{Here, }$g_1$ {is the pointwise stochastic constraint, while $g_i$}, {$i=2,\ldots,S$, are deterministic convex constraints. All deterministic feasibility restrictions, including $k\geq0$ and $\BFB\in\mathcal B$, are included among these constraints. For the KL reformulation, }the {exact pointwise representation for $k>0$ is
}{\[
    g_1^{\mathrm{KL}}(k,\BFbeta;\BFx,y)
    :=k\exp\!\left(\frac{\ell(\BFB^T\BFx,y)+R(\BFB)-\tau}{k}\right)-k.
\]}
{Indeed, $\mathbb E_{\mathbb Q}[g_1^{\mathrm{KL}}]\leq0$ is equivalent to the KL constraint in Theorem \ref{theorem:kl_reformulation}, with $\mathbb Q$ replacing $\phat$. Under the equivalence conditions of the }1-Wasserstein {reformulations, we may take
}{\[
    g_1^{\mathrm W}(k,\BFbeta;\BFx,y)
    :=\ell(\BFB^T\BFx,y)+R(\BFB)-\tau,
\]}
{while the remaining constraints include, for example, the pairwise norm constraints }in Theorem \ref{theorem:cross_entropy_reformulation}.

{We analyze Problem \eqref{eq:reformulation_for_convergence} through }the stochastic generalized {equation }(SGE) {associated with its Karush--Kuhn--Tucker }(KKT) system \citep{shapiro2021lectures}. {The following conditions provide the smoothness and local regularity needed for this analysis}.
\begin{assumptionAp}
    \label{asmp:smooth_regularity}
    \begin{enumerate}[(a)]
        \item (Smoothness) The {function $g_1(k,\BFbeta;\BFx,y)$ is twice continuously differentiable in $(k,\BFbeta)$ for $\ptrue$-almost every $(\BFx,y)$, and $g_i(k,\BFbeta)$, $i=2,\ldots,S$, are twice continuously differentiable}.

        \item ({Regularity) The population problem with $\mathbb Q=\ptrue$ satisfies the linear }independence constraint qualification (LICQ) and the strong second-order {sufficient }condition (SSOSC) {at $(k^*,\BFbeta^*)$, and its associated KKT multipliers satisfy strict complementarity}.
    \end{enumerate}
\end{assumptionAp}
Both assumptions are standard in the convex optimization.
{Assumption \ref{asmp:smooth_regularity} ensures that the KKT system is locally strongly regular at the population solution. For a nonsmooth loss }such as the hinge loss, the {analysis applies to a smooth approximation whenever the stated conditions hold. Let $\BFeta\in\mathbb R^S$ denote the multipliers associated with the constraints in Problem \eqref{eq:reformulation_for_convergence}. For a fixed observation $(\BFx,y)$, define the pointwise Lagrangian as
}{$$
    \mathcal{L}(k, \BFbeta, \BFeta; \BFx, y) = k + \eta_1 g_1(k,\BFbeta;\BFx,y) + \sum_{i = 2}^S \eta_i g_i(k,\BFbeta).
$$}
{Collect the primal and dual variables and define the corresponding population mapping by
}\begin{equation*}
    \BFw = \left(\begin{array}{c}
        k \\
        \BFbeta \\
        \BFeta
    \end{array} \right) \in \mathbb{R}^{1 + MC + S }, \ 
    \BFPsi(\BFw; \BFx, y) = \nabla_{\BFw} \mathcal{L}(\BFw; \BFx, y)\ \text{and}\ \BFPsi(\BFw) = \mathbb{E}_{\ptrue} [ \BFPsi(\BFw; \BFx, y) ].
\end{equation*}
For convenience, let $M_w = MC + S + 1$ be the dimension of vector $\BFw$. We define the projection operator $P_k \BFw = k$, $P_\beta \BFw = \BFbeta$, and $P_\eta \BFw = \BFeta$. {Define the complementarity mapping by
}\begin{align*}
    \BFGamma(\BFw) = \{\BFgamma \in \mathbb{R}^{M_w} \mid P_k \BFgamma = 0, P_\beta \BFgamma = 0, P_\eta \BFw\geq\BFzero, P_\eta \BFgamma \geq \BFzero, (P_\eta \BFgamma)\circ (P_\eta \BFw) = 0 \},
\end{align*}
where $\circ$ denotes the Hadamard product. 
{The population KKT system can then be written as the }SGE
\begin{equation}
    \label{eq:sge_true}
    \BFzero \in \BFPsi(\BFw)  + \BFGamma(\BFw).
\end{equation}
{For the empirical problem, define }$\hat{\BFPsi}_N(\BFw) = \mathbb{E}_{\phat_N} [ \BFPsi(\BFw; \BFx, y) ]${, so its KKT system is }$\BFzero \in \hat{\BFPsi}_N(\BFw) + \BFGamma(\BFw)$. Let $\BFw^*$ and $\hat{\BFw}_N$ denote {solutions to the population and empirical SGEs}, respectively. {We impose the following additional conditions}.
\begin{assumptionAp}
    \label{asmp:sge_bound_integrable}
    \begin{enumerate}[(a)]
        \item (Boundedness) There exists a compact set {$\mathcal{W}\subset\mathbb{R}^{M_w}$ such that $\BFw^*$ is the unique population SGE solution in $\mathcal W$ and $\hat{\BFw}_N\in\mathcal W$ for all sufficiently large $N$. For the KL reformulation, $\inf_{\BFw\in\mathcal W}P_k\BFw>0$. }\item (Integrable domination) {The functions $\sup_{\BFw\in\mathcal W}\|\BFPsi(\BFw;\BFx,y)\|$ and $\sup_{\BFw\in\mathcal W}\|\nabla_{\BFw}\BFPsi(\BFw;\BFx,y)\|$ }are dominated by {a $\ptrue$-integrable function on $\mathcal X\times\mathcal Y$}.
    \end{enumerate}
\end{assumptionAp}
Assumption \ref{asmp:sge_bound_integrable}(a) {localizes the relevant SGE solutions and excludes the singularity at $k=0$ in the KL formulation. }Assumption \ref{asmp:sge_bound_integrable}(b) {provides the uniform integrability needed to interchange differentiation and expectation and to apply the SAA convergence results}.

\paragraph{Asymptotics and {Finite-Sample Guarantees}} With the above assumptions, we can establish the convergence and asymptotics of the SGE.
\begin{propositionAp}
    \label{prop:convergence}
    Under {Assumptions }\ref{asmp:smooth_regularity} and \ref{asmp:sge_bound_integrable}, we have
    \begin{enumerate}[(a)]
        \item (Strong regularity) The SGE \eqref{eq:sge_true} is strongly regular at $\BFw^*$.    
        \item (Convergence) {The empirical solution }$\hat{\BFw}_N$ converges to $\BFw^*$ almost surely as $N \to \infty$.
        \item ({Asymptotic normality) Let $\BFeta_+^*$ collect the }positive components of $\BFeta^*$, {define $\BFw'^*:=(k^*,\BFbeta^{*T},\BFeta_+^{*T})^T$, and let }$\BFPsi'$ {be the corresponding reduced KKT mapping. Suppose
        }{\[
            A:=\nabla_{\BFw'}\BFPsi'(\BFw'^*)
\]}
        {is nonsingular and $\mathbb E_{\ptrue}[\|\BFPsi'(\BFw'^*;\BFx,y)\|^2]<\infty$. Then
        }{\[
            N^{1/2}(\hat{\BFw}_N'-\BFw'^*)
            \xrightarrow{d}
            \mathcal N\!\left(\BFzero,A^{-1}\Sigma'A^{-T}\right),
\]}
        where {$\hat{\BFw}_N'$ is the empirical reduced solution and $\Sigma':=\operatorname{Cov}_{\ptrue}(\BFPsi'(\BFw'^*;\BFx,y))$}.
    \end{enumerate}
\end{propositionAp}

Proposition \ref{prop:convergence} {establishes almost-sure consistency and asymptotic normality. We next derive exponential deviation bounds under stronger moment conditions}.
\begin{assumptionAp}
    \label{asmp:lipschitz_sge_finite_mgf}
    \begin{enumerate}[(a)]
        \item (Lipschitz {modulus}) There exists {a $\ptrue$-integrable function $\kappa_\BFPsi(\BFx,y)$ such that
        }{$$
            \|\BFPsi(\BFw_1; \BFx, y) - \BFPsi(\BFw_2; \BFx, y) \| \leq \kappa_\BFPsi(\BFx,y) \|\BFw_1 - \BFw_2\|, \ \forall \BFw_1, \BFw_2 \in \mathcal{W}.
$$}
        \item (Finite moment generating function) Define
        \begin{align*}
            M_i(t;\BFw) &:= \mathbb{E}_{\ptrue} \left[ \exp \left( t \left( \BFPsi_i(\BFw; \BFx, y) - \BFPsi_i(\BFw) \right) \right) \right], \\
            M_\kappa(t) &:= \mathbb{E}_{\ptrue} \left[ \exp \left( t \kappa_\BFPsi(\BFx,y) \right) \right].
\end{align*}
        {There exists a neighborhood $\mathcal T$ of zero such that $\sup_{\BFw\in\mathcal W}M_i(t;\BFw)<\infty$ for every $i\in[M_w]$ }and {$M_\kappa(t)<\infty$ for every $t\in\mathcal T$}.
    \end{enumerate}
\end{assumptionAp}

Assumption \ref{asmp:lipschitz_sge_finite_mgf} (a) also means the Lipschitz continuity of $\BFPsi(\BFw)$, which further refers to the Lipschitz continuity of the constraint $g_1$ and the loss function $\ell$. If only consider the cases of the KL-divergence and 1-Wasserstein distance that we have discussed, this assumption can be reduced to the Lipschitz continuity requirement of the loss function $\ell$ as Assumption \ref{asmp:lipschitz}. Assumption \ref{asmp:lipschitz_sge_finite_mgf} (b) is a standard assumption on the moment generating function for the large deviation theorem. {Assumption \ref{asmp:lipschitz_sge_finite_mgf} provides uniform control of the KKT sample mapping needed for the following deviation bounds.
}\begin{propositionAp}
    \label{prop:exponential_convergence}
    Suppose {Assumptions }\ref{asmp:smooth_regularity}, \ref{asmp:sge_bound_integrable}, {and }\ref{asmp:lipschitz_sge_finite_mgf} hold. Then, the following statements hold.
    \begin{enumerate}[(a)]
        \item For sufficiently small $\epsilon > 0$, there exist positive constants $\delta_1 (\epsilon)$ and $\delta_2 (\epsilon)$, which are independent of $N$, such that 
        \begin{equation}
            \label{eq:exp_converge_psi}
            (\ptrue)^{\otimes N} \left\{
                \sup_{\BFw \in \mathcal{W}} \left\| \hat{\BFPsi}_N(\BFw) - \BFPsi(\BFw) \right\| \geq \epsilon
                \right\} \leq \delta_1(\epsilon) \exp(-\delta_2(\epsilon) N).
\end{equation}

        \item Define the function 
        $$
            \rho(\epsilon) = \inf_{\BFw \in \mathcal{W}, \BFgamma \in \BFGamma(\BFw), \|\BFw - \BFw^*\| \geq \epsilon} \| \BFPsi(\BFw) + \BFgamma \|.
        $$
        Then, for sufficiently small $\epsilon > 0$, we have 
        {$$
            (\ptrue)^{\otimes N} \left\{
                \left\|\hat{\BFw}_N - \BFw^* \right\| \geq \epsilon
            \right\} \leq \delta_1(\rho(\epsilon)) \exp(-\delta_2(\rho(\epsilon)) N).
$$}
    \end{enumerate}
\end{propositionAp}

Proposition \ref{prop:exponential_convergence} shows that{, for any fixed tolerance, the deviation probabilities for the empirical KKT mapping and solution decay exponentially with $N$}. This is impressive as it indicates that a relatively small sample size can already lead to a very good estimation. The exponential convergence of $\hat{\BFw}_N$ to $\BFw^*$ is sound as $\BFw$ includes the fragility $k$ and the weight matrix $\BFB$, which ensures the model trained on the finite dataset can be very close to the model under the true distribution. We also highlight the exponential convergence of $\hat{\BFPsi}_N(\BFw)$ because $\hat{\BFPsi}_N(\BFw)$ is the gradient of the Lagrangian function. During the iterative update with batch gradient, the exponential convergence of $\hat{\BFPsi}_N(\BFw)$ provides a theoretical guarantee for the gradient estimation when the batch size is relatively limited. 

Our analysis so far shows that if the model admits a convex reformulation, we can establish the exponential convergence from the empirical model to the model under the true distribution with various mild assumptions. The conclusions are independent of any parameters such as the target $\tau$ and the regularization $R(\BFB)$. However, to further understand the optimal solution $k^*$ and $\BFB^*$ of the problem \eqref{eq:prob_true}, the target $\tau$ is the most important meta-parameter. For more discussion on how the target $\tau$ could affect the statistical properties of the model, we refer to \cite{li2024statistical}.

\subsection{Connection with DRO}
\label{appe:connection_dro}
This section is devoted to uncovering the connection between our FI-based training framework and the conventional DRO framework. As mentioned by \cite{long2023robust}, DRO and RS are closely related. We first reimplement the main reformulation results in the DRO framework for comparison. Then, we also propose a way to integrate the RS and DRO framework to counter the potential over-conservativeness. However, we highlight that the close connection with DRO does not undermine our contribution. The DRO results are also built on the new reformulation techniques developed in our paper. Moreover, starting from FI, we propose diverse insights about risk and generalization beyond the conventional interpretation of DRO.

\subsubsection{DRO Results}\label{sec:DRO_results}
Regarding the reformulation procedure, the two frameworks are quite similar and share the same key steps of addressing the c-transformed loss function {in }Proposition \ref{prop:convex_ot_reformulation}. The main difference lies in the uncertainty control parameter, where DRO uses the radius $\epsilon$ to control the uncertainty size, while our FI-based training uses the {training fragility }$k$ to control the fragility. Considering the similarity, we mainly reimplement the main reformulation results in the DRO framework for comparison.

Since literature has already investigated much about DRO reformulation, we mainly focus on novel results and give reference to some existing results. For the KL-divergence, we refer to \cite{hu2013kullback}. For the Wasserstein distance with piecewise convex loss, the reformulation refers to Theorem 4.2 of \cite{mohajerin2018data}. These are existing results. Then, we turn to the novel cross-entropy loss and {hinge-type loss. Define the Wasserstein ambiguity set by $\mathcal{P}(\phat, \epsilon):=\{\mathbb P\in\mathcal P(\mathcal X\times\mathcal Y):W(\mathbb P,\phat)\leq\epsilon\}$. By standard Wasserstein DRO duality, the corresponding problem can be written as
}\begin{equation*}
    \begin{aligned}
        &\phantom{=} \min_{\BFB \in \mathcal{B}} \sup_{\mathbb{P}\in \mathcal{P}(\phat, \epsilon)} \mathbb{E}_{\mathbb{P}}[\ell(\BFB^T\BFx, y)] + R(\BFB)\\
        &= \min_{\BFB \in \mathcal{B}} \sup_{\mathbb{P}\in \mathcal{P}(\phat, \epsilon)} \min_{\lambda \geq 0} \mathbb{E}_{\mathbb{P}}[\ell(\BFB^T\BFx, y)] + R(\BFB) + \lambda \left( \epsilon - D_{\mathrm{W}}(\mathbb{P}, \phat) \right)\\
        &= \min_{\BFB \in \mathcal{B}, \lambda \geq 0} \lambda\epsilon + R(\BFB) + \sup_{\mathbb{P}\in \mathcal{P}(\mathcal{X}, \mathcal{Y})} \mathbb{E}_{\mathbb{P}}[\ell(\BFB^T\BFx, y)] - \lambda D_{\mathrm{W}}(\mathbb{P}, \phat) \\
        &= \min_{\BFB \in \mathcal{B}, \lambda \geq 0} \lambda\epsilon + R(\BFB) + \frac{1}{N} \sum_{n \in [N]} \sup_{y_n \in \mathcal{Y}} \left\{ \sup_{\BFx \in \mathcal{X}} \left\{ \ell(\BFB^T\BFx, y_n) - \lambda c(\BFx, \hat{\BFx}_n) \right\} - \lambda \gamma \mathbb{I}(y_n \neq \hat{y}_n) \right\}.
    \end{aligned}
\end{equation*}
Notice that the reformulation is similar to the FI-based training if replacing the dual variable $\lambda$ with the {training }fragility $k$.

\begin{propositionAp}
    \label{prop:dro_reformulation}
    {Suppose $\mathcal X=\mathbb R^M$.
    }\begin{enumerate}[(a)]
        \item Consider the cross-entropy loss {$\ell(\BFB^T\hat{\BFx}_n, \hat{y}_n) = \ln \left(\sum_{i \in [C]} \exp(\BFbeta_i^T \hat{\BFx}_n)\right) - \BFe_{\hat{y}_n}^T \BFB^T \hat{\BFx}_n$. The following convex problem provides an upper approximation to the }DRO problem with 1-Wasserstein distance{:
        }\begin{equation*}
            \begin{aligned}
                \min_{\lambda\geq0, \BFB \in \mathcal{B}} &\ \lambda\epsilon + R(\BFB) + \frac{1}{N} \sum_{n \in [N]} \left[\ell(\BFB^T\hat{\BFx}_n, \hat{y}_n) + \lambda (\|\hat{\BFx}_n\| - \gamma)_+\right] \\
                \text{s.t.} \hspace{10pt} &\ \|\BFbeta_i - \BFbeta_j\|_* \leq \lambda, \ \forall i, j \in [C] \ \text{and}\ i < j.
            \end{aligned}
\end{equation*}
        When $\gamma \geq \max_{n\in[N]} \|\hat{\BFx}_n\|$, the {approximation is exact}.

        \item Consider the hinge-type loss $\ell(\BFB^T\BFx, y) = \max_{y' \neq y} \rho((\BFbeta_{y} - \BFbeta_{y'})^T \BFx)$. Suppose the surrogate loss function $\rho$ is convex and subdifferentiable, and {for some $\theta>0$, }$\sup_{u\in\mathbb{R}} \partial \rho(u) = 0$ and $\inf_{u\in\mathbb{R}} \partial \rho(u) = - \theta$. The {following convex problem provides an upper approximation to the }DRO problem with 1-Wasserstein distance{:
        }\begin{equation*}
            \begin{aligned}
                \min_{\lambda\geq0, \BFB \in \mathcal{B}} &\ \lambda\epsilon + R(\BFB) + \frac{1}{N} \sum_{n \in [N]} \left[\ell(\BFB^T\hat{\BFx}_n, \hat{y}_n) + \lambda (2 \|\hat{\BFx}_n\| - \gamma)_+\right] \\
                \text{s.t.} \hspace{10pt} &\ \|\BFbeta_i - \BFbeta_j\|_* \leq \frac{\lambda}{\theta}, \ \forall i, j \in [C] \ \text{and}\ i < j.
            \end{aligned}
\end{equation*}
        When $\gamma \geq 2 \max_{n\in[N]} \|\hat{\BFx}_n\|$, the {approximation is exact}.
    \end{enumerate}
\end{propositionAp}
Proposition \ref{prop:dro_reformulation} demonstrates the close relation between the FI-based reformulation and the DRO reformulation. The structure of regulating the weight difference remains the same in the two frameworks. The difference lies in the objective function, where the FI-based training minimizes the fragility $k$ while the DRO minimizes the worst-case expected loss.

\subsubsection{\texorpdfstring{Incorporating DRO and RS by Shrinking the Ambiguity Set}{Incorporating DRO and RS by Shrinking the Ambiguity Set}}\label{sec:shrinking_ambiguity}
A constant concern about the model with distributional ambiguity is the over-conservativeness. From the perspective of DRO, the ambiguity radius $\epsilon$ is a hyperparameter that controls the trade-off between the robustness and the in-sample performance. To counter the potential over-conservativeness, the value of $\epsilon$ should be carefully tuned. In RS and FI-based training, we allow a much larger space of ambiguity, but consider the fragility $k$, which is the target violation divided by the statistical distance. The fragility $k$ may not suffer from the expansion of the ambiguity set because it is normalized by the statistical distance. However, in case the ambiguity set may still be too large, we can control the ambiguity in FI-based training directly by integrating the RS and DRO framework. 

Instead of allowing the distribution $\mathbb{P}$ to be arbitrary in the general ambiguity set {$\mathcal{P}(\mathcal{X}\times\mathcal{Y})$}, we can restrict the ambiguity set to be the same as $\mathcal{P}(\phat, \epsilon)$ in DRO. Then, our problem becomes
\begin{equation}
    \label{eq:fi_dro}
    \begin{aligned}
        \min_{k\geq 0, \BFB \in \mathcal{B}} & \ k\\
		\text{s.t.}\  &\ \mathbb{E}_{\mathbb{P}} \left[\ell(\BFB^T\BFx, y) \right] + R(\BFB)\leq \tau + k W(\mathbb{P},\phat), &\ \forall \mathbb{P}\in \mathcal{P}(\phat, \epsilon).\\
    \end{aligned}
\end{equation}
{The same Wasserstein duality used above gives the key reformulation step:
}\begin{equation*}
    \begin{aligned}
        &\phantom{=} \sup_{\mathbb{P}\in \mathcal{P}(\phat, \epsilon)} \left\{\mathbb{E}_{\mathbb{P}}[\ell(\BFB^T\BFx, y)] - kW(\mathbb{P}, \phat)\right\}\\
        &= \min_{\lambda \geq 0} \lambda\epsilon + \frac{1}{N} \sum_{n \in [N]} \sup_{y_n \in \mathcal{Y}} \left\{ \sup_{\BFx \in \mathcal{X}} \left[ \ell(\BFB^T\BFx, y_n) - (\lambda + k) \|\BFx-\hat{\BFx}_n\| \right] - (\lambda + k) \gamma \mathbb{I}(y_n \neq \hat{y}_n) \right\}.
    \end{aligned}
\end{equation*}
Due to the close connection between the FI-based training and the DRO, their integration is straightforward and the reformulation is also similar. For example, we can reimplement the results of cross-entropy loss and {hinge-type }loss as 
\begin{propositionAp}
    \label{prop:dro_fi_reformulation}
    {Suppose $\mathcal X=\mathbb R^M$.
    }\begin{enumerate}[(a)]
        \item Consider the cross-entropy loss $\ell(\BFB^T\hat{\BFx}_n, \hat{y}_n) = \ln \left(\sum_{i \in [C]} \exp(\BFbeta_i^T \hat{\BFx}_n)\right) - \BFe_{\hat{y}_n}^T \BFB^T \hat{\BFx}_n$. The {following convex problem provides an upper approximation to Problem \eqref{eq:fi_dro} with the }1-Wasserstein distance{:
        }\begin{equation*}
            \begin{aligned}
                \min_{k\geq0, \lambda\geq0, \BFB \in \mathcal{B}} & \ k \\
                \text{s.t.} \hspace*{10pt} & \lambda \epsilon + \frac{1}{N} \sum_{n \in [N]} \left[\ell(\BFB^T\hat{\BFx}_n, \hat{y}_n) + (k + \lambda) (\|\hat{\BFx}_n\| - \gamma)_+\right] + R(\BFB) - \tau \leq 0,\\
                & \|\BFbeta_i - \BFbeta_j\|_* \leq k + \lambda, \ \forall i, j \in [C] \ \text{and}\ i < j.
            \end{aligned}
\end{equation*}
        When $\gamma \geq \max_{n\in[N]} \|\hat{\BFx}_n\|$, the {approximation is exact}.

        \item Consider the hinge-type loss $\ell(\BFB^T\BFx, y) = \max_{y' \neq y} \rho((\BFbeta_{y} - \BFbeta_{y'})^T \BFx)$. Suppose the surrogate loss function $\rho$ is convex and subdifferentiable, and {for some $\theta>0$, }$\sup_{u\in\mathbb{R}} \partial \rho(u) = 0$ and $\inf_{u\in\mathbb{R}} \partial \rho(u) = - \theta$. The {following convex problem provides an upper approximation to Problem \eqref{eq:fi_dro} with the }1-Wasserstein distance{:
        }\begin{equation*}
            \begin{aligned}
                \min_{k\geq0, \lambda\geq0, \BFB \in \mathcal{B}} & \ k \\
                \text{s.t.} \hspace*{10pt} & \lambda \epsilon + \frac{1}{N} \sum_{n \in [N]} \left[\ell(\BFB^T\hat{\BFx}_n, \hat{y}_n) + (k + \lambda) (2\|\hat{\BFx}_n\| - \gamma)_+\right] + R(\BFB) - \tau \leq 0,\\
                & \|\BFbeta_i - \BFbeta_j\|_* \leq \frac{k + \lambda}{\theta}, \ \forall i, j \in [C] \ \text{and}\ i < j.
            \end{aligned}
\end{equation*}
        When $\gamma \geq 2 \max_{n\in[N]} \|\hat{\BFx}_n\|$, the {approximation is exact}.
    \end{enumerate}
\end{propositionAp}

Another important implication of the restricted FI-based model \eqref{eq:fi_dro} {concerns }the generalization guarantee. {Because $\mathcal{P}(\mathcal{X}\times\mathcal{Y})$ is broad, the }bound in Lemma \ref{lemma:generalization} {may be loose in practice. Restricting $\mathbb{P}$ }to $\mathcal{P}(\phat, \epsilon)$ {can yield a tighter bound. Let $(\hat k_N,\hat{\BFB}_N)$ be an optimal solution to Problem \eqref{eq:prob_general} under $\phat_N$ and the 1-Wasserstein distance, and let }$k^\dagger_{N, \epsilon}$ {be the optimal value of Problem \eqref{eq:fi_dro} under $\phat_N$ with $\BFB=\hat{\BFB}_N$}. We have the following result.
\begin{propositionAp}
    \label{prop:generalization_fi_dro}
    {Under the conditions of Proposition \ref{prop:generalization_error}, we have $\hat k_N \geq k^\dagger_{N, \epsilon}$ }for any $N$. For {any $\delta\in(0,1)$ and all sufficiently large $N$, suppose that }$\epsilon > (1+\gamma) \left( \frac{1}{C_2 N} \log \left(\frac{C_1}{\delta}\right) \right)^{\frac{1}{M+1}}${. Then}, with probability at least $1 - \delta$ {under $(\ptrue)^{\otimes N}$}, we have
    \begin{equation}
        \label{eq:generalization_error_bound_fi_dro}
        \begin{aligned}
            & \mathbb{E}_{\ptrue}[\ell (\hat{\BFB}_N^T\BFx, y)] - \mathbb{E}_{\phat_N}[\ell (\hat{\BFB}_N^T\BFx, y)] \leq 
            \min \left\{
                (1+\gamma)\omega \|\hat{\BFB}_N^T\|_* \left(
                \frac{1}{C_2 N} \log \left(\frac{C_1}{\delta}\right)
            \right)^{\frac{1}{M+1}}, \right.\\
            &\hspace{150pt} \left.
            \tau -R(\hat{\BFB}_N)- \mathbb{E}_{\phat_N}[\ell (\hat{\BFB}_N^T\BFx, y)] + (1+\gamma) k^\dagger_{N, \epsilon} \left(
                \frac{1}{C_2 N} \log \left(\frac{C_1}{\delta}\right)
            \right)^{\frac{1}{M+1}}
            \right\},
        \end{aligned}
\end{equation}
\end{propositionAp}
Since {$\hat k_N \geq k^\dagger_{N, \epsilon}$}, the bound \eqref{eq:generalization_error_bound_fi_dro} is {no larger }than the bound in Proposition \ref{prop:generalization_error} {and may be strictly tighter when $k^\dagger_{N, \epsilon}<\hat k_N$}.

\section{Supplementary Experiments}
\label{appe:experiment}
\subsection{Supplementary to Heart Failure Prediction}
\label{appe:experiment_heart_failure_prediction}

\subsubsection{{Cost-Optimal Thresholds under Calibration}}
\label{appe:cost_optimal_selective_thresholds}

{When the predicted probabilities are calibrated and manual review is error-free, the cost-optimal automation and referral thresholds admit the following closed-form expressions.
}

\begin{lemmaAp}[Cost-optimal automation and referral thresholds]
    \label{lemma:cost_optimal_selective_thresholds}
    {Suppose that $p$ is calibrated, i.e., $\mathbb E[Y\mid p(\BFx)]=p(\BFx)$ almost surely, and that manual review is error-free. Define
    }{\[
        \theta_{\mathrm{neg}}^*=\min\left\{\frac{C_{\mathrm{man}}}{C_{\mathrm{FN}}},\frac{C_{\mathrm{FP}}}{C_{\mathrm{FP}}+C_{\mathrm{FN}}}\right\},
        \qquad
        \theta_{\mathrm{pos}}^*=\max\left\{1-\frac{C_{\mathrm{man}}}{C_{\mathrm{FP}}},\frac{C_{\mathrm{FP}}}{C_{\mathrm{FP}}+C_{\mathrm{FN}}}\right\}.
\]}
    {Then, the resulting threshold-based referral policy minimizes the conditional expected operational cost almost surely among an automated negative decision, manual review, and an automated positive decision.
}\end{lemmaAp}

{For example, if $C_{\mathrm{man}}=1$, $C_{\mathrm{FP}}=5$, and $C_{\mathrm{FN}}=20$, then Lemma \ref{lemma:cost_optimal_selective_thresholds} gives $\theta_{\mathrm{neg}}^*=0.05$ and $\theta_{\mathrm{pos}}^*=0.8$. The system therefore automates a negative decision only when $p(\BFx)<0.05$, refers patients with $0.05\leq p(\BFx)<0.8$ to manual review, and automates a positive decision when $p(\BFx)\geq0.8$. The more stringent requirement for automated negative decisions reflects their larger false-negative cost. The proof is provided in Appendix \ref{proof:lemma_cost_optimal_selective_thresholds}.
}

\subsubsection{Adversarial Training Details}
We concisely introduce the robust training baselines considered in our experiments: the loss-correction methods and data-augmentation methods. 

\textbf{Loss-function Baselines}
\begin{enumerate}
    \item Trimmed Loss \citep{shen2019learning}: This method enhances robustness by ignoring samples with high loss values, which are likely to be corrupted or mislabeled. The implementation is governed by a retention parameter, $\lambda \in (0, 1]$, which dictates the fraction of small-loss samples used for gradient updates in each batch. Therefore, $\lambda$ is also called the keep ratio. In experiment, we set $\lambda = 0.9$ to retain 90\% of samples with the smallest losses in each batch.

    \item Generalized Cross-Entropy (GCE) \citep{zhang2018generalized}: GCE provides a theoretical generalization between the noise-robust Mean Absolute Error (MAE) and the convergence-friendly Cross-Entropy (CE) loss. It utilizes a hyperparameter $q \in (0, 1]$ to control the trade-off between noise robustness (high $q$) and optimization difficulty. The loss is defined as $L_q(f(x), y) = \frac{1 - f_y(x)^q}{q}$. Notice that when $q \to 0$, GCE converges to CE, and when $q = 1$, GCE becomes MAE. In our experiments, we set $q = 0.7$ to balance robustness and convergence.

    \item Symmetric Cross-Entropy (SCE) \citep{wang2019symmetric}: SCE addresses the overfitting issue of CE in the presence of noisy labels by combining CE with a noise-tolerant Reverse Cross-Entropy (RCE) term. The total loss is formulated as $L = \alpha L_{CE} + \beta L_{RCE}$, where $L_{CE}$ is the standard cross-entropy loss and {$L_{RCE}(x,y) = - \sum_{i=1}^{C} f_i(x) \log q_i(y)$}, with $f_i(x)$ being the predicted probability for class $i$ {and $q_i(y)=\mathbb{I}\{i=y\}$ the one-hot label distribution. In our implementation, the zero entries inside the logarithm are clipped to $10^{-4}$ for numerical stability. }The implementation depends on the balancing hyperparameters $\alpha$ and $\beta$, which weight the contribution of the standard and reverse cross-entropy terms, respectively. We follow the conventional setting of $\alpha = 0.1$ and $\beta = 1.0$ in our experiments.
\end{enumerate}

\textbf{Data-augmentation Baselines}
\begin{enumerate}
    \item ERM Purify \citep{huang2019o2u}: Based on the O2U-Net framework, this method performs ``data purification'' by filtering out noisy samples before retraining. The key implementation detail here is the filtering threshold (or pruning rate) $\rho$, which determines the percentage of data removed from the training set before the final retraining phase. Even though it is similar to the trimmed loss method, ERM Purify conducts a two-stage training process, where the first stage identifies and removes noisy samples, and the second stage retrains the model on the purified dataset. In our experiments, we set $\rho = 0.1$, meaning that 10\% of samples with the highest losses are removed before retraining.

    \item Mixup \citep{zhang2017mixup}: Mixup trains the network on convex combinations of pairs of examples and their labels. For input pairs $(x_i, y_i)$ and $(x_j, y_j)$, a new training sample is generated as $\tilde{x} = \lambda x_i + (1-\lambda)x_j$ and $\tilde{y} = \lambda y_i + (1-\lambda)y_j$. The mixing coefficient $\lambda$ is sampled from a Beta distribution $\text{Beta}(\alpha, \alpha)$. The hyperparameter $\alpha$ controls the strength of interpolation between classes. In our experiments, we set $\alpha = 1.0$ to achieve a moderate level of mixing, which is also called the uniform mixup.

    \item Label Smoothing \citep{szegedy2016rethinking}: This regularization technique prevents the model from predicting training examples too confidently. Instead of using a ``one-hot'' target distribution, the label $y$ is smoothed using a parameter $\epsilon$. The new target becomes a mixture of the original ground truth (weighted by $1-\epsilon$) and a uniform distribution over the $K$ classes (weighted by $\epsilon/K$). This encourages the model to be less certain about its predictions, which can improve generalization. In our experiments, we set $\epsilon = 0.1$ to apply a moderate level of smoothing. It means that true labels are assigned a probability of $0.9$, while the remaining $0.1$ probability is uniformly distributed among all classes.
\end{enumerate}

All the adversarial methods are implemented {using the same linear classifier architecture as that used for ERM and }FI-based training. We use the Adam optimizer with 100 epochs and a learning rate of $0.01$ for all methods to ensure a fair comparison.

\subsection{Synthetic Data Experiment}
\label{appe:experiment_synthetic}

\subsubsection{Binary Classification with Label-Flipping Attack}\label{appe:experiment_synthetic_binary}
\paragraph{Setup} We consider a binary classification task subject to a label-flipping attack, with a balanced class distribution. The raw features $x_{+}, x_{-} \in \mathbb{R}^{2}$ are sampled from two Gaussian distributions with moderate overlap. These features are transformed using a polynomial feature map: for instance, the quadratic feature map defined as $\phi_{poly-2}((x_{1},x_{2})^{T}) = (1,x_{1},x_{2},x_{1}^{2},x_{2}^{2},x_{1}x_{2})^{T}$. The attack is introduced by flipping the label of a sample with probability $p_{flip}$ immediately after generation. This polluted data constitutes the training set, while the testing set remains unaffected by label flipping. As a result, the probability $p_{flip}$ represents the degree of distributional shift between the training and testing environments.

We use the ERM model as the benchmark, using the hinge loss defined in the equation \eqref{eq:hinge_type_loss}. This setup corresponds to a {linear support vector classifier on polynomial features}. In comparison, we evaluate our FI-based models using the same hinge loss. We implement two FI-based models: one based on the KL-divergence in Theorem \ref{theorem:kl_reformulation} and the other based on the {1-Wasserstein distance with the Euclidean ground norm }in Theorem \ref{theorem:hinge_type_reformulation}.

Regularization {$R(\BFB)$ }is selected via 5-fold cross-validation. We consider both L1 regularization ({$R_{1}(\BFB)=\alpha\sum_{i,j}|(\BFB)_{ij}|$}) and L2 regularization ({$R_{2}(\BFB)=\alpha\sum_{i,j}(\BFB)_{ij}^{2}$}), with $\alpha \in \{0.0001, 0.001, 0.01, 0.1\}$. To ensure a fair comparison, the FI-based models adopt the identical regularization settings used for the ERM model. The {training-loss }target $\tau$ in the FI-based models is determined by scaling the ERM training loss $\hat{L}_{ERM}$. Specifically, we set $\tau = \lambda\hat{L}_{ERM}$, where $\lambda \in \{1.05, 1.1, 1.15\}$ is referred to as the target ratio. {A larger }$\lambda$ {relaxes this training-loss target and thereby allows a smaller training fragility}.

\phantomsection\label{loc:synthetic_fi_evaluation_target}
We repeat the experiment 100 times, generating different training and testing sets for each run. The training sample size varies, while the testing sample size is fixed at 2,000. Model performance is evaluated on the testing set by averaging accuracy, AUC, and FI. {All reported FI values in the synthetic experiments are computed post-training using the KL-based ranking-error FI with target $0$; in the multi-class setting, we use the macro-averaged OVO FI in \eqref{eq:def_multiclass_fi}.
}

\paragraph{Effect of Sample Size} We first examine the impact of training sample size, varying it from 50 to 400. We consider a baseline setting with no distributional shift ($p_{flip}=0$). Even in the absence of data generating distributional shifts, robustness remains critical, particularly when sample size is small. Figure \ref{fig:synthetic_sample_size} presents the results. Larger values indicate better performance for Accuracy and AUC, while smaller values indicate better performance for FI. The Wasserstein model consistently outperforms the ERM model across all metrics. The KL-divergence model performs similarly to ERM in terms of Accuracy and AUC but achieves a superior (lower) FI. Notably, the performance gap between the Wasserstein and ERM models diminishes as the sample size increases; this suggests that larger datasets sufficiently capture the underlying distribution, thereby reducing the marginal benefit of robustness constraints.
\begin{figure}[htbp]
    \centering
    \subfloat[Accuracy]{\includegraphics[width=0.33\textwidth]{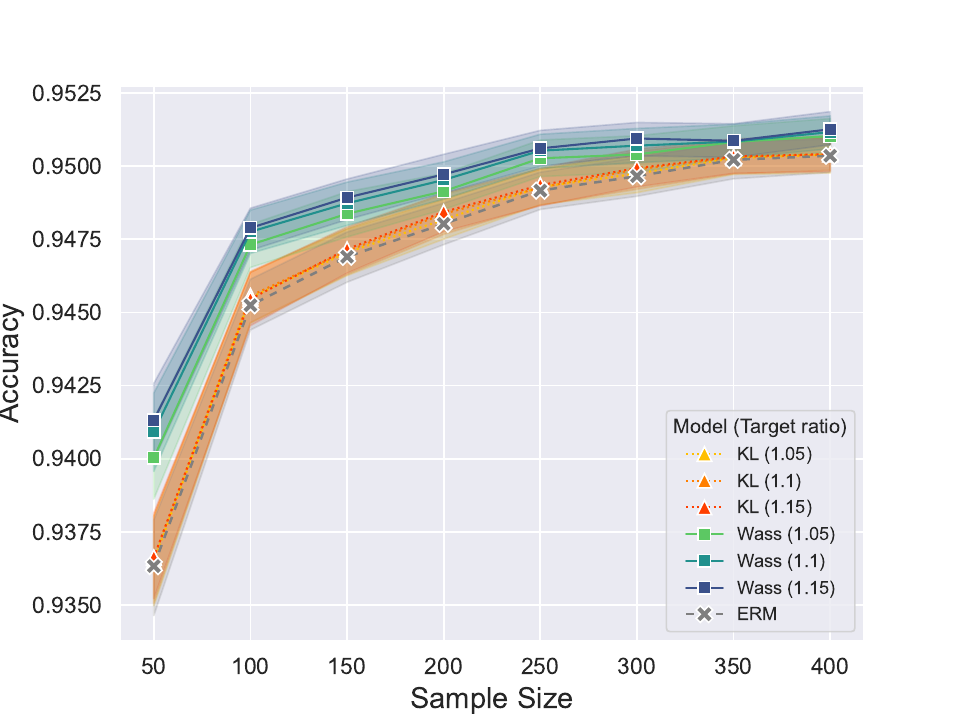}}
    \subfloat[AUC]{\includegraphics[width=0.33\textwidth]{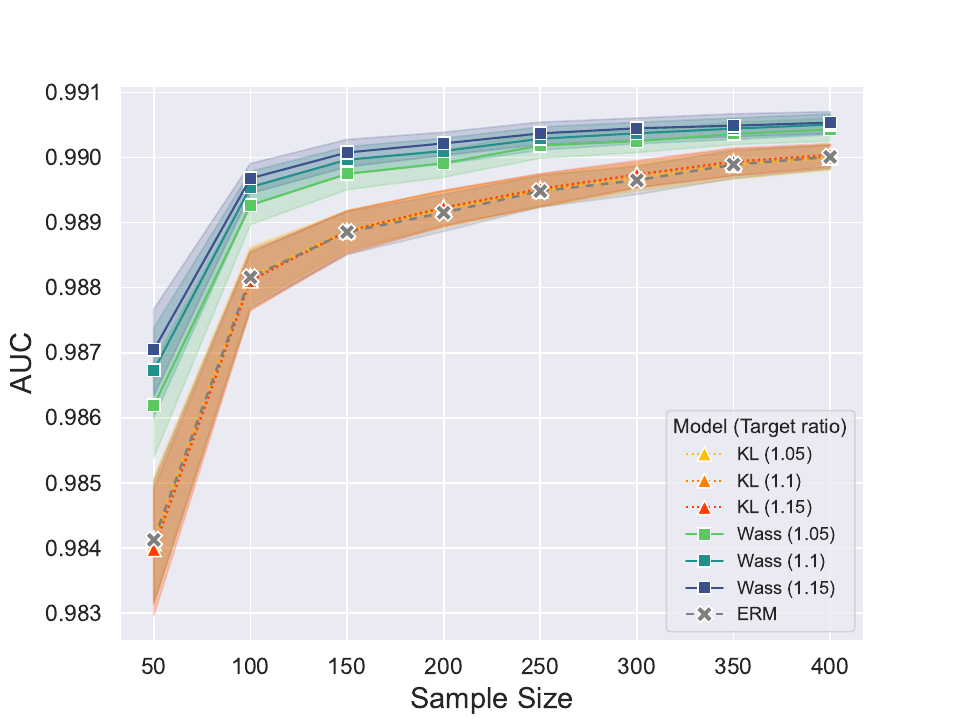}}
    \subfloat[FI]{\includegraphics[width=0.33\textwidth]{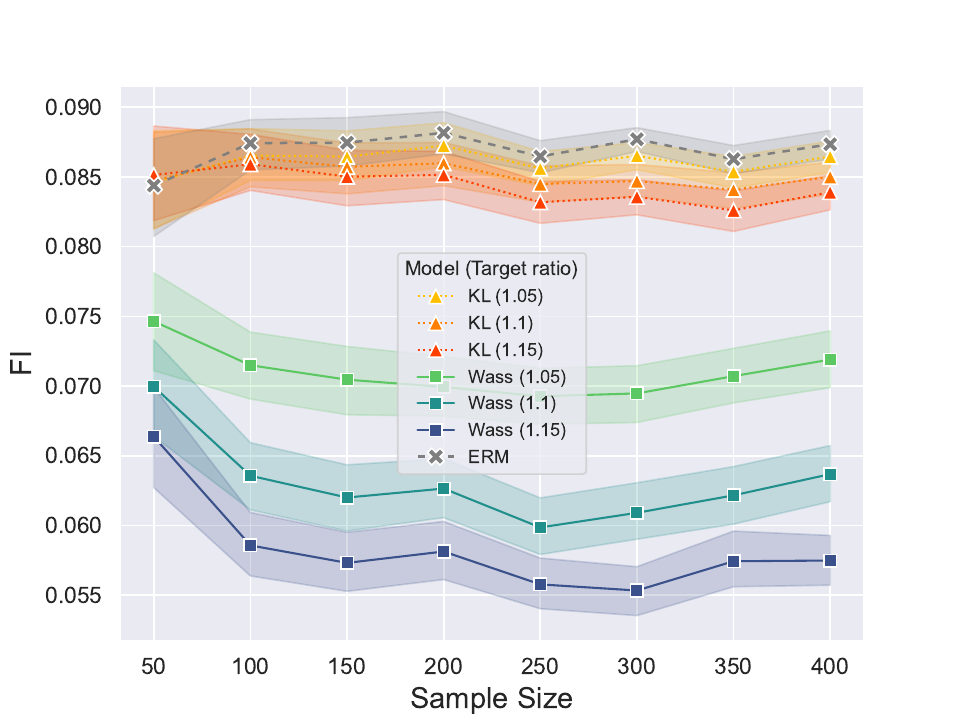}}
    \caption{The relationship between the training sample size and the accuracy, AUC, and FI when $p_{flip} = 0$. The values $1.05, 1.1, 1.15$ represent the target ratio $\lambda$ used in each model.}
    \label{fig:synthetic_sample_size}
\end{figure}

\paragraph{Effect of Distribution Shifts} We fix the sample size at 50 and vary the label-flipping rate $p_{flip}$ from 0 to 0.3 to simulate increasing degrees of distributional shift. We cap $p_{flip}$ at 0.3, as higher noise levels corrupt the training data too much to learn meaningful patterns.
\begin{figure}[htbp]
    \centering
    \subfloat[Accuracy]{\includegraphics[width=0.33\textwidth]{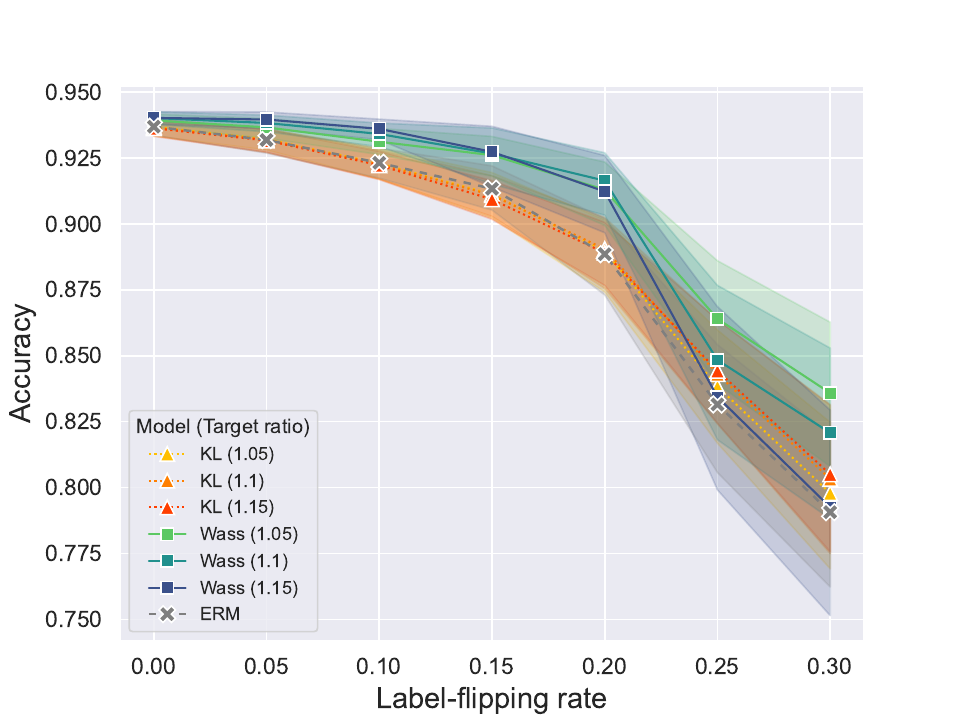}}
    \subfloat[AUC]{\includegraphics[width=0.33\textwidth]{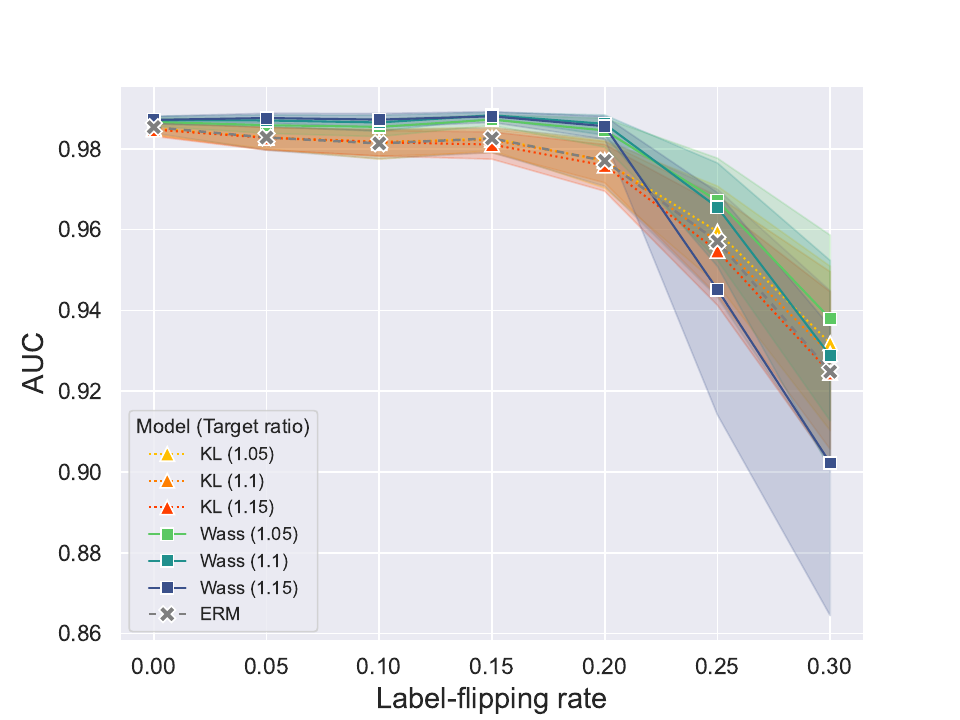}}
    \subfloat[FI]{\includegraphics[width=0.33\textwidth]{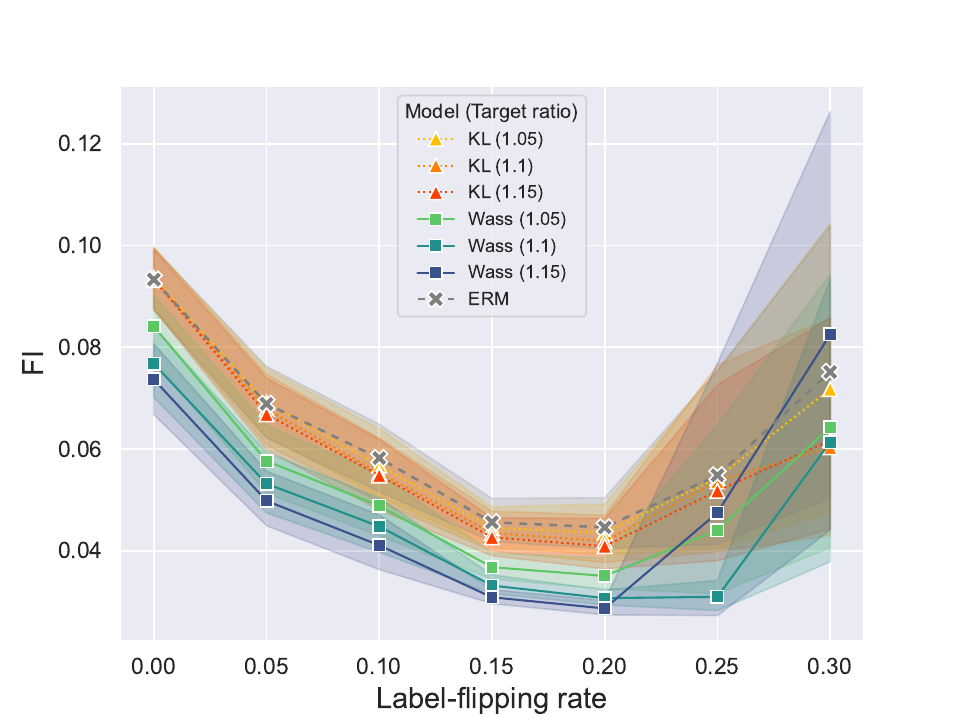}}
    \caption{The relationship between the label-flipping rate $p_{flip}$ and the accuracy, AUC, and FI when the sample size is 50. The values $1.05, 1.1, 1.15$ represent the target ratio $\lambda$ used in each model.}
    \label{fig:synthetic_flip_rate}
\end{figure}
Figure \ref{fig:synthetic_flip_rate} illustrates the performance metrics across different label-flipping rates. Both FI-based models demonstrate superior performance compared to the ERM model across most metrics. However, an exception occurs with the Wasserstein model when $\lambda=1.15$ under high $p_{flip}$ values, where it performs worse than the ERM baseline.

\paragraph{Effect of Target Ratio $\lambda$} A larger $\lambda$ {relaxes the training-loss target and allows a smaller training fragility, which can enhance }robustness against distributional shift. However, this does not imply that $\lambda$ should be as large as possible. Figure \ref{fig:synthetic_flip_rate} shows that when $p_{flip} $ is large, the Wasserstein model with $\lambda = 1.15$ performs worse than that with $\lambda = 1.05$, and can be even worse than the ERM model, indicating that an excessively large $\lambda$ can be detrimental to model performance.

The tradeoff between overfitting and underfitting explains this behavior. When $p_{flip} > 0$, the ERM model overfits the noisy training data. The Wasserstein model mitigates overfitting by setting a higher {training-loss target }$\tau$ via $\lambda > 1$ relative to $\hat{L}_{ERM}$. However, if $\lambda$ becomes too large, the model becomes overly conservative and consequently underfits and results in high bias. Careful problem-specific tuning of $\lambda$ is essential to balance overfitting and underfitting.

\paragraph{Comparison with Adversarial Training Methods} We further compare our FI-based models with adversarial training methods, which are widely used to enhance model robustness. We consider the same adversarial training baselines in \ref{appe:experiment_heart_failure_prediction}, which include both loss-function-based methods and data-augmentation-based methods.

Figure \ref{fig:synthetic_comparison_adversarial} presents the results. 
\begin{figure}[htbp]
    \centering
    \subfloat[Accuracy]{\includegraphics[width=0.33\textwidth]{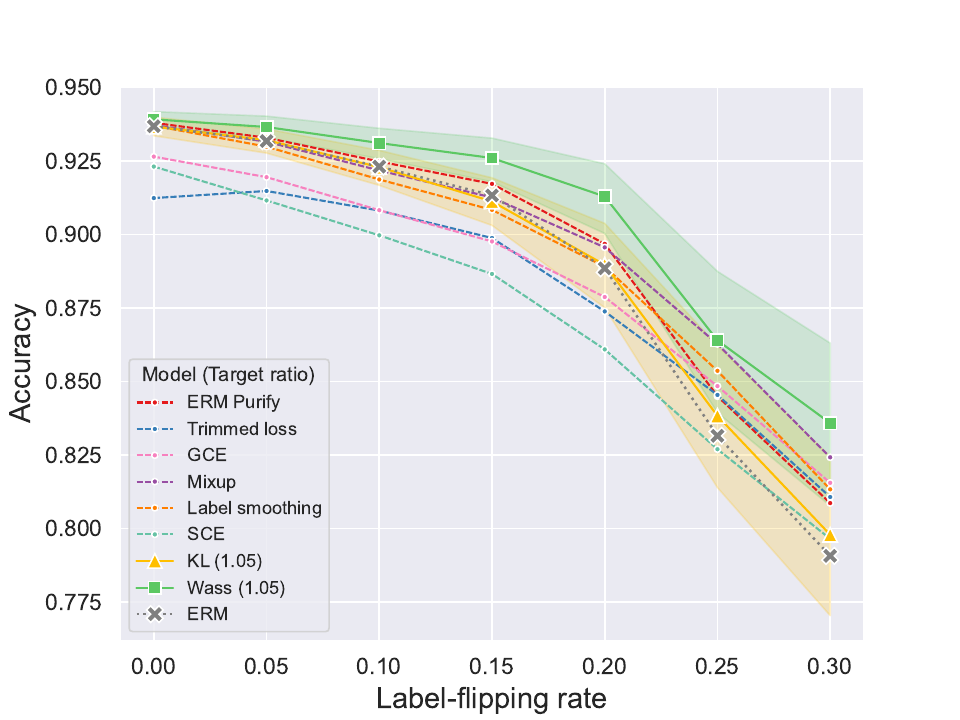}}
    \subfloat[AUC]{\includegraphics[width=0.33\textwidth]{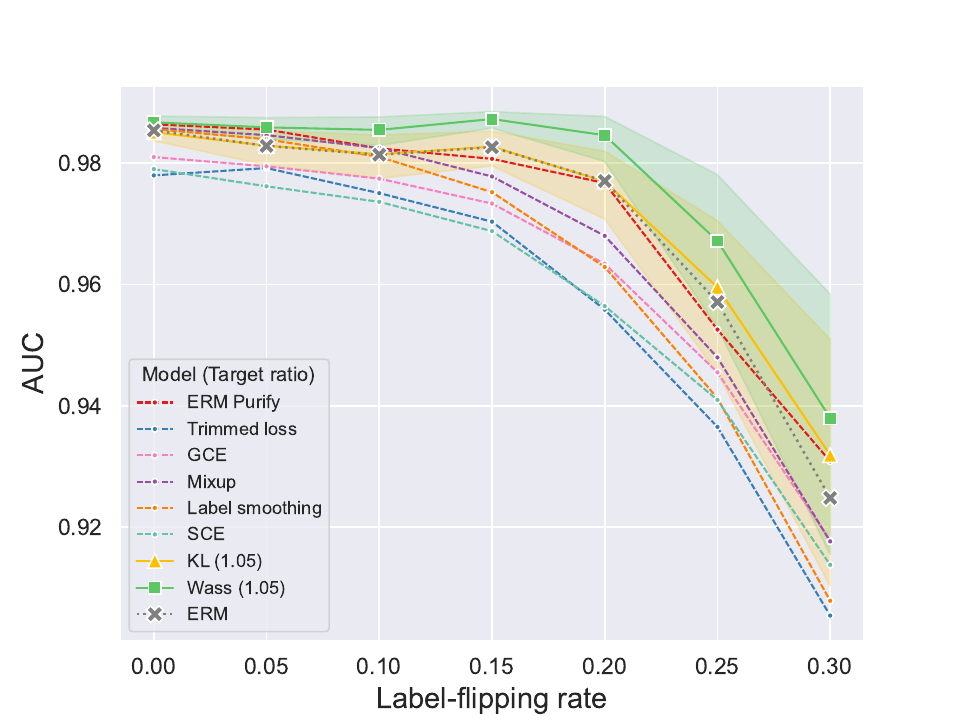}}
    \subfloat[FI]{\includegraphics[width=0.33\textwidth]{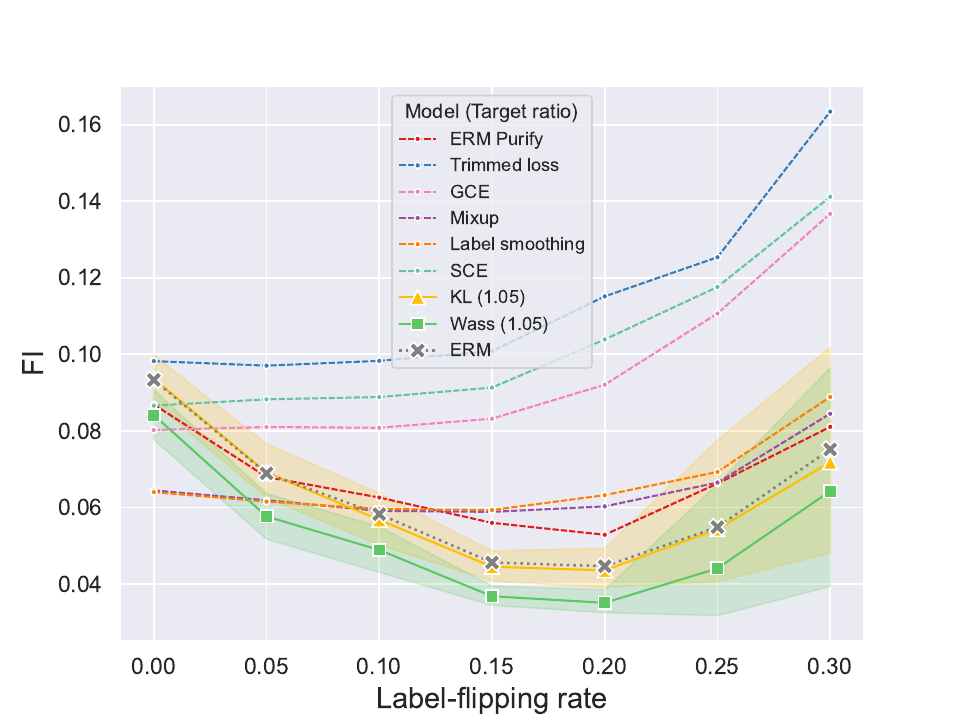}}
    \caption{The comparison between the FI-based models and adversarial training methods when the sample size is 50.}
    \label{fig:synthetic_comparison_adversarial}
\end{figure}
Overall, our FI-based models achieve performance comparable to these adversarial training methods across all metrics. Notably, the Wasserstein FI-based model outperforms most adversarial training methods in terms of FI, underscoring its effectiveness in controlling risk. Furthermore, as discussed in Appendix \ref{appe:experiment}, our framework is also compatible with data augmentation techniques, offering a pathway for potential performance improvements.

\paragraph{FI-Based Model with Data Augmentation} Some data-augmentation techniques can also be applied to our FI-based training framework to potentially improve the performance. For example, Mixup and Label Smoothing can be directly integrated into the FI-based training. We use the FI-inducing regularizer as our benchmark, and the implementation details are similar to those in the adversarial training baselines. Moreover, we also consider the Gaussian noise augmentation, which adds Gaussian noise to the input features during training, and the projected gradient descent (PGD) adversarial augmentation, which generates adversarial examples using PGD and includes them in the training set. The results of FI-based training with data augmentation on synthetic data are shown in Figure \ref{fig:synthetic_fi_data_augmentation}.
\begin{figure}[htbp]
    \centering
    \subfloat[Accuracy]{\includegraphics[width=0.34\textwidth]{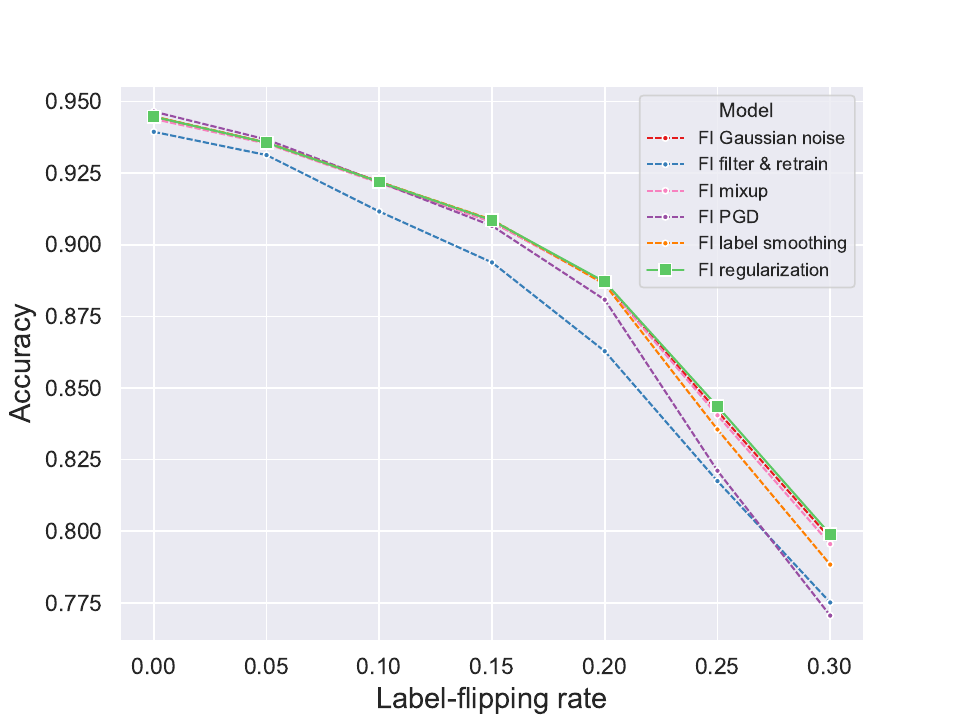}}
    \subfloat[AUC]{\includegraphics[width=0.34\textwidth]{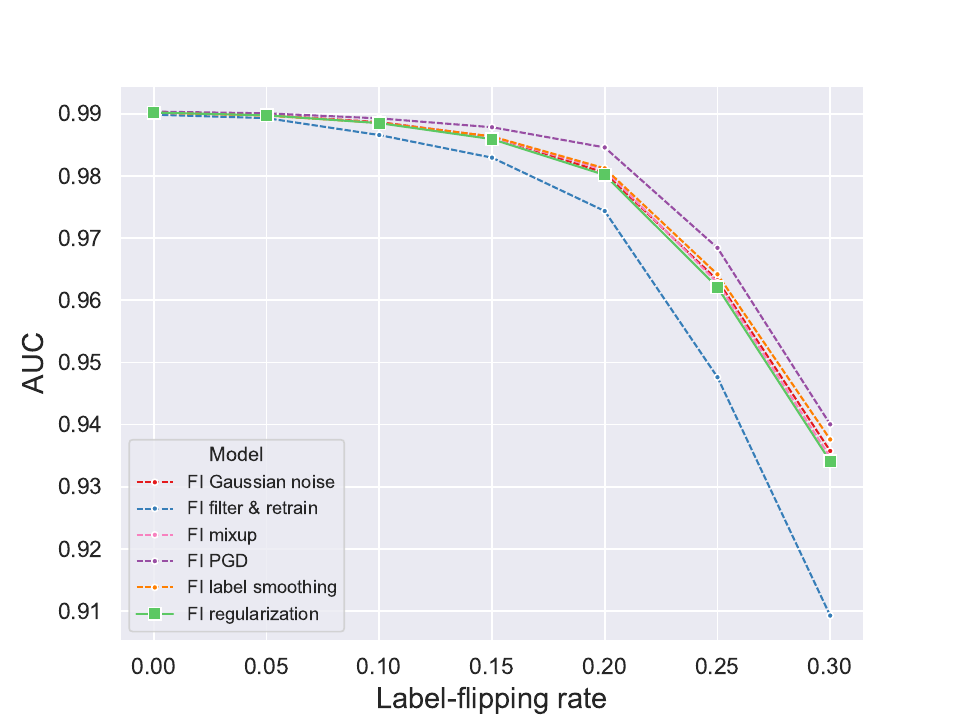}}
    \subfloat[FI]{\includegraphics[width=0.34\textwidth]{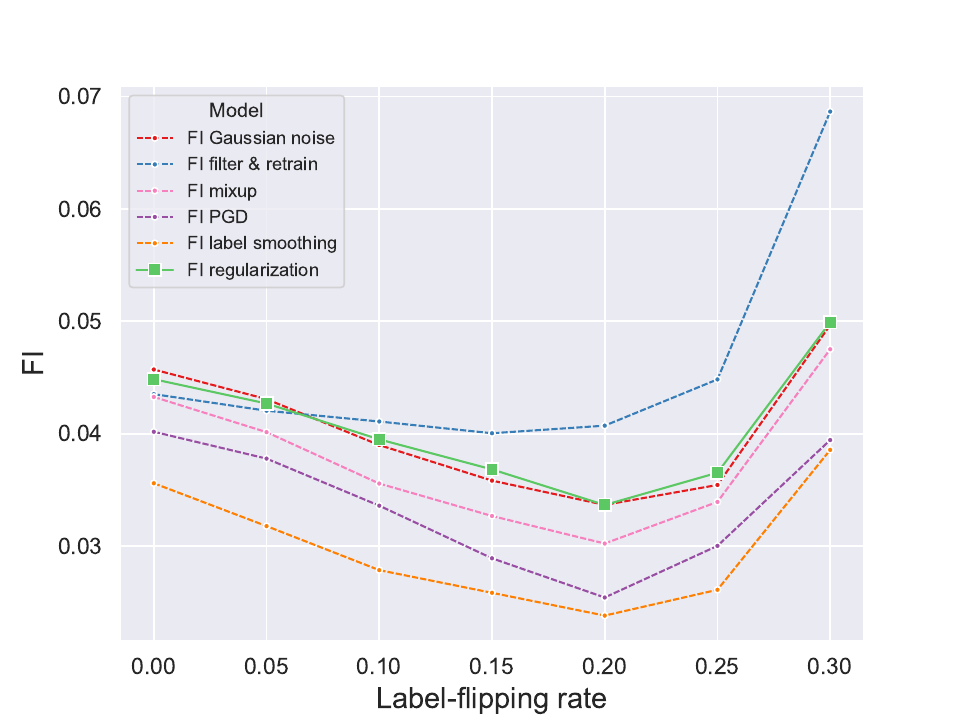}}
    \caption{The performance of FI-based training with data augmentation on synthetic data.}
    \label{fig:synthetic_fi_data_augmentation}
\end{figure}

For AUC, PGD can improve the performance of FI-based training when the label-flipping rate grows. More importantly, PGD, mixup, and label smoothing can all help reduce the FI of the model, indicating that these data-augmentation techniques can further enhance the robustness under distributional shift and reduce the risk of confidence misjudgment.

\subsubsection{Synthetic Data: Multi-Class Classification with Label-Flipping Attack}
\label{appe:experiment_synthetic_multi_class}

We also conduct experiments on the multi-classification case. We consider similar Gaussian clusters as the binary case, but we generate 4 classes. Following the same procedure of training and evaluation, we show the results of the multi-classification case on synthetic data. The results are shown in Figure \ref{fig:synthetic_4_class}.

The main patterns in Figure \ref{fig:synthetic_4_class} are similar to those in the binary case. The difference is mainly in the performance of the Wasserstein model as $p_{flip}$ changes. However, in the 4-class case, the Wasserstein model performs worse than ERM when $p_{flip}$ is very large. To our analysis, this is due to the bias caused by the underfitting of the Wasserstein model. Since the 4-class problem is more complicated, it is much easier for the Wasserstein model to be overconservative to unrealistic uncertainty and underfit the data. Therefore, the weaker multi-class performance of Wasserstein FI mainly comes from its full-support ambiguity, which can over-constrain margins at high $\tau$. KL FI avoids this by reweighting empirical evidence on the data manifold, yielding a better robustness-accuracy tradeoff in complex multi-class settings.
\begin{figure}[htbp]
    \centering
    \subfloat[Sample size v.s. Accuracy]{\includegraphics[width=0.32\linewidth]{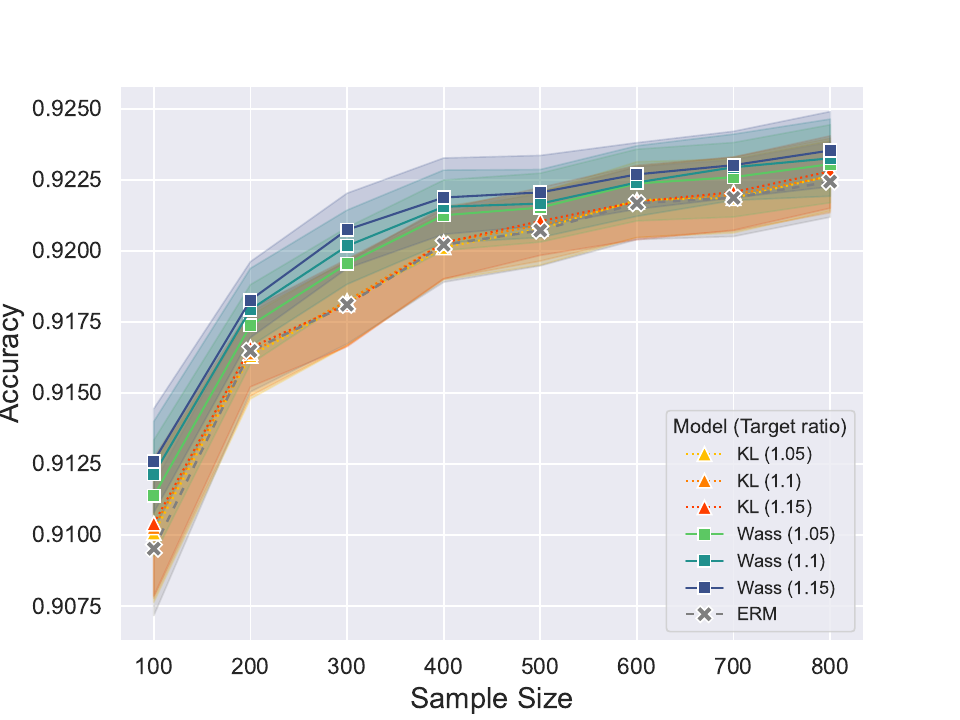}}
    \subfloat[Sample size v.s. AUC]{\includegraphics[width=0.32\linewidth]{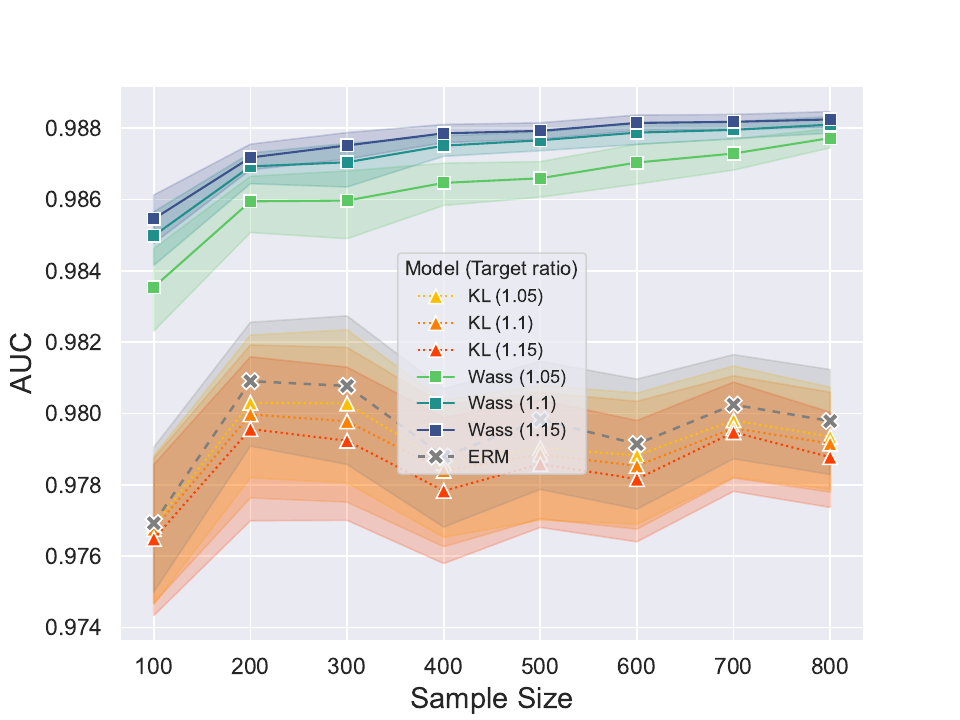}}
    \subfloat[Sample size v.s. FI]{\includegraphics[width=0.32\linewidth]{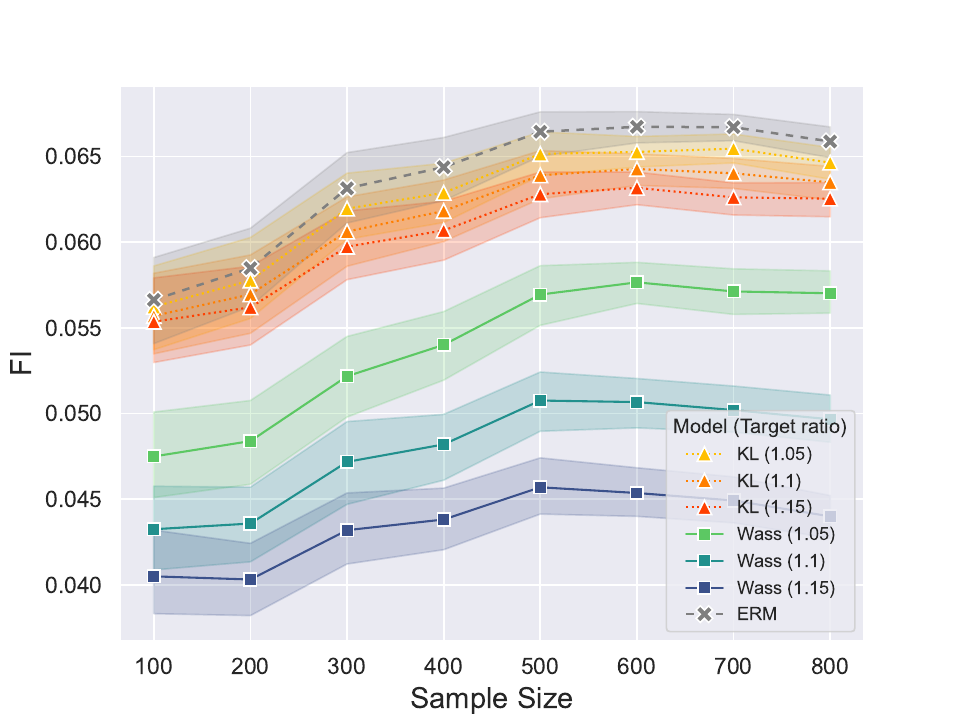}}

    \subfloat[$p_{flip}$ v.s. Accuracy]{\includegraphics[width=0.32\linewidth]{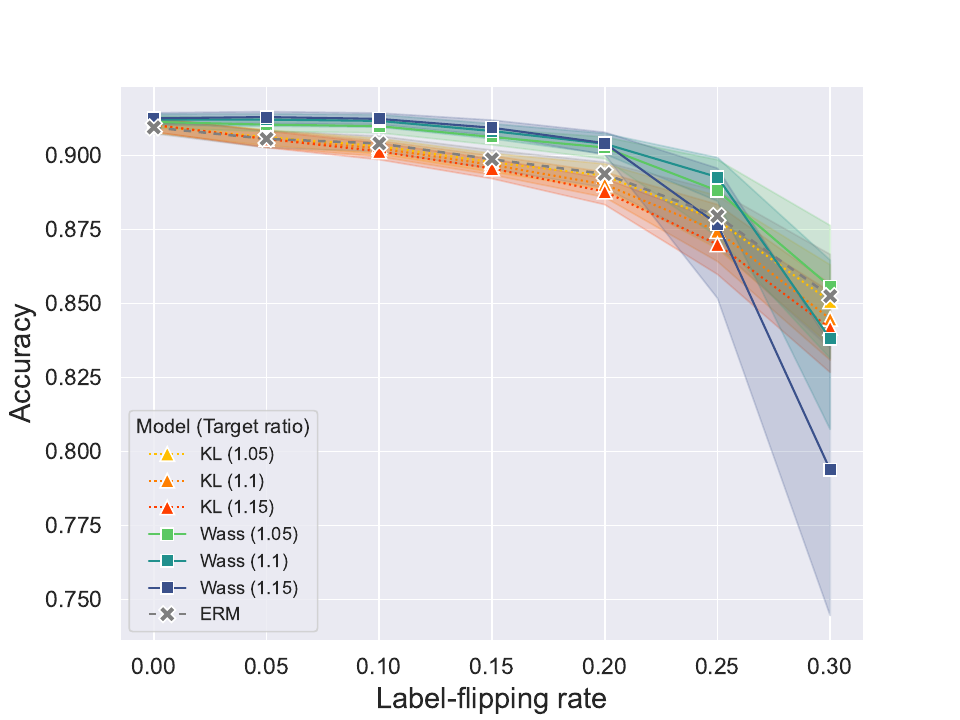}}
    \subfloat[$p_{flip}$ v.s. AUC]{\includegraphics[width=0.32\linewidth]{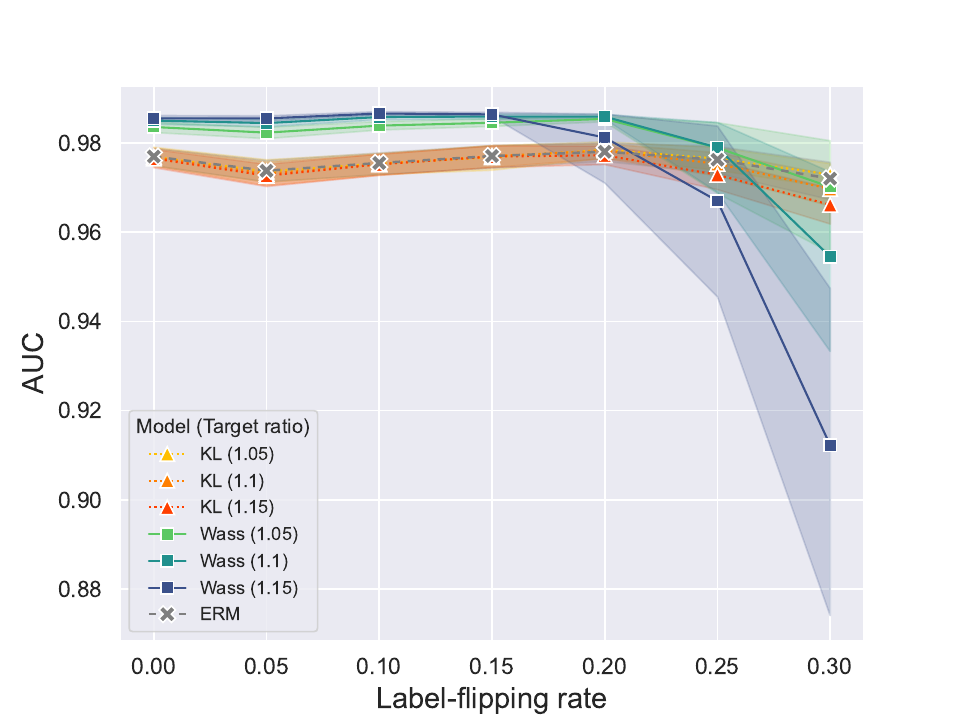}}
    \subfloat[$p_{flip}$ v.s. FI]{\includegraphics[width=0.32\linewidth]{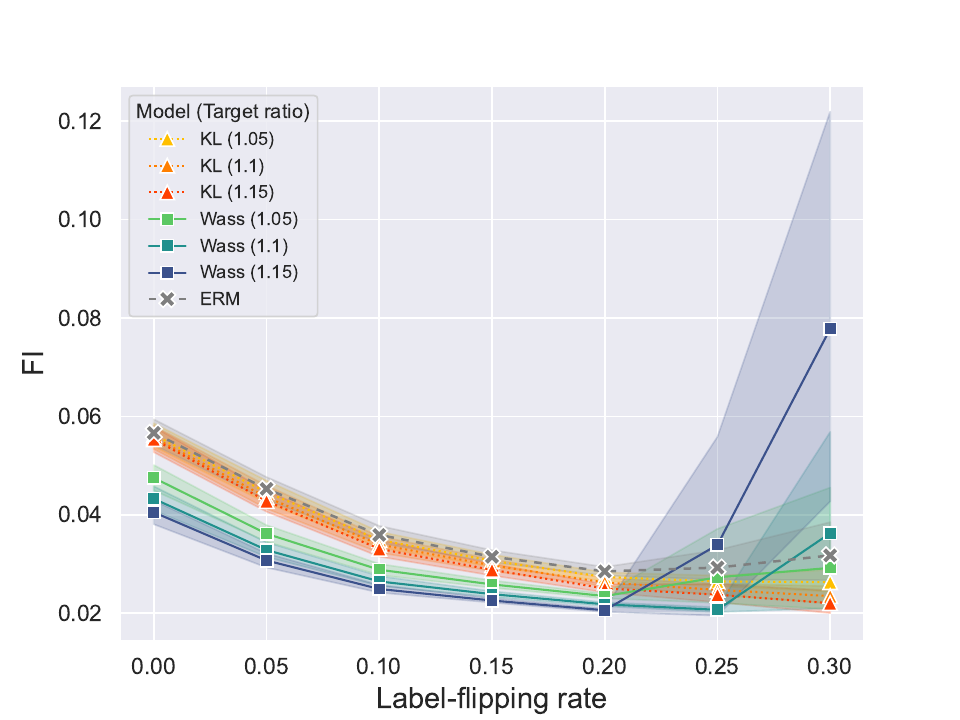}}
    \caption{The results of the FI-based methods and ERM on synthetic data with 4 classes.}
    \label{fig:synthetic_4_class}
\end{figure}

We also compare our methods with adversarial training baselines in the multi-classification case. The results are shown in Figure \ref{fig:synthetic_4_class_adv}. Even though the performance of the FI-based methods is still comparable with adversarial training baselines, the Wasserstein model performs slightly relatively poorly in this case. As mentioned, $\lambda = 1.05$ is probably too large for the Wasserstein model in the 4-class case, which leads to the underfitting issue and the performance degradation. This also suggests that the appropriate target ratio may depend on the exact problem setting, and should be tuned accordingly.
\begin{figure}[htbp]
    \centering
    \subfloat[Accuracy]{\includegraphics[width=0.32\linewidth]{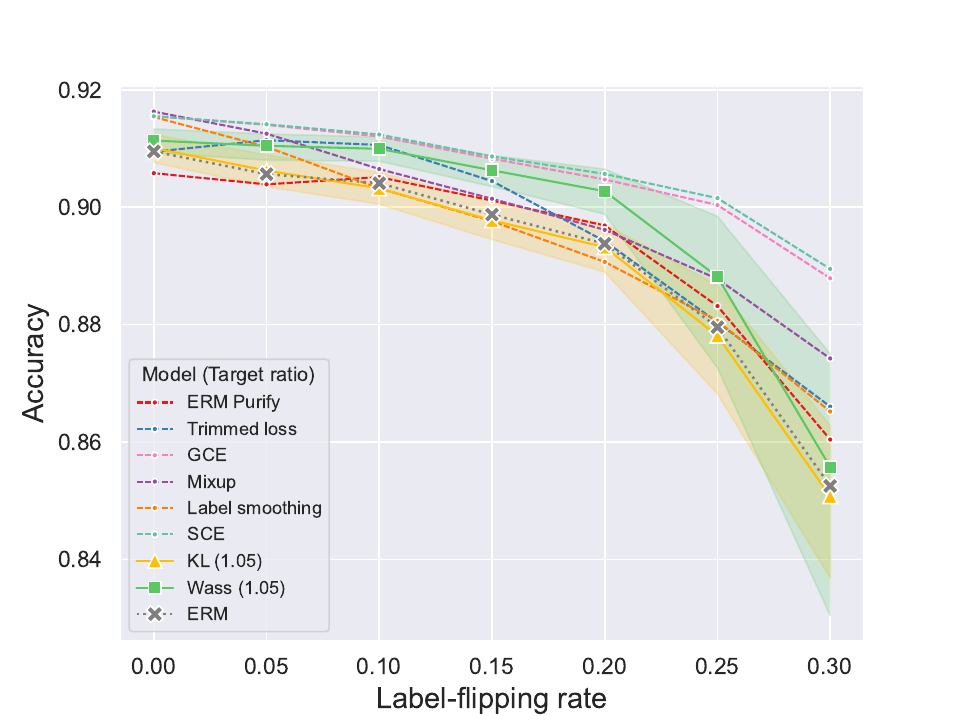}}
    \subfloat[AUC]{\includegraphics[width=0.32\linewidth]{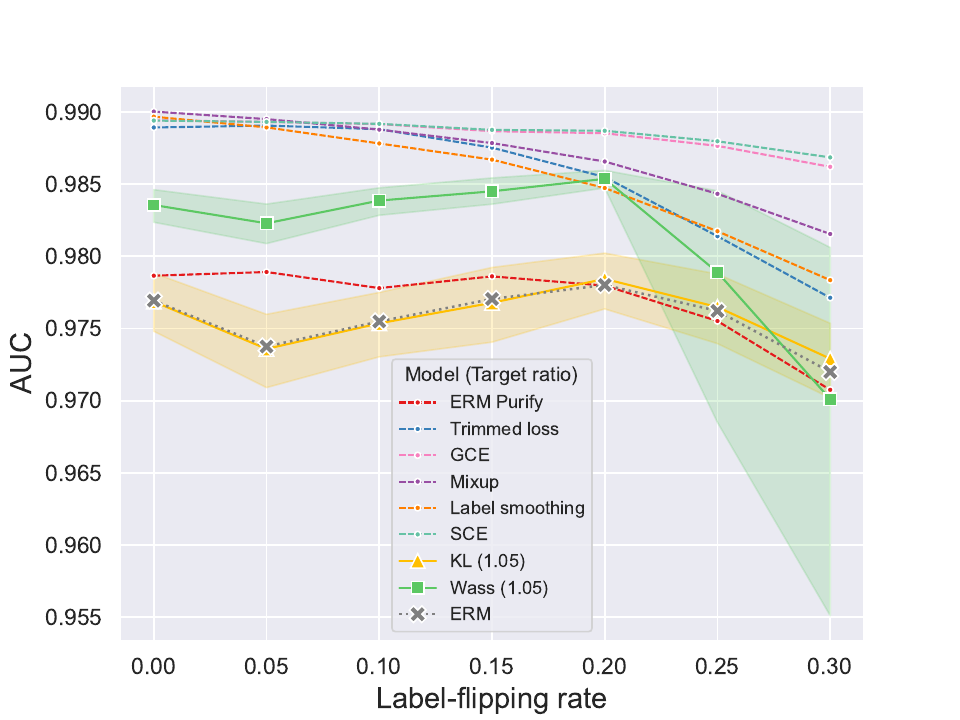}}
    \subfloat[FI]{\includegraphics[width=0.32\linewidth]{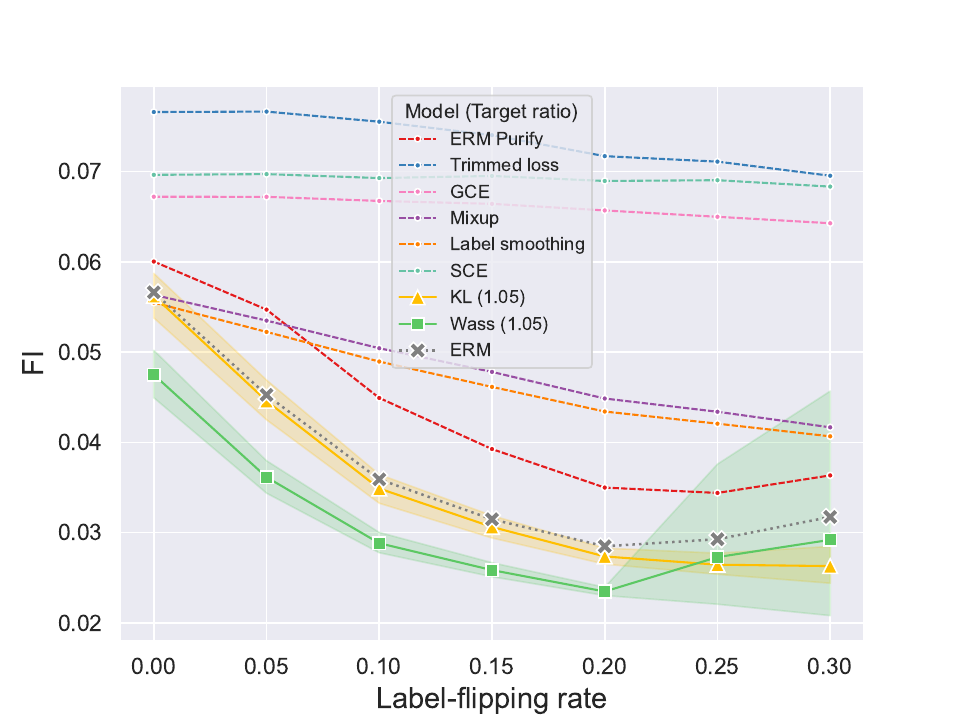}}
    \caption{The results of the FI-based methods and adversarial training baselines on synthetic data with 4 classes.}
    \label{fig:synthetic_4_class_adv}
\end{figure}

Finally, we also verify whether the data-augmentation techniques can help improve the performance of FI-based training in the multi-classification case. The results are shown in Figure \ref{fig:synthetic_4_class_fi_data_augmentation}. Similar to the binary case, PGD, mixup, and label smoothing can all help reduce the FI of the model, indicating that these data-augmentation techniques can further enhance the robustness under distributional shift and reduce the risk of confidence misjudgment.
\begin{figure}[htbp]
    \centering
    \subfloat[Accuracy]{\includegraphics[width=0.32\linewidth]{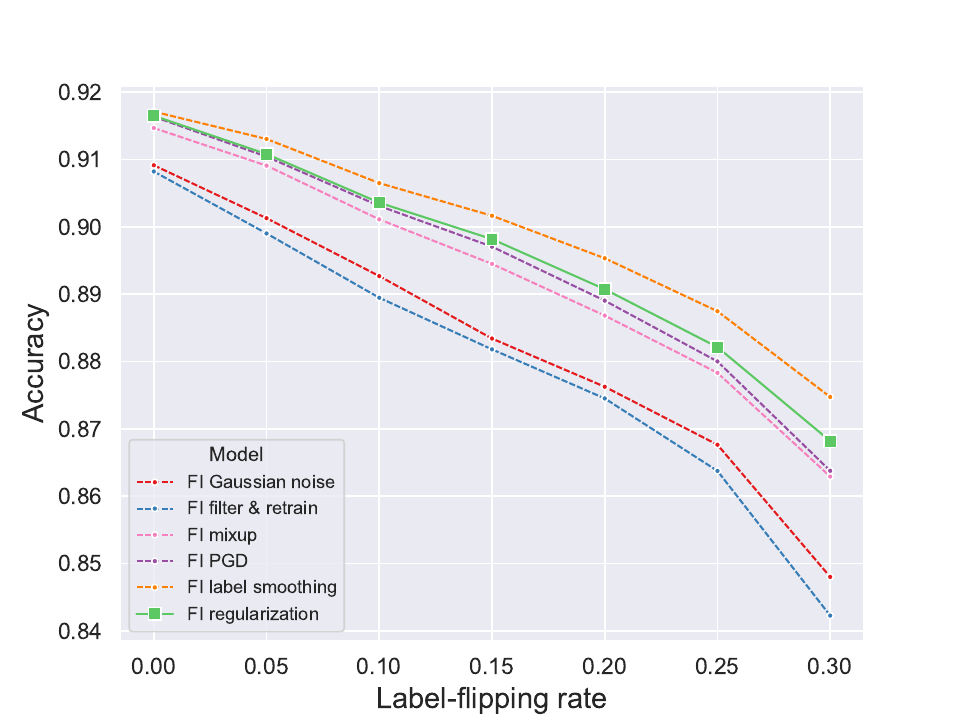}}
    \subfloat[AUC]{\includegraphics[width=0.32\linewidth]{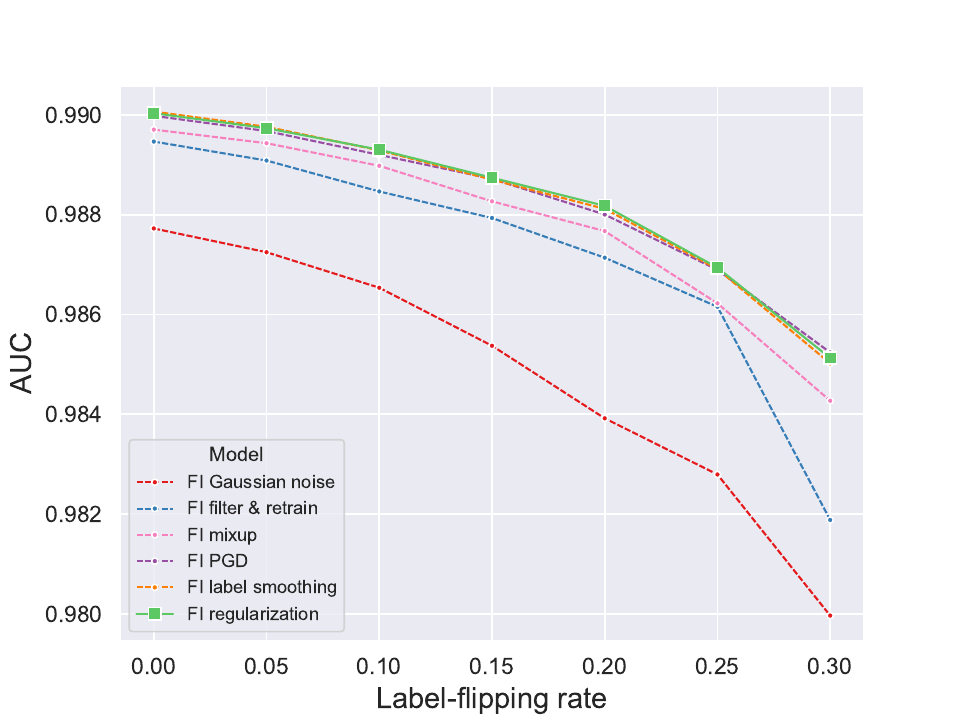}}
    \subfloat[FI]{\includegraphics[width=0.32\linewidth]{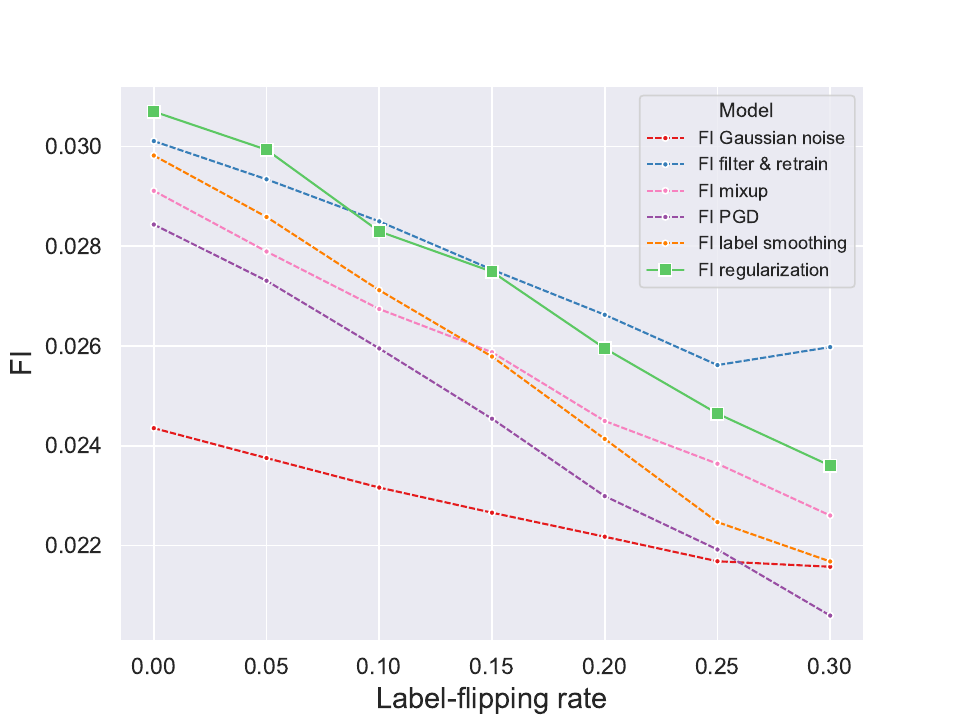}}
    \caption{The performance of FI-based training with data augmentation on synthetic data with 4 classes.}
    \label{fig:synthetic_4_class_fi_data_augmentation}
\end{figure}

\subsection{Supplementary Real-Data Experiments}
\label{appe:experiment_real_data}
In this section, we will perform numerical experiments based on real data to show how our fragility index behaves compared with other performance metrics. We select {8 }datasets from the UCI Machine Learning Repository \citep{uci_repository}. We list the information of these 8 datasets in Table \ref{tab:info_data}. 
\begin{table}[htbp]
    \centering
    \small
    \begin{tabular}{|c|c|c|c|}
        \hline
        Dataset                                                                                 & Abbr. & Size & Feature \\ \hline
        Breast Cancer Coimbra                                                                   & BCC          & 116  & 10      \\ \hline
        Liver Disorders                                                                         & LD           & 345  & 7       \\ \hline
        ILPD (Indian Liver Patient Dataset)                                                     & {ILPD         }& 583  & 10      \\ \hline
        Statlog (German Credit Data)                                                            & SG           & 1000 & 20      \\ \hline
        Breast Cancer Wisconsin (Prognostic)                                                    & BCW          & 198  & 34      \\ \hline
        Diabetes                                                                                & DI           & 768  & 8       \\ \hline
        Ionosphere                                                                              & IO           & 351  & 34      \\ \hline
        \begin{tabular}[c]{@{}c@{}}Connectionist Bench \\ (Sonar, Mines vs. Rocks)\end{tabular} & CB           & 208  & 60      \\ \hline
    \end{tabular}
    \caption{Information of the 8 datasets from UCI Repository.}
    \label{tab:info_data}
\end{table}
We follow the same training and evaluation process as the heart {failure prediction }dataset. For simplicity, we only consider the label-flipping rate with $p_{flip} = 0.0$ and $0.1$. The results are shown in Table \ref{tab:results_uci}. 

\begin{table}[htbp]
    \small
    \begin{tabular}{|c|c|c|c|c|c|c|c|}
        \hline
        \multirow{2}{*}{\textbf{Dataset}} & \multirow{2}{*}{\textbf{Model}} & \multicolumn{3}{c|}{$\bm{p_{flip} = 0.0}$} & \multicolumn{3}{c|}{$\bm{p_{flip} = 0.1}$}\\
        \cline{3-8}
        &  & \textbf{ACC} & \textbf{AUC} & \textbf{FI} & \textbf{ACC} & \textbf{AUC} & \textbf{FI} \\
        \hline
        \multirowcell{3}{BCC} & ERM & \makecell{0.695\\(0.671, 0.719)} & \makecell{0.789\\(0.766, 0.81)} & \makecell{0.154\\(0.13, 0.18)} & \makecell{0.635\\(0.603, 0.663)} & \makecell{0.717\\(0.69, 0.743)} & \makecell{0.115\\(0.098, 0.131)} \\
         & KL & \textbf{\makecell{0.702\\(0.675, 0.726)}} & \textbf{\makecell{0.789\\(0.766, 0.81)}} & \makecell{0.143\\(0.119, 0.171)} & \textbf{\makecell{0.638\\(0.608, 0.664)}} & \textbf{\makecell{0.719\\(0.689, 0.747)}} & \makecell{0.099\\(0.086, 0.113)} \\
         & Wass & \makecell{0.657\\(0.633, 0.678)} & \makecell{0.759\\(0.737, 0.783)} & \textbf{\makecell{0.111\\(0.095, 0.13)}} & \makecell{0.591\\(0.562, 0.619)} & \makecell{0.675\\(0.644, 0.704)} & \textbf{\makecell{0.079\\(0.068, 0.09)}} \\
        \hline
        \multirowcell{3}{BCW} & ERM & \makecell{0.756\\(0.74, 0.771)} & \makecell{0.645\\(0.595, 0.693)} & \makecell{0.034\\(0.018, 0.052)} & \makecell{0.75\\(0.735, 0.764)} & \makecell{0.611\\(0.56, 0.661)} & \makecell{0.029\\(0.019, 0.042)} \\
         & KL & \makecell{0.758\\(0.741, 0.772)} & \textbf{\makecell{0.752\\(0.727, 0.779)}} & \textbf{\makecell{0.032\\(0.019, 0.049)}} & \makecell{0.75\\(0.735, 0.764)} & \textbf{\makecell{0.711\\(0.676, 0.744)}} & \makecell{0.028\\(0.019, 0.039)} \\
         & Wass & \textbf{\makecell{0.758\\(0.743, 0.774)}} & \makecell{0.655\\(0.634, 0.677)} & \makecell{0.032\\(0.022, 0.045)} & \textbf{\makecell{0.758\\(0.742, 0.775)}} & \makecell{0.637\\(0.609, 0.665)} & \textbf{\makecell{0.025\\(0.021, 0.032)}} \\
        \hline
        \multirowcell{3}{ILPD} & ERM & \textbf{\makecell{0.712\\(0.703, 0.721)}} & \textbf{\makecell{0.691\\(0.669, 0.71)}} & \textbf{\makecell{0.013\\(0.013, 0.013)}} & \textbf{\makecell{0.712\\(0.703, 0.722)}} & \textbf{\makecell{0.703\\(0.683, 0.72)}} & \textbf{\makecell{0.013\\(0.013, 0.013)}} \\
         & KL & \makecell{0.712\\(0.704, 0.721)} & \makecell{0.678\\(0.658, 0.696)} & \makecell{0.013\\(0.013, 0.013)} & \makecell{0.712\\(0.703, 0.722)} & \makecell{0.68\\(0.661, 0.696)} & \makecell{0.013\\(0.013, 0.013)} \\
         & Wass & \makecell{0.712\\(0.703, 0.72)} & \makecell{0.568\\(0.551, 0.584)} & \makecell{0.033\\(0.032, 0.034)} & \makecell{0.712\\(0.703, 0.721)} & \makecell{0.556\\(0.539, 0.573)} & \makecell{0.035\\(0.035, 0.036)} \\
        \hline
        \multirowcell{3}{SG} & ERM & \makecell{0.757\\(0.75, 0.764)} & \makecell{0.791\\(0.784, 0.798)} & \makecell{0.173\\(0.163, 0.183)} & \makecell{0.727\\(0.716, 0.739)} & \makecell{0.762\\(0.747, 0.776)} & \makecell{0.091\\(0.072, 0.11)} \\
         & KL & \textbf{\makecell{0.759\\(0.751, 0.766)}} & \textbf{\makecell{0.792\\(0.785, 0.799)}} & \makecell{0.162\\(0.154, 0.171)} & \textbf{\makecell{0.728\\(0.717, 0.74)}} & \textbf{\makecell{0.764\\(0.752, 0.776)}} & \makecell{0.095\\(0.079, 0.112)} \\
         & Wass & \makecell{0.72\\(0.712, 0.727)} & \makecell{0.787\\(0.782, 0.793)} & \textbf{\makecell{0.083\\(0.074, 0.093)}} & \makecell{0.708\\(0.701, 0.715)} & \makecell{0.729\\(0.719, 0.74)} & \textbf{\makecell{0.047\\(0.042, 0.053)}} \\
        \hline
        \multirowcell{3}{LD}& ERM & \textbf{\makecell{0.681\\(0.669, 0.692)}} & \textbf{\makecell{0.72\\(0.706, 0.734)}} & \makecell{0.193\\(0.177, 0.211)} & \textbf{\makecell{0.626\\(0.605, 0.645)}} & \textbf{\makecell{0.654\\(0.631, 0.676)}} & \makecell{0.121\\(0.099, 0.142)} \\
         & KL & \makecell{0.674\\(0.662, 0.687)} & \makecell{0.715\\(0.702, 0.73)} & \makecell{0.161\\(0.147, 0.177)} & \makecell{0.624\\(0.604, 0.644)} & \makecell{0.643\\(0.617, 0.666)} & \makecell{0.103\\(0.084, 0.12)} \\
         & Wass & \makecell{0.59\\(0.57, 0.611)} & \makecell{0.689\\(0.671, 0.706)} & \textbf{\makecell{0.084\\(0.074, 0.095)}} & \makecell{0.581\\(0.564, 0.599)} & \makecell{0.576\\(0.545, 0.605)} & \textbf{\makecell{0.057\\(0.047, 0.067)}} \\
        \hline
        \multirowcell{3}{DI} & ERM & \textbf{\makecell{0.775\\(0.767, 0.782)}} & \makecell{0.83\\(0.822, 0.838)} & \makecell{0.226\\(0.21, 0.241)} & \textbf{\makecell{0.772\\(0.764, 0.781)}} & \makecell{0.827\\(0.817, 0.836)} & \makecell{0.185\\(0.172, 0.2)} \\
         & KL & \makecell{0.772\\(0.765, 0.78)} & \textbf{\makecell{0.831\\(0.822, 0.839)}} & \makecell{0.207\\(0.193, 0.223)} & \makecell{0.772\\(0.764, 0.78)} & \textbf{\makecell{0.828\\(0.819, 0.837)}} & \makecell{0.17\\(0.158, 0.183)} \\
         & Wass & \makecell{0.772\\(0.765, 0.779)} & \makecell{0.829\\(0.82, 0.837)} & \textbf{\makecell{0.154\\(0.142, 0.166)}} & \makecell{0.725\\(0.713, 0.736)} & \makecell{0.815\\(0.805, 0.825)} & \textbf{\makecell{0.116\\(0.107, 0.126)}} \\
        \hline
        \multirowcell{3}{IO}& ERM & \makecell{0.867\\(0.856, 0.879)} & \makecell{0.899\\(0.885, 0.914)} & \makecell{0.103\\(0.088, 0.12)} & \makecell{0.851\\(0.836, 0.863)} & \makecell{0.886\\(0.873, 0.9)} & \makecell{0.082\\(0.074, 0.09)} \\
         & KL & \textbf{\makecell{0.872\\(0.859, 0.884)}} & \textbf{\makecell{0.904\\(0.89, 0.918)}} & \makecell{0.099\\(0.084, 0.117)} & \textbf{\makecell{0.852\\(0.839, 0.865)}} & \makecell{0.889\\(0.876, 0.902)} & \makecell{0.08\\(0.073, 0.087)} \\
         & Wass & \makecell{0.843\\(0.828, 0.856)} & \makecell{0.898\\(0.883, 0.912)} & \textbf{\makecell{0.076\\(0.065, 0.087)}} & \makecell{0.788\\(0.769, 0.805)} & \textbf{\makecell{0.898\\(0.884, 0.911)}} & \textbf{\makecell{0.053\\(0.048, 0.058)}} \\
        \hline
        \multirowcell{3}{CB} & ERM & \makecell{0.743\\(0.729, 0.758)} & \makecell{0.829\\(0.811, 0.844)} & \makecell{0.16\\(0.14, 0.179)} & \makecell{0.732\\(0.714, 0.75)} & \makecell{0.806\\(0.788, 0.822)} & \makecell{0.135\\(0.12, 0.153)} \\
         & KL & \textbf{\makecell{0.748\\(0.731, 0.763)}} & \textbf{\makecell{0.833\\(0.817, 0.849)}} & \makecell{0.152\\(0.133, 0.173)} & \textbf{\makecell{0.74\\(0.725, 0.755)}} & \makecell{0.81\\(0.793, 0.826)} & \makecell{0.124\\(0.111, 0.14)} \\
         & Wass & \makecell{0.74\\(0.724, 0.753)} & \makecell{0.826\\(0.81, 0.843)} & \textbf{\makecell{0.117\\(0.105, 0.131)}} & \makecell{0.714\\(0.688, 0.737)} & \textbf{\makecell{0.811\\(0.793, 0.831)}} & \textbf{\makecell{0.091\\(0.082, 0.1)}} \\
         \hline
    \end{tabular}
    \caption{Results of the ERM, KL-divergence, and Wasserstein models on the 8 datasets from UCI Repository.}
    \label{tab:results_uci}
\end{table}
We observe that the KL-divergence model performs best in more cases than the other two models. This demonstrates the effectiveness of a robust reweighting strategy of the KL-divergence model. The Wasserstein model, on the other hand, is best in most FI, but not as good as the other two models regarding accuracy and AUC. This relates to the potential weakness of Wasserstein model as it may be too conservative because of considering the unrealistic support of the feature. 

\subsection{Training Details of the FI-Based ResNet on MedMNIST}
\label{appe:training_details_medmnist}
In this part, we illustrate the details of the hyperparameters and training process of the FI-based ResNet on MedMNIST. The training is conducted by PyTorch on an NVIDIA GeForce RTX 5090. We employ the default ResNet-18 architecture and the Adam optimizer. For the majority of the datasets, the default learning rate is set to $0.001$ and the weight decay, i.e., the L2 regularization, is set to $0.0001$. Both these two values are obtained by rounding from the Bayesian optimization hyperparameter tuning. Only for several extremely large datasets, we shrink the learning rate a little to control the volatility of the loss during training. 

The training epochs are determined by the early stopping strategy, i.e., if the validation loss does not decrease for 30 epochs, the training will stop. Then, we will select the model with the best validation loss during the training process as our final model. For the learning rate, we consider the ``reduce learning rate on plateau'' strategy with the patience of 5 epochs and a factor of 0.1, i.e., if the validation loss does not decrease for 5 epochs, the learning rate will be reduced by a factor of 0.1. 

The above hyperparameters are the same for both ERM and FI-based ResNet. For the FI-based model, we also need to specify the {hyperparameter $\lambda_0$ controlling }the FI-induced regularizer in \eqref{eq:fi_def_loss_nn}. We first use Bayesian optimization to find candidate values of $\lambda_0$. Then, based on the sense of the trade-off between accuracy and FI and the performance on the validation sets, we determine the final $\lambda_0$ for each dataset. 

For the hyperparameter $\alpha$, it is adaptively determined based on the fragility $k$, and follows the formula $\alpha = \frac{A}{k}$, where $A$ is a large parameter. Since the value of $k$ during training is often below 1, we usually set $A$ between $20-100$, which is already large enough to enforce that the constraints violation is small and the convergence process is stable.
Moreover, we initialize with a small $k_{init}$, and $k$ keeps ascending during the training epochs. By initializing with a small $k$, the penalty parameter $\alpha$ is large. This restricts the feasible region significantly, effectively convexifying the loss landscape or creating a strong "basin" around simple, robust solutions. Starting with a high constraint penalty amplifies this bias. It forces the network to learn only the most dominant, robust features (e.g., ``shapes'' rather than ``texture noise'') because the strict penalty makes memorizing outliers prohibitively expensive. In this way, Stochastic Gradient Descent (SGD) and its variants have an inherent ``simplicity bias'', meaning they tend to learn simple, linear, or low-frequency functions first before fitting complex noise. \citep{boursier2025early, refinetti2023neural}. However, we also admit this is only a heuristic based on experience, and there are potentially better ways to determine $\alpha$ and $k_{init}$.

\section{Mathematical Proof}
\label{appe:proof}

\subsection{Proof of Lemma \ref{lemma:root_finding}}\label{proof:lemma_root_finding}

{{For $k_2>k_1>0$ and every $\mathbb Q\in\mathcal P(\mathcal X^2)$, the nonnegativity of $D$ implies
}{\[
    \mathbb E_{\mathbb Q}[\varepsilon(p)]
    -k_2D(\mathbb Q,\widehat{\mathbb Q})
    \leq
    \mathbb E_{\mathbb Q}[\varepsilon(p)]
    -k_1D(\mathbb Q,\widehat{\mathbb Q}).
\]}
{Taking the supremum over $\mathbb Q$ shows that $G(k_2)\leq G(k_1)$. Moreover, for every $k>0$, the defining constraint of FI is equivalent to
}{\[
    \sup_{\mathbb Q\in\mathcal P(\mathcal X^2)}
    \left\{
        \mathbb E_{\mathbb Q}[\varepsilon(p)]
        -kD(\mathbb Q,\widehat{\mathbb Q})
    \right\}
    \leq\tau,
\]}
{or equivalently, $G(k)\leq0$. }Therefore,
{\[
    \mathrm{FI}(p;\tau)=\inf\{k>0:G(k)\leq0\}.
\]}}

\hfill \Halmos

\subsection{Proof of Lemma \ref{lemma:fi_kl}}\label{proof:lemma_fi_kl}

{For every $k>0$, the Donsker--Varadhan variational formula on $\mathcal X^2$ gives
}{\[
    \sup_{\mathbb Q\in\mathcal P(\mathcal X^2)}
    \left\{
        \mathbb E_{\mathbb Q}[\varepsilon(p)]
        -kD_{\mathrm{KL}}(\mathbb Q\Vert\widehat{\mathbb Q})
    \right\}
    =
    k\ln\mathbb E_{\widehat{\mathbb Q}}
    \left[\exp\left(\frac{\varepsilon(p)}{k}\right)\right].
\]}
{Because $\varepsilon(p)$ is bounded and the condition of the lemma implies that it is not constant under $\widehat{\mathbb Q}$, the expression on the right-hand side is continuous. For $0<k_1<k_2$, strict Jensen's inequality gives
}{\[
    \mathbb E_{\widehat{\mathbb Q}}
    \left[\exp\left(\frac{\varepsilon(p)}{k_2}\right)\right]
    <
    \left(
        \mathbb E_{\widehat{\mathbb Q}}
        \left[\exp\left(\frac{\varepsilon(p)}{k_1}\right)\right]
    \right)^{k_1/k_2},
\]}
{and taking logarithms and multiplying by $k_2$ shows that the preceding entropic expression is strictly decreasing in $k$. Moreover,
}{\[
    \lim_{k\downarrow0}
    k\ln\mathbb E_{\widehat{\mathbb Q}}
    \left[\exp\left(\frac{\varepsilon(p)}{k}\right)\right]
    =\esssup_{\widehat{\mathbb Q}}\varepsilon(p)
\]}
{and
}{\[
    \lim_{k\to\infty}
    k\ln\mathbb E_{\widehat{\mathbb Q}}
    \left[\exp\left(\frac{\varepsilon(p)}{k}\right)\right]
    =\mathbb E_{\widehat{\mathbb Q}}[\varepsilon(p)].
\]}
{Since $\mathbb E_{\widehat{\mathbb Q}}[\varepsilon(p)]<\tau<\esssup_{\widehat{\mathbb Q}}\varepsilon(p)$, there is a unique $k>0$ such that $G_{\mathrm{KL}}(k)=0$. Lemma \ref{lemma:root_finding} then shows that this root equals $\mathrm{FI}_{\mathrm{KL}}(p;\tau)$.
}

\hfill \Halmos

\subsection{Proof of Proposition \ref{prop:fi_kl_ub}}

{{Let $k^*=\mathrm{FI}_{\mathrm{KL}}(p;\tau)$. By Lemma \ref{lemma:fi_kl}, $0<k^*<\infty$ and
}{\[
    \mathbb E_{\widehat{\mathbb Q}}
    \left[\exp\left(\frac{\varepsilon(p)}{k^*}\right)\right]
    =
    \exp\left(\frac{\tau}{k^*}\right).
\]}
{Thus, for any $\theta\in\mathbb R$, Markov's inequality gives
}\begin{align*}
    \widehat{\mathbb Q}(\varepsilon(p)\geq\theta)
    =
    \widehat{\mathbb Q}\left(
        \exp\left(\frac{\varepsilon(p)}{k^*}\right)
        \geq
        \exp\left(\frac{\theta}{k^*}\right)
    \right)
    \leq
    \frac{
        \mathbb E_{\widehat{\mathbb Q}}
        \left[\exp\left(\varepsilon(p)/k^*\right)\right]
    }{
        \exp\left(\theta/k^*\right)
    }
    =
    \exp\left(\frac{\tau-\theta}{k^*}\right).
\end{align*}}
\hfill \Halmos

\subsection{Proof of Theorem \ref{theorem:multi_fi_properties}}

{
For each ordered class pair $(i,j)$, we call $k>0$
feasible for $\mathrm{FI}_{i|j}(p_i;\tau)$ if
\[
    \mathbb E_{\mathbb Q}[\varepsilon_{i|j}(p_i)]
    \leq
    \tau+kD(\mathbb Q,\widehat{\mathbb Q}_{ij}),
    \qquad
    \forall\mathbb Q\in\mathcal P(\mathcal X^2).
\]
The pairwise FI is the infimum of these feasible values,
with $\inf\varnothing=+\infty$.

\textbf{(a) Convexity under probability mixtures.}
Fix $\lambda\in(0,1)$ and let
$\boldsymbol p^\lambda
=\lambda\boldsymbol p+(1-\lambda)\widetilde{\boldsymbol p}$.
For every ordered class pair $(i,j)$,
\[
    \varepsilon_{i|j}(p_i^\lambda)
    =
    \lambda\varepsilon_{i|j}(p_i)
    +(1-\lambda)\varepsilon_{i|j}(\widetilde p_i).
\]
Suppose first that
$\mathrm{FI}_{i|j}(p_i;\tau)$ and
$\mathrm{FI}_{i|j}(\widetilde p_i;\tau)$ are finite.
Let $k>0$ and $\widetilde k>0$ be feasible for these two
pairwise FIs, respectively.
For every $\mathbb Q\in\mathcal P(\mathcal X^2)$,
\begin{align*}
    \mathbb E_{\mathbb Q}
    [\varepsilon_{i|j}(p_i^\lambda)]
    &=
    \lambda\mathbb E_{\mathbb Q}
    [\varepsilon_{i|j}(p_i)]
    +(1-\lambda)\mathbb E_{\mathbb Q}
    [\varepsilon_{i|j}(\widetilde p_i)]\\
    &\leq
    \tau+
    \bigl(\lambda k+(1-\lambda)\widetilde k\bigr)
    D(\mathbb Q,\widehat{\mathbb Q}_{ij}).
\end{align*}
Thus, $\lambda k+(1-\lambda)\widetilde k$ is feasible
for $\mathrm{FI}_{i|j}(p_i^\lambda;\tau)$.
Taking feasible sequences approaching the two infima gives
\[
    \mathrm{FI}_{i|j}(p_i^\lambda;\tau)
    \leq
    \lambda\mathrm{FI}_{i|j}(p_i;\tau)
    +(1-\lambda)
    \mathrm{FI}_{i|j}(\widetilde p_i;\tau).
\]
If either pairwise FI on the right-hand side is infinite,
the inequality holds immediately.
Averaging over all ordered class pairs as in
Equation \eqref{eq:def_multiclass_fi} proves the claim.

\textbf{(b) Prorobustness.}
Under the stated condition, for every ordered class pair
$(i,j)$ and every $\mathbb Q\in\mathcal P(\mathcal X^2)$,
\[
    \mathbb E_{\mathbb Q}[\varepsilon_{i|j}(p_i)]
    \leq\tau
    \leq
    \tau+kD(\mathbb Q,\widehat{\mathbb Q}_{ij}),
    \qquad \forall k>0,
\]
where the second inequality follows from $D\geq0$.
Hence every $k>0$ is feasible for every pairwise FI,
so $\mathrm{FI}_{i|j}(p_i;\tau)=0$ for all $i\neq j$.
Their arithmetic mean is therefore zero.

\textbf{(c) Antifragility.}
Suppose that, for some ordered class pair $(i,j)$,
$\mathbb E_{\widehat{\mathbb Q}_{ij}}
[\varepsilon_{i|j}(p_i)]>\tau$.
Evaluating the feasibility condition at
$\mathbb Q=\widehat{\mathbb Q}_{ij}$ would require
\[
    \mathbb E_{\widehat{\mathbb Q}_{ij}}
    [\varepsilon_{i|j}(p_i)]
    \leq
    \tau+kD(\widehat{\mathbb Q}_{ij},
             \widehat{\mathbb Q}_{ij})
    =\tau,
\]
which is impossible.
Thus, $\mathrm{FI}_{i|j}(p_i;\tau)=+\infty$.
Since all pairwise FIs are nonnegative and receive
positive aggregation weights,
$\mathrm{FI}(\boldsymbol p;\tau)=+\infty$.

\textbf{(d) $\tau$-FI tradeoff.}
If $\tau_1\geq\tau_2$, every $k>0$ feasible for
$\mathrm{FI}_{i|j}(p_i;\tau_2)$ is also feasible for
$\mathrm{FI}_{i|j}(p_i;\tau_1)$.
Taking infima yields
\[
    \mathrm{FI}_{i|j}(p_i;\tau_1)
    \leq
    \mathrm{FI}_{i|j}(p_i;\tau_2).
\]
Averaging these inequalities over all ordered class pairs
proves the claim.

\textbf{(e) Monotonicity.}
Throughout this part, all FIs are evaluated using
KL divergence.
For any bounded random variable $Z$ on the pair space
and any $k>0$, the variational formula for KL divergence gives
\[
    \sup_{\mathbb Q\in\mathcal P(\mathcal X^2)}
    \left\{
        \mathbb E_{\mathbb Q}[Z]
        -kD_{\mathrm{KL}}
        (\mathbb Q\Vert\widehat{\mathbb Q}_{ij})
    \right\}
    =
    k\log\mathbb E_{\widehat{\mathbb Q}_{ij}}
    [\exp(Z/k)].
\]
Since the ranking errors lie in $[-1,1]$,
this identity applies to both
$\varepsilon_{i|j}(p_i)$ and
$\varepsilon_{i|j}(\widetilde p_i)$.
In particular, $k>0$ is feasible for
$\mathrm{FI}_{i|j}(p_i;\tau)$ if and only if
\[
    k\log\mathbb E_{\widehat{\mathbb Q}_{ij}}
    \left[\exp\bigl(\varepsilon_{i|j}(p_i)/k\bigr)\right]
    \leq\tau.
\]
By the assumed first-order stochastic dominance and
the monotonicity of $z\mapsto\exp(z/k)$,
\[
    \mathbb E_{\widehat{\mathbb Q}_{ij}}
    \left[\exp\bigl(\varepsilon_{i|j}(p_i)/k\bigr)\right]
    \geq
    \mathbb E_{\widehat{\mathbb Q}_{ij}}
    \left[
        \exp\bigl(\varepsilon_{i|j}(\widetilde p_i)/k\bigr)
    \right]
\]
for every $k>0$.
Consequently, every $k>0$ feasible for
$\mathrm{FI}_{i|j}(p_i;\tau)$ is also feasible for
$\mathrm{FI}_{i|j}(\widetilde p_i;\tau)$.
Taking infima gives
\[
    \mathrm{FI}_{i|j}(p_i;\tau)
    \geq
    \mathrm{FI}_{i|j}(\widetilde p_i;\tau).
\]
Averaging over all ordered class pairs establishes
\[
    \mathrm{FI}_{\mathrm{KL}}(\boldsymbol p;\tau)
    \geq
    \mathrm{FI}_{\mathrm{KL}}
    (\widetilde{\boldsymbol p};\tau),
\]
completing the proof.
}

\subsection{Proof of Lemma \ref{lemma:lower_bound_k}}\label{proof:lemma_lower_bound_k}

{{Let $(k^*,\BFB^*)$ be an }optimal solution to {Problem \eqref{eq:prob_general}. Assumption \ref{asmp:tau} implies
}\begin{equation}
    \label{eq:lower_bound_k_dro_lower}
    \tau+\delta_1
    \leq
    \inf_{\BFB\in\mathcal B}
    \sup_{\substack{\mathbb P\in\mathcal P(\mathcal X\times\mathcal Y)\\
    D(\mathbb P,\hat{\mathbb P})\leq\delta_2}}
    \left\{
    \mathbb E_{\mathbb P}[\ell(\BFB^T\BFx,y)]+R(\BFB)
    \right\}
    \leq
    \sup_{\substack{\mathbb P\in\mathcal P(\mathcal X\times\mathcal Y)\\
    D(\mathbb P,\hat{\mathbb P})\leq\delta_2}}
    \left\{
    \mathbb E_{\mathbb P}[\ell(\BFB^{*T}\BFx,y)]+R(\BFB^*)
    \right\}.
\end{equation}
The {second inequality holds because $\BFB^*\in\mathcal B$. Next, the feasibility of $(k^*,\BFB^*)$ for Problem \eqref{eq:prob_general} implies that, for every $\mathbb P\in\mathcal P(\mathcal X\times\mathcal Y)$ satisfying $D(\mathbb P,\hat{\mathbb P})\leq\delta_2$,
}\begin{align*}
    \mathbb E_{\mathbb P}[\ell(\BFB^{*T}\BFx,y)]+R(\BFB^*)
    \leq \tau+k^*D(\mathbb P,\hat{\mathbb P})\leq \tau+k^*\delta_2.
\end{align*}
{Taking the supremum over the same distance ball gives
}\begin{equation}
    \label{eq:lower_bound_k_dro_upper}
    \sup_{\substack{\mathbb P\in\mathcal P(\mathcal X\times\mathcal Y)\\
    D(\mathbb P,\hat{\mathbb P})\leq\delta_2}}
    \left\{
    \mathbb E_{\mathbb P}[\ell(\BFB^{*T}\BFx,y)]+R(\BFB^*)
    \right\}
    \leq\tau+k^*\delta_2.
\end{equation}
{Combining \eqref{eq:lower_bound_k_dro_lower} and \eqref{eq:lower_bound_k_dro_upper} yields
}{\[
    \tau+\delta_1
    \leq
    \sup_{\substack{\mathbb P\in\mathcal P(\mathcal X\times\mathcal Y)\\
    D(\mathbb P,\hat{\mathbb P})\leq\delta_2}}
    \left\{
    \mathbb E_{\mathbb P}[\ell(\BFB^{*T}\BFx,y)]+R(\BFB^*)
    \right\}
    \leq \tau+k^*\delta_2,
\]}
{and therefore $\delta_1\leq k^*\delta_2$. Since $\delta_2>0$ by Assumption \ref{asmp:tau}, we conclude that $k^*\geq\delta_1/\delta_2>0$.}}\hfill \Halmos

\subsection{{Proof of Theorem \ref{theorem:kl_wasserstein_ovo_bound}}}\label{proof:theorem_kl_wasserstein_ovo_bound}

{{Recall that $\widehat{\mathbb{P}}_c$ denotes the empirical feature distribution conditional on $Y=c$ for each $c\in[C]$. For the KL case, Problem \eqref{eq:prob_general} uses $D_{\mathrm{KL}}(\mathbb P\Vert\hat{\mathbb P})$, whereas $\mathrm{FI}_{i|j}$ is defined using $D_{\mathrm{KL}}(\mathbb Q\Vert\widehat{\mathbb Q}_{ij})$. For the Wasserstein case, Problem \eqref{eq:prob_general} uses the distance $W$ in \eqref{eq:def_ot}. On $\mathcal X^2$, the additive extension of the same ground cost defines
}\begin{equation}
    \label{eq:pair_wasserstein_distance}
    W(\mathbb Q,\widehat{\mathbb Q}_{ij})
    :=
    \inf_{\pi\in\Pi(\mathbb Q,\widehat{\mathbb Q}_{ij})}
    \mathbb E_{\pi}
    \left[
    c(\BFx^i,i,\hat{\BFx}^{\,i},i)
    +c(\BFx^j,j,\hat{\BFx}^{\,j},j)
    \right].
\end{equation}
{Both occurrences of $W$ are standard Wasserstein distances; the underlying space determines whether the single-observation cost or its additive extension is used.
}}

{{Fix the feasible pair $(k,\BFB)$ from the theorem. Feasibility and Assumption \ref{asmp:tau} give
}\begin{align*}
    \tau+\delta_1
    \leq
    \inf_{\widetilde{\BFB}\in\mathcal B}
    \sup_{\substack{\mathbb P\in\mathcal P(\mathcal X\times\mathcal Y)\\
    D(\mathbb P,\hat{\mathbb P})\leq\delta_2}}
    \left\{
    \mathbb E_{\mathbb P}
    [\ell(\widetilde{\BFB}^{T}\BFx,Y)]+R(\widetilde{\BFB})
    \right\}
    \leq
    \sup_{\substack{\mathbb P\in\mathcal P(\mathcal X\times\mathcal Y)\\
    D(\mathbb P,\hat{\mathbb P})\leq\delta_2}}
    \left\{
    \mathbb E_{\mathbb P}[\ell(\BFB^T\BFx,Y)]+R(\BFB)
    \right\}
    \leq \tau+k\delta_2.
\end{align*}
{Hence, $k\geq\delta_1/\delta_2>0$.
}}

{{Fix an ordered pair $(i,j)$ with $i\neq j$, and let $\mathbb Q$ be a pair law whose discrepancy from $\widehat{\mathbb Q}_{ij}$ is finite under the selected divergence or distance. Denote its marginals by $\mathbb{Q}_i$ and $\mathbb{Q}_j$, and define
}{\[
    \mathcal L_{ij}(\mathbb Q)(d\BFx,c)
    :=
    \begin{cases}
        q_i\mathbb{Q}_i(d\BFx), & c=i,\\
        q_j\mathbb{Q}_j(d\BFx), & c=j,\\
        q_c\widehat{\mathbb{P}}_c(d\BFx), & c\notin\{i,j\}.
    \end{cases}
\]}
{Thus, $\mathcal L_{ij}(\mathbb Q)$ has the same label probabilities as $\hat{\mathbb P}$, replaces only the class-}$i$ {and class-$j$ feature conditionals, and satisfies
}\begin{align*}
    \mathbb E_{\mathcal L_{ij}(\mathbb Q)}
    [\ell(\BFB^T\BFx,Y)]
    ={}&
    q_i\mathbb E_{\mathbb{Q}_i}[\ell(\BFB^T\BFx,i)]
    +q_j\mathbb E_{\mathbb{Q}_j}[\ell(\BFB^T\BFx,j)]+\sum_{c\notin\{i,j\}}q_c
    \mathbb E_{\widehat{\mathbb{P}}_c}[\ell(\BFB^T\BFx,c)].
\end{align*}}

{{For any $(\BFx^i,\BFx^j)\in\mathcal X^2$, recall that
}{\[
    \varepsilon_{i|j}(p_i)
    =
    p_i(\BFx^j;\BFB)-p_i(\BFx^i;\BFB).
\]}
{Assumption \ref{asmp:probability_loss_dominance} gives
}\begin{align*}
    \varepsilon_{i|j}(p_i)
    &\leq
    1-p_i(\BFx^i;\BFB)
    \leq
    a\ell(\BFB^T\BFx^i,i)+b,\\
    \varepsilon_{i|j}(p_i)
    &\leq
    p_i(\BFx^j;\BFB)
    \leq
    1-p_j(\BFx^j;\BFB)
    \leq
    a\ell(\BFB^T\BFx^j,j)+b.
\end{align*}
{Taking their convex combination with weights $q_i/(q_i+q_j)$ and $q_j/(q_i+q_j)$ and then taking expectations under $\mathbb Q$ yields
}{\[
    \mathbb E_{\mathbb Q}[\varepsilon_{i|j}(p_i)]
    \leq
    b
    +\frac{a}{q_i+q_j}
    \left[
    q_i\mathbb E_{\mathbb{Q}_i}[\ell(\BFB^T\BFx,i)]
    +q_j\mathbb E_{\mathbb{Q}_j}[\ell(\BFB^T\BFx,j)]
    \right].
\]}
{Let $\Delta$ denote the discrepancy used in Problem \eqref{eq:prob_general}, namely $D_{\mathrm{KL}}$ in the KL case and $W$ in the Wasserstein case. Applying the feasibility of $(k,\BFB)$ at $\mathcal L_{ij}(\mathbb Q)$, together with the preceding two displays, yields the common loss-budget inequality
}{\[
    \mathbb E_{\mathbb Q}[\varepsilon_{i|j}(p_i)]
    \leq
    t_{ij}(\BFB)
    +\frac{ak}{q_i+q_j}
    \Delta\left(\mathcal L_{ij}(\mathbb Q),\hat{\mathbb P}\right).
\]}}

{{We next control the last term in each case. In the KL case, the chain rule gives
}{\[
    D_{\mathrm{KL}}
    \left(\mathcal L_{ij}(\mathbb Q)\Vert\hat{\mathbb P}\right)
    =
    q_iD_{\mathrm{KL}}(\mathbb{Q}_i\Vert\widehat{\mathbb{P}}_i)
    +q_jD_{\mathrm{KL}}(\mathbb{Q}_j\Vert\widehat{\mathbb{P}}_j).
\]}
{Moreover,
}\begin{align*}
    D_{\mathrm{KL}}
    \left(\mathbb Q\Vert\widehat{\mathbb Q}_{ij}\right)
    ={}&
    D_{\mathrm{KL}}(\mathbb{Q}_i\Vert\widehat{\mathbb{P}}_i)
    +D_{\mathrm{KL}}(\mathbb{Q}_j\Vert\widehat{\mathbb{P}}_j)+D_{\mathrm{KL}}(\mathbb Q\Vert \mathbb{Q}_i\otimes \mathbb{Q}_j),
\end{align*}
{and therefore
}{\[
    D_{\mathrm{KL}}
    \left(\mathcal L_{ij}(\mathbb Q)\Vert\hat{\mathbb P}\right)
    \leq
    \max\{q_i,q_j\}
    D_{\mathrm{KL}}
    \left(\mathbb Q\Vert\widehat{\mathbb Q}_{ij}\right).
\]}}

{In the Wasserstein case, for each $c\in\{i,j\}$, define
}{\[
    W(\mathbb{Q}_c,\widehat{\mathbb{P}}_c)
    :=
    \inf_{\pi_c\in\Pi(\mathbb{Q}_c,\widehat{\mathbb{P}}_c)}
    \mathbb E_{\pi_c}[\|\BFx-\hat{\BFx}\|].
\]}
{Mixing classwise couplings for classes $i$ and $j$ with identity couplings for the remaining classes and taking the infimum gives
}{\[
    W
    \left(\mathcal L_{ij}(\mathbb Q),\hat{\mathbb P}\right)
    \leq
    q_iW(\mathbb{Q}_i,\widehat{\mathbb{P}}_i)+q_jW(\mathbb{Q}_j,\widehat{\mathbb{P}}_j).
\]}
{Every coupling between $\mathbb Q$ and $\widehat{\mathbb Q}_{ij}$ projects onto admissible couplings of both coordinate marginals. Hence, the additive cost in \eqref{eq:pair_wasserstein_distance} implies
}{\[
    W
    \left(\mathbb Q,\widehat{\mathbb Q}_{ij}\right)
    \geq
    W(\mathbb{Q}_i,\widehat{\mathbb{P}}_i)+W(\mathbb{Q}_j,\widehat{\mathbb{P}}_j),
\]}
{so that
}{\[
    W
    \left(\mathcal L_{ij}(\mathbb Q),\hat{\mathbb P}\right)
    \leq
    \max\{q_i,q_j\}
    W
    \left(\mathbb Q,\widehat{\mathbb Q}_{ij}\right).
\]}

{{Substituting these bounds into the common loss-budget inequality gives
}{\[
    \mathbb E_{\mathbb Q}[\varepsilon_{i|j}(p_i)]
    \leq
    t_{ij}(\BFB)
    +\frac{a\max\{q_i,q_j\}}{q_i+q_j}k
    D_{\mathrm{KL}}
    \left(\mathbb Q\Vert\widehat{\mathbb Q}_{ij}\right),
\]}
{in the KL case, }and
{\[
    \mathbb E_{\mathbb Q}[\varepsilon_{i|j}(p_i)]
    \leq
    t_{ij}(\BFB)
    +\frac{a\max\{q_i,q_j\}}{q_i+q_j}k
    W
    \left(\mathbb Q,\widehat{\mathbb Q}_{ij}\right),
\]}
{in the Wasserstein case. If the pair discrepancy is infinite, the FI constraint is automatic because the preceding argument gives $k>0$. This establishes the stated bound for each ordered class pair.
}}

{{Finally, $t_{\max}(\BFB)\geq t_{ij}(\BFB)$ and FI is non-increasing in its target. Thus,
}\begin{align*}
    &\mathrm{FI}
    \left(
    p_1(\cdot;\BFB),\ldots,p_C(\cdot;\BFB);
    t_{\max}(\BFB)
    \right)\leq
    \frac{ak}{C(C-1)}
    \sum_{\substack{i,j\in[C]\\i\neq j}}
    \frac{\max\{q_i,q_j\}}{q_i+q_j}
    \leq ak,
\end{align*}
{which establishes \eqref{eq:kl_wasserstein_ovo_bound}. }}\hfill \Halmos

\subsection{Proof of Theorem \ref{theorem:kl_reformulation}}
{{By Lemma \ref{lemma:lower_bound_k}, every optimal solution under Assumption \ref{asmp:tau} has $k^*>0$, so restricting Problem \eqref{eq:prob_general_kl} to $k>0$ does not change its optimal solutions. For any $k>0$, the result follows }by applying the variational formula{:
}\begin{align*}
    &\sup_{\mathbb{P} \in \mathcal{P}(\supp{\phat})} \left\{
        \mathbb{E}_{\mathbb{P}}[\ell \left(\BFB^T\BFx, y \right)] - k D_{\mathrm{KL}}(\mathbb{P}\Vert\hat{\mathbb{P}})
    \right\} \\
    &\quad= k \sup_{\mathbb{P} \in \mathcal{P}(\supp{\phat})} \left\{
        \mathbb{E}_{\mathbb{P}}\left[\frac{\ell \left(\BFB^T\BFx, y \right)}{k}\right] - D_{\mathrm{KL}}(\mathbb{P}\Vert\hat{\mathbb{P}})
    \right\} \\
    &\quad= k \ln \left( \mathbb{E}_{\phat}\left[ \exp \left(\frac{\ell \left(\BFB^T\BFx, y \right)}{k}\right)\right] \right).
\end{align*}}
The convexity is guaranteed because it is the composition of the convex function $\ell(\cdot)$ and the perspective of the log-sum-exp function, which is convex and increasing. The worst-case distribution is also provided by \cite{donsker1975asymptotic}. 

\hfill \Halmos

\subsection{Proof of {Lemma \ref{lemma:1wass_reformulation}}}
{For the norm cost, its convex conjugate with respect to the first argument is
}\begin{equation*}
    c^{1*}(\BFzeta,\hat{\BFx})
    = \sup_{\BFx\in\mathbb{R}^M}
    \left\{\BFzeta^T\BFx-\|\BFx-\hat{\BFx}\|\right\}
    = \begin{cases}
        \BFzeta^T\hat{\BFx}, & \|\BFzeta\|_*\leq 1,\\
        +\infty, & \|\BFzeta\|_*>1.
    \end{cases}
\end{equation*}
{For $k>0$, Proposition \ref{prop:convex_ot_reformulation} and $\mathcal{X}=\mathbb{R}^M$ therefore imply that
}\begin{align*}
    \ell_c(\BFB,k,\hat{\BFx},y)
    &= \sup_{\BFzeta\in\dom{\ell^{1*}(\cdot,y)}}
    \left\{k c^{1*}\left(\frac{\BFB\BFzeta}{k},\hat{\BFx}\right)
    -\ell^{1*}(\BFzeta,y)\right\}\\
    &= \begin{cases}
        \displaystyle
        \sup_{\BFzeta\in\dom{\ell^{1*}(\cdot,y)}}
        \left\{\BFzeta^T\BFB^T\hat{\BFx}-\ell^{1*}(\BFzeta,y)\right\}
        =\ell(\BFB^T\hat{\BFx},y),
        & \displaystyle \sup_{\BFzeta\in\dom{\ell^{1*}(\cdot,y)}}
        \|\BFB\BFzeta\|_*\leq k,\\
        +\infty, & \text{otherwise},
    \end{cases}
\end{align*}
{where the finite-case equality follows from the Fenchel--Moreau representation of $\ell(\cdot,y)$. For $k=0$, the same expression follows directly from that representation: the supremum over $\BFx\in\mathbb{R}^M$ is finite if and only if $\BFB\BFzeta=\BFzero$ for every $\BFzeta\in\dom{\ell^{1*}(\cdot,y)}$, which is precisely $\sup_{\BFzeta\in\dom{\ell^{1*}(\cdot,y)}}\|\BFB\BFzeta\|_*\leq 0$}.

{Substituting this expression into the standard OT transformation in Appendix \ref{appe:fi_reformulation}, the robust constraint in Problem \eqref{eq:prob_general} is finite }if and only if
\begin{equation*}
    \sup_{\substack{y\in\mathcal Y\\
    \BFzeta\in\dom{\ell^{1*}(\cdot,y)}}}
    \|\BFB\BFzeta\|_*\leq k.
\end{equation*}
{Under this condition, that constraint reduces to
}\begin{equation*}
    \frac{1}{N}\sum_{n\in[N]}\max_{y_n\in\mathcal{Y}}
    \left\{\ell(\BFB^T\hat{\BFx}_n,y_n)
    -k\gamma\mathbb{I}(y_n\neq\hat{y}_n)\right\}
    +R(\BFB)-\tau\leq 0.
\end{equation*}
{Together with the objective $\min_{k\geq 0,\,\BFB\in\mathcal{B}} k$, these are exactly the two constraints in \eqref{eq:1wass_reformulation}. }\hfill \Halmos

\subsection{Proof of Theorem \ref{theorem:cross_entropy_reformulation}}
{According to (\cite{boyd2004convex} Example 3.25), we can calculate the convex conjugate of the cross-entropy loss given label $y$ as
$$
    \ell_{CE}^{1*}(\BFzeta, y) = \begin{cases}
        \sum_{i \in [C]} (\BFzeta_i + \mathbb{I}(i = y )) \ln (\BFzeta_i + \mathbb{I}(i = y )), & \text{if}\ \BFzeta + \BFe_y \geq 0 \ \text{and}\ \BFone^T(\BFzeta + \BFe_y) = 1, \\
        +\infty, & \text{otherwise}.
    \end{cases}
$$
Therefore, we have $\dom{\ell^{1*}} = \{\BFzeta \in \mathbb{R}^C \mid \BFzeta \geq -\BFe_y, \BFone^T \BFzeta = 0\} = \{\BFzeta' - \BFe_y \mid \BFzeta' \in \Delta^{C - 1}\}$, where $\Delta^{C - 1}$ is the $(C-1)$-simplex. There are $C$ extreme points in this domain, and they are ${\BFe_1 - \BFe_y, \dots, \BFe_C - \BFe_y}$. Hence, 
$$
    \sup_{\BFzeta \in \dom{\ell^{1*}}} \|\BFB\BFzeta\|_* \leq k \Leftrightarrow \|\BFbeta_i - \BFbeta_y\|_* \leq k, \ \forall i \in [C].
$$
These constraints are for the training sample with label $y$. Sum over all constraints induced by the training samples with all labels $y \in \mathcal{Y}$ and keep only the unique constraints, we have 
$$
    \|\BFbeta_i - \BFbeta_j\|_* \leq k, \ \forall i, j \in [C] \ \text{and}\ i < j.
$$
Then, for $y_n \neq \hat{y}_n$, we have 
\begin{align*}
    &\phantom{=} \ell(\BFB^T\hat{\BFx}_n, y_n) - k\gamma \mathbb{I}(y_n \neq \hat{y}_n) - \ell(\BFB^T\hat{\BFx}_n, \hat{y}_n) \\
    &= (\BFbeta_{\hat{y}_n} - \BFbeta_{y_n})^T \hat{\BFx}_n - k\gamma \mathbb{I}(y_n \neq \hat{y}_n) \\
    &\leq \|\BFbeta_{y_n} - \BFbeta_{\hat{y}_n}\|_* \|\hat{\BFx}_n\| - k\gamma \mathbb{I}(y_n \neq \hat{y}_n) \\
    &\leq k( \|\hat{\BFx}_n\| - \gamma)_+.
\end{align*}
The first inequality is due to the H\"{o}lder's inequality. The second inequality is due to the constraint $\|\BFbeta_i - \BFbeta_j\|_* \leq k$. Notice that we obtain 0 when $y_n = \hat{y}_n$. Therefore, we have 
{$$
    \max_{y_n \in \mathcal{Y}} \{ \ell(\BFB^T\hat{\BFx}_n, y_n) - k\gamma \mathbb{I}(y_n \neq \hat{y}_n)\} \leq \ell(\BFB^T\hat{\BFx}_n, \hat{y}_n) + k ( \|\hat{\BFx}_n\| - \gamma)_+.
$$}
This implies that any feasible solution to the constraint $\frac{1}{N} \sum_{n \in [N]} \ell(\BFB^T\hat{\BFx}_n, \hat{y}_n) + k (\|\hat{\BFx}_n\| - \gamma)_+ + R(\BFB) - \tau \leq 0$ is also feasible to the constraint $\frac{1}{N} \sum_{n \in [N]} \max_{y_n \in \mathcal{Y}} \{ \ell(\BFB^T\hat{\BFx}_n, y_n) - k\gamma \mathbb{I}(y_n \neq \hat{y}_n)\} + R(\BFB) - \tau \leq 0$.} 

For the equivalence condition, when $\gamma \geq \max_{n\in[N]} \|\hat{\BFx}_n\|$, we have for $y_n \neq \hat{y}_n$,
$$
    \ell(\BFB^T\hat{\BFx}_n, y_n) - k\gamma \mathbb{I}(y_n \neq \hat{y}_n) - \ell(\BFB^T\hat{\BFx}_n, \hat{y}_n) \leq k ( \|\hat{\BFx}_n\| - \gamma) \leq 0.
$$
{Therefore, 
{$$
    \max_{y_n \in \mathcal{Y}} \{ \ell(\BFB^T\hat{\BFx}_n, y_n) - k\gamma \mathbb{I}(y_n \neq \hat{y}_n)\} = \ell(\BFB^T\hat{\BFx}_n, \hat{y}_n) + k ( \|\hat{\BFx}_n\| - \gamma)_+ = \ell(\BFB^T\hat{\BFx}_n, \hat{y}_n).
$$}}
Hence, the reformulation is equivalent to the original problem. \hfill \Halmos

\subsection{{Proof of Corollary \ref{coro:flipping_label_attack}}}\label{proof:coro_flipping_label_attack}
{{For any label vector $\BFz=(z_n)_{n\in[N]}\in\mathcal Y^N$, define
}{\[
    L_{CE}(\BFB;\BFz)
    :=\frac{1}{N}\sum_{n\in[N]}\ell_{CE}(\BFB^T\hat{\BFx}_n,z_n).
\]}
{Let $\mathcal F:=\{n\in[N]:y_n^{\mathrm{true}}\neq\hat y_n\}$, so that $|\mathcal F|=N_{\mathrm{flip}}$. For the cross-entropy loss, the summands corresponding to $n\notin\mathcal F$ vanish, and hence
}\begin{align*}
    \left|L_{CE}(\BFB^*;\BFy^{\mathrm{true}})-L_{CE}(\BFB^*;\hat{\BFy})\right|
    &=\left|\frac{1}{N}\sum_{n\in\mathcal F}
    (\BFbeta_{\hat y_n}^*-\BFbeta_{y_n^{\mathrm{true}}}^*)^T\hat{\BFx}_n\right|\\
    &\leq\frac{1}{N}\sum_{n\in\mathcal F}
    \left|(\BFbeta_{\hat y_n}^*-\BFbeta_{y_n^{\mathrm{true}}}^*)^T\hat{\BFx}_n\right|\\
    &\leq\frac{1}{N}\sum_{n\in\mathcal F}
    \|\BFbeta_{\hat y_n}^*-\BFbeta_{y_n^{\mathrm{true}}}^*\|_*\,\|\hat{\BFx}_n\|\\
    &\leq\phi k^*\frac{N_{\mathrm{flip}}}{N}.
\end{align*}
{The first inequality follows from the triangle inequality, the second from }H\"{o}lder's inequality, {and the third from the constraints $\|\BFbeta_i^*-\BFbeta_j^*\|_*\leq k^*$ in Problem \eqref{eq:cross_entropy_reformulation}}.}

\hfill \Halmos

\subsection{Proof of Theorem \ref{theorem:hinge_type_reformulation}}
The proof mainly consists of three parts. First, we establish the equivalence between the effective domain of convex conjugate and the subderivative space. Then, we figure out the constraint $\|\BFbeta_i - \BFbeta_j\|_* \leq \frac{k}{\theta}$ for all $i, j \in [C]$ and $i < j$. Finally, we deal with the uncertainty of the label $\hat{y}_n$ similar to the proof of Theorem \ref{theorem:cross_entropy_reformulation}.

First, we consider the following Lemma. 
\begin{lemmaAp}
    \label{lemma:equivalence_dom_sub}
    Suppose $f:\mathbb{R}^n\to\mathbb{R}$ is convex and {subdifferentiable}. Then,
    {$$
        \mathrm{cl}(\dom{f^*})
        =\mathrm{cl}\left(\bigcup_{\BFx\in\mathbb{R}^n}\partial f(\BFx)\right).
$$}
\end{lemmaAp}
Lemma \ref{lemma:equivalence_dom_sub} {follows from standard convex-conjugacy results }\citep{rockafellar1970convex}. For completeness, we {provide a proof in Section }\ref{sec:proof_fenchel_young}.

Then, we try to figure out all the extreme points of $\text{cl}[\dom{\ell^{1*}}]$, where $\mathrm{cl}[\cdot]$ means the closure operator. For the sample with label $y$, consider the points of $\BFzero$ and $ \theta(\BFe_{y'} - \BFe_{y}), y' \in \mathcal{Y}/\{y\}$. We show that they are the extreme points of $\mathrm{cl}[\dom{\ell^{1*}}]$. Let $\mathcal{D} = \mathrm{ConvexHull} \{\BFzero, \theta(\BFe_{y'} - \BFe_{y}), y' \in \mathcal{Y}/\{y\}\}$; we aim to show 
$$
    \mathrm{cl}[\dom{\ell^{1*}}] = \mathcal{D}.
$$

We show $\mathrm{cl}[\dom{\ell^{1*}}] \subseteq \mathcal{D}$ first. Notice that our loss function 
$$
    \ell(\BFB^T\BFx, y) = \max_{y' \neq y} \rho((\BFbeta_{y} - \BFbeta_{y'})^T \BFx) = \max_{y' \neq y} \rho((\BFe_{y} - \BFe_{y'})^T \BFB^T\BFx).
$$
It is a piecewise convex function. In order to investigate its subgradient, we divide the discussion into two cases. First, consider the $\BFu$ such that $\ell(\BFu, y)$ is in the interior of piece $y'$. We have $\ell(\BFu, y) = \rho((\BFe_{y} - \BFe_{y'})^T \BFu) $, and 
{$$
    \partial_{\BFu}\ell(\BFu,y)
    =(\BFe_y-\BFe_{y'})\,
    \partial\rho\big((\BFe_y-\BFe_{y'})^T\BFu\big).
$$}
Since {$\partial\rho\big((\BFe_y-\BFe_{y'})^T\BFu\big)\subseteq[-\theta,0]$}, we have {$\partial_{\BFu}\ell(\BFu,y)\subseteq\mathcal D$}.

Then, we consider the junction $\BFu$ of two pieces. Without loss of generality, suppose it is the junction of piece $y_1'$ and $y_2'$. Then, we know its subgradient is the convex set as 
{$$
    \mathrm{ConvexHull}\!\left(
    \bigcup_{r\in\{1,2\}}
    (\BFe_y-\BFe_{y'_r})\,
    \partial\rho\big((\BFe_y-\BFe_{y'_r})^T\BFu\big)
    \right),
$$}which is a subset of $\mathcal{D}$. Consequently, we have $\mathrm{cl}[\dom{\ell^{1*}}] \subseteq \mathcal{D}$.

Then, we show $\mathcal{D} \subseteq \mathrm{cl}[\dom{\ell^{1*}}]$. Let $\BFzeta_\mathcal{D}$ denote one extreme point of $\mathcal{D}$. To show $\mathcal{D} \subseteq \mathrm{cl}[\dom{\ell^{1*}}]$, we need to find a sequence $\{\BFzeta_n\}$ such that $\BFzeta_n \in \dom{\ell^{1*}}$ and $\BFzeta_n \rightarrow \BFzeta_\mathcal{D}$ for arbitrary $\BFzeta_{\mathcal{D}}$. If $\BFzeta_\mathcal{D} = \BFzero$, this requirement is consistent with $\sup_{u\in\mathbb{R}} \partial \rho(u) = 0$; if $\BFzeta_\mathcal{D}$ is other extreme points, this requirement is consistent with $\inf_{u\in\mathbb{R}} \partial \rho(u) = - \theta$. Therefore, we have $\mathcal{D} \subseteq \mathrm{cl}[\dom{\ell^{1*}}]$. 

On top of that, reconsider the constraint in Lemma \ref{lemma:1wass_reformulation} as 
$$
    \sup_{\BFzeta \in \dom{\ell^{1*}}} \|\BFB\BFzeta\|_* \leq k \Leftrightarrow 
    \sup_{\BFzeta \in \mathrm{cl}[\dom{\ell^{1*}}]} \|\BFB\BFzeta\|_* \leq k \Leftrightarrow
    \theta\|\BFbeta_{y'} - \BFbeta_y\|_* \leq k, \ \forall y' \in \mathcal{Y}/\{y\}.
$$
The first equivalence results from the fact that $\inf_{u\in\mathbb{R}} \partial \rho(u)$ and $\sup_{u\in\mathbb{R}} \partial \rho(u)$ both exist and are bounded. The second equivalence is due to $\mathrm{cl}[\dom{\ell^{1*}}] = \mathcal{D}$. Sum over all constraints induced by the training samples with all labels $y \in \mathcal{Y}$ and keep only the unique constraints, we have
$$
    \|\BFbeta_i - \BFbeta_j\|_* \leq \frac{k}{\theta}, \ \forall i, j \in [C] \ \text{and}\ i < j.
$$

Then, we conduct the same analysis similar to the cross-entropy case. Let {$y_n^\dagger \in \arg \max_{y' \neq y_n} \rho((\BFbeta_{y_n} - \BFbeta_{y'})^T \hat{\BFx}_n)$ and $\hat{y}_n^\dagger \in \arg \max_{y' \neq \hat{y}_n} \rho((\BFbeta_{\hat{y}_n} - \BFbeta_{y'})^T \hat{\BFx}_n)$}. We have for $y_n \neq \hat{y}_n$,
\begin{align*}
    &\phantom{=} \ell(\BFB^T\hat{\BFx}_n, y_n) - k\gamma \mathbb{I}(y_n \neq \hat{y}_n) - \ell(\BFB^T\hat{\BFx}_n, \hat{y}_n) \\
    & = \max_{y' \neq y_n} \rho((\BFbeta_{y_n} - \BFbeta_{y'})^T \hat{\BFx}_n) - \max_{y' \neq \hat{y}_n} \rho((\BFbeta_{\hat{y}_n} - \BFbeta_{y'})^T \hat{\BFx}_n) - k\gamma \mathbb{I}(y_n \neq \hat{y}_n)\\
    & = \rho((\BFbeta_{y_n} - \BFbeta_{y_n^\dagger})^T \hat{\BFx}_n) - \rho((\BFbeta_{\hat{y}_n} - \BFbeta_{\hat{y}_n^\dagger})^T \hat{\BFx}_n) - k\gamma \mathbb{I}(y_n \neq \hat{y}_n)\\
    & \leq \theta |(\BFbeta_{y_n} - \BFbeta_{y_n^\dagger} - \BFbeta_{\hat{y}_n} + \BFbeta_{\hat{y}_n^\dagger})^T \hat{\BFx}_n| - k\gamma \mathbb{I}(y_n \neq \hat{y}_n)\\
    & \leq \theta (\|\BFbeta_{y_n} - \BFbeta_{y_n^\dagger}\|_* + \|\BFbeta_{\hat{y}_n} - \BFbeta_{\hat{y}_n^\dagger}\|_*) \|\hat{\BFx}_n\| - k\gamma \mathbb{I}(y_n \neq \hat{y}_n)\\
    &\leq 2k \|\hat{\BFx}_n\| - k\gamma \mathbb{I}(y_n \neq \hat{y}_n) \\
    &\leq k( 2\|\hat{\BFx}_n\| - \gamma).
\end{align*}
The first inequality is due to the $\theta$-Lipschitz property of $\rho(\cdot)$. The second inequality is due to the triangle inequality and the H\"{o}lder's inequality. The third inequality is due to the constraint $\|\BFbeta_i - \BFbeta_j\|_* \leq \frac{k}{\theta}$. Notice that we obtain 0 when $y_n = \hat{y}_n$. Therefore, we have 
{$$
    \max_{y_n \in \mathcal{Y}} \{ \ell(\BFB^T\hat{\BFx}_n, y_n) - k\gamma \mathbb{I}(y_n \neq \hat{y}_n)\} \leq \ell(\BFB^T\hat{\BFx}_n, \hat{y}_n) + k ( 2\|\hat{\BFx}_n\| - \gamma)_+.
$$}
This implies that any feasible solution to the constraint $\frac{1}{N} \sum_{n \in [N]} \ell(\BFB^T\hat{\BFx}_n, \hat{y}_n) + k (2\|\hat{\BFx}_n\| - \gamma)_+ + R(\BFB) - \tau \leq 0$ is also feasible to the constraint $\frac{1}{N} \sum_{n \in [N]} \max_{y_n \in \mathcal{Y}} \{ \ell(\BFB^T\hat{\BFx}_n, y_n) - k\gamma \mathbb{I}(y_n \neq \hat{y}_n)\} + R(\BFB) - \tau \leq 0$. 

For the equivalence condition, when $\gamma \geq 2 \max_{n\in[N]} \|\hat{\BFx}_n\|$, we have for $y_n \neq \hat{y}_n$,
$$
    \ell(\BFB^T\hat{\BFx}_n, y_n) - k\gamma \mathbb{I}(y_n \neq \hat{y}_n) - \ell(\BFB^T\hat{\BFx}_n, \hat{y}_n) \leq k ( 2\|\hat{\BFx}_n\| - \gamma) \leq 0.
$$
Therefore, 
{$$
    \max_{y_n \in \mathcal{Y}} \{ \ell(\BFB^T\hat{\BFx}_n, y_n) - k\gamma \mathbb{I}(y_n \neq \hat{y}_n)\} = \ell(\BFB^T\hat{\BFx}_n, \hat{y}_n) + k ( 2\|\hat{\BFx}_n\| - \gamma)_+ = \ell(\BFB^T\hat{\BFx}_n, \hat{y}_n).
$$}
Hence, the reformulation is equivalent to the original problem. \hfill \Halmos

\subsection{Proof of Theorem \ref{theorem:lipschitz_approx}}
We first show that $\|\BFzeta\|_* \leq \omega_1$ {for every $y\in\mathcal Y$ and every $\BFzeta \in \dom{\ell^{1*}(\cdot,y)}$. If $\BFp\in\partial_{\BFu}\ell(\BFu,y)$, then for every $\BFh\in\mathbb R^C$, the subgradient inequality and Assumption \ref{asmp:lipschitz} give
}{\[
    \BFp^T\BFh
    \leq \ell(\BFu+\BFh,y)-\ell(\BFu,y)
    \leq \omega_1\|\BFh\|.
\]}
{Taking the supremum over $\|\BFh\|\leq1$ yields $\|\BFp\|_*\leq\omega_1$. Thus, for every $y\in\mathcal Y$, the union of the subdifferentials of $\ell(\cdot,y)$ is contained in the closed dual-norm ball of radius $\omega_1$. Lemma \ref{lemma:equivalence_dom_sub} then implies that $\mathrm{cl}(\dom{\ell^{1*}(\cdot,y)})$, and hence $\dom{\ell^{1*}(\cdot,y)}$, is contained in the same ball}. Therefore,
{$$
    \sup_{\substack{y\in\mathcal Y\\
    \BFzeta \in \dom{\ell^{1*}(\cdot,y)}}}
    \|\BFB\BFzeta\|_* \leq \sup \{\|\omega_1 \BFB\BFzeta\|_* \mid \|\BFzeta\|_* = 1\} = \omega_1 \|\BFB\|_*.
$$}
As a result, the constraint {$\|\BFB\|_* \leq \frac{k}{\omega_1}$ is more conservative than $\sup_{\substack{y\in\mathcal Y\\ \BFzeta \in \dom{\ell^{1*}(\cdot,y)}}} \|\BFB\BFzeta\|_* \leq k$}.

Then, for $y_n \neq \hat{y}_n$, we have
\begin{align*}
    &\phantom{=} \ell(\BFB^T\hat{\BFx}_n, y_n) - k\gamma \mathbb{I}(y_n \neq \hat{y}_n) - \ell(\BFB^T\hat{\BFx}_n, \hat{y}_n) \\
    &\leq \omega_2 \|\BFB^T \hat{\BFx}_n\| - k\gamma \mathbb{I}(y_n \neq \hat{y}_n) \\
    &\leq \omega_2 \|\BFB\|_* \|\hat{\BFx}_n\| - k\gamma \mathbb{I}(y_n \neq \hat{y}_n) \\
    &\leq k \left(\frac{\omega_2}{\omega_1}\|\hat{\BFx}_n\| - \gamma\right).
\end{align*}
Similarly, since we obtain 0 when $y_n = \hat{y}_n$, we have 
{$$
    \max_{y_n \in \mathcal{Y}} \{ \ell(\BFB^T\hat{\BFx}_n, y_n) - k\gamma \mathbb{I}(y_n \neq \hat{y}_n)\} \leq \ell(\BFB^T\hat{\BFx}_n, \hat{y}_n) + k \left(\frac{\omega_2}{\omega_1}\|\hat{\BFx}_n\| - \gamma\right)_+.
$$}
{Summing the preceding pointwise inequality over $n\in[N]$, dividing by $N$, and adding $R(\BFB)-\tau$ gives
}\begin{align*}
&\frac{1}{N}\sum_{n\in[N]}
\max_{y_n\in\mathcal Y}
\left\{
    \ell(\BFB^T\hat{\BFx}_n,y_n)
    -k\gamma\mathbb I(y_n\neq\hat y_n)
\right\}
+R(\BFB)-\tau\\
&\qquad\leq
\frac{1}{N}\sum_{n\in[N]}
\left[
    \ell(\BFB^T\hat{\BFx}_n,\hat y_n)
    +k\left(
        \frac{\omega_2}{\omega_1}\|\hat{\BFx}_n\|-\gamma
    \right)_+
\right]
+R(\BFB)-\tau
\leq0,
\end{align*}
{where the last inequality is the first constraint of Problem \eqref{eq:lipschitz_reformulation}. Together with $\|\BFB\|_*\leq k/\omega_1$, which implies the norm constraint in Problem \eqref{eq:1wass_reformulation} by the preceding argument, this shows that every feasible pair for Problem \eqref{eq:lipschitz_reformulation} is feasible for Problem }\eqref{eq:1wass_reformulation}.

Then, we show the equivalence condition. When $\gamma \geq \frac{\omega_2}{\omega_1} \max_{n\in[N]} \|\hat{\BFx}_n\|$, we have
$$
    \ell(\BFB^T\hat{\BFx}_n, y_n) - k\gamma \mathbb{I}(y_n \neq \hat{y}_n) - \ell(\BFB^T\hat{\BFx}_n, \hat{y}_n) \leq k \left(\frac{\omega_2}{\omega_1}\|\hat{\BFx}_n\| - \gamma\right) \leq 0.
$$
Therefore, 
{$$
    \max_{y_n \in \mathcal{Y}} \{ \ell(\BFB^T\hat{\BFx}_n, y_n) - k\gamma \mathbb{I}(y_n \neq \hat{y}_n)\} = \ell(\BFB^T\hat{\BFx}_n, \hat{y}_n) + k \left(\frac{\omega_2}{\omega_1}\|\hat{\BFx}_n\| - \gamma\right)_+ = \ell(\BFB^T\hat{\BFx}_n, \hat{y}_n).
$$}

{It remains to ensure that the norm bound is exact. Under condition (b), for every $\BFv\in\mathcal V$, there exist $y\in\mathcal Y$ and $\BFu\in\dom{\ell(\cdot,y)}$ }such that {$\omega_1\BFv\in\partial_{\BFu}\ell(\BFu,y)$. The Fenchel--Young equality then implies that $\omega_1\BFv\in\dom{\ell^{1*}(\cdot,y)}$. Hence, every direction on the dual-norm sphere of radius $\omega_1$ is included in the conjugate domain for some $y\in\mathcal Y$, and therefore
}{\[
    \sup_{\substack{y\in\mathcal Y\\
    \BFzeta\in\dom{\ell^{1*}(\cdot,y)}}}
    \|\BFB\BFzeta\|_*
    =\omega_1\|\BFB\|_*,
\]}
{which proves exactness of the norm constraint. Together with condition }(a){, this establishes the claimed equivalence.
}

\hfill \Halmos

\subsection{{Proof of Proposition \ref{prop:fi_operational_cost_certificate}}}
{Fix any common threshold pair with $\ell\leq\theta_{\mathrm{neg}}\leq\theta_{\mathrm{pos}}\leq u$. The expected cost can be written as
\[
\begin{aligned}
    C(p;\theta_{\mathrm{neg}},\theta_{\mathrm{pos}})
    ={}&q_+\Big[(C_{\mathrm{FN}}-C_{\mathrm{man}})
        \mathbb{P}_+\bigl(p(\BFx)<\theta_{\mathrm{neg}}\bigr)
        +C_{\mathrm{man}}\mathbb{P}_+\bigl(p(\BFx)<\theta_{\mathrm{pos}}\bigr)\Big]\\
    &+q_-\Big[C_{\mathrm{man}}\mathbb{P}_-\bigl(p(\BFx)\geq\theta_{\mathrm{neg}}\bigr)
        +(C_{\mathrm{FP}}-C_{\mathrm{man}})
        \mathbb{P}_-\bigl(p(\BFx)\geq\theta_{\mathrm{pos}}\bigr)\Big].
\end{aligned}
\]
All four coefficients are nonnegative. Both thresholds lie in each controlled interval, so the assumed probability inequalities compare every term and give the claimed cost ordering. The decomposition also holds when the thresholds coincide or the score distributions have atoms at either threshold.

For the FI comparison, let $U_r=p_r(\BFx^+)$ and $V_r=p_r(\BFx^-)$ for $r\in\{1,2\}$, where $\BFx^+\sim \mathbb{P}_+$ and $\BFx^-\sim \mathbb{P}_-$ are independent. Define
\[
\begin{aligned}
    \Delta_+(t)&:=\mathbb{P}(U_1<t)-\mathbb{P}(U_2<t),\\
    \Delta_-(t)&:=\mathbb{P}(V_1\geq t)-\mathbb{P}(V_2\geq t).
\end{aligned}
\]
Both functions are left-continuous and take values in $[-1,1]$. Integrating the bounded scores gives, for every $s>0$,
\[
\begin{aligned}
    \mathbb E[e^{-sU_1}]-\mathbb E[e^{-sU_2}]
        &=s\int_0^1 e^{-st}\Delta_+(t)\,dt,\\
    \mathbb E[e^{sV_1}]-\mathbb E[e^{sV_2}]
        &=s\int_0^1 e^{st}\Delta_-(t)\,dt.
\end{aligned}
\]
By assumption, $\Delta_+\geq0$ on $[0,u]$ and is positive at some point there. Left continuity yields $0\leq a<b<u$ and $\delta>0$ such that $\Delta_+(t)\geq\delta$ on $[a,b]$. This also applies when the strict inequality occurs at $u$; it cannot occur at zero. Consequently,
\[
    \mathbb E[e^{-sU_1}]-\mathbb E[e^{-sU_2}]
    \geq \delta(e^{-sa}-e^{-sb})-e^{-su}>0
\]
for all sufficiently large $s$, since $b<u$.

For the negative class, $\Delta_-\geq0$ on $[\ell,1]$ and is positive somewhere on that interval. There exist $0\leq c<d\leq1$ and $\eta>0$ such that $\Delta_-\geq0$ on $[c,1]$ and $\Delta_-\geq\eta$ on $[c,d]$. If a strict inequality holds above $\ell$, choose $c$ and $d$ in a sufficiently small interval to the left of that point, with $c\geq\ell$. If it holds only at $\ell$, then $\ell>0$ and left continuity provides such an interval $[c,\ell]$, with $d=\ell$. Thus,
\[
    \mathbb E[e^{sV_1}]-\mathbb E[e^{sV_2}]
    \geq\eta(e^{sd}-e^{sc})-(e^{sc}-1)>0
\]
for all sufficiently large $s$, since $c<d$. Independence therefore gives an $s_0>0$ such that, for every $s\geq s_0$,
\[
\begin{aligned}
    \mathbb E[e^{s\varepsilon(p_2)}]
    &=\mathbb E[e^{sV_2}]\mathbb E[e^{-sU_2}]\\
    &<\mathbb E[e^{sV_1}]\mathbb E[e^{-sU_1}]
    =\mathbb E[e^{s\varepsilon(p_1)}].
\end{aligned}
\]

For $k>0$, write $H_r(k):=k\ln\mathbb E[e^{\varepsilon(p_r)/k}]$, $r\in\{1,2\}$. The KL variational representation used in Lemma \ref{lemma:fi_kl} gives
\[
    \mathrm{FI}_{\mathrm{KL}}(p_r;\tau)
    =\inf\{k>0\mid H_r(k)\leq\tau\}.
\]
The exponential-moment comparison implies $H_2(k)<H_1(k)$
for $0<k\leq 1/s_0$. Set $\kappa=1/s_0$ and
$\tau_0=H_1(\kappa)$. For every $\tau\geq\tau_0$,
$\kappa$ is feasible in the definition of
$\mathrm{FI}_{\mathrm{KL}}(p_1;\tau)$. Hence, restricting
the infimum to $0<k\leq\kappa$ does not change its value.
Every $k$ in this interval satisfying $H_1(k)\leq\tau$
also satisfies $H_2(k)<H_1(k)\leq\tau$ and is therefore
feasible for $\mathrm{FI}_{\mathrm{KL}}(p_2;\tau)$.
Consequently,
\[
    \mathrm{FI}_{\mathrm{KL}}(p_2;\tau)
    \leq
    \inf\{k\in(0,\kappa]\mid H_1(k)\leq\tau\}
    =
    \mathrm{FI}_{\mathrm{KL}}(p_1;\tau).
\]

Since $\varepsilon(p_1)$ is nonconstant,
\[
    \mathbb E[\varepsilon(p_1)]
    <\tau_0=H_1(\kappa)<M_1.
\]
For every $\tau\in[\tau_0,M_1)$, Lemma \ref{lemma:fi_kl}
gives a positive finite
$k_1=\mathrm{FI}_{\mathrm{KL}}(p_1;\tau)\leq\kappa$
satisfying $H_1(k_1)=\tau$.
Hence $H_2(k_1)<\tau$. By continuity of $H_2$, there
exists $k\in(0,k_1)$ such that $H_2(k)<\tau$.
Therefore,
\[
    0\leq\mathrm{FI}_{\mathrm{KL}}(p_2;\tau)
    \leq k<k_1
    =\mathrm{FI}_{\mathrm{KL}}(p_1;\tau)<\infty.
\]

The choice of $\tau_0$ depends only on the score
distributions and not on the decision thresholds.}\hfill \Halmos

\subsection{Proof of Proposition \ref{prop:nn_fi_tradeoff}}

{For $\lambda_1 > \lambda_2>0$, let $(B_1,k_1)$, with $k_1\geq0$, }be a minimizer for $\lambda_1$ and {$(B_2,k_2)$, with $k_2\geq0$, }be a minimizer for $\lambda_2${, and define
}{\[
R_r:=k_r+\frac{\alpha}{2}\sum_{i,j\in[C],\,i>j}
\bigl(\|\BFbeta_{r,i}-\BFbeta_{r,j}\|_*-k_r\bigr)_+^2,
\qquad r\in\{1,2\},
\]}
{where $\BFbeta_{r,i}$ denotes the $i$-th column of $B_r$. By optimality, after multiplying the two objectives by their respective positive values of $\lambda_r$, we have
}{\[
R_1+\lambda_1L_{ERM}(B_1)\leq R_2+\lambda_1L_{ERM}(B_2)
\]}
{and
}{\[
R_2+\lambda_2L_{ERM}(B_2)\leq R_1+\lambda_2L_{ERM}(B_1).
\]}
{Adding these inequalities yields
}{\[
(\lambda_1-\lambda_2)L_{ERM}(B_1)
\leq
(\lambda_1-\lambda_2)L_{ERM}(B_2).
\]}
Therefore, {$L^*(\lambda_0)$ is non-increasing }in $\lambda_0$.

{Dividing the two optimality inequalities by $\lambda_1$ and $\lambda_2$, respectively, and adding them gives
\[
\left(\frac{1}{\lambda_1}-\frac{1}{\lambda_2}\right)(R_1-R_2)\leq0,
\]
so $R_1\geq R_2$. For $r\in\{1,2\}$, let $v_{r,ij}:=(\|\BFbeta_{r,i}-\BFbeta_{r,j}\|_*-k_r)_+$. Optimality with respect to $k\geq0$ gives
\[
1-\alpha\sum_{i>j}v_{r,ij}
\begin{cases}
=0,& k_r>0,\\
\geq0,& k_r=0.
\end{cases}
\]
Thus $\sum_{i>j}v_{r,ij}\leq1/\alpha$, and
\[
0\leq R_r-k_r
=\frac{\alpha}{2}\sum_{i>j}v_{r,ij}^2
\leq\frac{\alpha}{2}\left(\sum_{i>j}v_{r,ij}\right)^2
\leq\frac{1}{2\alpha}.
\]
Consequently,
\[
k_1\geq R_1-\frac{1}{2\alpha}
\geq R_2-\frac{1}{2\alpha}
\geq k_2-\frac{1}{2\alpha},
\]
which proves the stated bound.}

$\hfill \square$

\subsection{Proof of Proposition \ref{prop:fi_sum_bound}}
\label{appendix:proof_fi_sum_bound}

{{Fix $i\in[C]$ and let $w_{ij}:=N_j/N_{\lnot i}$ for every $j\neq i$. For $k>0$, define
}\begin{align*}
    H_{i|j}(k)
    &:={}
    k\ln\left(
        \mathbb E_{\widehat{\mathbb Q}_{ij}}
        \left[\exp\left(\frac{\varepsilon_{i|j}(p_i)}{k}\right)\right]
    \right),\\
    H_i(k)
    &:={}
    k\ln\left(
        \mathbb E_{\widehat{\mathbb Q}_i}
        \left[\exp\left(\frac{\varepsilon_i(p_i)}{k}\right)\right]
    \right).
\end{align*}
{Because $\widehat{\mathbb Q}_i =\sum_{j\neq i}w_{ij}\widehat{\mathbb Q}_{ij}$, we have
}\begin{equation}
    \label{eq:ova_ovo_entropic_mixture}
    H_i(k)
    =k\ln\left(
        \sum_{j\neq i}w_{ij}
        \exp\left(\frac{H_{i|j}(k)}{k}\right)
    \right).
\end{equation}
{By the KL reformulation and Lemma \ref{lemma:root_finding},
}\begin{align*}
    \mathrm{FI}_{i|j}(p_i;\tau)
    &=\inf\{k>0:H_{i|j}(k)\leq\tau\},\\
    \mathrm{FI}_i(p_i;\tau)
    &=\inf\{k>0:H_i(k)\leq\tau\}.
\end{align*}}

{Let
\begin{equation*}
    k_{\min}:=\min_{j\neq i}\mathrm{FI}_{i|j}(p_i;\tau),
    \qquad
    k_{\max}:=\max_{j\neq i}\mathrm{FI}_{i|j}(p_i;\tau).
\end{equation*}
{For any $k>k_{\max}$, the monotonicity of each $H_{i|j}$ implies $H_{i|j}(k)\leq\tau$ for every $j\neq i$. Therefore,
}\begin{equation*}
    H_i(k)
    \leq
    k\ln\left(
        \sum_{j\neq i}w_{ij}\exp(\tau/k)
    \right)
    =\tau,
\end{equation*}
{which gives $\mathrm{FI}_i(p_i;\tau)\leq k_{\max}$.
}}

{{Conversely, for any $0<k<k_{\min}$}, we have {$H_{i|j}(k)>\tau$ for every $j\neq i$. }Since {all classes are represented in the empirical sample, $w_{ij}>0$ for every $j\neq i$, and thus
}\begin{equation*}
    H_i(k)
    >
    k\ln\left(
        \sum_{j\neq i}w_{ij}\exp(\tau/k)
    \right)
    =\tau.
\end{equation*}
{Hence, $\mathrm{FI}_i(p_i;\tau)\geq k_{\min}$}.}

{Finally, if all nominal pairwise distributions $\widehat{\mathbb Q}_{ij}$ induce the same distribution of $\varepsilon_{i|j}(p_i)$, then all $H_{i|j}$ coincide and \eqref{eq:ova_ovo_entropic_mixture} shows that $H_i$ is equal to this common function. Consequently, $\mathrm{FI}_i(p_i;\tau)=\mathrm{FI}_{i|j}(p_i;\tau)$ for every $j\neq i$, so both bounds are attained. This proves their tightness.
}

\hfill \Halmos

\subsection{Proof of Proposition \ref{prop:fi_wass_reformulation_binary_calculation}}

{
Fix $k>0$. Optimizing over the first marginal of a transport coupling gives
\begin{align*}
    G(k)+\tau
    &={}\sup_{\mathbb Q\in\mathcal P(\mathcal X^2)}
        \left\{\mathbb E_{\mathbb Q}[\varepsilon(p)]
        -kW(\mathbb Q,\widehat{\mathbb Q})\right\}\\
    &={}\sup_{\pi:\,\pi_2=\widehat{\mathbb Q}}
        \mathbb E_\pi\!\left[
            p(\BFx^-)-p(\BFx^+)
            -k\left(c_+(\BFx^+,\widehat{\BFx}^{\, +})
            +c_-(\BFx^-,\widehat{\BFx}^{\, -})\right)
        \right]\\
    &={}\mathbb E_{\widehat{\mathbb Q}}\!\left[
        \sup_{\BFx^+,\BFx^-\in\mathcal X}
        \left\{p(\BFx^-)-p(\BFx^+)
        -k\left(c_+(\BFx^+,\widehat{\BFx}^{\, +})
        +c_-(\BFx^-,\widehat{\BFx}^{\, -})\right)\right\}
        \right].
\end{align*}
The nominal pair law has finite support, so its conditional transported distributions can be optimized separately at each atom. Approximate maximizers suffice if a supremum is not attained. The additive cost separates the supremum over $(\BFx^+,\BFx^-)$ into the sum of the two coordinatewise suprema. Taking their expectations under the nominal class marginals proves \eqref{eq:binary_pair_wasserstein_reformulation}.

Now specialize to the norm costs and logistic classifier in the proposition. For a fixed nominal observation $\widehat{\BFx}$, write $z=\BFw^\top\widehat{\BFx}+b$. By H\"older's inequality, every $\BFx$ with $\BFw^\top\BFx+b=t$ satisfies
\[
    |t-z|\leq L\|\BFx-\widehat{\BFx}\|.
\]
The dual norm is attained on the unit sphere in finite dimensions. Hence there exists $\BFv$ with $\|\BFv\|=1$ and $\BFw^\top\BFv=L$, and the point
$\BFx=\widehat{\BFx}+(t-z)\BFv/L$ attains equality. Because $\mathcal X=\mathbb R^M$,
\[
    \inf_{\BFx:\,\BFw^\top\BFx+b=t}
        \|\BFx-\widehat{\BFx}\|
    =\frac{|t-z|}{L}.
\]
Consequently, with $\lambda=k/L$,
\[
    \sup_{\BFx\in\mathbb R^M}
        \{p(\BFx)-k\|\BFx-\widehat{\BFx}\|\}
    =\sup_{t\in\mathbb R}\{\sigma(t)-\lambda|t-z|\}
    =\sup_{t\geq z}\{\sigma(t)-\lambda(t-z)\}.
\]
The last equality uses the monotonicity of $\sigma$, since every $t<z$ is worse than $t=z$.

If $\lambda\geq1/4$, the bound $\sigma'(t)\leq1/4$ shows that the maximum is attained at $z$ and equals $\sigma(z)$. If $0<\lambda<1/4$, the stationary points of $\sigma(t)-\lambda(t-z)$ are $-a_\lambda$ and $a_\lambda$. Its derivative is negative on $(-\infty,-a_\lambda)$, positive on $(-a_\lambda,a_\lambda)$, and negative on $(a_\lambda,\infty)$. Thus $-a_\lambda$ is a local minimum, and the maximum on $[z,\infty)$ is attained at $z$ or, when $z<a_\lambda$, at $a_\lambda$. This proves
\[
    \sup_{\BFx\in\mathbb R^M}
        \{p(\BFx)-k\|\BFx-\widehat{\BFx}\|\}
    =H_{k/L}(z).
\]
Using $-\sigma(t)=\sigma(-t)-1$ and applying the same calculation to $-\BFw$ and $-b$ gives
\[
    \sup_{\BFx\in\mathbb R^M}
        \{-p(\BFx)-k\|\BFx-\widehat{\BFx}\|\}
    =H_{k/L}(-z)-1.
\]
Substitution into \eqref{eq:binary_pair_wasserstein_reformulation} yields \eqref{eq:logistic_wasserstein_fi_explicit}. The infimum characterization of FI follows from its definition, as in Lemma \ref{lemma:root_finding}.

It remains to establish the search boundaries. Evaluating at $\mathbb Q=\widehat{\mathbb Q}$ gives
$G(k)\geq\mathbb E_{\widehat{\mathbb Q}}[\varepsilon(p)]-\tau$, so no $k$ is feasible when the target is below the nominal mean. For $\tau\geq1$, the bound $\varepsilon(p)\leq1$ makes every $k>0$ feasible. For $k\geq L/4$, both pointwise transforms equal their nominal values, and therefore
\[
    G(k)=\mathbb E_{\widehat{\mathbb Q}}[\varepsilon(p)]-\tau.
\]
For each finite $z$, the supremum defining $H_\lambda(z)$ is at most $1$ and has limit $1$ as $\lambda\downarrow0$. Indeed, evaluating at any fixed $t\geq z$ gives a limiting lower bound $\sigma(t)$, which tends to $1$ as $t\to\infty$. Thus $G(k)\to1-\tau$ as $k\downarrow0$. The explicit branches of $H_\lambda(z)$ agree at $z=a_\lambda$ and continuously at $\lambda=1/4$, so $G$ is continuous for $k>0$, and it is non-increasing because the transport costs are nonnegative. If the target lies between the nominal mean and $1$, with the upper endpoint excluded, $G$ is positive near zero and nonpositive at $L/4$. Its smallest feasible value is therefore attained in $(0,L/4]$, proving the stated bounds and the validity of bracketing bisection even when the zero set is a plateau.

Finally, restricting the feature domain can only decrease each pointwise supremum. Any $k$ feasible for the full-space expression is consequently feasible for the restricted-domain FI, which proves the stated upper-bound interpretation. If $\BFw=\BFzero$, then $\varepsilon(p)=0$ for every pair, so $G(k)=-\tau$ for every $k>0$ and the constant-score convention follows.
\hfill \Halmos
}

\subsection{{Proof of Lemma \ref{lemma:common_losses_probability_dominance}}}

{By the softmax definition,
}{\[
    \ell_{CE}(\BFB^T\BFx,y)
    =
    \ln\left(\sum_{i\in[C]}\exp(\BFbeta_i^T\BFx)\right)
    -\BFbeta_y^T\BFx
    =-\ln p_y(\BFx;\BFB).
\]}
{Since $p_y(\BFx;\BFB)\in(0,1]$, the elementary inequality $1-z\leq-\ln z$ for $z\in(0,1]$ gives
}{\[
    1-p_y(\BFx;\BFB)
    \leq
    -\ln p_y(\BFx;\BFB)
    =\ell_{CE}(\BFB^T\BFx,y).
\]}
{Thus, one may take $a=1$ and $b=0$.
}

{{For the margin-$\kappa$ multi-class hinge loss with $\kappa>0$, define
}{\[
    \ell_{hinge,\kappa}(\BFB^T\BFx,y)
    :=
    \max_{y'\neq y}
    \left(\kappa-(\BFbeta_y-\BFbeta_{y'})^T\BFx\right)_+,
    \qquad
    m_{yy'}:=(\BFbeta_y-\BFbeta_{y'})^T\BFx.
\]}
{For every $y'\neq y$, the definition of $\ell_{hinge,\kappa}$ implies $m_{yy'}\geq\kappa-\ell_{hinge,\kappa}(\BFB^T\BFx,y)$. Therefore,
}\begin{align*}
    1-p_y(\BFx;\BFB)
    &=
    \frac{\sum_{y'\neq y}\exp(-m_{yy'})}{1+\sum_{y'\neq y}\exp(-m_{yy'})}\\
    &\leq
    \frac{(C-1)\exp\left(\ell_{hinge,\kappa}(\BFB^T\BFx,y)-\kappa\right)}{1+(C-1)\exp\left(\ell_{hinge,\kappa}(\BFB^T\BFx,y)-\kappa\right)}\\
    &=
    \sigma\!\left(\ell_{hinge,\kappa}(\BFB^T\BFx,y)+\ln(C-1)-\kappa\right),
    \qquad
    \sigma(t):=\frac{1}{1+\exp(-t)}.
\end{align*}
{Since $\sigma'(t)=\sigma(t)(1-\sigma(t))\leq1/4$ and $\ell_{hinge,\kappa}\geq0$, it follows that
}\begin{align*}
    1-p_y(\BFx;\BFB)
    &\leq
    \frac{1}{4}\ell_{hinge,\kappa}(\BFB^T\BFx,y)
    +\sigma\!\left(\ln(C-1)-\kappa\right)\\
    &=
    \frac{1}{4}\ell_{hinge,\kappa}(\BFB^T\BFx,y)
    +\frac{(C-1)e^{-\kappa}}{1+(C-1)e^{-\kappa}}.
\end{align*}
{Hence the margin-$\kappa$ hinge loss satisfies Assumption \ref{asmp:probability_loss_dominance} with $a=1/4$ and $b=(C-1)e^{-\kappa}/(1+(C-1)e^{-\kappa})$. }}\hfill \Halmos

\subsection{{Proof of {Theorem \ref{theorem:ce_entropy_ovo_bound}}}}
\label{proof:theorem_ce_entropy_ovo_bound}

{Fix an ordered pair $(i,j)$ with $i\neq j$, and write $r:=q_i+q_j$ and $w:=q_i/r$. Let $\mathbb Q$ be any pair law with finite pair discrepancy under the divergence or distance in the theorem, and let $\mathbb{Q}_i$ and $\mathbb{Q}_j$ be its marginals. Use the lifted joint law $\mathcal L_{ij}(\mathbb Q)$ defined in the proof of Theorem \ref{theorem:kl_wasserstein_ovo_bound}.
}

{{For any $(\BFx^i,\BFx^j)\in\mathcal X^2$, set $u:=p_i(\BFx^i;\BFB)$ and $v:=p_j(\BFx^j;\BFB)$. Since $\ell_{\mathrm{CE}}(\BFB^T\BFx,c)=-\ln p_c(\BFx;\BFB)$ and $p_i(\BFx^j;\BFB)\leq1-v$, the ranking error $\varepsilon_{i|j}(p_i)$ satisfies
}{\[
    \varepsilon_{i|j}(p_i)
    \leq 1-u-v.
\]}
{Moreover, $-\ln z\geq1-z$ for every $z>0$, and hence
}\begin{align*}
    &w\ell_{\mathrm{CE}}(\BFB^T\BFx^i,i)
    +(1-w)\ell_{\mathrm{CE}}(\BFB^T\BFx^j,j)
    -H_{\mathrm b}(w)\\
    &\quad=
    w\left[-\ln\left(\frac{u}{w}\right)\right]
    +(1-w)\left[-\ln\left(\frac{v}{1-w}\right)\right]\\
    &\quad\geq
    w\left(1-\frac{u}{w}\right)
    +(1-w)\left(1-\frac{v}{1-w}\right)
    =1-u-v.
\end{align*}
{Combining the preceding two displays and taking expectation under $\mathbb Q$ gives
}\begin{align*}
    \mathbb E_{\mathbb Q}[\varepsilon_{i|j}(p_i)]
    \leq{}&
    \frac{1}{r}
    \left[
    q_i\mathbb E_{\mathbb{Q}_i}
    [\ell_{\mathrm{CE}}(\BFB^T\BFx,i)]
    +q_j\mathbb E_{\mathbb{Q}_j}
    [\ell_{\mathrm{CE}}(\BFB^T\BFx,j)]
    \right]
    -H_{\mathrm b}(w).
\end{align*}}

{Applying the feasibility of $(k,\BFB)$ at $\mathcal L_{ij}(\mathbb Q)$ and using \eqref{eq:ce_entropy_target} therefore yields
}{\[
    \mathbb E_{\mathbb Q}[\varepsilon_{i|j}(p_i)]
    \leq
    t^{\mathrm{CE}}_{ij}(\BFB)
    +\frac{k}{q_i+q_j}
    \Delta\left(
    \mathcal L_{ij}(\mathbb Q),\hat{\mathbb P}
    \right),
\]}
{where $\Delta$ is $D_{\mathrm{KL}}$ in the KL case and $W$ in the Wasserstein case. The KL chain-rule and Wasserstein coupling arguments in the proof of Theorem \ref{theorem:kl_wasserstein_ovo_bound} give, in the respective cases,
}{\[
    \Delta\left(
    \mathcal L_{ij}(\mathbb Q),\hat{\mathbb P}
    \right)
    \leq
    \max\{q_i,q_j\}
    D\left(\mathbb Q,\widehat{\mathbb Q}_{ij}\right).
\]}
{Consequently,
}{\[
    \mathbb E_{\mathbb Q}[\varepsilon_{i|j}(p_i)]
    \leq
    t^{\mathrm{CE}}_{ij}(\BFB)
    +\frac{\max\{q_i,q_j\}}{q_i+q_j}k
    D\left(\mathbb Q,\widehat{\mathbb Q}_{ij}\right).
\]}
{The constraint is automatic when the pair discrepancy is infinite because $k>0$. This proves \eqref{eq:ce_entropy_pair_bound}. Finally, $t^{\mathrm{CE}}_{\max}(\BFB)\geq t^{\mathrm{CE}}_{ij}(\BFB)$ for every ordered pair, and FI is non-increasing in its target. Averaging \eqref{eq:ce_entropy_pair_bound} over ordered pairs proves \eqref{eq:ce_entropy_macro_bound}. }\hfill \Halmos

\subsection{Proof of Proposition \ref{prop:convex_ot_reformulation}}
{For fixed $k$, $\hat{\BFx}$, and $y$, the function inside the supremum in \eqref{eq:c_transformed_loss} is convex in $\BFB$. Hence, its pointwise supremum is also convex in $\BFB$.
}

{For the conjugate representation, fix $k>0$ and let $\delta_\mathcal X$ denote the indicator function of $\mathcal X$. By the Fenchel--Moreau representation of the loss,
}

\begin{align*}
    &\phantom{=} \sup_{\BFx \in \mathcal{X}} \left\{ \ell(\BFB^T\BFx, y) - k c(\BFx, \hat{\BFx}) \right\} \\ 
    &= \sup_{\BFx \in \mathbb{R}^M} \left\{ \ell(\BFB^T \BFx, y) - k c(\BFx, \hat{\BFx}) - \delta_\mathcal{X}(\BFx) \right\} \\
    &= \sup_{\BFx \in \mathbb{R}^M} \sup_{\BFzeta \in \dom{\ell^{1*}}} \left\{ \BFzeta^T \BFB^T \BFx - \ell^{1*}(\BFzeta, y) - k c(\BFx, \hat{\BFx}) - \delta_\mathcal{X}(\BFx) \right\} \\
    &= \sup_{\BFzeta \in \dom{\ell^{1*}}} \sup_{\BFx \in \mathbb{R}^M}  \left\{ \BFzeta^T \BFB^T \BFx  - (k c(\BFx, \hat{\BFx}) + \delta_\mathcal{X}(\BFx)) - \ell^{1*}(\BFzeta, y) \right\} \\
    &= \sup_{\BFzeta \in \dom{\ell^{1*}}} \inf_{\BFtheta \in \dom{\delta^*_{\mathcal{X}}}} \left\{
        kc^{1*}((\BFB\BFzeta - \BFtheta)/k, \hat{\BFx}) + \delta_\mathcal{X}^*(\BFtheta) - \ell^{1*}(\BFzeta, y) 
    \right\},
\end{align*}
{where the third line uses $\ell(\BFu,y)=\sup_{\BFzeta\in\dom{\ell^{1*}}}\{\BFzeta^T\BFu-\ell^{1*}(\BFzeta,y)\}$. The }last equality follows from the {infimal-convolution identity for the conjugate of $kc(\cdot,\hat{\BFx})+\delta_\mathcal X$. The relative-interior condition in Assumption \ref{asmp:wass_convexity} ensures that this identity is exact, and $[kc(\cdot,\hat{\BFx})]^{1*}(\BFzeta)=kc^{1*}(\BFzeta/k,\hat{\BFx})$}.

When the {feature-support constraint is removed by taking $\mathcal{X}=\mathbb{R}^M$}, we have 
$$
    \delta_\mathcal{X}^*(\BFtheta) = \sup_{\BFx \in \mathcal{X}} \BFtheta^T \BFx = \begin{cases}
        0, & \BFtheta = \BFzero, \\
        +\infty, & \BFtheta \neq \BFzero.
    \end{cases}
$$
Hence, it reduces to 
$$
\sup_{\BFzeta \in \dom{\ell^{1*}}} \inf_{\BFtheta \in \dom{\delta^*_{\mathcal{X}}}} \left\{
    kc^{1*}((\BFB\BFzeta - \BFtheta)/k, \hat{\BFx}) + \delta_\mathcal{X}^*(\BFtheta) - \ell^{1*}(\BFzeta, y) 
\right\} = \sup_{\BFzeta \in \dom{\ell^{1*}}} \left\{
    kc^{1*}((\BFB\BFzeta)/k, \hat{\BFx}) - \ell^{1*}(\BFzeta, y)
\right\}.
$$

\hfill \Halmos

\subsection{{Proof of {Proposition
\ref{prop:cross_entropy_exact_all_gamma}}}} {{The proof of Theorem \ref{theorem:cross_entropy_reformulation} establishes the exact equivalence
}{\[
    \sup_{y\in[C]}\sup_{\BFzeta\in\dom{\ell^{1*}(\cdot,y)}}
    \|\BFB\BFzeta\|_*\leq k
    \quad\Longleftrightarrow\quad
    \|\BFbeta_i-\BFbeta_j\|_*\leq k,
    \quad \forall i,j\in[C],\ i<j.
\]}
{It remains only to represent the finite maximum over candidate labels in Problem \eqref{eq:1wass_reformulation}. For each $n\in[N]$, }the {epigraph identity
}\begin{align*}
    s_n\geq
    \max_{y\in[C]}\big\{
        \ell_{CE}(\BFB^T\hat{\BFx}_n,y)
        -k\gamma\mathbb I(y\neq\hat y_n)
    \big\}
    \quad\Longleftrightarrow\quad
    s_n\geq
    \ell_{CE}(\BFB^T\hat{\BFx}_n,y)
    -k\gamma\mathbb I(y\neq\hat y_n),
    \quad \forall y\in[C],
\end{align*}
{converts that maximum into the $NC$ label-wise inequalities in \eqref{eq:cross_entropy_exact_all_gamma}. Combining these inequalities with the aggregate constraint }and the {pairwise norm characterization gives exactly Problem \eqref{eq:1wass_reformulation}. Each label-wise loss inequality is convex in $(k,\BFB,s_n)$, and all remaining constraints are convex. No norm-based upper bound on the label-wise loss difference is used, so the equivalence holds for every $\gamma\geq0$. }}\hfill \Halmos

\subsection{{Proof of Proposition
\ref{prop:hinge_exact_all_gamma}}} {{The proof of Theorem \ref{theorem:hinge_type_reformulation} establishes the exact conjugate-domain characterization
}{\[
    \sup_{y\in[C]}\sup_{\BFzeta\in\dom{\ell^{1*}(\cdot,y)}}
    \|\BFB\BFzeta\|_*\leq k
    \quad\Longleftrightarrow\quad
    \|\BFbeta_i-\BFbeta_j\|_*\leq\frac{k}{\theta},
    \quad \forall i,j\in[C],\ i<j.
\]}
{For each $n\in[N]$, epigraphing the remaining finite maximum over $y\in[C]$ gives
}\begin{align*}
    s_n\geq
    \max_{y\in[C]}\big\{
        \ell_{hinge}(\BFB^T\hat{\BFx}_n,y)
        -k\gamma\mathbb I(y\neq\hat y_n)
    \big\}
    \quad\Longleftrightarrow\quad
    s_n\geq
    \ell_{hinge}(\BFB^T\hat{\BFx}_n,y)
    -k\gamma\mathbb I(y\neq\hat y_n),
    \quad \forall y\in[C].
\end{align*}
{Together with the aggregate constraint and the pairwise norm inequalities, this is exactly Problem \eqref{eq:1wass_reformulation}. Since $\ell_{hinge}(\BFB^T\hat{\BFx}_n,y)$ is the maximum of the $C-1$ convex functions indexed by $y'\neq y$, each label-wise inequality is convex; fully expanding this maximum gives one scalar inequality for every $(n,y,y')$ with $y'\neq y$. Again, no high-$\gamma$ bound is used, so the equivalence holds for every $\gamma\geq0$. }}\hfill \Halmos

\subsection{Proof of Theorem \ref{theorem:piecewise_reformulation}}
{By Lemma \ref{lemma:lower_bound_k}, every optimal solution under Assumption \ref{asmp:tau} has $k>0$. Thus, restricting the following epigraph formulation to $k>0$ does not change the optimal solutions of Problem \eqref{eq:prob_general}.
}

{Using }the piecewise linear {representation of $\ell(\BFu,y)$, we first write the training problem in epigraph form as
}

\begin{align*}
    \min_{k>0,\BFB\in\mathcal B,\BFs} & \ k \\
    \text{s.t.} & \ \frac{1}{N} \sum_{n \in [N]} s_n + R(\BFB) - \tau \leq 0\\
    & \ \max_{y\in\mathcal Y}\left\{
    \max_{i\in[K_y]}\sup_{\BFx\in\mathcal X}
    \left\{\BFa_{yi}^T\BFB^T\BFx+b_{yi}-kc(\BFx,\hat{\BFx}_n)\right\}
    -k\gamma\mathbb I(y\neq\hat y_n)\right\}\leq s_n,
    & \forall n\in[N].
\end{align*}
{For every $i\in[K_y]$, $n\in[N]$, and $y\in\mathcal Y$, Proposition \ref{prop:convex_ot_reformulation} gives
}\begin{align*}
    &\phantom{=}\sup_{\BFx\in\mathcal X}
    \left\{\BFa_{yi}^T\BFB^T\BFx+b_{yi}-kc(\BFx,\hat{\BFx}_n)\right\}\\
    &=\sup_{\BFx\in\mathbb R^M}
    \left\{\BFa_{yi}^T\BFB^T\BFx+b_{yi}-kc(\BFx,\hat{\BFx}_n)
    -\delta_\mathcal X(\BFx)\right\}\\
    &=\inf_{\BFv_{iny}\in\dom{\delta_\mathcal X^*}}
    \left\{b_{yi}+\delta_\mathcal X^*(\BFv_{iny})
    +kc^{1*}((\BFB\BFa_{yi}-\BFv_{iny})/k,\hat{\BFx}_n)\right\}.
\end{align*}
{Substituting this identity and enumerating $i\in[K_y]$ and $y\in\mathcal Y$ yields
}\begin{align*}
    \min_{k>0, \BFB \in \mathcal{B}, \BFs, \BFv_{iny}} & \ k \\
            \text{s.t.} \hspace*{20pt} & \frac{1}{N} \sum_{n \in [N]} s_n + R(\BFB) - \tau \leq 0, \\
            & b_{yi} + \delta_\mathcal{X}^*(\BFv_{iny}) + k c^{1*}((\BFB \BFa_{yi} - \BFv_{iny})/k, \hat{\BFx}_n ) -  k \gamma \mathbb{I}(y \neq \hat{y}_n) \leq s_n, \\
            &\hspace{200pt} \forall i \in [K_y], n \in [N], y \in \mathcal{Y}. 
\end{align*}
{The term $kc^{1*}((\BFB\BFa_{yi}-\BFv_{iny})/k,\hat{\BFx}_n)$ is the perspective }of the convex {function $c^{1*}(\cdot,\hat{\BFx}_n)$, and $\delta_\mathcal X^*$ is convex as a support function}. Therefore, the {resulting }formulation is convex.

\hfill \Halmos

\subsection{Proof of Corollary \ref{cor:eg_uncertainty}}

\textbf{(a)} {For the polyhedral support set, its support function admits the linear-programming representation
}{$$
    \delta_\mathcal{X}^*(\BFz) = \max_\BFx \left\{
        \BFz^T \BFx: \BFA \BFx \leq \BFd
    \right\} = \min_\BFlambda \left\{
        \BFlambda^T \BFd: \BFlambda \geq 0, \BFA^T \BFlambda = \BFz
    \right\}
$$}
{Substituting this identity into Theorem \ref{theorem:piecewise_reformulation} gives
}\begin{align*}
    \min_{k>0,\BFB \in \mathcal{B}, \BFs, \BFlambda_{iny}} & \ k\\
    \text{s.t.} &\frac{1}{N} \sum_{n \in [N]} s_n + R(\BFB) - \tau \leq 0\\
    & \ b_{yi} + \BFd^T \BFlambda_{iny}  + k c^{1*}((\BFB\BFa_{yi} - \BFA^T \BFlambda_{iny})/k, \hat{\BFx}_n ) -  k \gamma \mathbb{I}(y \neq \hat{y}_n) \leq s_n, & \forall i \in [K_y], n \in [N], y \in \mathcal{Y}, \\
    & \BFlambda_{iny} \geq 0, & \forall i \in [K_y], n \in [N], y \in \mathcal{Y}.
\end{align*}

\textbf{(b)} {Under the stated strict-feasibility condition, convex duality gives
}\begin{align*}
    \delta_\mathcal{X}^*(\BFz) &= \max_\BFx \left\{
        \BFz^T \BFx: f_j(\BFx) \leq 0, j \in [J]
    \right\} \\
    &= \min_\BFlambda \left\{
        \left[\sum_{j\in[J]} \lambda_j f_j\right]^*(\BFz): \BFlambda \geq 0
    \right\} \\
    &= \min_{\BFlambda, \BFz_1, \dots, \BFz_J}\left\{
        \sum_{j\in[J]} \lambda_j f_j^*\left(\frac{\BFz_j}{\lambda_j}\right): \BFlambda \geq 0, \sum_{j\in[J]} \BFz_j = \BFz 
    \right\}.
\end{align*}
Therefore, {Theorem \ref{theorem:piecewise_reformulation} yields
}\begin{align*}
    \min_{k>0,\BFB \in \mathcal{B}, \BFs, \BFlambda_{iny}, \BFz_{inyj}, \BFv_{iny}} & \ k\\
    \text{s.t.} & \frac{1}{N} \sum_{n \in [N]} s_n + R(\BFB) - \tau \leq 0\\
    & \ b_{yi} + \sum_{j\in[J]} \lambda_{inyj} f_j^*\left(\frac{\BFz_{inyj}}{\lambda_{inyj}}\right) \\
    & \quad + k c^{1*}((\BFB \BFa_{yi} - \BFv_{iny})/k, \hat{\BFx}_n ) -  k \gamma \mathbb{I}(y \neq \hat{y}_n) \leq s_n,
    && \forall i \in [K_y], n \in [N], y \in \mathcal{Y}, \\
    & \sum_{j\in[J]} \BFz_{inyj} = \BFv_{iny},
    && \forall i \in [K_y], n \in [N], y \in \mathcal{Y},\\
    & \BFlambda_{iny} \geq 0,
    && \forall i \in [K_y], n \in [N], y \in \mathcal{Y}.
\end{align*}

\hfill \Halmos

\subsection{Proof of Lemma \ref{lemma:generalization}}

{Feasibility of $(k^*,\BFB^*)$ in Problem \eqref{eq:prob_general} gives, for every $\mathbb P\in\mathcal P(\mathcal X\times\mathcal Y)$,
}{\[
    \mathbb E_{\mathbb P}[\ell(\BFB^{*T}\BFx,y)]+R(\BFB^*)
    \leq
    \tau+k^*D(\mathbb P,\hat{\mathbb P}).
\]}
{Subtracting $\mathbb E_{\hat{\mathbb P}}[\ell(\BFB^{*T}\BFx,y)]$ from both sides yields the result}.

\hfill \Halmos

\subsection{Proof of Proposition \ref{prop:generalization_error}}
As a consequence of the light-tail assumption, we have the following bound on the Wasserstein distance between $\phat_N$ and $\ptrue$. 
\begin{lemmaAp}
    \label{lemma:wass_bound}
    (Theorem 2, \cite{fournier2015rate}) If Assumption \ref{asmp:light_tail} holds, we have 
    \begin{equation*}
        (\ptrue)^{\otimes N}\left\{
            D_{\mathrm{W}}^\dagger(\phat_N, \ptrue) \geq \theta
        \right\} \leq \begin{cases}
            C_1 \exp(-C_2 N \theta^{\max\{m,2\}}), & \ \text{if}\ \theta \leq 1, \\
            C_1 \exp(-C_2 N \theta^a), & \ \text{if}\ \theta > 1,
        \end{cases}
\end{equation*}
    for all $N \geq 1$, $m\neq 2$, and $\theta > 0$, where $m$ is the dimension of $(\BFx, y)$, and {$C_1$ and $C_2$ }are positive constants {depending }on $a${, }$m${, and the exponential moment in Assumption \ref{asmp:light_tail}}.
\end{lemmaAp}

{The }Wasserstein distance in Lemma \ref{lemma:wass_bound} is
{\[
    D_{\mathrm{W}}^\dagger(\phat_N, \ptrue)
    :=\inf_{\pi \in \Pi(\phat_N,\ptrue)}
    \mathbb{E}_{\pi} [\|(\BFx, y) - (\hat{\BFx}, \hat{y})\|].
\]}
{This differs from the ground-cost-based distance in Equation \eqref{eq:def_ot}, but the two distances satisfy
}\begin{align*}
    W(\phat_N,\ptrue)
    &=\inf_{\pi\in\Pi(\phat_N,\ptrue)}
    \mathbb E_\pi[\|\BFx-\hat{\BFx}\|+\gamma\mathbb I(y\neq\hat y)]\\
    &\leq(1+\gamma)D_{\mathrm W}^\dagger(\phat_N,\ptrue).
\end{align*}
{Indeed, under the product norm in Assumption \ref{asmp:light_tail}, both $\|\BFx-\hat{\BFx}\|$ and $\mathbb I(y\neq\hat y)$ are bounded by $\|(\BFx,y)-(\hat{\BFx},\hat y)\|$ for $y,\hat y\in[C]$}.

{We next apply the Kantorovich--Rubinstein duality to the metric $c$ in Equation \eqref{eq:cost_1wass}}.

\begin{lemmaAp}
    \label{lemma:kr_duality}
    {Suppose that $\phat_N$ and $\ptrue$ have finite first moments with respect to $c$. For the metric $c$ in Equation \eqref{eq:cost_1wass},
    }\begin{equation}
        \label{eq:kr_duality}
        W(\phat_N, \ptrue) = \sup_{\|f\|_L \leq 1} \left\{
            \mathbb{E}_{\ptrue}[f(\BFx, y)] - \mathbb{E}_{\phat_N}[f(\BFx, y)]
        \right\},
\end{equation}
    where
    {\[
        \|f\|_L := \inf\left\{\omega \geq 0 \colon
        |f(\BFx,y)-f(\BFx',y')|
        \leq \omega c(\BFx,y,\BFx',y'),\ 
        \forall (\BFx,y),(\BFx',y')\in\mathcal X\times\mathcal Y\right\}.
\]}
\end{lemmaAp}

For completeness, we provide a proof {of Lemma \ref{lemma:kr_duality} in Section \ref{sec:proof_kr_duality}. Using Assumption \ref{asmp:lipschitz}, the Lipschitz modulus of $\ell(\hat{\BFB}_N^T\BFx,y)$ with respect to $c$ satisfies
}\begin{align*}
    &\phantom{=} \left| \ell(\hat{\BFB}_N^T \BFx, y) - \ell(\hat{\BFB}_N^T \BFx', y') \right| \\
    & \leq |\ell(\hat{\BFB}_N^T \BFx, y) - \ell(\hat{\BFB}_N^T \BFx', y)| + |\ell(\hat{\BFB}_N^T \BFx', y) - \ell(\hat{\BFB}_N^T \BFx', y')| \\
    & \leq \omega_1 \|\hat{\BFB}_N^T\|_* \|\BFx - \BFx'\| + \omega_2 \|\hat{\BFB}_N^T\|_* \|\BFx'\| \mathbb{I} (y \neq y') \\
    & \leq \|\hat{\BFB}_N^T\|_* \max\left\{\omega_1, \omega_2 \frac{\sup_{(\BFx, y) \in \supp{\ptrue}} \|\BFx\|}{\gamma}\right\} (\|\BFx - \BFx'\| + \gamma \mathbb{I} (y \neq y')) \\
    & = \|\hat{\BFB}_N^T\|_* \max\left\{\omega_1, \omega_2 \frac{\sup_{(\BFx, y) \in \supp{\ptrue}} \|\BFx\|}{\gamma}\right\} c(\BFx, y, \BFx', y').
\end{align*}

Let {$\omega = \max\left\{\omega_1, \omega_2 \frac{\sup_{(\BFx, y) \in \supp{\ptrue}} \|\BFx\|}{\gamma}\right\}$}. Then,

\begin{align*}
    &\mathbb{E}_{\ptrue}[\ell(\hat{\BFB}_N^T \BFx, y)] - \mathbb{E}_{\phat_N}[\ell(\hat{\BFB}_N^T \BFx, y)]  \\
    &\leq \sup_{ \|f\|_L \leq \|\hat{\BFB}_N^T\|_* \omega} 
    \left\{
        \mathbb{E}_{\ptrue}[f(\BFx, y)] - \mathbb{E}_{\phat_N}[f(\BFx, y)]
    \right\} \\
    &= \|\hat{\BFB}_N^T\|_* \omega W(\phat_N, \ptrue) \\
    & \leq (1 + \gamma) \|\hat{\BFB}_N^T\|_* \omega D_{\mathrm{W}}^\dagger(\phat_N, \ptrue).
\end{align*}

{Since $M\geq2$, the ambient dimension $m=M+1$ is at least three. Thus, Lemma \ref{lemma:wass_bound} implies, for any $\epsilon\in(0,1)$ and all sufficiently large $N$,
}\begin{align*}
    (\ptrue)^{\otimes N}\left\{
         D_{\mathrm{W}}^\dagger(\phat_N, \ptrue) \leq \left(\frac{1}{NC_2} \log\left(\frac{C_1}{\epsilon}\right)\right)^{\frac{1}{M+1}} 
    \right\} 
    \geq 
    1- \epsilon.
\end{align*}
Consequently, with probability at least $1-\epsilon$, we have
{$$
    \mathbb{E}_{\ptrue}[\ell(\hat{\BFB}_N^T \BFx, y)] - \mathbb{E}_{\phat_N}[\ell(\hat{\BFB}_N^T \BFx, y)] \leq (1+\gamma)\omega \|\hat{\BFB}_N^T\|_* D_{\mathrm{W}}^\dagger(\phat_N, \ptrue) \leq (1+\gamma)\omega \|\hat{\BFB}_N^T\|_* \left(\frac{1}{NC_2} \log\left(\frac{C_1}{\epsilon}\right)\right)^{\frac{1}{M+1}}.
$$}

The second term in the final bound {follows from Lemma \ref{lemma:generalization}:
}\begin{align*}
    &\mathbb{E}_{\ptrue}[\ell(\hat{\BFB}_N^T \BFx, y)] - \mathbb{E}_{\phat_N}[\ell(\hat{\BFB}_N^T \BFx, y)]\\
    &\leq \tau-R(\hat{\BFB}_N)-\mathbb{E}_{\phat_N}[\ell (\hat{\BFB}_N^T\BFx, y)]
    +\hat k_N W(\ptrue,\phat_N) \\
    &\leq \tau-R(\hat{\BFB}_N)-\mathbb{E}_{\phat_N}[\ell (\hat{\BFB}_N^T\BFx, y)]
    +(1+\gamma)\hat k_N D_{\mathrm{W}}^\dagger(\ptrue,\phat_N).
\end{align*}
{Consequently, with }probability at least $1-\epsilon$,
{\[
    \mathbb{E}_{\ptrue}[\ell(\hat{\BFB}_N^T \BFx, y)] - \mathbb{E}_{\phat_N}[\ell(\hat{\BFB}_N^T \BFx, y)]
    \leq \tau-R(\hat{\BFB}_N)-\mathbb{E}_{\phat_N}[\ell (\hat{\BFB}_N^T\BFx, y)]
    +(1+\gamma)\hat k_N
    \left(\frac{1}{NC_2} \log\left(\frac{C_1}{\epsilon}\right)\right)^{\frac{1}{M+1}}.
\]}

\hfill \Halmos

\subsection{Proof of Proposition \ref{prop:convergence}}
(a) For the strong regularity, we follow the definition in \cite{shapiro2021lectures}, which further dates back to \cite{robinson1980strongly}. We exhibit the definition here for completeness. 

\begin{definitionAp} 
    \label{def:strong_regularity}
    (Strong regularity, \cite{robinson1980strongly}) Suppose that the mapping $\BFPsi(\BFw)$ is continuously differentiable. We say that a solution $\BFw^*$ of the SGE \eqref{eq:sge_true} is strongly regular if there exists a neighborhood $\mathcal{N}_1$ of $\BFzero \in \mathbb{R}^{M_w}$ and $\mathcal{N}_2$ of $\BFw^*$, such that for every $\BFdelta \in \mathcal{N}_1$, the (linearized) perturbed SGE
    \begin{equation*}
        \BFzero \in \BFdelta + \BFPsi(\BFw^*) + \nabla \BFPsi(\BFw^*) (\BFw - \BFw^*) + \BFGamma(\BFw) 
    \end{equation*}
    has a unique solution in $\mathcal{N}_2$. Let $\BFw(\BFdelta)$ denote this solution. Then, the mapping $\BFdelta \to \BFw(\BFdelta)$ is Lipschitz continuous on $\mathcal{N}_1$.
\end{definitionAp}

According to \cite{izmailov2003karush}, the strong regularity of the smooth KKT system is equivalent to LICQ and SSOSC, which are guaranteed by Assumption \ref{asmp:smooth_regularity} (b). 

(b) The convergence of $\hat{\BFw}_N$ is a direct result of the following theorem of \cite{shapiro2021lectures}. 
\begin{lemmaAp}
    \label{lemma:convergence_shapiro}
    (Theorem 5.15 of \cite{shapiro2021lectures}) Let $\BFC$ be a compact set and $\BFw^*$ be a unique solution of the SGE \eqref{eq:sge_true} in $\BFC$. Suppose that:
    \begin{enumerate}[(a)]
        \item the set $\BFGamma(\BFw)$ is closed;
        \item for {$\ptrue$-almost every $(\BFx,y)$}, the mapping {$\BFw \mapsto \BFPsi(\BFw;\BFx,y)$ }is continuously differentiable on $\BFC$, and {$\sup_{\BFw\in\BFC}\|\BFPsi(\BFw;\BFx,y)\|$ and $\sup_{\BFw\in\BFC}\|\nabla_{\BFw}\BFPsi(\BFw;\BFx,y)\|$ }are dominated by {a $\ptrue$-integrable }function;
        \item the solution $\BFw^*$ is strongly regular;
        \item $\hat{\BFPsi}_N(\BFw)$ and $\nabla \hat{\BFPsi}_N(\BFw)$ converge with probability 1 to $\BFPsi(\BFw)$ and $\nabla \BFPsi(\BFw)$, respectively, uniformly on $\BFC$.
    \end{enumerate} 
    Then, with probability 1, $\hat{\BFw}_N$ converges to $\BFw^*$ as $N \to \infty$.
\end{lemmaAp}
The condition (a) is obvious from the definition of $\BFGamma(\BFw)$. The condition (b) is guaranteed by Assumption \ref{asmp:sge_bound_integrable} (b). The condition (c) is the part (a) of the Proposition. {For }condition (d), {the samples are iid under $\ptrue$, }and {Assumption \ref{asmp:sge_bound_integrable}(b), together with }Theorem 7.48 {of }\cite{shapiro2021lectures}{, yields the almost-sure uniform convergence of $\hat{\BFPsi}_N$ and $\nabla\hat{\BFPsi}_N$ to their population counterparts}. Hence, $\hat{\BFw}_N$ converges to $\BFw^*$ as $N \to \infty$.

(c) {Strict complementarity and the almost-sure convergence in part (b) imply local stability of the active set}. {Thus, for all sufficiently large }$N$ {almost surely, the reduced empirical solution satisfies
}{\[
    \hat{\BFPsi}_N'(\hat{\BFw}_N')=\BFzero.
\]}
{A first-order expansion around $\BFw'^*$ gives
}{\[
    \sqrt N(\hat{\BFw}_N'-\BFw'^*)
    =-A^{-1}\sqrt N\bigl(\hat{\BFPsi}_N'(\BFw'^*)-\BFPsi'(\BFw'^*)\bigr)+o_p(1).
\]}
{Because $\BFPsi'(\BFw'^*)=\BFzero$, the multivariate }central limit theorem {yields
}{\[
    \sqrt N\,\hat{\BFPsi}_N'(\BFw'^*)
    \xrightarrow{d}\mathcal N(\BFzero,\Sigma').
\]}
{The stated result then follows from Slutsky's theorem.
}

\hfill \Halmos

\subsection{Proof of Proposition \ref{prop:exponential_convergence}}
\textbf{(a)} This is a direct result of the uniform Large Deviation Theorem. Assumption \ref{asmp:sge_bound_integrable} {provides the compact set $\mathcal W$ and integrable domination, while Assumption \ref{asmp:lipschitz_sge_finite_mgf} controls the corresponding }moment-generating {functions}. By Theorem 7.65 of \cite{shapiro2021lectures}, we have that for sufficiently small $\epsilon > 0$, there {exist }positive constants $\delta_{1i} (\epsilon)$ and $\delta_{2i} (\epsilon)$ for each $i \in [M_w]$ such that 
{$$
    (\ptrue)^{\otimes N} \left\{
        \sup_{\BFw \in \mathcal{W}} \left\| (\hat{\BFPsi}_N)_i(\BFw) - \BFPsi_i(\BFw) \right\| \geq \epsilon
    \right\} \leq \delta_{1i}(\epsilon) \exp(-\delta_{2i}(\epsilon) N).
$$}
According to Theorem 2.9 of \cite{chen2019convergence}, we can obtain \eqref{eq:exp_converge_psi} by {combining }these inequalities. As to the determination of {the functions }$\delta_{1i}(\epsilon)$ and $\delta_{2i}(\epsilon)$, we refer to \cite{shapiro2021lectures} for the details. More nuanced assumptions are required to obtain the explicit form of $\delta_{1i}(\epsilon)$ and $\delta_{2i}(\epsilon)$.

\textbf{(b)} Suppose $\| \hat{\BFw}_N - \BFw^* \| \geq \epsilon$. Then, we have
\begin{align*}
    0 & = \inf_{\BFgamma \in \BFGamma(\hat{\BFw}_N)} \| \hat{\BFPsi}_N(\hat{\BFw}_N) + \BFgamma \| \\
    & = \inf_{\BFgamma \in \BFGamma(\hat{\BFw}_N)} \| \hat{\BFPsi}_N(\hat{\BFw}_N) + \BFgamma + \BFPsi(\hat{\BFw}_N) - \BFPsi(\hat{\BFw}_N) \|  \\
    & \geq \inf_{\BFgamma \in \BFGamma(\hat{\BFw}_N)} \| \BFPsi(\hat{\BFw}_N) + \BFgamma \| - \| \hat{\BFPsi}_N(\hat{\BFw}_N) - \BFPsi(\hat{\BFw}_N) \| \\
    & \geq \rho(\epsilon) - \sup_{\BFw \in \mathcal{W}} \left\| \hat{\BFPsi}_N(\BFw) - \BFPsi(\BFw) \right\|.
\end{align*}
The first inequality is due to the triangle inequality of $\|\BFa - \BFb\| \geq \|\BFa\| - \|\BFb\|$. The second inequality is the composition of 
\begin{gather*}
    \inf_{\BFgamma \in \BFGamma(\hat{\BFw}_N)} \| \BFPsi(\hat{\BFw}_N) + \BFgamma \| \geq \rho(\epsilon),\\
    \| \hat{\BFPsi}_N(\hat{\BFw}_N) - \BFPsi(\hat{\BFw}_N) \| \leq \sup_{\BFw \in \mathcal{W}} \left\| \hat{\BFPsi}_N(\BFw) - \BFPsi(\BFw) \right\|.
\end{gather*}

Therefore, if $\| \hat{\BFw}_N - \BFw^* \| \geq \epsilon$ holds, we must have $\sup_{\BFw \in \mathcal{W}} \left\| \hat{\BFPsi}_N(\BFw) - \BFPsi(\BFw) \right\| \geq \rho(\epsilon)$. Notice that for any $\epsilon > 0$, we have $\rho(\epsilon) > 0$ as a result of the uniqueness of the solution $\BFw^*$, which guarantees that the inequality $\sup_{\BFw \in \mathcal{W}} \left\| \hat{\BFPsi}_N(\BFw) - \BFPsi(\BFw) \right\| \geq \rho(\epsilon)$ is nontrivial. Therefore,  
{$$
    (\ptrue)^{\otimes N} \left\{
        \left\|\hat{\BFw}_N - \BFw^* \right\| \geq \epsilon
    \right\} \leq (\ptrue)^{\otimes N} \left\{
        \sup_{\BFw \in \mathcal{W}} \left\| \hat{\BFPsi}_N(\BFw) - \BFPsi(\BFw) \right\| \geq \rho(\epsilon)
    \right\} \leq \delta_1(\rho(\epsilon)) \exp(-\delta_2(\rho(\epsilon)) N).
$$}
\hfill \Halmos

\subsection{Proof of Proposition \ref{prop:dro_reformulation} and \ref{prop:dro_fi_reformulation}}
{The stated upper approximations and exactness conditions follow by applying the arguments used in the proofs of Theorems \ref{theorem:cross_entropy_reformulation} and \ref{theorem:hinge_type_reformulation}.
}

\hfill \Halmos

\subsection{Proof of Proposition \ref{prop:generalization_fi_dro}}
{Since $\mathcal P(\phat_N,\epsilon)\subseteq\mathcal P(\mathcal X\times\mathcal Y)$, the solution $(\hat k_N,\hat{\BFB}_N)$ is feasible for Problem \eqref{eq:fi_dro} }with {$\BFB=\hat{\BFB}_N$. Hence, $k^\dagger_{N,\epsilon}\leq\hat k_N$. Let
}{\[
    r_N(\delta):=\left(\frac{1}{C_2N}\log\left(\frac{C_1}{\delta}\right)\right)^{\frac{1}{M+1}}.
\]}
{The concentration argument in the proof of Proposition \ref{prop:generalization_error} shows that, with probability at least $1-\delta$ under $(\ptrue)^{\otimes N}$,
}{\[
    W(\ptrue,\phat_N)\leq(1+\gamma)r_N(\delta)<\epsilon.
\]}
{Thus, $\ptrue\in\mathcal P(\phat_N,\epsilon)$ on this event. Feasibility of $k^\dagger_{N,\epsilon}$ then yields
}\begin{align*}
    &\mathbb E_{\ptrue}[\ell(\hat{\BFB}_N^T\BFx,y)]-\mathbb E_{\phat_N}[\ell(\hat{\BFB}_N^T\BFx,y)]\\
    &\leq \tau-R(\hat{\BFB}_N)-\mathbb E_{\phat_N}[\ell(\hat{\BFB}_N^T\BFx,y)]
    +k^\dagger_{N,\epsilon}W(\ptrue,\phat_N)\\
    &\leq \tau-R(\hat{\BFB}_N)-\mathbb E_{\phat_N}[\ell(\hat{\BFB}_N^T\BFx,y)]
    +(1+\gamma)k^\dagger_{N,\epsilon}r_N(\delta).
\end{align*}
{Combining this inequality with the first bound established in the proof }of Proposition \ref{prop:generalization_error} {gives \eqref{eq:generalization_error_bound_fi_dro}}.

\hfill \Halmos

\subsection{{Proof of Lemma \ref{lemma:cost_optimal_selective_thresholds}}}
\label{proof:lemma_cost_optimal_selective_thresholds}
{{Fix a score value $s\in[0,1]$ at which calibration holds, so that $\mathbb P(Y=1\mid p(\BFx)=s)=s$. An automated negative decision incurs $C_{\mathrm{FN}}$ only when $Y=1$, whereas an automated positive decision incurs $C_{\mathrm{FP}}$ only when $Y=0$. Manual review incurs the fixed per-patient cost $C_{\mathrm{man}}$ regardless of the true class and, because it is assumed to be error-free, incurs no additional misclassification cost}. Thus, {conditional on $p(\BFx)=s$, the three expected costs are
}{\[
    R_-(s)=sC_{\mathrm{FN}},
    \qquad
    R_{\mathrm{man}}(s)
    =sC_{\mathrm{man}}+(1-s)C_{\mathrm{man}}
    =C_{\mathrm{man}},
    \qquad
    R_+(s)=(1-s)C_{\mathrm{FP}}.
\]}
{An automated negative decision is optimal exactly when $R_-(s)\leq R_{\mathrm{man}}(s)$ and $R_-(s)\leq R_+(s)$. These inequalities are equivalent to
}{\[
    s\leq\frac{C_{\mathrm{man}}}{C_{\mathrm{FN}}}
    \qquad\text{and}\qquad
    s\leq\frac{C_{\mathrm{FP}}}{C_{\mathrm{FP}}+C_{\mathrm{FN}}},
\]}
{and hence to $s\leq\theta_{\mathrm{neg}}^*$. Similarly, an automated positive decision is optimal exactly when $R_+(s)\leq R_{\mathrm{man}}(s)$ and $R_+(s)\leq R_-(s)$, or equivalently,
}{\[
    s\geq1-\frac{C_{\mathrm{man}}}{C_{\mathrm{FP}}}
    \qquad\text{and}\qquad
    s\geq\frac{C_{\mathrm{FP}}}{C_{\mathrm{FP}}+C_{\mathrm{FN}}},
\]}
{which is equivalent to $s\geq\theta_{\mathrm{pos}}^*$.
}}

{{Let
}{\[
    \theta_{\mathrm{cls}}:=\frac{C_{\mathrm{FP}}}{C_{\mathrm{FP}}+C_{\mathrm{FN}}},
    \qquad
    \bar C:=
    \frac{C_{\mathrm{FP}}C_{\mathrm{FN}}}{C_{\mathrm{FP}}+C_{\mathrm{FN}}}.
\]}
{If $C_{\mathrm{man}}<\bar C$, then
}{\[
    \theta_{\mathrm{neg}}^*=\frac{C_{\mathrm{man}}}{C_{\mathrm{FN}}}
    <\theta_{\mathrm{cls}}<
    1-\frac{C_{\mathrm{man}}}{C_{\mathrm{FP}}}
    =\theta_{\mathrm{pos}}^*.
\]}
{In this case, the automated negative decision minimizes the conditional cost below $\theta_{\mathrm{neg}}^*$, manual review minimizes it between $\theta_{\mathrm{neg}}^*$ and $\theta_{\mathrm{pos}}^*$, and the automated positive decision minimizes it above $\theta_{\mathrm{pos}}^*$. At each boundary, manual review ties with the adjacent automated decision. If $C_{\mathrm{man}}\geq\bar C$, then $\theta_{\mathrm{neg}}^*=\theta_{\mathrm{pos}}^*=\theta_{\mathrm{cls}}$ and the review region is empty: the automated negative and positive decisions are optimal for $s<\theta_{\mathrm{cls}}$ and $s\geq\theta_{\mathrm{cls}}$, respectively, and they tie at $s=\theta_{\mathrm{cls}}$}. Therefore, {under the half-open review convention $\theta_{\mathrm{neg}}^*\leq s<\theta_{\mathrm{pos}}^*$, the policy minimizes the conditional expected cost for every score value $s$ at which calibration holds. Taking expectations over $p(\BFx)$ establishes its optimality for the overall expected operational cost.
}}

\hfill \Halmos

\subsection{Proof of Lemma \ref{lemma:equivalence_dom_sub}}
\label{sec:proof_fenchel_young}
Let $\mathcal R(\partial f):=\bigcup_{\BFx\in\mathbb R^n}\partial f(\BFx)$. If $\BFp\in\partial f(\BFx^\dagger)$, the equality condition in the Fenchel--Young inequality gives
\[
    f^*(\BFp)=\BFp^T\BFx^\dagger-f(\BFx^\dagger)<\infty.
\]
Hence, $\mathcal R(\partial f)\subseteq\dom{f^*}$.

Conversely, let $\BFp\in\dom{f^*}$. For every $\epsilon>0$, the definition of the supremum provides $\BFx_\epsilon\in\mathbb R^n$ such that
\[
    \BFp^T\BFx_\epsilon-f(\BFx_\epsilon)
    \geq f^*(\BFp)-\epsilon,
\]
which is equivalent to $\BFp\in\partial_\epsilon f(\BFx_\epsilon)$. By the Br{{\o}}ndsted--Rockafellar approximation theorem \citep{rockafellar1970convex}{, there exist $\widetilde{\BFx}_\epsilon\in\mathbb R^n$ and $\widetilde{\BFp}_\epsilon\in\partial f(\widetilde{\BFx}_\epsilon)$ such that $\|\widetilde{\BFp}_\epsilon-\BFp\|_2\leq\sqrt{\epsilon}$. Therefore, $\BFp\in\mathrm{cl}(\mathcal R(\partial f))$. We conclude that
}{\[
    \mathcal R(\partial f)
    \subseteq\dom{f^*}
    \subseteq\mathrm{cl}(\mathcal R(\partial f)),
\]}
{and taking closures proves the claim}.

\hfill \Halmos

\subsection{Proof of Lemma \ref{lemma:kr_duality}}
\label{sec:proof_kr_duality}
{Let $\mathcal Z:=\mathcal X\times\mathcal Y$, write $z=(\BFx,y)$, and set $\mathbb P=\ptrue$ and $\widehat{\mathbb P}=\phat_N$. Because $c$ is a metric, $W(\phat_N,\ptrue)=W(\mathbb P,\widehat{\mathbb P})$. Standard Kantorovich duality }\citep{kantorovich1958space} {gives
}{\[
    W(\mathbb P,\widehat{\mathbb P})
    =
    \sup_{\substack{f,g:\\ f(z)+g(z')\leq c(z,z')}}
    \left\{
        \mathbb E_{\mathbb P}[f(z)]
        +\mathbb E_{\widehat{\mathbb P}}[g(z)]
    \right\}.
\]}

{First, let $h$ satisfy $\|h\|_L\leq1$. Then $h(z)-h(z')\leq c(z,z')$, so the pair $f=h$ }and {$g=-h$ is feasible in the preceding dual problem. Hence,
}{\[
    \mathbb E_{\mathbb P}[h(z)]
    -\mathbb E_{\widehat{\mathbb P}}[h(z)]
    \leq W(\mathbb P,\widehat{\mathbb P}).
\]}
{Taking the supremum over all such $h$ yields
}{\[
    \sup_{\|h\|_L\leq1}
    \left\{
        \mathbb E_{\mathbb P}[h(z)]
        -\mathbb E_{\widehat{\mathbb P}}[h(z)]
    \right\}
    \leq W(\mathbb P,\widehat{\mathbb P}).
\]}

{Conversely, fix any feasible pair $(f,g)$ in the Kantorovich dual and define
}{\[
    \widetilde h(z)
    :=\inf_{z'\in\mathcal Z}\{c(z,z')-g(z')\}.
\]}
{Feasibility implies $f(z)\leq\widetilde h(z)$, while choosing $z'=z$ gives $\widetilde h(z)\leq-g(z)$. Moreover, the triangle inequality gives, for every $z,\bar z\in\mathcal Z$,
}{\[
    \widetilde h(z)
    \leq c(z,\bar z)+\widetilde h(\bar z).
\]}
{Interchanging $z$ and $\bar z$ therefore yields
}{\[
    |\widetilde h(z)-\widetilde h(\bar z)|
    \leq c(z,\bar z),
\]}
{so $\|\widetilde h\|_L\leq1$. The finite-first-moment condition ensures that these Lipschitz functions are integrable after an immaterial additive normalization. Consequently},
\begin{align*}
    \mathbb E_{\mathbb P}[f(z)]
    +\mathbb E_{\widehat{\mathbb P}}[g(z)]
    &\leq
    \mathbb E_{\mathbb P}[\widetilde h(z)]
    -\mathbb E_{\widehat{\mathbb P}}[\widetilde h(z)]\\
    &\leq
    \sup_{\|h\|_L\leq1}
    \left\{
        \mathbb E_{\mathbb P}[h(z)]
        -\mathbb E_{\widehat{\mathbb P}}[h(z)]
    \right\}.
\end{align*}
{Taking the supremum over all feasible pairs $(f,g)$ gives the reverse inequality. Combining the two inequalities proves \eqref{eq:kr_duality}}.

\hfill \Halmos
\end{APPENDIX}

\putbib

\end{bibunit}

\end{document}